\documentclass[11pt,letter]{article}
\usepackage{epsfig,graphics}
\usepackage{graphicx,subcaption}
\usepackage{amssymb,amsmath,array,multirow,float}
\usepackage[authoryear]{natbib}
\usepackage{bm,algorithm,algorithmicx}
\usepackage{algpseudocode}
\usepackage{color}
\usepackage{calligra,mathabx}
\usepackage{mathrsfs}
\usepackage{tikz}
\usepackage{enumitem}
\usepackage{booktabs}

\usepackage[framemethod=tikz]{mdframed}
\usepackage{hyperref}
\hypersetup{colorlinks=true,citecolor=blue,urlcolor=blue}
\usepackage[capitalize]{cleveref}

\DeclareMathAlphabet{\mathcalligra}{T1}{calligra}{m}{n}
\DeclareFontShape{T1}{calligra}{m}{n}{<->s*[2.2]callig15}{}

\renewcommand{\theequation} {\arabic{equation}}

\usepackage{amsmath, amsthm, amssymb, amsfonts}

\usepackage{sectsty}
\sectionfont{\Large}
\subsectionfont{\large}

\usepackage{comment}
\usepackage[all,cmtip]{xy}

\newcommand{\dbip}[1]{\langle\kern-1.0ex~\langle \,#1 \,\rangle\kern-1.0ex~\rangle_\Hb}
\newcommand{\ldbip}[1]{\langle\kern-1.0ex~\langle \,#1 \,\rangle_\Hb}
\newcommand{\rdbip}[2]{\langle #1 \rangle\kern-1.0ex~\rangle_{\Hb, \Hb^{#2}}}

\newcommand{\ip}[1]{\langle\,#1\,\rangle}

\newcommand{\norm}[1]{\Vert#1\Vert}
\newcommand{\normbig}[1]{\left\Vert#1\right\Vert}

\renewcommand{\bold}{\boldsymbol}

\def\Var{\mbox{Var}}

\def\diag{\mbox{diag}}

\def\cv{\mathbf c}

\def\fv{\mathbf f}
\def\gv{\mathbf g}

\def\gv{\mathbf g}

\def\uv{\mathbf u}

\def\xv{\mathbf x}

\def\Xv{\mathbf X}

\def\Zv{\mathbf Z}

\newcommand{\alphav}{\mbox{\boldmath{$\alpha$}}}
\newcommand{\betav}{\mbox{\boldmath{$\beta$}}}

\newcommand{\deltav}{\mbox{\boldmath{$\delta$}}}

\newcommand{\etav}{\mbox{\boldmath{$\eta$}}}

\newcommand{\sumin}{\sum_{i=1}^{n}}

\newcommand{\Bc}{\mathcal{B}}

\newcommand{\Gc}{\mathcal{G}}

\newcommand{\Mc}{\mathcal{M}}

\newcommand{\Qc}{\mathcal{Q}}

\newcommand{\Sc}{\mathcal{S}}
\newcommand{\Tc}{\mathcal{T}}
\newcommand{\Uc}{\mathcal{U}}

\newcommand{\Wc}{\mathcal{W}}

\newcommand{\Bs}{\mathscr{B}}
\newcommand{\Cs}{\mathscr{C}}

\newcommand{\Ms}{\mathscr{M}}

\newcommand{\Bb}{\mathbb{B}}

\newcommand{\Eb}{\mathbb{E}}

\newcommand{\Hb}{\mathbb{H}}

\newcommand{\Nb}{\mathbb{N}}

\newcommand{\Pb}{\mathbb{P}}

\newcommand{\Rb}{\mathbb{R}}

\newcommand{\Ub}{\mathbb{U}}
\newcommand{\Vb}{\mathbb{V}}
\newcommand{\Wb}{\mathbb{W}}
\newcommand{\Xb}{\mathbb{X}}

\newcommand{\bz}{\bold{0}}

\newcommand{\sumjd}{\sum_{j=1}^d}

\newcommand{\hPi}{{\widehat\Pi}}

\newcommand{\hf}{{\widehat f}}

\newcommand{\tf}{{\widetilde f}}

\newcommand{\red}{\textcolor{red}}

\newcommand{\rmv}{{\rm v}}
\newcommand{\rmc}{{\rm c}}

\newcommand{\rmd}{{\rm d}}

\newcommand{\rmtp}{{\rm tp}}

\newcommand{\rmop}{{\rm op}}

\newcommand{\rmtl}{{\rm TL}}

\newcommand{\sumj}{\sum_{j=1}^d}
\newcommand{\sumknotj}{\sum_{k=1,\not=j}^d}
\newcommand{\dxj}{\, \rmd x_j}
\newcommand{\dxk}{\, \rmd x_k}
\newcommand{\dxl}{\, \rmd x_l}
\newcommand{\dxv}{\, \rmd \xv}
\newcommand{\rmpen}{{\rm pen}}

\newcommand{\hp}{{\widehat p}}
\newcommand{\tp}{{\widetilde p}}
\newcommand{\hatf}{{\widehat f}}
\newcommand{\hatfv}{{\widehat \fv}}

\newcommand{\hatm}{{\widehat m}}

\newcommand{\sH}{{\mathscr H}}

\newcommand{\rmadd}{{\rm add}}
\newcommand{\ve}{{\varepsilon}}
\newcommand{\sumjs}{{\sum_{j\in\Sc_\bfnum{0}}}}
\newcommand{\sumjnots}{{\sum_{j\not\in\Sc_\bfnum{0}}}}
\newcommand{\sF}{{\mathscr F}}
\newcommand{\tPi}{{\widetilde \Pi}}

\newcommand{\sD}{{\mathscr D}}
\newcommand{\cD}{{\mathcal D}}

\newcommand{\cS}{{\mathcal S}}

\newcommand{\rmmin}{{\rm min}}
\newcommand{\rmmax}{{\rm max}}
\newcommand{\sumiiprime}{{\underset{1\le i\ne i'\le n}{\sum\sum}}}

\newcommand{\sG}{{\mathscr G}}
\newcommand{\cA}{{\mathcal A}}

\newcommand{\bfnum}[1]{{\bf #1}}
\newcommand{\bfsnum}[2]{{{\bf #1}| #2}}
\newcommand{\hM}{{\widehat M}}
\newcommand{\tM}{{\widetilde M}}
\newcommand{\Mnum}[1]{{M_\bfnum{#1}}}
\newcommand{\hMnum}[1]{{\hM_\bfnum{#1}}}
\newcommand{\tMnum}[1]{{\tM_\bfnum{#1}}}
\newcommand{\suminum}[1]{{\sum_{i=1}^{n_\bfnum{#1}}}}

\newcommand{\duj}{{\, \rmd u_j}}
\newcommand{\duk}{{\, \rmd u_k}}

\newcommand{\pnum}[1]{{p_\bfnum{#1}}}
\newcommand{\ab}{{\mathbf{a}}}
\newcommand{\sumaac}{{\sum_{\ab\in\cA}}}

\newcommand{\KL}[2]{K\left(#1\,\middle\|\,#2\right)}
\newcommand{\chisqdiv}[2]{\chi^2\left(#1\,\middle\|\,#2\right)}

\newcommand{\opnumnum}[2]{{\bfnum{#1}|\rmop,{#2}}}

\newcommand{\sk}{{\mathfrak s}}
\newcommand{\rmprod}{{\rm prod}}

\newcommand{\hc}{{\widehat \rmc}}
\newcommand{\tc}{{\widetilde \rmc}}
\newcommand{\rmI}{{\rm I}}
\newcommand{\sumjk}{{\underset{1\le j<k\le d}{\sum\sum}}}

\newcommand{\angl}[1]{{\langle #1 \rangle}}
\newcommand{\rmuniv}{{\rm univ}}
\newcommand{\rmbiv}{{\rm biv}}
\newcommand{\hL}{{\widehat L}}
\newcommand{\hdelta}{{\widehat \delta}}
\newcommand{\hdeltav}{{\widehat \deltav}}
\newcommand{\tnu}{{\widetilde \nu}}

\newtheorem{lemma}{\sc Lemma}

\newtheorem{theorem}{\sc Theorem}
\newtheorem{remark}{\sc Remark}

\newtheorem{proposition}{\sc Proposition}

\newtheorem{corollary}{\sc Corollary}

\numberwithin{equation}{section}

\begin{document}
	
\begin{center}
{\LARGE \bf Minimax optimal transfer learning for high-dimensional additive regression (draft version)}
\author{Seung Hyun Moon}

\vspace{20pt}

{\large Seung Hyun Moon$^a$}

\bigskip
{\it $^a$Seoul National University, Seoul, South Korea}

\bigskip

\today

\end{center}
	
\begin{abstract}
This paper studies high-dimensional additive regression under the transfer learning framework, where one observes samples from a target population together with auxiliary samples from different but potentially related regression models. We first introduce a target-only estimation procedure based on the smooth backfitting estimator with local linear smoothing. In contrast to previous work, we establish general error bounds under sub-Weibull($\alpha$) noise, thereby accommodating heavy-tailed error distributions. In the sub-exponential case ($\alpha=1$), we show that the estimator attains the minimax lower bound under regularity conditions, which requires a substantial departure from existing proof strategies. We then develop a novel two-stage estimation method within a transfer learning framework, and provide theoretical guarantees at both the population and empirical levels. Error bounds are derived for each stage under general tail conditions, and we further demonstrate that the minimax optimal rate is achieved when the auxiliary and target distributions are sufficiently close. All theoretical results are supported by simulation studies and real data analysis.
\end{abstract}

\clearpage

\section{Introduction}

Many human tasks benefit from prior experience when that experience is related to the task at hand. This phenomenon, whereby knowledge from previous tasks is transferred to new ones, has motivated the machine learning technique known as transfer learning. From a statistical perspective, consider the problem of analyzing a regression relationship when the available data are limited. Transfer learning (\cite{torrey2010transfer}), one of the most widely used techniques in machine learning, can provide a solution. In this framework, one typically leverages related estimates obtained from large but non-identically distributed \textit{auxiliary samples}, and then refines these estimates to obtain improved estimators from the smaller \textit{target sample}. Transfer learning has been shown to be effective in a wide range of real-world applications, including computer vision (\cite{kolesnikov2020big, bu2021gaia}), natural language processing (\cite{lee2020biobert, yuan2020parameter}), and bioinformatics (\cite{vorontsov2024foundation, gao2020deep}), among others.

Recently, the theoretical properties of transfer-learned estimators have been extensively investigated across a range of statistical problems. There exists a rich collection of works on classification (\cite{reeve2021adaptive, cai2021transfer, qin2025adaptive, fan2023robust}), high-dimensional linear regression (\cite{li2022transfer, tian2023transfer}), non- or semi-parametric regression (\cite{liu2023augmented, hu2023optimal, cai2024transfer}), piecewise constant mean estimation (\cite{wang2025tlpiece}), and graphical models (\cite{li2023tlgraphical}). Despite this growing literature, to the best of our knowledge, no work has addressed nonparametric regression in the high-dimensional regime where the number of covariates $d$ diverges. This gap motivates the present study.

There are few works on sparse high-dimensional additive modeling itself. Within this line of research, studies assuming $\ell_1$-type sparsity include spline-based approaches (\cite{Meier2009}), RKHS-based approaches (\cite{Raskutti2012}), and more recently kernel smoothing-based methods (\cite{Lee2024}). In particular, \cite{Raskutti2012} established the minimax optimality of the proposed estimator, and \cite{yuan2016minimax} further extended this by considering $\ell_q$-type sparsity in RKHS-based high-dimensional additive model estimation, also proving minimax optimality. While RKHS-based estimators are theoretically appealing, their practical applicability is limited. For instance, the analysis in this line of work does not provide an explicit algorithm for implementation. To overcome this limitation, \cite{Lee2024} proposed an efficient kernel-smoothing-based procedure. However, the aforementioned study employs a Nadaraya–Watson type estimator, which is known to fall short of achieving minimax optimality even in low-dimensional settings. To overcome this limitation, it is necessary to develop an estimator based on local linear smoothing, which attains minimax optimality. Moreover, such a refinement is inevitable for constructing minimax optimal transfer-learned estimators.

Accordingly, the contributions of this paper can be summarized in three parts. First, we establish improved error bounds under conditions weaker than those in \cite{Lee2024}. In particular, we introduce the notion of sub-Weibull noise (\cite{arun2022moving}) to capture heavy-tailed errors, and by combining $U$-statistics (\cite{chakrabortty2018tail}) with a new theoretical approach, we demonstrate that the resulting improvement is not merely a consequence of extending to local linear estimation but instead yields fundamentally sharper bounds. To illustrate this briefly, consider the additive regression model  
\begin{align*}
f_\bfnum{0}(\xv) := \Eb(Y_\bfnum{0} \mid \Xv_\bfnum{0}=\xv) = \Eb(Y_\bfnum{0}) + f_\bfsnum{0}{1}(x_1) + \cdots + f_\bfsnum{0}{d}(x_d),
\end{align*}
where only $|\cS_\bfnum{0}|$ of the component functions $f_\bfsnum{0}{j}$ are nonzero. Throughout, the subscript $\bfnum{0}$ is used to indicate the target population. In \cite{Lee2024}, the error bound is shown to satisfy  
\begin{align*}
\norm{\hf_\bfnum{0}^{\rm Lee} - f_\bfnum{0}}^2 \lesssim |\cS_\bfnum{0}|\left(h_\bfnum{0}^3 + \frac{\log d}{n_\bfnum{0}h_\bfnum{0}}\right),
\end{align*}
where $\hf_\bfnum{0}^{\rm Lee}$ denotes the Nadaraya–Watson type fLasso–SBF estimator for $f_\bfnum{0}$ proposed in \cite{Lee2024} and $h_\bfnum{0}$ is the bandwidth. Roughly speaking, the term $h_\bfnum{0}^{3}$ arises from smoothing bias, whereas the term $\frac{\log d}{n_\bfnum{0}h_\bfnum{0}}$ corresponds to the variance contribution. A natural extension to the local linear smoothing approach yields  
\begin{align} \label{normal-ll}
\norm{\hf_\bfnum{0} - f_\bfnum{0}}^2 \lesssim |\cS_\bfnum{0}|\left(h_\bfnum{0}^4 + \frac{\log d}{n_\bfnum{0}h_\bfnum{0}}\right),
\end{align}
where $\hf_\bfnum{0}$ denotes the locally linear fLasso–SBF estimator for $f_\bfnum{0}$ proposed in this paper. However, in Theorem~\ref{thm:ll-bound-error-emp} we establish that  
\begin{align} \label{new-ll}
\norm{\hf_\bfnum{0} - f_\bfnum{0}}^2 \lesssim |\cS_\bfnum{0}|\left(h_\bfnum{0}^4 + \frac{1}{n_\bfnum{0}h_\bfnum{0}} + (\log n_\bfnum{0})^{3}\frac{\log d}{n_\bfnum{0}}\right),
\end{align}
under assumptions similar to, but weaker than, those in \cite{Lee2024}. If $h_\bfnum{0} \sim n_\bfnum{0}^{-1/5}$, the bounds in \eqref{normal-ll} and \eqref{new-ll} coincide when $d$ is fixed, whereas for diverging $d$, the bound in \eqref{new-ll} is substantially sharper.

Second, building on the proposed target-only estimator, we develop a novel two-stage transfer learning procedure and establish its theoretical properties. To develop the theory, we incorporate the notions of functional similarity and probabilistic structural similarity between the target and auxiliary populations, concepts that have also been adopted in the study of transfer learning for linear regression (\cite{li2022transfer, tian2023transfer}). 
However, we found that there is a substantial difference between the parametric and nonparametric approaches. To demonstrate this, suppose that for some informative set $\cA$ we have access to $|\cA|$ auxiliary samples. In the parametric setting, where for each $\ab\in\cA$ we assume the linear relationship $\Eb(Y_\bfnum{a} \mid \Xv_\bfnum{a}) = \Xv_\bfnum{a}\betav_\bfnum{a}$, one first estimates the minimizer of the weighted average loss functional
\begin{align*}
\sumaac \frac{n_\ab}{n_\cA}\Eb\left(\left(Y_\bfnum{a} - \Xv_\bfnum{a}\alphav\right)^2\right),
\end{align*}
where $n_\cA=\sumaac n_\ab$. The minimizer is well defined as an element of $\Rb^d$, the space in which all $\beta_\bfnum{a}$, $\ab\in\{\bfnum{0}\}\cup\cA$, reside.  
In this paper, however, we assume an additive regression model for each auxiliary population, given by
\begin{align*}
f_\bfnum{a}(\xv) := \Eb(Y_\bfnum{a}\mid \Xv_\bfnum{a} = \xv) = \Eb(Y_\bfnum{a}) + f_\bfsnum{a}{1}(x_1)+\cdots+f_\bfsnum{a}{d}(x_d).
\end{align*}
Under the transfer learning framework, the target of first-stage estimator is typically defined as the minimizer of the weighted average loss functional
\begin{align*}
\sumaac \frac{n_\ab}{n_\cA}\Eb\left(\left(Y_\bfnum{a} - \Eb(Y_\bfnum{a}) - g(X_\bfnum{a})\right)^2\right),
\end{align*}
where the minimization is taken in the $L^2$ space. However, there is no guarantee that the minimizer is bounded or differentiable, even when all $f_\bfnum{a}$ are smooth. Consequently, the minimizer does not necessarily belong to the function space in which all $f_\bfnum{a}$, $\ab \in \{\bfnum{0}\}\cup \cA$, reside.
This motivates a fundamentally different approach from parametric transfer learning. In Section~\ref{sec:tl}, we address this issue through notions of similarity. Our results are established under sub-Weibull error distributions.

Third, we derive minimax lower bounds for both the target-only sparse high-dimensional additive regression and its extension under the transfer learning framework. Although minimax lower bounds for sparse high-dimensional additive regression have been obtained in RKHS-based settings, our result is the first to establish such bounds within the H\"older class without recourse to basis expansion. Moreover, to the best of our knowledge, the minimax lower bound under transfer learning for sparse high-dimensional additive regression has not been studied previously and is established here for the first time. Consequently, we found that our estimators for both the target-only and the transfer learning framwork are minimax optimal under mild regularity conditions.

The organization of the paper is as follows. In Section~\ref{sec:ll}, we introduce a local linear estimator for sparse high-dimensional additive regression and establish its minimax optimality. Section~\ref{sec:tl} develops a novel two-stage transfer learning algorithm together with its population-level analysis. We also derive error bounds for each stage and show that the transfer-learned estimator attains the minimax lower bound when the probabilistic structures of the target and auxiliary populations are sufficiently close. Finally, Section~\ref{sec:numerical} presents simulation results and a real data application.

\smallskip\noindent
{\it Notations. }
In the statements of the assumptions and throughout this paper, we use the term \textit{absolute constant} to refer to a positive constant that is independent of the sample size. 
For a stochastic sequence $\{Z_n\}$ and a deterministic sequence $\{a_n > 0\}$, we write $Z_n \lesssim a_n$ if there exists an absolute constant $0 < C < \infty$ such that $|Z_n|/a_n \le C$ with probability tending to one. We write $Z_n \ll a_n$ if $Z_n = o_p(a_n)$.
For two deterministic sequences $\{a_n > 0\}$ and $\{b_n > 0\}$, we write $a_n \lesssim b_n$ if there exists an absolute constant $0 < C < \infty$ such that $a_n / b_n \le C$ for all $n$, and $a_n \ll b_n$ if $a_n / b_n \to 0$ as $n \to \infty$. We write $a_n \sim b_n$ if both $a_n \lesssim b_n$ and $b_n \lesssim a_n$ hold.
For scalars $a$ and $b$, we let $a \vee b$ denote $\max\{a, b\}$ and $a \wedge b$ denote $\min\{a, b\}$. We also write $(a)_+ := a \vee 0$. For a given $d \in \mathbb{N}$ and $\ell=1,2$, we let $[d]^\ell$ denote the collection of all ordered subsequences of length $\ell$ from $\{1,\ldots, d\}$.

Let $L^2([0,1]^d)$ denote the space of square-integrable functions on $[0,1]^d$.  
We define $L^{2,\rmtp}([0,1]^d)$ as the space of full function tuples $g^\rmtp = (g^0, g^1, \ldots, g^d)$ such that each $g^0$ and $g^j$ for $j\in[d]$ belongs to $L^2([0,1]^d)$.  
We refer to a function tuple $g_j^\rmtp$ for $j \in [d]$ as the \textit{$j$-th univariate function tuple} if it takes the form
\[
g_j^\rmtp = (g^0, 0_{j-1}^\top, g^j, 0_{d-j}^\top),
\]
where $g^0, g^j : [0,1]^d \to \mathbb{R}$ are such that $g^0(\xv) = g_j(x_j)$ and $g^j(\xv) = g_j^{(1)}(x_j)$ for some univariate functions $g_j$ and $g_j^{(1)}$.  
We denote the space of all such $j$-th univariate function tuples by $\sH_j^\rmtp$, and define their additive space as $\sH_\rmadd^\rmtp := \sH_1^\rmtp + \cdots + \sH_d^\rmtp$. Let $\sH_\rmprod^\rmtp$ denote the product space of the univariate spaces $\sH_j^\rmtp$.
For each $j \in [d]$, define the matrix
\[
U_j := \begin{pmatrix}
1 & 0_{j-1}^\top & 0 & 0_{d-j}^\top \\
0 & 0_{j-1}^\top & 1 & 0_{d-j}^\top
\end{pmatrix}.
\]
Corresponding to this structure, we define the \textit{$j$-th univariate function vector} $g_j^\rmv := (g_j, g_j^{(1)})$ for each $j \in [d]$, which has a one-to-one correspondence with the $j$-th univariate function tuple $g_j^\rmtp$ through the relation
\begin{align} \label{relationship-tuple-vector}
g_j^\rmtp = U_j^\top \cdot g_j^\rmv \quad \text{and} \quad g_j^\rmv = U_j \cdot g_j^\rmtp.
\end{align}
Finally, we denote the Cartesian product and the sum of the generic univariate function tuples $g_j^\rmtp$ by $\gv^\rmtp := (g_j^\rmtp : j \in [d]) \in \sH_\rmprod^\rmtp$ and $g_+^\rmtp := \sumj g_j^\rmtp \in \sH_\rmadd^\rmtp$, respectively.

\section{High-dimensional Locally Linear Additive Regression} \label{sec:ll}

Let $\Xv_\bfnum{0}=(X_\bfsnum{0}{1}, \ldots, X_\bfsnum{0}{d})$ be the covariate vector of the target population taking values in $[0,1]^d$ and $Y_\bfnum{0}$ be the associated response variable. We consider an additive model for the target population. Additive regression assumes that the mean function $f_\bfnum{0}:=\Eb(Y_\bfnum{0}|\Xv_\bfnum{0}=\cdot)$ admits 
\begin{align} \label{model}
f_\bfnum{0}(\xv) = \Eb(Y_\bfnum{0}) + f_\bfsnum{0}{1}(x_1)+\cdots f_\bfsnum{0}{d}(x_d)
\end{align}
for some square integrable univariate functions $f_\bfsnum{0}{j}$ satisfying the constraints
\begin{align} \label{constr-pop}
\int_0^1 f_\bfsnum{0}{j}(x_j)p_\bfsnum{0}{j}(x_j)\dxj = 0, \quad j\in[d], 
\end{align}
where $\xv=(x_1, \ldots, x_d)^\top$ and $p_\bfsnum{0}{j}$ denotes the marginal density of $X_\bfsnum{0}{j}$. 

Suppose that we observe $n_\bfnum{0}$ i.i.d. copies of $(\Xv_\bfnum{0}, Y_\bfnum{0})$. We denote each observed target sample by $(\Xv_\bfnum{0}^{i}, Y_\bfnum{0}^{i})$ for $1 \le i \le n_\bfnum{0}$, where $\Xv_\bfnum{0}^{i} = (X_\bfsnum{0}{1}^i, \ldots, X_\bfsnum{0}{d}^i)$. In our high-dimensional additive regression framework, we allow the number of covariates $d$ to diverge to infinity as the sample size $n_\bfnum{0}$ increases. We impose a sparsity condition, meaning that $f_\bfsnum{0}{j} \equiv 0$ for all but a relatively small number of indices $j$.

\subsection{Kernel scheme} \label{subsec:ll-kernel}
We introduce the normalized kernel scheme, which has played an important role in the smooth backfitting literature.
Let $K:\Rb\to\Rb_{\ge 0}$ be a baseline kernel supported on $[-1,1]$ and $K_h$ be defined by $K_h(u)=h^{-1}K(u/h)$. We take $K$ such that $K$ vanishes outside $[-1,1]$, is nonnegative, symmetric, bounded, Lipschitz continuous with Lipschitz contant $L_K$ and $\int K=1$. Then, we define $K_h(\cdot, \cdot):[0,1]^2\to\Rb$ by 
\begin{align*}
K_h(u,v):= \frac{K_h(u-v)}{\int_0^1 K_h(w-v)\,\rmd w}, \quad u,v\in[0,1]. 
\end{align*}
By definition, it follows that $\int_0^1 K_h(u,v)\,\rmd u=1$ for all $v\in[0,1]$. This is known as the \textit{normalization property}, which is considered desirable. For example, see \cite{mammen1999existence, Yu2008, jeon2020additive}, among others. We also note that $K_h(u,v)=K_h(u-v)$ for all $v\in[0,1]$ if $u\in[2h,1-2h]$ and 
\begin{align*}
K_h(u-v)\le K_h(u,v)\le 2K_h(u-v), \quad u,v\in[0,1]
\end{align*} 

\subsection{Projection operators} \label{subsec:ll-projection}
Define the inner product $\ip{\cdot, \cdot}_M$ associated with a $(d+1)\times (d+1)$ matrix function $M$ on $[0,1]^d$ by
\begin{align*}
\ip{g^\rmtp, \eta^\rmtp}_M := \int_{[0,1]^d} g^\rmtp(\xv)^\top M(\xv)\, \eta^\rmtp(\xv)\, \dxv, 
\quad g^\rmtp, \eta^\rmtp \in L^{2,\rmtp}([0,1]^d). 
\end{align*}
The corresponding norm $\norm{\cdot}_M$ is defined as the norm induced by this inner product. We introduce several matrix functions that serve the role of $M$ in the above definition. Let $p_\bfnum{0}$ denote the joint density function of $\Xv_\bfnum{0}$. Define a matrix function $M_\bfnum{0}(\uv):= \diag(1, \mu_2 1_d)\cdot p_\bfnum{0}(\uv)$, where $\mu_2 = \int_{-1}^1 v^2K(v)\, \rmd v$. The inner product structure induced by the matrix function $M_\bfnum{0}$ reflects the underlying probabilistic structure. Let $\Zv_\bfnum{0}^{i}(\uv):= (1, (X_\bfsnum{0}{1}^i-u_1)/h_\bfsnum{0}{1}, \ldots, (X_\bfsnum{0}{d}^i-u_d)/h_\bfsnum{0}{d})^\top$ be the vector-valued function on $[0,1]^d$, where $h_\bfsnum{0}{j}$ denotes the bandwidth for the $j$-th covariate from the target sample. We allow $h_\bfsnum{0}{j}$ to vary across $j$. Define the matrix function $\hM_\bfnum{0}$ by 
\begin{align*}
\hM_\bfnum{0}(\uv):= n_\bfnum{0}^{-1}\suminum{0} \Zv_\bfnum{0}^{i}(\uv)\Zv_\bfnum{0}^{i}(\uv)^\top \prod_{l=1}^d K_{h_\bfsnum{0}{l}}(u_l, \Xv_\bfsnum{0}{l}^i). 
\end{align*}
The inner product structure induced by the matrix function $\hM_\bfnum{0}$ approximates that of $M_\bfnum{0}$. Finally, let $\tM_\bfnum{0}$ denote the expectation of the matrix function $\hM_\bfnum{0}$, i.e., $\tM_\bfnum{0}(\uv) := \Eb(\hM_\bfnum{0}(\uv))$.

Since we are considering an additive model, our main focus is on the additive space $\sH_\rmadd^\rmtp$. For any $g^\rmtp, \eta^\rmtp \in \sH_\rmadd^\rmtp$ with respective additive components $g_j^\rmtp, \eta_j^\rmtp \in \sH_j^\rmtp$, the inner product $\ip{g^\rmtp, \eta^\rmtp}_M$ involves only the terms $\ip{g_j^\rmtp, \eta_j^\rmtp}_M$ for $j \in [d]$ and $\ip{g_j^\rmtp, \eta_k^\rmtp}_M$ for $(j,k) \in [d]^2$. Using the relationship in \eqref{relationship-tuple-vector}, we further obtain the following reduced expressions:
\begin{align*}
\ip{g_j^\rmtp, \eta_j^\rmtp}_M &= \int_0^1 g_j^\rmv(x_j)^\top \cdot \int_{[0,1]^{d-1}} U_j M(\xv) U_j^\top \dxv_{-j} \cdot \eta_j^\rmv(x_j)\dxj, \quad j \in [d], \\
\ip{g_j^\rmtp, \eta_k^\rmtp}_M &= \int_0^1 g_j^\rmv(x_j)^\top \cdot \int_{[0,1]^{d-2}} U_j M(\xv) U_k^\top \dxv_{-\{j,k\}}\cdot \eta_k^\rmv(x_k)\dxj\dxk, \quad (j,k) \in [d]^2,
\end{align*}
for $M = M_\bfnum{0}, \hM_\bfnum{0}, \tM_\bfnum{0}$. To simplify notation, we define the following expressions for each value of $M$. We write
\begin{align*}
M_\bfsnum{0}{jj}(u_j) &:= \int_{[0,1]^{d-1}} U_j M_\bfnum{0}(\uv) U_j^\top \, d\uv_{-j} = \diag(1, \mu_2) \cdot p_\bfsnum{0}{j}(u_j), \quad j \in [d], \\
M_\bfsnum{0}{jk}(u_j, u_k) &:= \int_{[0,1]^{d-2}} U_j M_\bfnum{0}(\uv) U_k^\top \, d\uv_{-\{j,k\}} = \diag(1, 0) \cdot p_\bfsnum{0}{jk}(u_j, u_k), \quad (j,k) \in [d]^2,
\end{align*}
where $p_\bfsnum{0}{jk}$ denotes the marginal bivariate density function of $(X_\bfsnum{0}{j}, X_\bfsnum{0}{k})$. Similarly, we denote the empirical versions by
\begin{align*}
\hM_\bfsnum{0}{jj}(u_j) &:= \int_{[0,1]^{d-1}} U_j \hM_\bfnum{0}(\uv) U_j^\top \, d\uv_{-j} \\
&= \frac{1}{n_\bfnum{0}} \sum_{i=1}^{n_\bfnum{0}} Z_\bfsnum{0}{j}^i(u_j) Z_\bfsnum{0}{j}^i(u_j)^\top K_{h_\bfsnum{0}{j}}(u_j, X_\bfsnum{0}{j}^i), \quad j \in [d], \\
\hM_\bfsnum{0}{jk}(u_j, u_k) &:= \int_{[0,1]^{d-2}} U_j \hM_\bfnum{0}(\uv) U_k^\top \, d\uv_{-\{j,k\}} \\
&= \frac{1}{n_\bfnum{0}} \sum_{i=1}^{n_\bfnum{0}} Z_\bfsnum{0}{j}^i(u_j) Z_\bfsnum{0}{k}^i(u_k)^\top K_{h_\bfsnum{0}{j}}(u_j, X_\bfsnum{0}{j}^i) K_{h_\bfsnum{0}{k}}(u_k, X_\bfsnum{0}{k}^i), \quad (j,k) \in [d]^2,
\end{align*}
where $Z_\bfsnum{0}{j}^i(u_j) := U_j \cdot \Zv_\bfnum{0}^{i}(\uv) = (1, (X_\bfsnum{0}{j}^i - u_j)/h_\bfsnum{0}{j})^\top$ for $j \in [d]$. Here, we have utilized the normalization property. Define $\tM_\bfsnum{0}{jj}:=\Eb(\hM_\bfsnum{0}{jj})$ and define $\tM_\bfsnum{0}{jk}:=\Eb(\hM_\bfsnum{0}{jk})$.

We conclude this section by describing a set of projection operators that act on the additive space $\sH_\rmadd^\rmtp$, each associated with a specific inner product. Let $\Rb^\rmtp$ denote the space of constant function tuples, i.e., $\Rb^\rmtp := \{ (c, 0_d^\top)^\top : c \in \Rb \}$.

\smallskip\noindent
{\it Projection operators onto univariate spaces $\sH_j^\rmtp$. } For each $j \in [d]$, define the projection operator $\Pi_\bfsnum{0}{j} : \sH_\rmadd^\rmtp \to \sH_j^\rmtp$ by
\[
\Pi_\bfsnum{0}{j}(g_+^\rmtp)(u_j) := g_j^\rmtp(u_j) + U_j^\top \cdot \left( \sumknotj \int_0^1 M_\bfsnum{0}{jj}(u_j)^{-1} M_\bfsnum{0}{jk}(u_j, u_k) g_k^\rmv(u_k) \duk \right),
\]
where $g_+^\rmtp = \sumj g_j^\rmtp \in \sH_\rmadd^\rmtp$. This operator satisfies the orthogonality condition
\[
\ip{g_+^\rmtp - \Pi_\bfsnum{0}{j}(g_+^\rmtp), \eta_j^\rmtp}_\Mnum{0} = 0, \quad \forall \, g_+^\rmtp \in \sH_\rmadd^\rmtp, \, \eta_j^\rmtp \in \sH_j^\rmtp,
\]
and hence legitimately defines a projection operator under the inner product $\ip{\cdot,\cdot}_\Mnum{0}$. In the same manner, we define $\hPi_\bfsnum{0}{j}$ and $\tPi_\bfsnum{0}{j}$ by replacing $M_\bfnum{0}$ with $\hM_\bfnum{0}$ and $\tM_\bfnum{0}$, respectively. These operators also satisfy orthogonality in the respective empirical and expected inner product spaces.

\smallskip\noindent 
{\it Projection operators onto constant space $\Rb^\rmtp$. } In addition to projections onto the univariate spaces, we define a projection operator onto the space $\Rb^\rmtp$. Let $p_\bfsnum{0}{j}^\rmv := (p_\bfsnum{0}{j}, 0)^\top$. Then, the projection operator $\Pi_\bfsnum{0}{0} : \sH_\rmadd^\rmtp \to \Rb^\rmtp$ is given by
\[
\Pi_\bfsnum{0}{0}(g_+^\rmtp) := U_j^\top \cdot \left( \sumj \int_0^1 g_j^\rmv(u_j)^\top p_\bfsnum{0}{j}^\rmv(u_j) \duj, \, 0_d^\top \right)^\top,
\]
where $g_+^\rmtp = \sumj g_j^\rmtp \in \sH_\rmadd^\rmtp$. This operator is also a projection with respect to the inner product structure. Define 
\[
\hp_\bfsnum{0}{j}^\rmv(u_j) := \frac{1}{n_\bfnum{0}} \sum_{i=1}^{n_\bfnum{0}} Z_\bfsnum{0}{j}^i(u_j) K_{h_\bfsnum{0}{j}}(u_j, X_\bfsnum{0}{j}^i),
\]
and put $\tp_\bfsnum{0}{j}^\rmv(u_j) := \Eb (\hp_\bfsnum{0}{j}^\rmv(u_j))$. 
Similarly, we define the operators $\hPi_\bfsnum{0}{0}$ and $\tPi_\bfsnum{0}{0}$ by replacing $p_\bfsnum{0}{j}^\rmv$ in $\Pi_\bfsnum{0}{0}$ with $\hp_\bfsnum{0}{j}^\rmv$ and $\tp_\bfsnum{0}{j}^\rmv$, respectively.

\subsection{Estimation} \label{subsec:ll-estimation}

In this section, we propose \textit{LL-fLasso-SBF estimator}, which is specifically tailored for the locally linear high-dimensional additive regression model.
In the case of unpenalized estimation, we typically minimize the empirical loss functional
\begin{align*}
\hL_\bfnum{0}(\gv^\rmtp) := \frac{1}{2n_\bfnum{0}} \int_{[0,1]^d}\suminum{0}\Big(Y_\bfnum{0}^{i} - \bar Y_\bfnum{0} - \sumj Z_\bfsnum{0}{j}^i(x_j)^\top g_j^\rmv(x_j)\Big)^2 \prod_{l=1}^d K_{h_\bfsnum{0}{l}}(x_l, X_\bfsnum{0}{l}^i)\dxl, 
\end{align*}
where $\bar Y_\bfnum{0}= \frac{1}{n_\bfnum{0}}\suminum{0} Y_\bfnum{0}^{i}$, over the function tuples $\gv^\rmtp=(g_j^\rmtp: j\in[d]) \in \sH_\rmprod^\rmtp$. This minimization procedure is applicable when $d$ is fixed, and it is shown in \cite{jeon2022locally} that the minimizer of $\hL_\bfnum{0}$ is well-defined with probability tending to one.
{However, in our setting, as in \cite{Lee2024}, direct minimization of $\hL_\bfnum{0}$ often becomes infeasible since we allow $d$ to exceed $n_\bfnum{0}$. To address this challenge, we adopt a penalized regression framework developed in \cite{Lee2024}, adapted to the locally linear estimation context.} Specifically, we introduce a penalty term into the loss functional $\hL_\bfnum{0}$, leading to the penalized loss functional $\hL_\bfnum{0}^\rmpen$ defined by
\begin{align*}
\hL_\bfnum{0}^\rmpen(\gv^\rmtp):= \hL_\bfnum{0}(\gv^\rmtp) + \lambda_\bfnum{0}\sumj \norm{g_j^\rmtp}_\hMnum{0},
\end{align*}
where $\lambda_\bfnum{0}$ is a penalty parameter. We minimize $\hL_\bfnum{0}^\rmpen$ over function tuples $\gv^\rmtp$ subject to the following constraints:
\begin{align} \label{constraint-estimation}
\int_0^1 g_j^\rmv(x_j)^\top \hp_\bfsnum{0}{j}^\rmv(x_j)\dxj = 0, \quad j \in [d].
\end{align}
These constraints ensure that the resulting estimator lies in the orthogonal complement of the constant function tuple space $\Rb^\rmtp$ with respect to the inner product $\ip{\cdot, \cdot}_\hMnum{0}$.

Let $\hatfv_\bfnum{0}^\rmtp = (\hatf_\bfsnum{0}{j}^\rmtp : j \in [d])$ denote the minimizer of $\hL_\bfnum{0}^\rmpen$. To compute $\hatfv_\bfnum{0}^\rmtp$, we employ an iterative algorithm in which each component function tuple $\hatf_\bfsnum{0}{j}^\rmtp$ is updated sequentially. A detailed analysis of this algorithm is provided in \cite{Lee2024} for the Nadaraya--Watson type estimation. Since the locally linear case requires only trivial modifications, we provide only a sketch of the algorithm here.
Suppose that at a given iteration, we have a current estimator $(\hatf_\bfsnum{0}{j}^{\rmtp, \rm OLD} : j \in [d])$ satisfying the constraints in \eqref{constraint-estimation}. The updated estimator $\hatf_\bfsnum{0}{j}^{\rmtp, \rm NEW}$ is then obtained by minimizing
\begin{align*}
\hL_\bfsnum{0}{j}^\rmpen(g_j^\rmtp) := \hL_\bfnum{0}\Big(\hatf_\bfsnum{0}{1}^{\rmtp, \rm OLD}, \ldots, \hatf_\bfsnum{0}{j-1}^{\rmtp, \rm OLD}, g_j^\rmtp, \hatf_\bfsnum{0}{j+1}^{\rmtp, \rm OLD}, \ldots, \hatf_\bfsnum{0}{d}^{\rmtp, \rm OLD}\Big) + \lambda_\bfnum{0} \norm{g_j^\rmtp}_\hMnum{0}
\end{align*}
over function tuples $g_j^\rmtp \in \sH_j^\rmtp$.
The minimization of $\hL_\bfsnum{0}{j}^\rmpen$ can be carried out via a two-stage procedure. Define the unpenalized functional $\hL_\bfsnum{0}{j}(g_j^\rmtp) := \hL_\bfsnum{0}{j}^\rmpen(g_j^\rmtp) - \lambda_\bfnum{0} \norm{g_j^\rmtp}_\hMnum{0}$, and let $\hatf_\bfsnum{0}{j}^{\rmtp, *}$ denote the minimizer of $\hL_\bfsnum{0}{j}$. This unpenalized minimization can be implemented using standard smooth backfitting techniques. Then, the updated estimator $\hatf_\bfsnum{0}{j}^{\rmtp, \rm NEW}$ is given by
\begin{align*}
\hatf_\bfsnum{0}{j}^{\rmtp, \rm NEW} = \left(1 - \frac{\lambda_\bfnum{0}}{\norm{\hatf_\bfsnum{0}{j}^{\rmtp, *}}_\hMnum{0}}\right)_+ \hatf_\bfsnum{0}{j}^{\rmtp, *}.
\end{align*}

\begin{remark}
As a desirable property established in \cite{Lee2024}, the local linear fLasso-SBF estimator $\hatfv_\bfnum{0}^\rmtp$ automatically satisfies the constraints in \eqref{constraint-estimation}. This follows from the fact that each $g_j^\rmtp \in \sH_j^\rmtp$ for $j \in [d]$, when satisfying the constraints in \eqref{constraint-estimation}, is orthogonal under the inner product $\ip{\cdot, \cdot}_\hMnum{0}$ to the constant function tuple space $\Rb^\rmtp$.
\end{remark}

\subsection{Theory} \label{subsec:ll-theory}

In this section, we present the $L^2$ error bound for the LL-fLasso-SBF estimator $\hatfv_\bfnum{0}^\rmtp$. Specifically, under conditions that are similar to or weaker than those in \cite{Lee2024}, we show that the estimator $\hatfv_\bfnum{0}^\rmtp$ achieves minimax optimality. Define the univariate function vector $f_\bfsnum{0}{j}^\rmv := (f_\bfsnum{0}{j}, h_\bfnum{0} f_\bfsnum{0}{j}')^\top$ and let $f_\bfsnum{0}{j}^\rmtp$ denote the corresponding univariate function tuple. We also set $\fv_\bfnum{0}^\rmtp := (f_\bfsnum{0}{j}^\rmtp : j \in [d])$.

\subsubsection{Assumptions} \label{subsubsec:ll-assumptions}

{To establish the theoretical results, we impose a set of assumptions, grouped according to their respective roles in the analysis. All assumptions are stated using notation without the subscript $\bz$, as they will be applied analogously for the auxiliary populations in the transfer learning framework discussed in Section~\ref{sec:tl} below. For instance, we denote the marginal univariate and bivariate density functions by $p_j$ and $p_{jk}$, respectively. This convention allows us to present the assumptions in a unified and generalizable form.} For generic $n$, $h$, $d$ and a given $\alpha > 0$, define
\begin{align*}
A(n,h,d;\alpha) &:= \frac{(\log d)^{\frac{1}{2}}}{nh^{\frac{1}{2}}} + \frac{\log d}{n} + \frac{(\log n)^{\frac{1}{\alpha}}(\log d)^{\frac{1}{2}+\frac{1}{\alpha\wedge 1}}}{n^{\frac{3}{2}}h^{\frac{1}{2}}} + \frac{(\log n)^{\frac{1}{2} + \frac{1}{\alpha}}(\log d)^{\frac{1}{\alpha\wedge 1}}}{n^{\frac{3}{2}}h^{\frac{1}{2}}}\\
&\quad  + \frac{(\log n)^{1+\frac{1}{\alpha\wedge 1} + \frac{2}{\alpha}}(\log d)^{\frac{1}{\alpha\wedge 1}}}{n^2h} + \frac{(\log n)^{\frac{1}{\alpha\wedge 1} + \frac{2}{\alpha}}(\log d)^{\frac{2}{\alpha\wedge 1}}}{n^2h}.
\end{align*}
Also, define
\begin{align*}
B(n,h,d) &:= \frac{(\log d)^{\frac{1}{2}}}{nh^{\frac{1}{2}}} + \frac{\log d}{n} + \frac{(\log d)^{\frac{3}{2}}}{n^{\frac{3}{2}}h^{\frac{1}{2}}} + \frac{(\log d)^2}{n^2 h}.
\end{align*}
We note that $B(n,h,d) \lesssim A(n,h,d;\alpha)$ for all $\alpha > 0$. The quantities $A(n,h,d;\alpha)$ and $B(n,h,d)$ are frequently introduced to simplify the expression of the error bounds.

\smallskip\noindent 
{\it (P) Assumptions on the probability density functions. }
\begin{itemize}
    \item[(P1)] \textit{Univariate densities.}  
    The marginal univariate density functions $p_j$ satisfy
    \begin{align*}
    C_{p,L}^\rmuniv \le \min_{j\in[d]}\inf_{x_j\in[0,1]}p_j(x_j) \le \max_{j\in[d]}\sup_{x_j\in[0,1]}p_j(x_j) \le C_{p,U}^\rmuniv
    \end{align*}
    for some absolute constants $0 <  C_{p,L}^\rmuniv\le C_{p,U}^\rmuniv < \infty$, and are continuous on $[0,1]$.

    \item[(P2)] \textit{Bivariate densities.}  
    The marginal bivariate density functions $p_{jk}$ satisfy
    \begin{align*}
    \max_{(j,k)\in[d]^2}\sup_{x_j,x_k\in[0,1]}p_{jk}(x_j,x_k) & \le C_{p,U}^{\rmbiv,1}, \\
    \max_{(j,k)\in[d]^2}\sup\left\{\frac{|p_{jk}(x_j,x_k) - p_{jk}(x_j', x_k')|}{|x_j - x_j'| + |x_k - x_k'|}: x_j \neq x_j' \text{ or } x_k \neq x_k'\right\} & \le C_{p,U}^{\rmbiv,2}
    \end{align*}
    for some absolute constants $0 < C_{p,U}^{\rmbiv,1}, C_{p,U}^{\rmbiv,2} < \infty$.
\end{itemize}

\smallskip\noindent
{\it (F) Assumptions on the component functions. }
\begin{itemize}
    \item[(F)] For each $j \in [d]$, the component function $f_j$ is twice differentiable on $[0,1]$. Moreover, for each $\ell = 0,1,2$, its $\ell$-th derivative satisfies
    \begin{align*}
    \max_{j\in[d]} \sup_{x_j \in [0,1]} |f_j^{(\ell)}(x_j)| \le C_{f,U}^\ell
    \end{align*}
    for some absolute constants $0 < C_{f,U}^\ell < \infty$.
\end{itemize}

\smallskip\noindent 
{\it (R-$\alpha$) Assumption on the residuals. }
\begin{itemize}
    \item[(R-$\alpha$)] Given a value of $\alpha>0$, the error term $\ve := Y - \Eb(Y | \Xv)$ satisfies
    \begin{align*}
    \Eb\left( \exp\left( |\ve|^\alpha / C_\ve^\alpha \right) | \Xv \right) \le 2 \quad \text{a.s.,}
    \end{align*}
    for some absolute constant $C_\ve > 0$. 
\end{itemize}

\smallskip\noindent 
{\it (B-$\alpha$) Assumptions on the bandwidths and the number of covariates. }
\begin{itemize}
\item[(B-$\alpha$)] The bandwidths $h_j$ are assumed to satisfy $C_{h,L} h_j \le h \le C_{h,U} h_j$ for all $j \in [d]$, for some absolute constants $0 < C_{h,L} \le C_{h,U} < \infty$. We refer to $h$ as the \textit{reference bandwidth}. In addition, we assume that $h = n^{-\zeta}$ for some $\zeta<\frac{1}{4}$, and that the number of covariates $d$ is sufficiently large so that $A(n,h,d;\alpha), B(n,h^2,d) = o(1)$ for a fixed $\alpha > 0$.
\end{itemize}

Most of our assumptions align closely with those in \cite{Lee2024}, but we highlight two key distinctions. First, our assumption (R-$\alpha$) allows the residuals $\ve := Y - \Eb(Y | \Xv)$ to follow a sub-Weibull distribution characterized by a tail parameter $\alpha$, thereby generalizing the sub-exponential framework adopted in \cite{Lee2024}. See \cite{arun2022moving} for the detailed discussion for sub-Weibull random variables and references therein. Specifically, (R-1) corresponds to the sub-exponential case ($\alpha = 1$), while (R-2), corresponding to the sub-Gaussian setting. Notably, when $\alpha < 1$, the sub-Weibull class captures a broad range of heavy-tailed distributions. 
Second, under the general condition (R-$\alpha$), the assumption (B-$\alpha$) characterizes the bandwidth size and the admissible growth rate of $d$ required for our analysis under various tail behaviors. In particular, under sub-exponential noise assumption when $\alpha\ge1$, our assumption (B-1) is satisfied if and only if $\log d = o(nh)$, which is obviously weaker than the condition $\log d = o(nh^2)$ required in \cite{Lee2024}. The latter condition arises from the conjunction of their assumption (A5) and the sparsity constraint imposed in their Theorem~2.

\subsubsection{Norm compatibility} \label{subsubsec:ll-norm-compat}

Analogous to the restricted eigenvalue condition commonly used in the theory of high-dimensional linear regression, our framework also requires a norm compatibility condition between the additive and product spaces, as previously introduced in \cite{Lee2024}. Define the active index set for the target population as
\begin{align*}
\cS_\bfnum{0}:= \{j \in [d] : \norm{f_\bfsnum{0}{j}^\rmtp}_\Mnum{0} \ne 0\}.
\end{align*}
For a given constant $0 < C < \infty$, define $\phi_\bfnum{0}(C)$ as the largest positive number, possibly depending on the sample size $n_\bfnum{0}$, such that
\begin{align} \label{eq:norm-compatibility}
\left\Vert \sumj g_j^\rmtp \right\Vert_\tMnum{0}^2 \ge \phi_\bfnum{0}(C) \left( \sumjs \norm{g_j^\rmtp}_\tMnum{0}^2 \right)
\end{align}
for all $\gv^\rmtp = (g_j^\rmtp : j \in [d]) \in \sH_\rmprod^\rmtp$ satisfying $\int_0^1 g_j^\rmv(x_j)^\top \tp_\bfsnum{0}{j}(x_j)\dxj=0$ for all $j \in [d]$ and
\begin{align*}
\sumjnots \norm{g_j^\rmtp}_\tMnum{0} \le C \left( \sumjs \norm{g_j^\rmtp}_\tMnum{0} \right).
\end{align*}
We note that $\phi_\bfnum{0}(C)$ is a non-decreasing function in $C$. {However, even if the value of $C$ is given, the existence of a strictly positive value of $\phi_\bfnum{0}(C)$ in \eqref{eq:norm-compatibility} is not guaranteed in general. This condition is closely related to the compatibility between the additive space $\sH_\rmadd^\rmtp$ and the product space $\sH_\rmprod^\rmtp$ and to ensure such compatibility it is common to impose structural assumptions such as exponential mixing among covariates. In particular, we establish Proposition~\ref{prop:suff-cond-norm-cmpt}, which serves as a locally linear analogue of Proposition~1 in \cite{Lee2024}, in the supplementary material.
}

\subsubsection{Error bound} \label{subsubsec:ll-error-bound}

In this section, we present the error bound for the proposed LL-fLasso-SBF estimator $\hatfv_\bfnum{0}^\rmtp$. Let $\hatf_\bfnum{0}^\rmtp :=  (\bar Y_\bfnum{0},0_d^\top)^\top + \sumj \hatf_\bfsnum{0}{j}^\rmtp$ and let $f_\bfnum{0}^\rmtp :=  (\Eb(Y_\bfnum{0}), 0_d^\top)^\top + \sumj f_\bfsnum{0}{j}^\rmtp$. Define the univariate function vector
\begin{align*}
\hatm_\bfsnum{0}{j}^\rmv(u_j) := \hM_\bfsnum{0}{jj}(u_j)^{-1} \cdot \frac{1}{n_\bfnum{0}} \suminum{0} Z_\bfsnum{0}{j}^i(u_j) K_{h_\bfsnum{0}{j}}(u_j, X_\bfsnum{0}{j}^i) Y_\bfnum{0}^{i},
\end{align*}
whose first component corresponds to the marginal local linear estimator of $\Eb(Y_\bfnum{0} | X_\bfsnum{0}{j} = x_j)$. The corresponding univariate function tuple is denoted by $\hatm_\bfsnum{0}{j}^\rmtp$. Define
\begin{align*}
\Delta_\bfsnum{0}{j}^\rmtp := \hatm_\bfsnum{0}{j}^\rmtp - \hPi_\bfsnum{0}{j}(f_\bfnum{0}^\rmtp).
\end{align*}
In the unpenalized framework, the identity
\begin{align*}
\Delta_\bfsnum{0}{j}^\rmtp = \hPi_\bfsnum{0}{j}(\hatf_\bfnum{0}^\rmtp - f_\bfnum{0}^\rmtp)
\end{align*}
holds, so the magnitude of $\Delta_\bfsnum{0}{j}^\rmtp$ determines the convergence rate of the SBF estimator. In the penalized setting, however, $\Delta_\bfsnum{0}{j}$ additionally reflects the influence of the penalty parameter $\lambda_\bfnum{0}$. Consequently, in our theoretical analysis, $\Delta_\bfsnum{0}{j}^\rmtp$ competes with the penalty term associated with $\lambda_\bfnum{0}$ and ultimately governs its asymptotic order. The following lemma provides an upper bound of $\Delta_\bfsnum{0}{j}^\rmtp$. 

\begin{lemma} \label{lem:ll-bound-Delta}
Assume that conditions (P1)--(P2) and (F) hold for the target population. Also, for some fixed $\alpha>0$, conditions (R-$\alpha$) and (B-$\alpha$) hold with the reference bandwidth of $h_\bfsnum{0}{j}$ denoted by $h_\bfnum{0}$. Then, it holds that
\begin{align*}
\max_{j\in[d]} \norm{\Delta_\bfsnum{0}{j}^\rmtp}_\hMnum{0}^2 \lesssim |\cS_\bfnum{0}|^2h_\bfnum{0}^4 + \frac{1}{n_\bfnum{0} h_\bfnum{0}} + A(n_\bfnum{0}, h_\bfnum{0}, d; \alpha).
\end{align*}
\end{lemma}

Let $\Delta_\bfnum{0} := \max_{j \in [d]} \norm{\Delta_\bfsnum{0}{j}^\rmtp}_\hMnum{0}$. The following theorem provides the $L^2$ error bound for the LL-fLasso-SBF estimator $\hatfv_\bfnum{0}^\rmtp$ under the empirical norm $\norm{\cdot}_\hMnum{0}$.

\begin{theorem} \label{thm:ll-bound-error-emp}
Assume the conditions in Lemma~\ref{lem:ll-bound-Delta}. Also, assume that the additive model is sufficiently sparse so that 
\begin{align*}
|\cS_\bfnum{0}| \lesssim h_\bfnum{0}^{-2} \left(\frac{1}{n_\bfnum{0}h_\bfnum{0}} + A(n_\bfnum{0}, h_\bfnum{0}, d;\alpha)\right)^{\frac{1}{2}}, \quad |\cS_\bfnum{0}|\ll \left(\frac{1}{n_\bfnum{0}h_\bfnum{0}^2}+B(n_\bfnum{0}, h_\bfnum{0}^2, d)\right)^{-\frac{1}{2}},
\end{align*}
and $|\cS_\bfnum{0}|\ll n_\bfnum{0}$. Suppose that the penalty parameter $\lambda_\bfnum{0}$ is chosen to satisfy
\begin{align*}
\mathfrak{C}_\bfnum{0} \Delta_\bfnum{0} \le \lambda_\bfnum{0} \lesssim \left( \frac{1}{n_\bfnum{0} h_\bfnum{0}} + A(n_\bfnum{0}, h_\bfnum{0}, d; \alpha) \right)^{\frac{1}{2}}
\end{align*}
for a sufficiently large absolute constant $\mathfrak{C}_\bfnum{0} > 1$. If there exists an absolute constant $C_\bfnum{0} > 2 \cdot \frac{\mathfrak{C}_\bfnum{0}+1}{\mathfrak{C}_\bfnum{0}-1}$ such that $\phi_\bfnum{0}(C_\bfnum{0}) > 0$ for all $n_\bfnum{0}$, then it holds that
\begin{align*}
\sum_{j=1}^d \norm{\hatf_\bfsnum{0}{j}^\rmtp - f_\bfsnum{0}{j}^\rmtp}_\hMnum{0} \lesssim |\cS_\bfnum{0}| \left( h_\bfnum{0}^4 + \frac{1}{n_\bfnum{0}h_\bfnum{0}} + A(n_\bfnum{0}, h_\bfnum{0}, d;\alpha)\right)^{\frac{1}{2}}.
\end{align*}
Furthermore, it follows that
\begin{align*}
\norm{\hatf_\bfnum{0}^\rmtp - f_\bfnum{0}^\rmtp}_\hMnum{0}^2 \lesssim |\cS_\bfnum{0}| \left( h_\bfnum{0}^4  + \frac{1}{n_\bfnum{0}h_\bfnum{0}} + A(n_\bfnum{0}, h_\bfnum{0}, d; \alpha) \right).
\end{align*}
\end{theorem}

Under assumption (P1), the norms $\norm{\cdot}_\hMnum{0}$ and $\norm{\cdot}_\Mnum{0}$ are equivalent on each univariate space $\sH_j^\rmtp$. Consequently, Theorem~\ref{thm:ll-bound-error-emp} implies that
\begin{align*}
\sumj \norm{\hatf_\bfsnum{0}{j}^\rmtp - f_\bfsnum{0}{j}^\rmtp}_\Mnum{0} \lesssim |\cS_\bfnum{0}| \left( h_\bfnum{0}^4 + \frac{1}{n_\bfnum{0}h_\bfnum{0}} + A(n_\bfnum{0}, h_\bfnum{0}, d;\alpha)\right)^{\frac{1}{2}}.
\end{align*}
However, this equivalence does not generally extend to the additive space $\sH_\rmadd^\rmtp$. The following corollary shows that, under a suitable mixing condition on the covariates, the two norms are also equivalent on $\sH_\rmadd^\rmtp$.

\begin{corollary} \label{cor:ll-bound-error-pop}
Assume the conditions in Theorem~\ref{thm:ll-bound-error-emp} hold. Further, suppose the mixing condition in Proposition~\ref{prop:suff-cond-norm-cmpt} is satisfied. Then, if $\sqrt{h_\bfnum{0}}|\cS_\bfnum{0}| \ll 1$,
it follows that
\begin{align*}
\norm{\hatf_\bfnum{0}^\rmtp - f_\bfnum{0}^\rmtp}_\Mnum{0}^2 \lesssim |\cS_\bfnum{0}| \left( h_\bfnum{0}^4  + \frac{1}{n_\bfnum{0}h_\bfnum{0}} + A(n_\bfnum{0}, h_\bfnum{0}, d; \alpha) \right).
\end{align*}
\end{corollary}

\begin{remark}
We observe that when $\alpha \ge 1$, under the additional conditions $h_\bfnum{0} \sim n^{-\frac{1}{5}}$ and $\log d = o(n_\bfnum{0}h_\bfnum{0})$, Corollary~\ref{cor:ll-bound-error-pop} yields
\begin{align*}
\norm{\hatf_\bfnum{0}^\rmtp - f_\bfnum{0}^\rmtp}_\Mnum{0}^2 \lesssim |\cS_\bfnum{0}| \left( n_\bfnum{0}^{-\frac{4}{5}} + (\log n_\bfnum{0})^3 \frac{\log d}{n_\bfnum{0}} \right).
\end{align*}
This result implies that our estimator achieves the minimax lower bound in Theorem~\ref{thm:ll-minimax} below when $\beta = 2$ up to logarithmic factors. 
\end{remark}

\subsection{Minimax lower bound} \label{subsec:ll-minimax}

This section is devoted to establishing a minimax lower bound for estimating regression function $f_\bfnum{0}$ in \eqref{model}, with respect to the $L^2$ norm weighted by the density $p_\bfnum{0}$, defined as
\begin{align*}
\norm{g}_{p_\bfnum{0}}^2 := \int_{[0,1]^d} g(\xv)^2 p_\bfnum{0}(\xv)\dxv, \quad g \in L^2([0,1]^d).
\end{align*}
Our theoretical framework is based on the general H\"older class, which offers a perspective distinct from prior minimax results that focus on reproducing kernel Hilbert spaces (RKHS), as seen in \cite{Raskutti2012, yuan2016minimax}. Unlike RKHS, the H\"older class does not admit a basis representation, and one of the key technical contributions of this section is to address the associated challenges that arise from this structural difference.

Recall that the H\"older class $\Sigma(\beta, L)$ on $[0,1]$ with smoothness parameter $\beta > 0$ and constant $L > 0$ is defined by
\begin{align*}
\Sigma(\beta, L) := \left\{ g : [0,1] \to \Rb : \sup_{x,x' \in [0,1]} \frac{|g^{(\lfloor \beta \rfloor)}(x) - g^{(\lfloor \beta \rfloor)}(x')|}{|x - x'|^{\beta - \lfloor \beta \rfloor}} \le L \right\},
\end{align*}
where $\lfloor \beta \rfloor$ denotes the greatest integer less than or equal to $\beta$.
For each $j \in [d]$, we define the function class $\sF_\bfsnum{0}{j}(\beta,L)$ as the collection of functions $g_j \in \Sigma(\beta, L)$ satisfying the constraint $\Eb[g_j(X_\bfsnum{0}{j})] = 0$. For a given index set $\cS \subset [d]$, we define the corresponding sparse additive function class as
\begin{align*}
\sF_\bfsnum{0}{\rmadd}(\cS,\beta,L) := \left\{ g = \sum_{j \in \cS} g_j : g_j \in \sF_\bfsnum{0}{j}(\beta,L) \text{ for all } j \in \cS \right\}.
\end{align*}
Then, for a fixed cardinality $s \le \lfloor d/8 \rfloor$, we define the $s$-sparse additive function class as
\begin{align*}
\sF_\bfsnum{0}{\rmadd}^s(\beta,L) := \bigcup_{|\cS| = s} \sF_\bfsnum{0}{\rmadd}(\cS,\beta,L).
\end{align*}

We derive a minimax lower bound under the assumption that the true regression function $f_\bfnum{0}$ lies in the $s$-sparse additive function class $\sF_\bfsnum{0}{\rmadd}^s$. To this end, we impose the following norm inequality:
\begin{align} \label{ll-norm-ineq-minimx}
C_{\sF,L}\sum_{j=1}^d \norm{g_j}_{p_\bfnum{0}}^2 \le \left\Vert \sumj g_j \right\Vert_{p_\bfnum{0}}^2 \le C_{\sF,U}\sumj \norm{g_j}_{p_\bfnum{0}}^2, \quad \sumj g_j \in \sF_\bfsnum{0}{\rmadd}^s,
\end{align}
for some absolute constants $0 < C_{\sF,L} \le C_{\sF,U} < \infty$. This type of inequality frequently arises in the minimax theory of high-dimensional additive regression (see, e.g., \cite{Raskutti2012, yuan2016minimax}).
In the RKHS framework, however, it is often difficult to directly verify such norm inequalities, as RKHS-based approaches typically focus on the structure of the function space itself, often disregarding the probabilistic structure of the covariates. For this reason, for example, \cite{yuan2016minimax} does not provide any explicit sufficient condition for \eqref{ll-norm-ineq-minimx}.
In contrast, we can establish, along the lines of the proof of Proposition~1 in \cite{Lee2024}, that the norm inequality in \eqref{ll-norm-ineq-minimx} holds under a mixing condition of the form
\begin{align*}
\int_{[0,1]^2}\bigl(p_\bfsnum{0}{jk}(x_j,x_k) - p_\bfsnum{0}{j}(x_j) p_\bfsnum{0}{k}(x_k)\bigr)^2 \, \dxj \dxk \le \varphi \cdot \psi^{|j-k|},
\end{align*}
for all $(j,k)\in[d]$ after some appropriate permutation of indices $1,2,\ldots,d$. 
In this case, the constants $C_{\sF,L}$ and $C_{\sF,U}$ in \eqref{ll-norm-ineq-minimx} can be specified as
\begin{align*}
C_{\sF,L} = \frac{C_{p,L}^\rmuniv - \sqrt{\psi}\,(C_{p,L}^\rmuniv + 2\sqrt{\varphi})}{(1-\sqrt{\psi})C_{p,L}^\rmuniv}, 
\quad
C_{\sF,U} = \frac{C_{p,L}^\rmuniv - \sqrt{\psi}\,(C_{p,L}^\rmuniv - 2\sqrt{\varphi})}{(1-\sqrt{\psi})C_{p,L}^\rmuniv}.
\end{align*}

Before presenting the main result, we introduce an assumption on the conditional distribution of $\ve_\bfnum{0}$ given $\Xv_\bfnum{0}$. This assumption is less restrictive than the fixed design Gaussian setting considered in previous studies and is widely adopted in the literature. For consistency with the presentation of other assumptions, we express the following condition using generic notation.

\smallskip\noindent
{\it Assumptions on the residuals (Minimax theory). }
\begin{itemize}
\item[(M)] The random variable $\ve$, conditional on $\Xv$, admits a density $p_{\ve | \Xv}$ with respect to the Lebesgue measure on $\Rb$. Moreover, there exist absolute constants $0 < c_\ve, v_\ve < \infty$ such that for all $|v| \le v_\ve$, it holds that
\begin{align*}
\int_\Rb p_{\ve| \Xv}(u) \cdot \log \frac{p_{\ve | \Xv}(u)}{p_{\ve | \Xv}(u + v)}\, \rmd u \le c_\ve v_\ve^2, \quad \text{almost surely}.
\end{align*}
\end{itemize}

\begin{theorem} \label{thm:ll-minimax}
Assume that conditions (P1) and (M) hold for the target population with $\ve_\bfnum{0}:= Y_\bfnum{0}-\Eb(Y_\bfnum{0}|\Xv_\bfnum{0})$, and that the norm inequality in \eqref{ll-norm-ineq-minimx} is satisfied. Assume
\begin{align} \label{ll-restriction-s}
s\left(n^{-\frac{\beta}{2\beta+1}} + \sqrt{\frac{\log (d/s)}{n}}\right) \ll 1.
\end{align}
Then there exists a constant $0<C_{\sF, \beta, L}<\infty$, depending only on $C_{\sF,L}, C_{\sF,U},\beta$ and $L$, such that
\begin{align*}
\liminf_{n_\bfnum{0}\to\infty}\inf_{\tf} \sup_{f_\bfnum{0} \in \sF_\bfsnum{0}{\rmadd}^s(\beta,L)} \Pb_f\left( \norm{\tf - f_\bfnum{0}}_{p_\bfnum{0}}^2 \ge C_{\sF, \beta, L}\cdot s \left( n^{-\frac{2\beta}{2\beta+1}} + \frac{\log (d/s)}{n} \right) \right) \ge \frac{1}{2},
\end{align*}
where $\Pb_f$ denotes the probability measure under which the true regression function for the target population is $f_\bfnum{0}$, and the infimum is taken over all measurable functions of the target samples.
\end{theorem}

\begin{remark}
The restrictive assumption \eqref{ll-restriction-s} on $s$ can be eliminated under the additional assumption that the error $\ve_\bfnum{0}$ follows a normal distribution as in \cite{Raskutti2012, yuan2016minimax}. Also, we observe that the minimax lower bound in Theorem~\ref{thm:ll-minimax} coincides with the result in \cite{Raskutti2012}. In the probabilistic argument, the two terms on the right-hand side can be interpreted as follows: the first term corresponds to the cost due to nonparametric estimation, while the second term reflects the combinatorial complexity of selecting $s$ active indices from $d$ covariates.
\end{remark}

\section{Transfer Learning Framework} \label{sec:tl}

In this section, we introduce a novel transfer learning algorithm for high-dimensional additive modeling, along with its theoretical guarantees, which differ fundamentally from those established for target-only estimation in Section~\ref{sec:ll}. Let $\cA = \{ \ab : \ab \ne \bz \}$ denote a collection of auxiliary indices, to be specified later. In the transfer learning framework, we additionally assume access to $n_\bfnum{a}$ i.i.d.\ copies of $(\Xv_\bfnum{a}, Y_\bfnum{a})$ for each $\ab \in \cA$, referred to as the \textit{$\ab$-th auxiliary samples}. Suppose that the additive regression function of each $\ab$-th auxiliary population is given by
\begin{align*}
f_\bfnum{a}(\xv) = \Eb(Y_\bfnum{a}) + f_\bfsnum{a}{1}(x_1) + \cdots + f_\bfsnum{a}{d}(x_d),
\end{align*}
for some square-integrable univariate functions $f_\bfsnum{a}{j}$ satisfying the constraints
\begin{align} \label{constraint-auxiliary}
\int_0^1 f_\bfsnum{a}{j}(x_j)\, p_\bfsnum{a}{j}(x_j)\, \mathrm{d}x_j = 0, \quad j \in [d],
\end{align}
where $\xv = (x_1, \ldots, x_d)$ and $p_\bfsnum{a}{j}$ denotes the marginal density of $X_\bfsnum{a}{j}$.

Within this framework, one can expect to enhance the efficiency of the estimator for both the mean regression function and the component functions of the target population by leveraging appropriate \textit{similarity} between the target and auxiliary populations. Analogous to parametric frameworks such as those studied in \cite{li2022transfer,tian2023transfer}, we consider two types of similarity measures: (i) functional similarity and (ii) probabilistic structural similarity. Unlike the parametric setting, these two notions of similarity are intricately connected in our nonparametric framework. This is because each component function $f_\bfsnum{0}{j}$ of the target population satisfies the constraint in \eqref{constr-pop} with respect to its marginal density functions $p_\bfsnum{0}{j}$, while each auxiliary component function $f_\bfsnum{a}{j}$ must satisfy the analogous constraint in \eqref{constraint-auxiliary} with respect to $p_\bfsnum{a}{j}$. Intuitively, the component functions $f_\bfsnum{0}{j}$ and $f_\bfsnum{a}{j}$ can be similar only if the marginal density functions $p_\bfsnum{0}{j}$ and $p_\bfsnum{a}{j}$ are sufficiently close.

In the following sections, unless otherwise specified, notations with the subscript $\ab$ are defined as their counterparts with subscript $\bz$, which correspond to the target population (or sample). Define
\begin{align*}
p_\cA := \sumaac w_\bfnum{a} p_\bfnum{a}, \quad \text{where} \quad n_\bfnum{\cA} := \sumaac n_\bfnum{a} \quad \text{and} \quad w_\bfnum{a} = \frac{n_\bfnum{a}}{n_\cA}.
\end{align*}
In this framework, we assume $n_\cA\gg n_\bfnum{0}$. Define $M_\bfnum{\cA} := \sumaac w_\bfnum{a} M_\bfnum{a}$. In a similar fashion, we define $\hp_\bfnum{\cA}$, $\tp_\bfnum{\cA}$, $\hM_\bfnum{\cA}$, and $\tM_\bfnum{\cA}$ as the weighted averages of $\hp_\bfnum{a}$, $\tp_\bfnum{a}$, $\hM_\bfnum{a}$, and $\tM_\bfnum{a}$ with weights $w_\bfnum{a}$, respectively, but evaluated using a unified bandwidths $h_\bfsnum{\cA}{j}$, which may differ from the bandwidths $h_\bfsnum{0}{j}$ used in the target-only estimation.
Furthermore, for each $j \in \{0\} \cup [d]$, define the projection operators $\Pi_\bfsnum{\cA}{j}$, $\hPi_\bfsnum{\cA}{j}$, and $\tPi_\bfsnum{\cA}{j}$ analogously to $\Pi_\bfsnum{0}{j}$, $\hPi_\bfsnum{0}{j}$, and $\tPi_\bfsnum{0}{j}$, with $M_\bfnum{0}$, $\hM_\bfnum{0}$, and $\tM_\bfnum{0}$ replaced by $M_\bfnum{\cA}$, $\hM_\bfnum{\cA}$, and $\tM_\bfnum{\cA}$, respectively.
We emphasize that the projection operators $\Pi_\bfsnum{\cA}{j}$, $\hPi_\bfsnum{\cA}{j}$, and $\tPi_\bfsnum{\cA}{j}$ are not equal to the weighted averages of their counterparts indexed by $\ab$.

\subsection{Estimation} \label{subsec:tl-estimation}

We propose a two-stage transfer learning algorithm to construct the \textit{transfer-learned LL-fLasso-SBF estimator} $\hatfv_\bfnum{0}^{\rmtp, \rmtl} = (\hatf_\bfsnum{0}{j}^{\rmtp, \rmtl} : j \in [d])$. For each $\ab \in \{\bz\} \cup \cA$, define the loss functional $\hL_\bfnum{a}$ by
\begin{align*}
\hL_\bfnum{a}(\gv^\rmtp) := \frac{1}{2n_\bfnum{a}} \int_{[0,1]^d} \sum_{i=1}^{n_\bfnum{a}} \left( Y_\bfnum{a}^{i} - \bar Y_\bfnum{a} - \sum_{j=1}^d Z_\bfsnum{a}{j}^i(x_j)^\top g_j^\rmv(x_j) \right)^2 \prod_{l=1}^d K_{h_\bfsnum{\cA}{l}}(x_l, X_\bfsnum{a}{l}^i)\, \mathrm{d}x_l.
\end{align*}

\smallskip\noindent
{\it Step 1: Fitting the aggregated estimator. }
In the first stage, we obtain the estimator $\hatfv_\bfnum{\cA}^{\rmtp} = (\hatf_\bfsnum{\cA}{j}^\rmtp : j \in [d])$ as the minimizer of the penalized squared loss functional
\begin{align*}
\hL_\bfnum{\cA}^{\rmpen, \rmtl1}(\gv^\rmtp) := \sum_{\ab \in \cA} w_\bfnum{a} \hL_\bfnum{a}(\gv^\rmtp) + \lambda_\cA^{\rmtl1} \sum_{j=1}^d \norm{g_j^\rmtp}_{\hMnum{\cA}},
\end{align*}
over $\gv^\rmtp \in \sH_\rmprod^\rmtp$, subject to the constraint
\begin{align*}
\int_0^1 g_j^\rmv(x_j)^\top \hp_\bfsnum{\cA}{j}(x_j)\, \mathrm{d}x_j = 0.
\end{align*}
Here, $\lambda_\cA^{\rmtl1}$ denotes the penalty parameter used in the first stage.

\smallskip\noindent
{\it Step 2: Centering the aggregated estimator. }
Before proceeding to the second stage, we adjust $\hatfv_\bfnum{\cA}^\rmtp$ so that it satisfies the empirical constraints associated with the target sample. Specifically, we define the centered estimator $\hatfv_\bfnum{\cA}^{\rmtp, \hc} := (\hatf_\bfsnum{\cA}{j}^{\rmtp, \hc} : j \in [d])$ by
\begin{align*}
\hatf_\bfsnum{\cA}{j}^{\rmtp, \hc} := \hatf_\bfsnum{\cA}{j}^\rmtp - \hPi_\bfsnum{0}{0}(\hatf_\bfsnum{\cA}{j}^\rmtp), \quad j \in [d].
\end{align*}

\smallskip\noindent
{\it Step 3: De-biasing the aggregated estimator. }
In the second stage, we obtain the minimizer of 
\begin{align*}
\hL_\bfnum{\cA}^{\rmpen, \rmtl2}(\gv^\rmtp) := \hL_\bfnum{0}(\hatfv_\bfnum{\cA}^{\rmtp, \hc} + \gv^\rmtp) + \lambda_\cA^{\rmtl2} \sum_{j=1}^d \norm{g_j^\rmtp}_{\hMnum{0}},
\end{align*}
subject to the constraint
\begin{align*}
\int_0^1 g_j^\rmv(x_j)^\top \hp_\bfsnum{0}{j}^\rmv(x_j)\, \mathrm{d}x_j = 0, \quad j \in [d].
\end{align*}
Note that the bandwidths $h_\bfsnum{0}{j}$ used in the definition of $\hL_\bfnum{0}$ in this stage coincide with those employed in the target-only estimation. Let the minimizer of $\hL_\bfnum{\cA}^{\rmpen, \rmtl2}$ be denoted by $\hdeltav_\bfnum{\cA}^\rmtp$.

\smallskip\noindent
{\it Step 4: Getting final estimator. }
The final transfer-learned LL-fLasso-SBF estimator $\hatfv_\bfnum{0}^{\rmtp, \rmtl}$ is then given by
\begin{align*}
\hatfv_\bfnum{0}^{\rmtp, \rmtl} := \hatfv_\bfnum{\cA}^\rmtp + \hdeltav_\bfnum{\cA}^\rmtp.
\end{align*}

\begin{remark}
In the first stage, one may also include the target sample when constructing the aggregated estimator. All theoretical results remain valid under this modification. For notational simplicity, however, we do not aggregate the target sample in the first stage. Incorporating the target sample at this stage may complicate the theoretical analysis, since the bandwidths used in the first and second stages need not coincide. In the empirical study, we aggregated the target sample with all auxiliary samples in the first stage.
\end{remark}

\subsection{Population-level analysis} \label{subsec:tl-popul-anal}

\subsubsection{True objective of $\hatfv_\bfnum{\cA}^\rmtp$} \label{subsubsec:tl-true-objective}

{To derive the $L^2$ error bound for the two-stage estimator, a common strategy is to bound the error at each stage separately and then combine the results. Within this approach, it is essential to identify the \textit{true objective} for the estimator $\hatfv_\bfnum{\cA}^\rmtp$ obtained in the first stage. In parametric transfer learning settings, it is natural to define the true objective of the aggregated estimator as the minimizer of a weighted average of loss functionals. This approach is straightforward because the estimands are finite-dimensional vectors. However, in the context of locally linear estimation within nonparametric analysis, the target includes not only the component functions themselves but also their first derivatives. Consequently, additional consideration is required in defining the true objective for the aggregated estimator.}

Define the population-level loss functionals $L_\bfnum{a}$ for each $\ab \in \cA$ by
\begin{align*}
L_\bfnum{a}(\gv^\rmtp) := \int_{[0,1]^d} \left( \sum_{j=1}^d g_j^\rmtp(x_j) - \sum_{j=1}^d f_\bfsnum{a}{j}^\rmtp(x_j) \right)^\top M_\bfnum{a}(\xv) \left( \sum_{j=1}^d g_j^\rmtp(x_j) - \sum_{j=1}^d f_\bfsnum{a}{j}^\rmtp(x_j) \right) \, \mathrm{d}\xv.
\end{align*}
We define the true objective $\fv_\bfnum{\cA}^\rmtp := (f_\bfsnum{\cA}{j}^\rmtp : j \in [d])$ of the estimator $\hatfv_\bfnum{\cA}^\rmtp$ as the minimizer of the aggregated loss functional
\begin{align*}
L_\bfnum{\cA}(\gv^\rmtp) := \sum_{\ab \in \cA} w_\bfnum{a} \, L_\bfnum{a}(\gv^\rmtp),
\end{align*}
subject to the constraints
\begin{align} \label{constraint-aggr}
\int_0^1 f_\bfsnum{\cA}{j}^\rmv(x_j)^\top p_\bfsnum{\cA}{j}^\rmv(x_j) \, \mathrm{d}x_j = 0, \quad j \in [d].
\end{align}
Notably, this approach does not require $f_\bfsnum{\cA}{j}$ to be differentiable.

\begin{remark}
{Suppose that $\widecheck\fv_\cA := (\widecheck f_\bfsnum{\cA}{j} : j \in [d])$ is the minimizer of the weighted average of the population-level loss functionals:
\begin{align*}
L_\cA(\gv) := \sum_{\ab \in \cA} w_\bfnum{a} \, \Eb\left[ \left( Y_\bfnum{a} - \Eb(Y_\bfnum{a}) - \sum_{j=1}^d g_j(X_\bfsnum{a}{j}) \right)^2 \right],
\end{align*}
subject to the normalization constraints $\int_0^1 \widecheck f_\bfsnum{\cA}{j}(x_j) p_\bfsnum{\cA}{j}(x_j)\, \mathrm{d}x_j = 0$ for all $j \in [d]$. Based on this minimizer, we define the corresponding function tuple $\widecheck \fv_\bfnum{\cA}^\rmtp := (\widecheck f_\bfsnum{\cA}{j}^\rmtp : j \in [d])$ by
\begin{align*}
\widecheck f_\bfsnum{\cA}{j}^\rmtp := \left( \widecheck f_\bfsnum{\cA}{j}, \, 0_{j-1}^\top, \, h_\bfsnum{\cA}{j} \widecheck f_\bfsnum{\cA}{j}', \, 0_{d-j}^\top \right)^\top.
\end{align*}
This construction requires that each component $\widecheck f_\bfsnum{\cA}{j}$ be differentiable. However, even if each $f_\bfsnum{a}{j}$ is smooth, the differentiability of $\widecheck f_\bfsnum{\cA}{j}$ cannot be ensured without further structural assumptions on the projection operators $\Pi_\bfsnum{a}{j}$. In fact, under general conditions, even continuity or boundedness of $\widecheck f_\bfsnum{\cA}{j}$ may not be guaranteed. For this reason, we propose an alternative formulation of the true objective for the estimator $\hatfv_\bfnum{\cA}^\rmtp$, which avoids direct reliance on differentiability.}
\end{remark}

\smallskip\noindent
{\it Existence and uniqueness of $\fv_\bfnum{\cA}^\rmtp$. }
It is important to verify that our proposed function tuple $\fv_\bfnum{\cA}^\rmtp$ is well-defined. To this end, we modify the definition of the projection operator $\Pi_\bfsnum{a}{j} : \sH_\rmadd^\rmtp \to \sH_j^\rmtp$ for $\ab\in\cA$ as
\begin{align*}
\Pi_\bfsnum{a}{j}(g_+^\rmtp)(x_j) &= g_j^\rmtp(x_j) \\
&\quad + U_j \cdot \left( \sum_{k =1,\ne j}^d \int_0^1 \left( M_\bfsnum{a}{jj}(x_j)^{-1} M_\bfsnum{a}{jk}(x_j,x_k) - \diag(1,0) \cdot p_\bfsnum{a}{k}(x_k) \right) g_k^\rmv(x_k) \, \mathrm{d}x_k \right),
\end{align*}
where $g_+^\rmtp = \sum_{j=1}^d g_j^\rmtp \in \sH_\rmadd^\rmtp$. We also refine the definition of $\Pi_\bfsnum{\cA}{j}$ analogously by replacing $M_\bfnum{a}$ and $p_\bfnum{a}$ with $M_\bfnum{\cA}$ and $p_\bfnum{\cA}$, respectively. These revised definitions of $\Pi_\bfsnum{a}{j}$ and $\Pi_\bfsnum{\cA}{j}$ coincide with the original ones when the univariate function tuples $g_j^\rmtp \in \sH_j^\rmtp$ satisfy the constraints in \eqref{constraint-auxiliary} and \eqref{constraint-aggr}, respectively.
For each $\ab \in \cA$, we define the operator $\Pi_\bfnum{a}^{\ominus,\rmtp} : \sH_\rmprod^\rmtp \to \sH_\rmprod^\rmtp$ by
\begin{align*}
\Pi_\bfnum{a}^{\ominus,\rmtp}(\gv^\rmtp) := \left( \Pi_\bfsnum{a}{1} \left( \sum_{k =2}^d g_k^\rmtp \right), \ldots, \Pi_\bfsnum{a}{d} \left( \sum_{k=1}^{d-1} g_k^\rmtp \right) \right)^\top, \quad \gv^\rmtp = (g_j^\rmtp : j \in [d]) \in \sH_\rmprod^\rmtp.
\end{align*}
Also, define the operator $\Mc_\bfnum{a}^\rmtp : \sH_\rmprod^\rmtp \to \sH_\rmprod^\rmtp$ by
\begin{align*}
\Mc_\bfnum{a}^\rmtp(\gv^\rmtp) := \left( U_1^\top \cdot M_\bfsnum{a}{11} g_1^\rmv, \ldots, U_d^\top \cdot M_\bfsnum{a}{dd} g_d^\rmv \right)^\top, \quad \gv^\rmtp = (g_j^\rmtp : j \in [d]) \in \sH_\rmprod^\rmtp.
\end{align*}
The operators $\Pi_\bfnum{\cA}^{\ominus,\rmtp}$ and $\Mc_\bfnum{\cA}^\rmtp$ are defined analogously by replacing $\Pi_\bfsnum{a}{j}$ and $M_\bfsnum{a}{jj}$ with $\Pi_\bfsnum{\cA}{j}$ and $M_\bfsnum{\cA}{jj}$, respectively.

Recall that $\fv_\cA^\rmtp = (f_\bfsnum{\cA}{j}^\rmtp : j \in [d])$ is a minimizer of $L_\cA$ subject to the constraints in \eqref{constraint-aggr}. Since $L_\cA$ is convex and continuous over $\sH_\rmprod^\rmtp$, Theorem~5.3.19 of \cite{han2009theoretical} ensures that the directional Fréchet derivative, denoted by $\partial L_\cA(\fv_\cA^\rmtp; \etav^\rmtp)$, vanishes for all directions $\etav^\rmtp \in \sH_\rmprod^\rmtp$. After some straightforward calculations, we obtain the following fundamental identity:
\begin{align} \label{fundamental-identity}
\Mc_\cA^\rmtp(\rmI^\rmtp + \Pi_\bfnum{\cA}^{\ominus,\rmtp})(\fv_\bfnum{\cA}^\rmtp) 
= \sum_{\ab \in \cA} w_\bfnum{a}  \Mc_\bfnum{a}^\rmtp(\rmI^\rmtp + \Pi_\bfnum{a}^{\ominus,\rmtp})(\fv_\bfnum{a}^\rmtp),
\end{align}
where $\rmI^\rmtp : \sH_\rmprod^\rmtp \to \sH_\rmprod^\rmtp$ denotes the identity operator, and $\fv_\bfnum{a}^\rmtp = (f_\bfsnum{a}{j}^\rmtp : j \in [d])$ with
\[
f_\bfsnum{a}{j}^\rmtp := \left( f_\bfsnum{a}{j}, \, 0_{j-1}^\top, \, h_\bfsnum{\cA}{j} f_\bfsnum{a}{j}', \, 0_{d-j}^\top \right)^\top.
\]
This identity holds under the assumption that $\fv_\cA^\rmtp$ satisfies the constraint in \eqref{constraint-aggr}, which is guaranteed since each $\fv_\bfnum{a}^\rmtp$ satisfies the corresponding constraint in \eqref{constraint-auxiliary}.

\begin{remark}
{It is legitimate to assume the existence of a minimizer $\fv_\cA^\rmtp$ satisfying the constraint in \eqref{constraint-aggr}. In particular, such an assumption is justified if $\sum_{j=1}^d \Pi_\bfsnum{\cA}{0}(f_j^\rmtp) = 0$ holds. To formalize this, define $\cv^\rmtp := (c_j^\rmtp : j \in [d])$ where $c_j^\rmtp := (\Pi_\bfsnum{\cA}{0}(f_\bfsnum{\cA}{j}^\rmtp), 0_d^\top)^\top$. If $\sum_{j=1}^d \Pi_\bfsnum{\cA}{0}(f_\bfsnum{\cA}{j}^\rmtp) \ne 0$, then the loss functional $L_\cA$ satisfies
\[
L_\cA(\fv_\cA^\rmtp) = L_\cA(\fv_\cA^\rmtp - \cv^\rmtp) + \left\Vert \sum_{j=1}^d \Pi_\bfsnum{\cA}{0}(f_j^\rmtp) \right\Vert_{\Mnum{\cA}}^2 > L_\cA(\fv_\cA^\rmtp - \cv^\rmtp),
\]
where the first equality follows from the orthogonality condition $f_\bfsnum{\cA}{j}^\rmtp - c_j^\rmtp \perp \Rb^\rmtp$ with respect to the inner product $\ip{\cdot, \cdot}_\Mnum{\cA}$, and the fact that $\Pi_\bfsnum{a}{0}(f_\bfsnum{a}{j}^\rmtp) = 0$ for all $\ab \in \cA$ and $j \in [d]$. Since the centered tuple $\fv_\cA^\rmtp - \cv^\rmtp$ satisfies the constraint in \eqref{constraint-aggr}, the original tuple $\fv_\cA^\rmtp$ cannot be optimal. Hence, without loss of generality, we may assume that any minimizer $\fv_\cA^\rmtp$ satisfies $\sum_{j=1}^d \Pi_\bfsnum{\cA}{0}(f_\bfsnum{\cA}{j}^\rmtp) = 0$.}
\end{remark}

From \eqref{fundamental-identity}, it can be easily verified that invertibility of the operator $\Mc_\bfnum{\cA}^\rmtp(\rmI^\rmtp+\Pi_\cA^{\ominus,\rmtp})$ determines the well-definedness of $\fv_\bfnum{\cA}^\rmtp$. The following result demonstrate the sufficient condition to make this operator invertible.

\begin{itemize}
\item[(T1)] For each $\ab\in\{\bz\}\cup \cA$ and for any non-zero function tuple $\gv:=(g_j:j\in[d])$, satisfying the constraints in \eqref{constraint-auxiliary}, it holds that 
\begin{align*}
\Eb\left(\left(\sumj g_j(X_\bfsnum{a}{j})\right)^2\right)>0.
\end{align*}
\end{itemize}

We note that assumption (T1) is minimal and closely related to the model identifiability condition in additive regression. Specifically, (T1) implies that if $\sumj g_j(X_\bfsnum{a}{j})=0$, then each $g_j$ must vanish. This property is also required in additive regression models with fixed $d$ and is established under sufficient conditions of (T1) in Lemma~S.8 of \cite{jeon2020additive}. In particular, \cite{Lee2024} implicitly assumed the invertibility of $\rmI^\rmtp + \Pi_\bfnum{0}^{\ominus,\rmtp}$, which follows from our assumption (T1) via Proposition~\ref{prop:invertibility-PicA}.

\begin{proposition} \label{prop:invertibility-PicA}

Assume that conditions (P1)--(P2) hold for all target and auxiliary populations, and that (T1) are also satisfied. Then, the operators $\rmI^\rmtp + \Pi_\bfnum{a}^{\ominus,\rmtp}$ for all $\ab \in \{\bz\} \cup \cA$, as well as $\rmI^\rmtp + \Pi_\cA^{\ominus,\rmtp}$, are invertible. 
\end{proposition}

\subsubsection{Analysis of the impact of simlarities} \label{subsubsec:tl-impact-similarity}

In this section, we investigate the population-level impact of probabilistic and functional similarities on our regression framework.

\smallskip\noindent 
{\it Probabilistic structural similarity. }
We present a theoretical result concerning the role of probabilistic similarity. To this end, we introduce an additional assumption. To formally represent this, we introduce additional assumptions. For $r=1,2$, we define the $L^r$-type operator norm for a linear operator $\Qc:\sH_\rmprod^\rmtp\to\sH_\rmprod^\rmtp$ by
\begin{align*}
\norm{\Qc}_\opnumnum{0}{r} : =\sup\left\{\left(\sumj \norm{[\Qc(\gv^\rmtp)]_j}_\Mnum{0}^r\right)^{\frac{1}{r}} : \gv^\rmtp = (g_j^\rmtp:j\in[d])\in\sH_\rmprod^\rmtp \text{ with } \sumj \norm{g_j^\rmtp}_\Mnum{0}^r\le 1\right\},
\end{align*}
where $[\Qc(\gv^\rmtp)]_j$ denotes the $j$-th component tuple of $\Qc(\gv^\rmtp)$. Let $\sk := \norm{(\rmI^\rmtp + \Pi_\bfnum{0}^{\ominus,\rmtp})^{-1}}_{\opnumnum{0}{1}}$, and define a measure of probabilistic structural similarity by
\begin{align*}
\eta_{p,1} := \max_{\ab \in \cA} \norm{\Mc_\bfnum{a}^\rmtp (\rmI^\rmtp + \Pi_\bfnum{a}^{\ominus,\rmtp}) - \Mc_\bfnum{0}^\rmtp (\rmI^\rmtp + \Pi_\bfnum{0}^{\ominus,\rmtp})}_{\opnumnum{0}{1}}.
\end{align*}

\begin{itemize}
\item[(T2)] There exists a constant $\gamma \in [0,1)$ such that $\sk \eta_{p,1} \le \gamma$.
\end{itemize}

Our assumption (T2) guarantees that the probabilistic discrepancy between the target and auxiliary populations remains sufficiently small. It is noteworthy that $\eta_{p,1}$ vanishes if $p_\bfsnum{a}{jk} \equiv p_\bfsnum{0}{jk}$ for all $\ab \in \cA$ and $(j,k) \in [d]^2$. Although this type of assumption is introduced here for the first time, it is conceptually similar to conditions commonly found in the parametric transfer learning literature (\cite{li2022transfer}), where the similarity between covariance matrices is controlled. Such covariance-based conditions effectively serve as analogues to projection operator conditions in their analyses.

\begin{proposition} \label{prop:bound-PicA}
Assume that conditions (P1)--(P2) hold for auxiliary populations, and that (T1)--(T2) are also satisfied. Then, it holds that
\begin{align*}
\norm{(\rmI^\rmtp + \Pi_\bfnum{\cA}^{\ominus,\rmtp})^{-1} (\Mc_\bfnum{\cA}^\rmtp)^{-1}}_{\opnumnum{0}{1}} \le \frac{\sk}{1 - \sk \eta_{p,1}} \le \frac{\sk}{1 - \gamma}.
\end{align*}
\end{proposition}

It is often straightforward to obtain a bound for the weighted average of operators when operator norm bounds for all individual operators are available. For example, observing that $\Mc_\bfnum{\cA}^\rmtp(\rmI^\rmtp + \Pi_\bfnum{\cA}^{\ominus,\rmtp}) = \sum_{\ab \in \cA} w_\bfnum{a} \Mc_\bfnum{a}^\rmtp(\rmI^\rmtp + \Pi_\bfnum{a}^{\ominus,\rmtp})$, we may deduce that
\begin{align*}
\norm{\Mc_\bfnum{\cA}^\rmtp(\rmI^\rmtp + \Pi_\bfnum{\cA}^{\ominus,\rmtp}) - \Mc_\bfnum{0}^\rmtp(\rmI^\rmtp + \Pi_\bfnum{0}^{\ominus,\rmtp})}_{\opnumnum{0}{1}} \le \eta_{p,1}.
\end{align*}
However, obtaining a norm bound for the inverse of the aggregated operator is generally more challenging. The lemma above demonstrates that if the probabilistic structural similarity is sufficiently small, then the operator norm of the inverse of $\Mc_\bfnum{\cA}^\rmtp(\rmI^\rmtp + \Pi_\bfnum{\cA}^{\ominus,\rmtp})$ can be effectively controlled.

\smallskip\noindent
{\it Homogeneous regime. } 
We often refer to the case in which $p_\bfsnum{a}{jk} \equiv p_\bfsnum{0}{jk}$ for all $\ab \in \cA$ and $(j,k) \in [d]^2$ as the \textit{homogeneous} regime. When we denote a probabilistic similarity measure by $\eta_{p,\ell}$ for $\ell \in \Nb$, it implicitly means that the measure $\eta_{p,\ell}$ shares the vanishing property with $\eta_{p,1}$ under the homogeneous regime. Homogeneity is not a particularly strong assumption since even under this condition it does not necessarily follow that $p_\bfnum{a} \equiv p_\bfnum{0}$ for all $\ab \in \cA$. The following remark provides a simple example that illustrates this point.

\begin{remark}
Consider the following discrete example with $d = 3$. Let the joint distribution be defined as $p_{123}(x_1, x_2, x_3) = p_1(x_1)p_2(x_2)p_3(x_3)$, where $\Pb(X_j = 1) = 0.5$ and $\Pb(X_j = 0) = 0.5$ for each $j = 1, 2, 3$. Define an alternative distribution $q_{123}(x_1, x_2, x_3)$ by
\begin{align*}
q_{123}(x_1, x_2, x_3) = 
\begin{cases}
0.25 &\quad \text{if } \mathrm{mod}_2(x_1 + x_2 + x_3) = 0, \\
0 &\quad \text{otherwise}.
\end{cases}
\end{align*}
It is straightforward to verify that $p_{jk} \equiv q_{jk}$ for all $(j,k) \in [3]^2$. However, the full joint distributions $p_{123}$ and $q_{123}$ are not equal.
\end{remark}

\smallskip\noindent 
{\it Functional similarity. }
Define the functional deviations $\deltav_\cA^\rmtp := \fv_\bfnum{0}^\rmtp - \fv_\bfnum{\cA}^\rmtp$ and $\deltav_\bfnum{a}^\rmtp := \fv_\bfnum{0}^\rmtp - \fv_\bfnum{a}^\rmtp$. Let $\delta_\bfsnum{\cA}{j}^\rmtp$ and $\delta_\bfsnum{a}{j}^\rmtp$ denote the $j$-th univariate function tuple of $\deltav_\cA^\rmtp$ and $\deltav_\bfnum{a}^\rmtp$, respectively. Define the corresponding univariate function vectors by $\delta_\bfsnum{\cA}{j}^\rmv := (\delta_\bfsnum{\cA}{j}, \delta_\bfsnum{\cA}{j}^{(1)})^\top$ and $\delta_\bfsnum{a}{j}^\rmv := (\delta_\bfsnum{a}{j}, \delta_\bfsnum{a}{j}^{(1)})^\top$. We note that $\delta_\bfsnum{a}{j}^{(1)} = h_\bfsnum{\cA}{j} \, \delta_\bfsnum{a}{j}'$, whereas $\delta_\bfsnum{\cA}{j}$ may not be differentiable. 

We refer to the set $\cA$ as an \textit{$\eta_\delta$-informative set} if it satisfies
\begin{align} \label{def-cA-eta-delta}
\max_{\ab \in \cA} \left( \sumj \norm{\delta_\bfsnum{a}{j}^\rmtp}_{\Mnum{0}} \right) \le \eta_\delta.
\end{align}
The condition in \eqref{def-cA-eta-delta} ensures that not only the magnitude of each $\delta_\bfsnum{a}{j}$ is controlled, but also that of its scaled derivative, $h_\bfsnum{\cA}{j} \delta_\bfsnum{a}{j}^{(1)}$. In particular, it implies that the influence of the derivative term is not significantly greater than that of the component function itself.
Subtracting $\Mc_\bfnum{\cA}^\rmtp(\rmI^\rmtp + \Pi_\cA^\rmtp)(\fv_\bfnum{0}^{\ominus,\rmtp})$ from both sides of \eqref{fundamental-identity} yields
\begin{align} \label{fundamental-identity-delta}
\Mc_\cA^\rmtp(\rmI^\rmtp + \Pi_\bfnum{\cA}^{\ominus,\rmtp})(\deltav_\bfnum{\cA}^\rmtp) 
= \sum_{\ab \in \cA} w_\bfnum{a} \, \Mc_\bfnum{a}^\rmtp(\rmI^\rmtp + \Pi_\bfnum{a}^{\ominus,\rmtp})(\deltav_\bfnum{a}^\rmtp).
\end{align}
Under the homogeneous regime, \eqref{fundamental-identity-delta} reduces to
\begin{align*}
\delta_\cA^\rmtp = \sum_{\ab \in \cA} w_\bfnum{a} \, \delta_\bfnum{a}^\rmtp,
\end{align*}
indicating that the aggregated deviation $\deltav_\cA^\rmtp$ is simply a weighted average of the individual deviations $\deltav_\bfnum{a}^\rmtp$. Moreover, in this case, the differentiability of each $\delta_\bfsnum{\cA}{j}$ is guaranteed, enabling more straightforward analysis. However, this simplification is generally hard to satisfy in practice.
The following lemma demonstrates that $\deltav_\bfnum{\cA}^\rmtp$ behaves approximately as a weighted average of $\deltav_\bfnum{a}^\rmtp$ when the probabilistic structures of the target and auxiliary populations are sufficiently similar.

\begin{proposition} \label{prop:functional-similarity}
Assume that conditions (P1)--(P2) hold for all target and auxiliary populations, and that (T1)--(T2) are also satisfied. For any $\eta_\delta$-informative set $\cA$, it holds that 
\begin{align*}
\sumj \left\Vert \delta_\bfsnum{\cA}{j}^\rmtp - \sumaac w_\bfnum{a} \delta_\bfsnum{a}{j}^\rmtp \right\Vert_\Mnum{0} \le \frac{2\sk\eta_{p,1}}{1-\sk\eta_{p,1}}\eta_\delta \le 2\gamma\eta_\delta. 
\end{align*}
\end{proposition}

\subsection{Empirical-level analysis}

In what follows, we assume that (T1)--(T2) hold. We are now ready to analyze the transfer-learned LL-fLasso-SBF estimator $\hatfv_\bfnum{0}^{\rmtp, \rmtl}$ introduced in Section~\ref{subsec:tl-estimation}. Throughout this analysis, we assume that $\cA$ is a $\eta_\delta$-informative set for some $\eta_\delta=o(1)$ and that $|\cA| < \infty$. However, we do not impose independence assumptions, neither between the target and auxiliary samples nor within the auxiliary samples themselves. Furthermore, we assume that all probabilistic similarity measures satisfy $\eta_{p,\ell}=o(1)$ for $\ell=1,2,3$, where $\eta_{p,2}$ and $\eta_{p,3}$ will be introduced later.

\subsubsection{Assumptions}

To accommodate the transfer learning framework, we introduce additional assumptions on the density functions, expressed in terms of generic notation for broader applicability. Notably, differentiability of the density functions is a standard assumption in Nadaraya–Watson estimation, whereas locally linear estimation does not require it. Although our setting follows the structure of locally linear estimation, these two assumptions are technically necessary because we do not assume differentiability of the component functions $f_\bfsnum{\cA}{j}$.

\smallskip\noindent 
{\it Modified versions of assumptions on density functions (Transfer learning). }
\begin{itemize}
\item[(P1$'$)]
The marginal univariate density functions $p_j$ satisfy (P1) and are continuously differentiable on $[0,1]$ with Lipschitz continuous and uniformly bounded derivatives:
\begin{align*}
\max_{j\in[d]}\sup_{x_j\in[0,1]}|\partial p_j(x_j)/\partial x_j| \le C_{p,1}^\rmuniv,
\end{align*}
for some absolute constant $0 < C_{p,1}^\rmuniv < \infty$.

\item[(P2$'$)]
The marginal bivariate density functions $p_{jk}$ satisfy (P2) and are continuously partially differentiable on $[0,1]^2$ with Lipschitz continuous and uniformly bounded partial derivatives:
\begin{align*}
\max_{(j,k)\in[d]^2}\sup_{x_j,x_k\in[0,1]}\max\left( \left|\frac{\partial p_{jk}(x_j,x_k)}{\partial x_j}\right|,\left|\frac{\partial p_{jk}(x_j,x_k)}{\partial x_k}\right| \right) \le C_{p,1}^\rmbiv,
\end{align*}
for some absolute constant $0<C_{p,1}^\rmbiv<\infty$. 
\end{itemize}

\subsubsection{Norm compatibility}

As we mentioned earlier we analyze the errors arising from the first and second stages separately.
The analogous notion of norm compatibility between $\sH_\rmadd^\rmtp$ and $\sH_\rmprod^\rmtp$ in terms of $\norm{\cdot}_\tMnum{\cA}$ is also needed for the analysis of the first-stage estimator $\hatfv_\bfnum{\cA}^\rmtp$. For a given constant $0 < C < \infty$ define
\begin{align*}
\phi_\cA(C)&:= \inf\Biggr\{\frac{\left\Vert\sumj g_j^\rmtp\right\Vert_\tMnum{\cA}^2}{\sumjs \norm{g_j^\rmtp}_\tMnum{\cA}^2}: \sumjnots \norm{g_j^\rmtp}_\tMnum{\cA} \le C\sumjs \norm{g_j^\rmtp}_\tMnum{\cA}, \, \sumjs \norm{g_j^\rmtp}_\tMnum{\cA} \ne 0, \\
&\hspace{7cm} \int_0^1 g_j^\rmv(x_j)^\top \tp_\bfsnum{\cA}{j}(x_j)\dxj = 0, \, j \in [d] \Biggr\}
\end{align*}
which is defined analogously to $\phi_\bfnum{0}$. We present a proposition that provides a sufficient condition ensuring the strict positivity of $\phi_\cA(C)$ for a given value of $C$. It is important to note that this result is not a direct consequence of Proposition~\ref{prop:suff-cond-norm-cmpt}. That is, although $p_\bfsnum{\cA}{j} = \sumaac w_\bfnum{a} p_\bfsnum{a}{j}$ and $p_\bfsnum{\cA}{jk} = \sumaac w_\bfnum{a} p_\bfsnum{a}{jk}$, it does not follow that
\begin{align*}
&\int_{[0,1]^2} \left(p_\bfsnum{\cA}{jk}(x_j,x_k) - p_\bfsnum{\cA}{j}(x_j) p_\bfsnum{\cA}{k}(x_k)\right)^2 \, \dxj \, \dxk \\
&\hspace{4cm} \le \sumaac w_\bfnum{a} \int_{[0,1]^2} \left(p_\bfsnum{a}{jk}(x_j,x_k) - p_\bfsnum{a}{j}(x_j) p_\bfsnum{a}{k}(x_k)\right)^2 \, \dxj \, \dxk
\end{align*}
in general. We define an additional measure of probabilistic similarity as
\begin{align*}
\eta_{p,2} := \max_{\ab \in \cA} \max_{j \in [d]} \chisqdiv{P_\bfsnum{a}{j}}{P_\bfsnum{0}{j}} = \max_{\ab\in\cA}\max_{j\in[d]}\int_0^1 \frac{(p_\bfsnum{a}{j}(x_j) - p_\bfsnum{0}{j}(x_j))^2}{p_\bfsnum{0}{j}(x_j)}\dxj,
\end{align*}
where $P_\bfsnum{a}{j}$ denotes the marginal distribution of $X_\bfsnum{a}{j}$ for $\ab \in \{\bz\} \cup \cA$, and $\chisqdiv{\cdot}{\cdot}$ denotes the chi-square divergence between probability measures.

\begin{proposition} \label{prop:tl-norm-compt}

Assume that conditions (P1)--(P2) hold for both of target and auxiliary populations. Furthermore, for some fixed $\alpha>0$, condition (B-$\alpha$) holds with the reference bandwidth of $h_\bfsnum{\cA}{j}$ denoted by $h_\bfnum{\cA}$. Suppose that $\eta_{p,2}=o(1)$ and there exist absolute constants $\varphi>0$ and $0<\psi<(\frac{(C_{p,L}^\rmuniv)^2}{(C_{p,L}^\rmuniv)^2+9\sqrt{\varphi}C_{p,U}^\rmuniv})^2$ such that after some permutation of the indices $1,2,\ldots, d$, we have 
\begin{align} \label{tl:mixing-cond}
\max_{\ab\in\cA}\int_{[0,1]^2}(p_\bfsnum{a}{jk}(x_j,x_k) - p_\bfsnum{a}{j}(x_j) p_\bfsnum{a}{k}(x_k))^2\dxj\dxk \le \varphi\cdot \psi^{|j-k|},
\end{align}
for all $(j,k)\in[d]^2$. Then, there exists an absolute constant $0<C_\cA<\infty$ such that if $\gv^\rmtp=(g_j^\rmtp:j\in[d])$ satisfies the constraints $\int_0^1 g_j^\rmv(x_j)^\top \tp_\bfsnum{\cA}{j}(x_j) \dxj=0$ for $j\in[d]$, and 
\begin{align*}
\sum_{j \not\in \cS_\bfnum{0}} \norm{g_j^\rmtp}_\tMnum{\cA} \le C \sum_{j \in \cS_\bfnum{0}} \norm{g_j^\rmtp}_\tMnum{\cA},
\end{align*}
then 
\begin{align*}
\left\Vert\sumj g_j^\rmtp\right\Vert_\tMnum{\cA}^2 &\ge \Biggr(\frac{(C_{p,L}^\rmuniv\mu_2)^2 - \sqrt{\psi}((C_{p,L}^\rmuniv\mu_2)^2 + 9\sqrt{\varphi}C_{p,U}^\rmuniv)}{(1-\sqrt{\psi})(C_{p,L}^\rmuniv\mu_2)^2}  \\
&\hspace{3cm} -C_\cA \left(1+ \sqrt{\eta_{p,2} + h_\bfnum{\cA}}\right)\sqrt{h_\bfnum{\cA}}|\cS_\bfnum{0}|\Biggr)\sumjd \norm{g_j^\rmtp}_\tMnum{\cA}^2. 
\end{align*}
\end{proposition}

\subsubsection{Error bound} \label{tl-error-bound}

We organize the theoretical results in three stages. First, we present the result for the first-stage estimation. Second, we provide the result for the second-stage estimation. Finally, we combine the two to establish the error bound for transfer-learned LL-fLasso-SBF estimator $\hatfv_\bfnum{0}^{\rmtp, \rmtl}$. 

\smallskip\noindent 
{\it Error bound for first-stage estimation. } 
To establish the error bound of the first-stage estimator $\hatfv_\bfnum{\cA}^\rmtp$ we adopt an approach similar to that used in the target-only estimation described in Section~\ref{subsubsec:ll-error-bound}. Although the structure is similar the technical proof is entirely distinct from that of the target-only case as we do not assume the differentiability of the component functions $f_\bfsnum{\cA}{j}$. Define the univariate function vector $\hatm_\bfsnum{\cA}{j}^\rmv$ by
\begin{align*}
\hatm_\bfsnum{\cA}{j}^\rmv(u_j) := \hM_\bfsnum{\cA}{jj}(u_j)^{-1}\left(\sumaac w_\bfnum{a}\cdot \frac{1}{n_\bfnum{a}}\suminum{a} Z_\bfsnum{a}{j}^i(u_j) K_{h_\bfsnum{\cA}{j}}(u_j, X_\bfsnum{a}{j}^i)(Y_\bfnum{a}^{i}-\bar Y_\bfnum{a})\right)
\end{align*}
and define the corresponding univariate function tuple $\hatm_\bfsnum{\cA}{j}^\rmtp$ in the usual way. Let $f_\bfnum{\cA}^\rmtp := \sumj f_\bfsnum{\cA}{j}^\rmtp$ and define $\Delta_\bfsnum{\cA}{j}^\rmtp := \hatm_\bfsnum{\cA}{j}^\rmtp - \hPi_\bfsnum{\cA}{j}(f_\bfnum{\cA}^\rmtp)$. Put $\hatf_\cA^\rmtp := \sumj \hatf_\bfsnum{\cA}{j}^\rmtp$. Since the equality $\Delta_\bfsnum{\cA}{j}^\rmtp = \hatm_\bfsnum{\cA}{j}^\rmtp - \hPi_\bfsnum{\cA}{j}(\hatf_\bfnum{\cA}^\rmtp - f_\bfnum{\cA}^\rmtp)$ holds in the unpenalized scheme it is also important to consider the magnitude of $\norm{\Delta_\bfsnum{\cA}{j}^\rmtp}_\hMnum{\cA}$ in order to control the size of the penalty parameter $\lambda_\cA^{\rmtl1}$. Recall that $\cS_\bfnum{a}$ denotes the active index set of the $\ab$-th auxiliary population. Let $|\cS_\cA| := \max_{\ab\in\cA}|\cS_\bfnum{a}|$. Define an additional probabilistic similarity measure by
\begin{align*}
\eta_{p,3} &:= \max_{\ab\in\cA}\Biggr(\max_{j\in[d]}\sup_{x_j\in[0,1]}\left|\frac{\partial_jp_\bfsnum{a}{j}(x_j)}{\partial x_j} - \frac{\partial_jp_\bfsnum{0}{j}(x_j)}{\partial x_j}\right| \\
&\hspace{4cm} \vee \max_{1\le j\ne k\le d}\Biggr(\sup_{x_j,x_k\in[0,1]} \left|\frac{\partial (p_\bfsnum{a}{jk}(x_j,x_k) - p_\bfsnum{0}{jk}(x_j,x_k))}{\partial x_j}\right|\Biggr)\Biggr).
\end{align*}
We note that the assumption that $\eta_{p,3}$ is small imposes a stronger condition than the corresponding assumptions on $\eta_{p,1}$ or $\eta_{p,2}$, as $\eta_{p,3}$ quantifies the deviation between the derivatives of the density functions. Our first result demonstrates the upper bound for $\Delta_\bfsnum{\cA}{j}$ in terms of similarity measures.

\begin{lemma} \label{lem:tl-nonasymp-bound-Delta}
Assume that conditions (P1$'$)--(P2$'$) and (F) hold for the auxiliary populations. Also suppose that for some fixed $\alpha > 0$ the conditions (R-$\alpha$) and (B-$\alpha$) hold with the sample size $n_\bfnum{\cA}$ and with the reference bandwidth of $h_\bfsnum{\cA}{j}$ denoted by $h_\bfnum{\cA}$. Then, if $|\cS_\bfnum{a}|\ll n_\bfnum{a}$ for all $\ab\in\cA$, it holds that
\begin{align*}
\max_{j\in[d]}\norm{\Delta_\bfsnum{\cA}{j}^\rmtp}_\hMnum{\cA} &\lesssim |\cS_\cA| h_\bfnum{\cA}^2 + \left(\frac{1}{n_\bfnum{\cA}h_\bfnum{\cA}} + A(n_\cA, h_\cA, d;\alpha)\right)^{\frac{1}{2}} \\
&\quad+\left(\left(\frac{1}{n_\cA h_\cA^2
} + B(n_\bfnum{\cA}, h_\bfnum{\cA}^2, d)\right)^{\frac{1}{2}} + h_\cA\eta_{p,3} + \eta_{p,1} + \eta_{p,2}\right) \eta_\delta+ \eta_{p,\delta}
\end{align*}
where
\begin{align*}
\eta_{p,\delta} := \frac{2\sk\eta_{p,1}}{1-\sk\eta_{p,1}}\eta_\delta.
\end{align*}
\end{lemma}

Put $\Delta_\cA := \max_{j \in [d]} \norm{\Delta_\bfsnum{\cA}{j}^\rmtp}_\hMnum{\cA}$. It is important to note that when $h_\cA \eta_{p,3} \sim \eta_{p,1} + \eta_{p,2}$, the term $\eta_{p,3}$ does not influence the magnitude of $\Delta_\cA$.
Given a subset $S\subset[d]$, define partial sums of $\eta_{\delta}$ and $\eta_{p,\delta}$ as measures of similarity by
\begin{align*}
\eta_{\delta,S} &:= \max_{\ab \in \cA} \left(\sum_{j\in S} \norm{\delta_\bfsnum{a}{j}^\rmtp}_\Mnum{0}\right), \\
\eta_{p,\delta, S} &:= \sum_{j\in S} \left\Vert \delta_\bfsnum{\cA}{j}^\rmtp - \sumaac w_\bfnum{a} \delta_\bfsnum{a}{j}^\rmtp \right\Vert_\Mnum{0}.
\end{align*}
It is immediate that for any subset $S\subset[d]$, one has $\eta_{\delta, S}\le \eta_\delta$ and $\eta_{p,\delta, S}\le \eta_{p,\delta}$. In the following theorem, we establish an error bound for the first-stage estimator $\hatfv_\bfnum{\cA}^\rmtp$.

\begin{theorem} \label{thm:tl-aggr}
Assume the conditions in Lemma~\ref{lem:tl-nonasymp-bound-Delta}. Also suppose that the additive models for the target and auxiliary populations are sufficiently sparse so that
\begin{align*}
|\cS_\bfnum{0}|\vee |\cS_\cA| \lesssim h_\bfnum{\cA}^{-2} \left(\frac{1}{n_\bfnum{\cA} h_\bfnum{\cA}} + A(n_\bfnum{\cA}, h_\bfnum{\cA}, d; \alpha)\right)^{\frac{1}{2}}, \quad
|\cS_\bfnum{0}| \ll \left(\frac{1}{n_\bfnum{\cA} h_\bfnum{\cA}^2} + B(n_\bfnum{\cA}, h_\bfnum{\cA}^2, d)\right)^{-\frac{1}{2}}.
\end{align*}
Suppose that the penalty parameter $\lambda_\cA^{\rmtl1}$ is chosen to satisfy
\begin{align*}
\mathfrak{C}_\cA\Delta_\bfnum{\cA} \le \lambda_\cA^{\rmtl1} &\lesssim \left(h_\cA^4 + \frac{1}{n_\bfnum{\cA}h_\bfnum{\cA}} + A(n_\cA, h_\cA, d;\alpha)\right)^{\frac{1}{2}} \\
&\quad+\left(\left(\frac{1}{n_\cA h_\cA^2
} + B(n_\bfnum{\cA}, h_\bfnum{\cA}^2, d)\right)^{\frac{1}{2}} + h_\cA\eta_{p,3} + \eta_{p,1} + \eta_{p,2}\right) \eta_\delta+ \eta_{p,\delta},
\end{align*}
for a sufficiently large constant $\mathfrak{C}_\cA>1$. If there exists an absolute constant $C_\cA>2\cdot \frac{\mathfrak{C}_\cA+2}{\mathfrak{C}_\cA-1}$ such that $\phi_\cA(C_\cA)$ is bounded away from zero, then it holds that 
\begin{align*}
\sumj \norm{\hatf_\bfsnum{\cA}{j}^\rmtp - f_\bfsnum{\cA}{j}^\rmtp}_\hMnum{\cA} \lesssim |\cS_\bfnum{0}|\lambda_\cA^{\rmtl1} + \eta_{p,\delta,\cS_\bfnum{0}} + \eta_{p,2}\eta_{\delta,\cS_\bfnum{0}} + \eta_{\delta, \cS_\bfnum{0}^c} + \eta_{p,\delta, \cS_\bfnum{0}^c}.
\end{align*}
Furthermore, it follows that 
\begin{align*}
\norm{\hatf_\bfnum{\cA}^\rmtp - f_\bfnum{\cA}^\rmtp}_\hMnum{\cA}^2 &\lesssim |\cS_\bfnum{0}|(\lambda_\cA^{\rmtl1})^2 + \lambda_\cA^{\rmtl1}(\eta_{p,\delta,\cS_\bfnum{0}} + \eta_{p,2}\eta_{\delta,\cS_\bfnum{0}}) \\
&\quad+ \left(\lambda_\cA^{\rmtl1}(\eta_{\delta,\cS_\bfnum{0}^c} + \eta_{p,\delta,\cS_\bfnum{0}^c})\wedge (\eta_{\delta,\cS_\bfnum{0}^c} + \eta_{p,\delta,\cS_\bfnum{0}^c})^2\right).
\end{align*}
\end{theorem}

\smallskip\noindent 
{\it Error bound for second-stage estimation. }
Next we investigate the error bound for $\hdeltav_\bfnum{\cA}^\rmtp$ relative to $\deltav_\bfnum{\cA}^\rmtp$. Notably $\hdeltav_\bfnum{\cA}^\rmtp$ satisfies the empirical constraints associated with the target sample while $\deltav_\bfnum{\cA}^\rmtp$ does not satisfy the corresponding constraints of the target population. This distinction contrasts with much of the existing literature which typically bounds the estimation error relative to \textit{fake} target. By \textit{fake}, we mean that the true target of $\hdeltav_\bfnum{\cA}^\rmtp$ is given by $\deltav_\bfnum{\cA}^{\rmtp, \rmc} := (\delta_\bfsnum{\cA}{j}^{\rmtp, \rmc} : j \in [d])$ with
\begin{align*}
\delta_\bfsnum{\cA}{j}^{\rmtp,\rmc} := \delta_\bfsnum{\cA}{j}^\rmtp - \Pi_\bfsnum{0}{0}(\delta_\bfsnum{\cA}{j}^\rmtp).
\end{align*}
To address this discrepancy, we explicitly utilize the probabilistic structural similarity between populations. Let $\hdelta_\bfnum{\cA}^\rmtp := (\bar Y_\bfnum{0}, 0_d^\top)^\top + \sumj \hdelta_\bfsnum{\cA}{j}^\rmtp$ and $\delta_\bfnum{\cA}^\rmtp := (\Eb(Y_\bfnum{0}), 0_d^\top)^\top + \sumj \delta_\bfsnum{\cA}{j}^\rmtp$. Recall also the definition of $\Delta_\bfnum{0}$ given in Section~\ref{subsubsec:ll-error-bound}.

\begin{theorem} \label{thm:tl-debias}
Assume that conditions (P1$'$)--(P2$'$) and (F) hold for the target populations. Also suppose that for some fixed $\alpha > 0$ the conditions (R-$\alpha$) and (B-$\alpha$) hold with the sample size $n_\bfnum{0}$ and with the reference bandwidth of $h_\bfsnum{0}{j}$ denoted by $h_\bfnum{0}$. Also, assume that the additive model for the target population is sufficiently sparse so that 
\begin{align*}
|\cS_\bfnum{0}|(\lambda_\cA^{\rmtl2} + \sqrt{h_\bfnum{0}}) \lesssim 1 ,
\end{align*}
with the penalty parameter $\lambda_\cA^{\rmtl2}$ chosen to satisfy 
\begin{align*}
\mathfrak{C}_\bfnum{0}' \Delta_\bfnum{0} \le \lambda_\cA^{\rmtl2} \lesssim \left(h_\bfnum{0}^4 + \frac{1}{n_\bfnum{0}h_\bfnum{0}} + A(n_\bfnum{0}, h_\bfnum{0}, d;\alpha)\right)^{\frac{1}{2}}
\end{align*}
for a sufficiently large absolute constant $\mathfrak{C}_\bfnum{0}'>1$. Then, if 
\begin{align} \label{add-condition-thm5}
h_\bfnum{0}\eta_\delta^2 \wedge (|\cS_\bfnum{0}|\vee|\cS_\bfnum{\cA}|)^2h_\bfnum{0}^4\lesssim \lambda_\cA^{\rmtl2}\eta_\delta,
\end{align}
it holds that 
\begin{align*}
\sumj \norm{\hdelta_\bfsnum{\cA}{j}^\rmtp - \delta_\bfsnum{\cA}{j}^\rmtp}_\hMnum{0} \lesssim \frac{1}{\lambda_\cA^{\rmtl2}}\norm{\hf_\bfnum{\cA}^\rmtp - f_\bfnum{\cA}^\rmtp - \hPi_\bfsnum{0}{0}(\hf_\bfnum{\cA}^\rmtp - f_\bfnum{\cA}^\rmtp)}_\hMnum{0}^2 + \eta_\delta + \eta_{p,\delta}^* 
\end{align*}
where 
\begin{align*}
\eta_{p,\delta}^* := \eta_{p,\delta} + \frac{1}{\lambda_\cA^{\rmtl2}}\cdot (\eta_{p,\delta} + |\cS_\bfnum{0}|\eta_{p,2})\cdot (|\cS_\bfnum{0}|\lambda_\cA^{\rmtl2}\vee (\eta_{p,\delta} + |\cS_\bfnum{0}|\eta_{p,2})).
\end{align*}
Furthermore, it follows that 
\begin{align*}
\norm{\hdelta_\bfnum{\cA}^\rmtp - \delta_\bfnum{\cA}^\rmtp}_\hMnum{0}^2 &\lesssim \norm{\hatf_\bfnum{\cA}^\rmtp - f_\bfnum{\cA}^\rmtp- \hPi_\bfsnum{0}{0}(\hf_\bfnum{\cA}^\rmtp - f_\bfnum{\cA}^\rmtp)}_\hMnum{0}^2 +  \lambda_\cA^{\rmtl2}(\eta_\delta+\eta_{p,\delta}^*) \wedge (\eta_\delta+\eta_{p,\delta}^*)^2. 
\end{align*}
\end{theorem}

It is noteworthy that the assumption in \eqref{add-condition-thm5} is not restrictive. This condition is satisfied if and only if
\begin{align*}
\eta_\delta \lesssim \frac{\lambda_\cA^{\rmtl2}}{h_\bfnum{0}} \quad \text{or} \quad \eta_\delta \gtrsim \frac{(|\cS_\bfnum{0}|\vee|\cS_\bfnum{\cA}|)^2 h_\bfnum{0}^4}{\lambda_\cA^{\rmtl2}}.
\end{align*}
A sufficient condition under which the requirement is automatically fulfilled is $\eta_\delta \lesssim h_\bfnum{0}$. In this case, we have
\begin{align*}
h_\bfnum{0} \eta_\delta^2 \lesssim h_\bfnum{0}^2 \eta_\delta \le \lambda_\cA^{\rmtl2} \eta_\delta.
\end{align*}
In particular, the assumption becomes redundant when $\lambda_\cA^{\rmtl2} \gtrsim (|\cS_\bfnum{0}|\vee|\cS_\bfnum{\cA}|) h_\bfnum{0}^{5/2}$.

\smallskip\noindent
{\it Error bound for total estimation. }
From the two-stage estimation procedure, we construct the transfer-learned LL-fLasso-SBF estimator as $\hatfv_\bfnum{0}^{\rmtp,\rmtl} := \hatfv_\bfnum{\cA}^{\rmtp} + \hdeltav_\bfnum{\cA}^\rmtp$. Let $\hatf_\bfnum{0}^{\rmtp, \rmtl} := (\bar Y_\bfnum{0}, 0_d^\top )^\top + \sumj \hatf_\bfsnum{0}{j}^{\rmtp, \rmtl}$, and recall that $f_\bfnum{0}^\rmtp = (\Eb(Y), 0_d^\top )^\top + \sumj f_\bfsnum{0}{j}^\rmtp$. The following corollary establishes an error bound for the transfer-learned LL-fLasso-SBF estimator $\hatfv_\bfnum{0}^{\rmtp,\rmtl}$ measured in the target population norm $\norm{\cdot}_\Mnum{0}$. For theoretical simplicity, we focus on the homogeneous regime, under which all measures $\eta_{p,\ell}$ for $\ell = 1, 2, 3$, as well as $\eta_{p,\delta}$ and $\eta_{p,\delta}^*$ vanish.

\begin{corollary} \label{cor:tl-total}
Assume the conditions in Theorems~\ref{thm:tl-aggr} and~\ref{thm:tl-debias}, and suppose that the mixing conditions in Propositions~\ref{prop:suff-cond-norm-cmpt} and~\ref{prop:tl-norm-compt} are satisfied. In addition, assume the following:
\begin{itemize}
\item $\lambda_\cA^{\rmtl1}\lesssim \lambda_\cA^{\rmtl2}$;
\item $|\cS_\bfnum{0}|\ll \left(h_\bfnum{\cA}+h_\bfnum{0}\right)^{-\frac{1}{2}}$;
\item $\left(h_\bfnum{\cA} \vee \left(\frac{1}{n_\cA h_\cA^2} + B(n_\cA, h_\cA^2, d)\right)^{\frac{1}{2}}\right)\eta_\delta^2 \lesssim \lambda_\cA^{\rmtl1}\eta_\delta$;
\item $\left(h_\bfnum{0} \vee \left(\frac{1}{n_\bfnum{0} h_\bfnum{0}^2} + B(n_\bfnum{0}, h_\bfnum{0}^2, d)\right)^{\frac{1}{2}}\right)\eta_\delta^2 \lesssim \lambda_\cA^{\rmtl2}\eta_\delta$.
\end{itemize}
Then, under the homogeneous regime, it holds that
\begin{align*}
\norm{\hatf_\bfnum{0}^{\rmtp, \rmtl} - f_\bfnum{0}^{\rmtp}}_\Mnum{0}^2 &\lesssim |\cS_\bfnum{0}|\left(h_\bfnum{\cA}^4 + \frac{1}{n_\bfnum{\cA}h_\bfnum{\cA}} + A(n_\bfnum{\cA}, h_\bfnum{\cA}, d;\alpha)\right) \\
&\hspace{2cm} + \left(h_\bfnum{0}^4 + \frac{1}{n_\bfnum{0}h_\bfnum{0}} + A(n_\bfnum{0}, h_\bfnum{0}, d;\alpha)\right)^{\frac{1}{2}}\eta_\delta \wedge \eta_\delta^2. 
\end{align*}
\end{corollary}

\begin{remark}
The additional assumption on the functional similarity measure $\eta_\delta$ in Corollary~\ref{cor:tl-total} is not particularly restrictive. Additional conditions on functional similarity have been imposed in \cite{li2022transfer} and \cite{tian2023transfer} to ensure the validity of their theoretical results. 
\end{remark}

Under mild regularity conditions, the error bound established in Corollary~\ref{cor:tl-total} matches the minimax lower bound. To see this, consider the case where the error distribution is sub-exponential ($\alpha = 1$) and the bandwidths satisfy $h_\bfnum{\cA} \sim n_\bfnum{\cA}^{-1/5}$ and $h_\bfnum{0} \sim n_\bfnum{0}^{-1/5}$. In this setting, the bound reduces to
\begin{align} \label{bound-final}
\begin{aligned}
\norm{\hatf_\bfnum{0}^{\rmtp, \rmtl} - f_\bfnum{0}^{\rmtp}}_\Mnum{0}^2 &\lesssim |\cS_\bfnum{0}|\left(n_\cA^{-\frac{4}{5}} + (\log n_\cA)^3\frac{\log d}{n_\cA}\right) + \left(n_\bfnum{0}^{-\frac{4}{5}} + (\log n_\bfnum{0})^3\frac{\log d}{n_\bfnum{0}}\right)^{\frac{1}{2}}\eta_\delta \wedge \eta_\delta^2.
\end{aligned}
\end{align}
Consequently, if
\begin{align} \label{add-cond}
\eta_\delta \lesssim |\cS_\bfnum{0}| \left(n_\bfnum{0}^{-\frac{4}{5}} + (\log n_\bfnum{0})^3\frac{\log d}{n_\bfnum{0}}\right)^{\frac{1}{2}},
\end{align}
then the bound in \eqref{bound-final} matches the minimax lower bound in Theorem~\ref{thm:tl-minimax} below when $\beta = 2$, up to a logarithmic factor.

\subsection{Minimax lower bound} \label{subsec:minimax-lower-bound}

In this section, we establish the minimax lower bound under the transfer learning framework. Recall the sparse additive function class $\sF_\bfsnum{0}{\rmadd}^s(\beta,L)$ introduced in Section~\ref{subsec:ll-minimax}. For each $\ab \in \cA$, we additionally define the function class $\sF_\bfsnum{a}{\rmadd}(\beta,L) := \sF_\bfsnum{a}{1}(\beta,L) + \cdots + \sF_\bfsnum{a}{d}(\beta,L)$, where each $\sF_\bfsnum{a}{j}(\beta,L)$ is defined analogously to $\sF_\bfsnum{0}{j}(\beta,L)$ but with the norm $\norm{\cdot}_{p_\bfnum{0}}$ replaced by $\norm{\cdot}_{p_\bfnum{a}}$. Let $\bigotimes_{\ab \in \cA} \sF_\bfsnum{a}{\rmadd}(\beta,L)$ denote the product space of these auxiliary function classes. Given a sparsity parameter $s$, define the following class of functions:
\begin{align*}
\sF_\bfsnum{0}{\rmadd}^{s, \rmtl}(\beta,L) &:= \Biggr\{ (g_\bfnum{0}, (g_\bfnum{a} : \ab \in \cA)) \in \sF_\bfsnum{0}{\rmadd}^s(\beta,L) \times \bigotimes_{\ab \in \cA} \sF_\bfsnum{a}{\rmadd}(\beta,L) : \\
&\hspace{7cm} \max_{\ab \in \cA} \left( \sumj \norm{g_\bfsnum{a}{j} - g_\bfsnum{0}{j}}_{p_\bfnum{0}} \right) \le \eta_\delta \Biggr\}.
\end{align*}
Clearly, $\sF_\bfsnum{0}{\rmadd}^{s, \rmtl}$ characterizes the class of functions relevant to the transfer learning framework. For generic numbers $n,s,d$, simply write 
\begin{align*}
C(n,s,d;\beta) = n^{-\frac{2\beta}{2\beta+1}} + \frac{\log (d/s)}{n}.
\end{align*}

\begin{theorem} \label{thm:tl-minimax}
Assume the conditions of Theorem~\ref{thm:ll-minimax} hold for all target and auxiliary populations, where $\ve_\bfnum{a} := Y_\bfnum{a} - \Eb(Y_\bfnum{a} \mid \Xv_\bfnum{a})$ for each $\ab \in \cA$. Then, there exists a constant $C_{\sF,\beta,L}'$, depending only on $C_{\sF,L},C_{\sF,U},\beta$ and $L$, such that
\begin{align*}
\liminf_{n_\bfnum{0}\to\infty}\inf_{\tf}\sup_{(f_\bfnum{0}, (f_\bfnum{a}:\ab\in\cA))\in \sF_\bfsnum{0}{\rmadd}^{s,\rmtl}(\beta,L)} \Pb_f\Big(&\norm{\tf-f_\bfnum{0}}_{p_\bfnum{0}}^2 \ge C_{\sF,\beta,L}' \cdot \Big\{sC(n_\bfnum{\cA}, s,d;\beta) \\
&\quad + sC(n_\bfnum{0}, s, d;\beta)\wedge C(n_\bfnum{0},s,d;\beta)^{\frac{1}{2}}\eta_\delta\wedge \eta_\delta^2 \Big\}\Big) \ge \frac{1}{2},
\end{align*}
where $\Pb_f$ denotes the probability measure under which the true regression function for the target population and the auxiliary populations are $f_\bfnum{0}$ and $f_\bfnum{a}$, respectively, and the infimum is taken over all measurable functions of the target and auxiliary samples.
\end{theorem}

\section{Numerical Evidences} \label{sec:numerical}

\subsection{Simulation}

In this section, we evaluate the finite-sample performance of the proposed transfer learning estimator in comparison with benchmark methods. We set $n_\bfnum{0}\in\{100,300\}$ for the target sample and $n_\bfnum{1}=n_\bfnum{2}=200$ for the auxiliary samples, so that two auxiliary datasets are available for the transfer learning algorithm. Specifically, we compare the performance of our estimator with the Nadaraya–Watson estimator of \cite{Lee2024} and with local linear estimators based on the same target sample size. The results of the Nadaraya–Watson estimator and the local linear estimators are denoted by ``NW'' and ``LL'' respectively, while the transfer learning estimator is denoted by ``TL''. We adopt the rule-of-thumb bandwidth introduced in \cite{Lee2024}, and each simulation is repeated $M=50$ times.

\subsubsection{Choice of penalty parameters}

For the Nadaraya–Watson and local linear estimators, we apply the BIC criterion of \cite{Lee2024}. In contrast, we select $\lambda_\cA^{\rm TL1}$ and $\lambda_\cA^{\rm TL2}$ using a BIC criterion adapted to our transfer learning framework. Specifically, let $(\hatf_\bfsnum{0}{j}^{\rmtl,\lambda_1,\lambda_2}: j \in [d])$ denote the transfer-learned component estimators, and let $\widehat \cS_\bfnum{0}^{\lambda_1,\lambda_2}$ denote the estimated active index set when $(\lambda_\cA^{\rmtl1},\lambda_\cA^{\rmtl2})=(\lambda_1,\lambda_2)$. The penalty parameters are chosen to minimize  
\begin{align*}
\log\left(\frac{1}{2n_\bfnum{0}}\suminum{0} \left(Y_\bfnum{0}^{i} - \sumj \hatf_\bfsnum{0}{j}^{\rmtl,\lambda_1,\lambda_2}(\Xv_\bfnum{0}^{i})\right)^2\right) + \sum_{j \in \widehat \cS_\bfnum{0}^{\lambda_1,\lambda_2}} \frac{\log (n_\bfnum{0}h_\bfsnum{0}{j})}{n_\bfnum{0}h_\bfsnum{0}{j}}.
\end{align*}
The minimization is carried out via a two-dimensional grid search.

\subsubsection{Similarity measure}

We examine the effectiveness of transfer learning by varying the probabilistic structural similarity and functional similarity measures introduced in the theoretical development.  

\smallskip\noindent
{\it Probabilistic structural similarity. }
We generate $\Xv_\bfnum{0}^{i}=(X_\bfsnum{0}{1}^i, \ldots, X_\bfsnum{0}{d}^i)$ following the procedure of \cite{Lee2024}. For each $j\in[d]$, let $U_j$ and $V$ be independent random variables uniformly distributed on $[0,1]$. Given $t \ge 0$, each component of $\Xv_\bfnum{0}^{i}$ is generated according to the distribution of $\Xv_\bfnum{0}=(X_\bfsnum{0}{1},\ldots,X_\bfsnum{0}{d})$ defined by  
\begin{align*}
X_\bfsnum{0}{j} = \frac{U_j + tV}{1+t}.
\end{align*}
As $t$ increases, the dependence among the covariates becomes stronger. Let $\Xv_\bfnum{0}'$ be an independent copy of $\Xv_\bfnum{0}$. For $\ab\in\{\bfnum{1},\bfnum{2}\}$, the auxiliary samples $\Xv_\bfnum{a}^{i}=(X_\bfsnum{a}{1}^i,\ldots,X_\bfsnum{a}{d}^i)$ are generated according to the distribution of $\Xv_\bfnum{a}=(X_\bfsnum{a}{1},\ldots,X_\bfsnum{a}{d})$ defined by  
\begin{align*}
X_\bfsnum{a}{j} = \begin{cases}
X_\bfsnum{0}{j}, & \text{if } W \le 1-\Delta_p, \\
\frac{X_\bfsnum{0}{j} + X_\bfsnum{0}{j}'}{2}, & \text{if } W > 1-\Delta_p,
\end{cases}
\end{align*}
where $W \sim \mathrm{Unif}[0,1]$ is independent of $U_j$ and $V$, and $\Delta_p \ge 0$. Clearly, the probabilistic dissimilarity increases with $\Delta_p$.

\smallskip\noindent 
{\it Functional similarity. }
The target responses are generated as  
\begin{align*}
Y_\bfnum{0}^{i} = \sumj f_\bfsnum{0}{j}(X_\bfsnum{0}{j}^i) + \ve_\bfnum{0}^{i}, \quad i \in [n_\bfnum{0}],
\end{align*}
where $\ve_\bfnum{0}^{i} \sim N(0,1)$. We assume that among the $d$ component functions, only $|\cS_\bfnum{0}|=12$ are active. Specifically, we set  
\begin{align*}
f_\bfsnum{0}{1}(u) = u-a_1, \quad f_\bfsnum{0}{2}(u) = (2u-1)^2 - a_2, \quad f_\bfsnum{0}{3}(u) = \frac{\sin(2\pi u)}{2-\sin(2\pi u)}-a_3, \\
f_\bfsnum{0}{4}(u) = \tfrac{1}{10}\sin(2\pi u) + \tfrac{2}{10}\sin(2\pi u) + \tfrac{3}{10}\sin^2(2\pi u) + \tfrac{4}{10}\cos^3(2\pi u) + \tfrac{5}{10}\sin^3(2\pi u),
\end{align*}
$f_\bfsnum{0}{j}(u) = \tfrac{3}{2}\, f_\bfsnum{0}{j-4}(u)$ for $5 \le j \le 8$ and $f_\bfsnum{0}{j}(u) = 2\, f_\bfsnum{0}{j-8}(u)$ for $9 \le j \le 12$. Here $a_j$ is chosen such that $\Eb(f_\bfsnum{0}{j}(X_\bfsnum{0}{j}))=0$ for $1 \le j \le 4$. For $j \ge 13$, we set $f_\bfsnum{0}{j} \equiv 0$.  

For the auxiliary samples, we generate
\begin{align*}
Y_\bfnum{a}^{i} = \sumj f_\bfsnum{a}{j}(X_\bfsnum{a}{j}^i) + \ve_\bfnum{a}^{i}, \quad i \in [n_\bfnum{a}],
\end{align*}
where $\ve_\bfnum{a}^{i} \sim N(0,1)$. The component functions $f_\bfsnum{a}{j}$ for $\ab \in \{\bfnum{1},\bfnum{2}\}$ coincide with $f_\bfsnum{0}{j}$ except in the cases summarized in Table~\ref{tab:functions}. In particular, $f_\bfsnum{a}{13} \not\equiv 0$ for $\ab \in \{\bfnum{1},\bfnum{2}\}$, whereas $f_\bfsnum{0}{13} \equiv 0$. Under this data-generation scheme, the functional dissimilarity between populations increases with $\Delta_f$.

\begin{table}[H]
\centering
\caption{Modified component functions for auxiliary samples.}
\label{tab:functions}
\begin{tabular}{cll}
\toprule
Population & Modified function & Index set \\ 
\midrule
 &
$f_\bfsnum{1}{j}(u)=f_\bfsnum{0}{j}(u)+\Delta_f \cdot f_\bfsnum{0}{j-3}(u)$ &
$j \in \{5,6,7\}$ \\

$\ab=\bfnum{1}$& $f_\bfsnum{1}{j}(u)=f_\bfsnum{0}{j}(u)+\Delta_f \cdot f_\bfsnum{0}{j-7}(u)$ &
$j \in \{8\}$ \\

& $f_\bfsnum{1}{j}(u)=\Delta_f \cdot \big(f_\bfsnum{1}{5}(u)+f_\bfsnum{1}{6}(u)+f_\bfsnum{1}{7}(u)+f_\bfsnum{1}{8}(u)\big)$ &
$j \in \{13\}$ \\[1ex]

 &
$f_\bfsnum{2}{j}(u)=f_\bfsnum{0}{j}(u)+\Delta_f \cdot f_\bfsnum{0}{j-7}(u)$ &
$j \in \{9,10,11\}$ \\

$\ab=\bfnum{2}$& $f_\bfsnum{2}{j}(u)=f_\bfsnum{0}{j}(u)+\Delta_f \cdot f_\bfsnum{0}{j-11}(u)$ &
$j \in \{12\}$ \\

& $f_\bfsnum{2}{j}(u)=\Delta_f \cdot \big(f_\bfsnum{2}{9}(u)+f_\bfsnum{2}{10}(u)+f_\bfsnum{2}{11}(u)+f_\bfsnum{2}{12}(u)\big)$ &
$j \in \{13\}$ \\
\bottomrule
\end{tabular}
\end{table}

\subsubsection{Simulation results}

\begin{figure}[ht]
  \centering

  \begin{subfigure}{\textwidth}
    \centering
    \includegraphics[width=\linewidth]{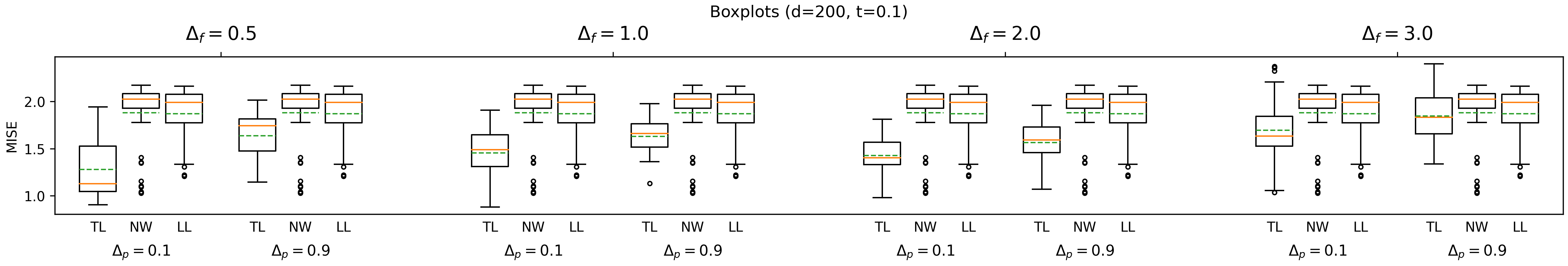}
  \end{subfigure}

  \begin{subfigure}{\textwidth}
    \centering
    \includegraphics[width=\linewidth]{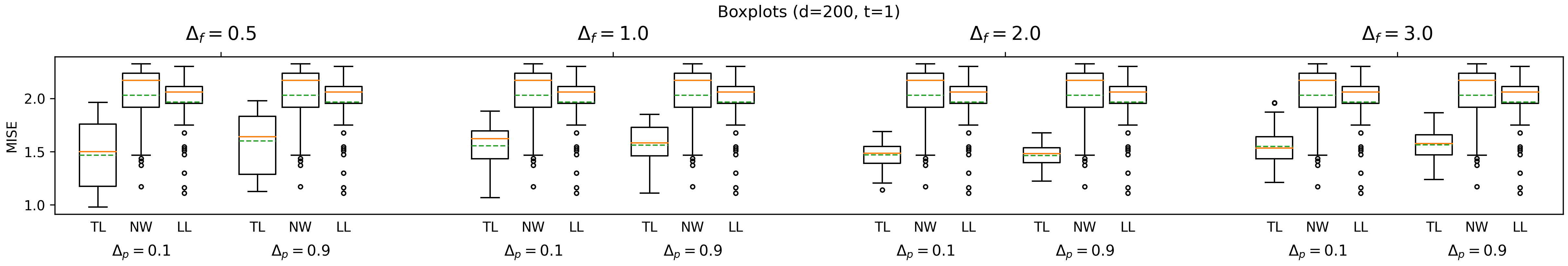}
  \end{subfigure}

  \begin{subfigure}{\textwidth}
    \centering
    \includegraphics[width=\linewidth]{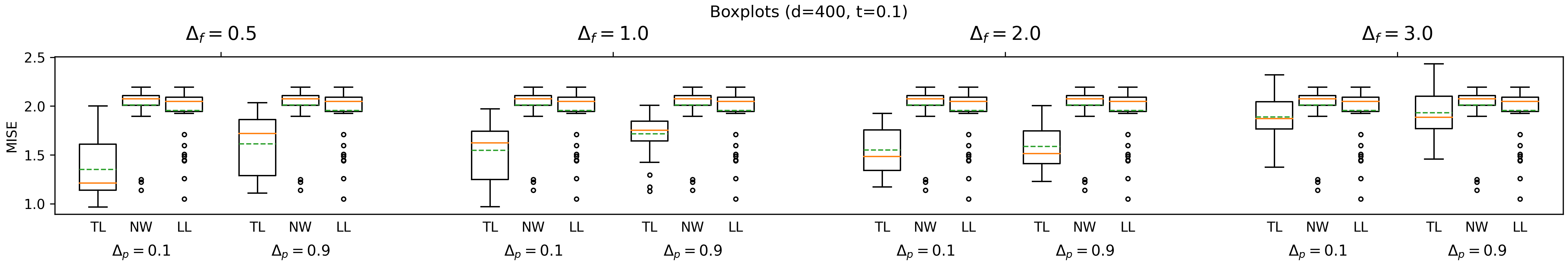}
  \end{subfigure}

  \begin{subfigure}{\textwidth}
    \centering
    \includegraphics[width=\linewidth]{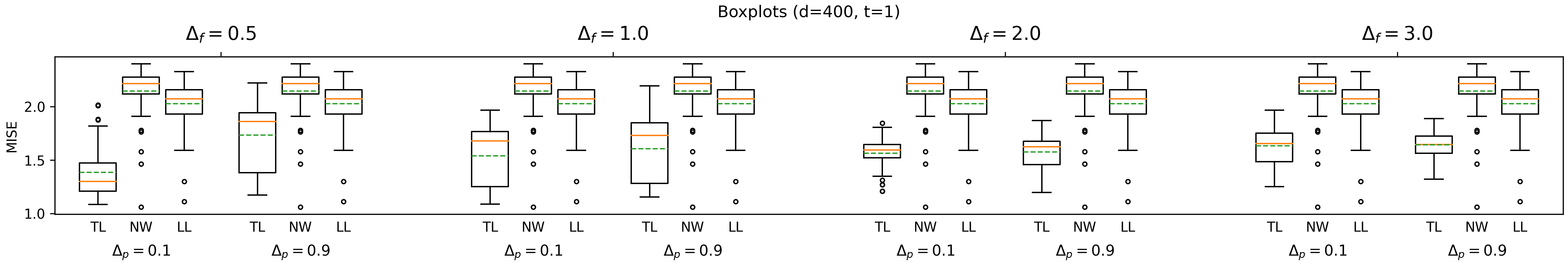}
  \end{subfigure}

  \caption{Boxplots of prediction errors across 32 scenarios when $n_\bfnum{0}=100$.}
\label{fig:simul1}
\end{figure}

\begin{figure}[ht]
  \centering

  \begin{subfigure}{\textwidth}
    \centering
    \includegraphics[width=\linewidth]{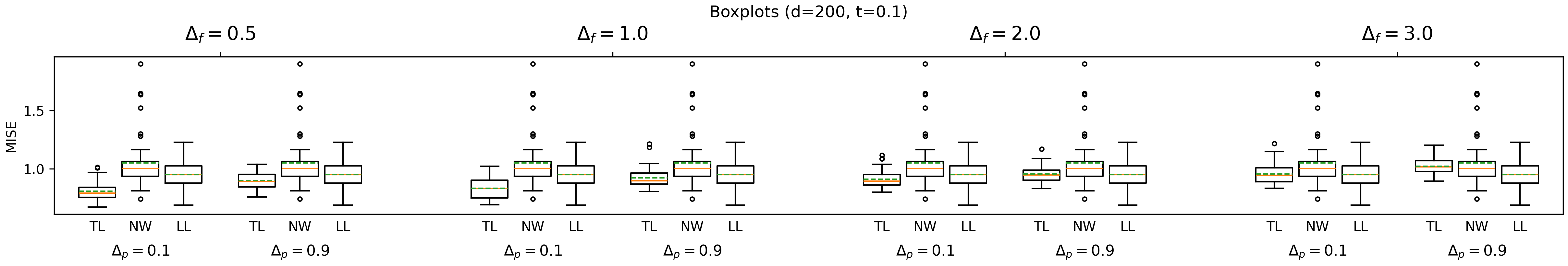}
  \end{subfigure}

  \begin{subfigure}{\textwidth}
    \centering
    \includegraphics[width=\linewidth]{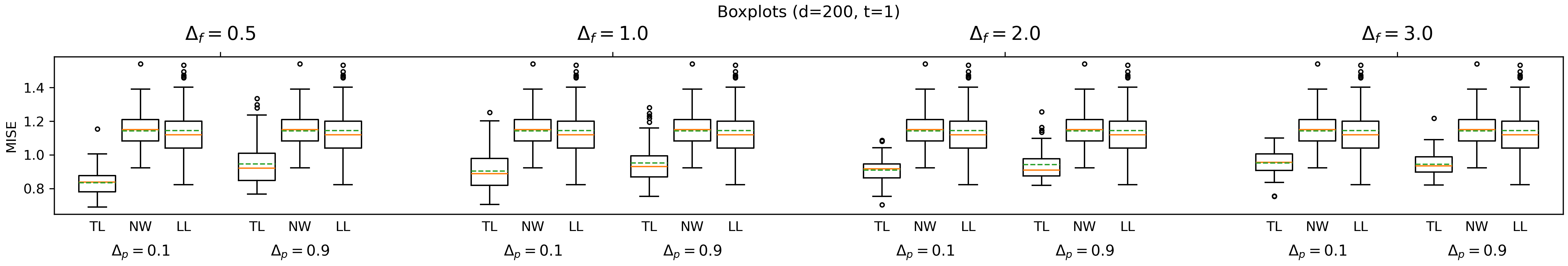}
  \end{subfigure}

  \begin{subfigure}{\textwidth}
    \centering
    \includegraphics[width=\linewidth]{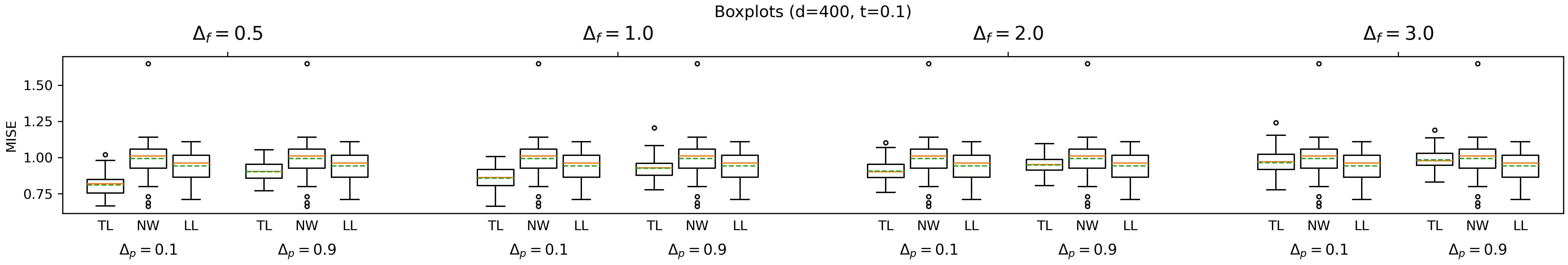}
  \end{subfigure}

  \begin{subfigure}{\textwidth}
    \centering
    \includegraphics[width=\linewidth]{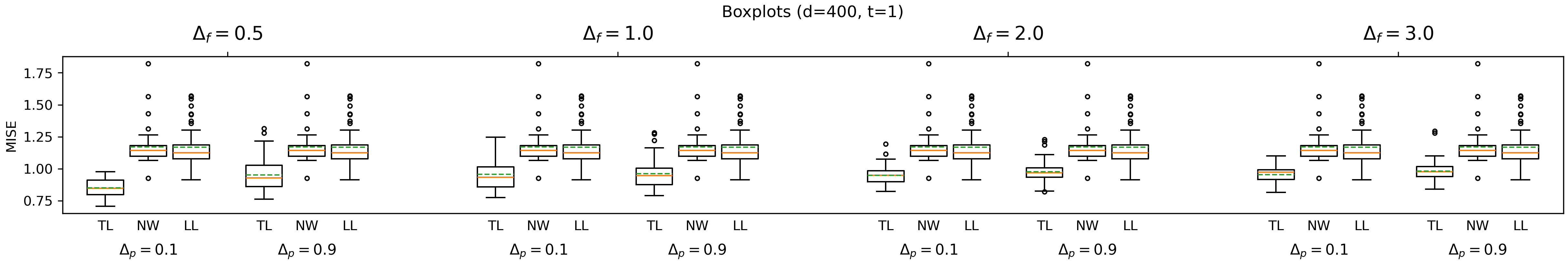}
  \end{subfigure}

  \caption{Boxplots of prediction errors across 32 scenarios when $n_\bfnum{0}=300$.}
\label{fig:simul2}
\end{figure}

To compare performance, we computed the mean integrated squared error (MISE). Specifically, for a generic regression function estimator $\tf_\bfnum{0}$, we defined  
\begin{align*}
{\rm MISE}(\tf_\bfnum{0}) := \int_{[0,1]^d} \left(\tf_\bfnum{0}(\xv) - f_\bfnum{0}(\xv)\right)^2 p_\bfnum{0}(\xv)\dxv.
\end{align*}
The values of MISE were computed for the NW, LL and TL estimators.
The results are summarized in boxplots of $M=50$ values of MISE. The target samples were generated for $d \in \{200,400\}$ and $t \in \{0.1,1.0\}$. For the auxiliary samples, we chose $\Delta_p \in \{0.1,0.9\}$ and $\Delta_f \in \{0.5,1.0,2.0,3.0\}$. Note that neither the Nadaraya–Watson estimator nor the local linear estimator is affected by $\Delta_p$ or $\Delta_f$, and that increasing either parameter enlarges the corresponding dissimilarity. In total, the combinations of $(d,t,\Delta_p,\Delta_f)$ yield 32 scenarios for each $n_\bfnum{0}\in\{100,300\}$. These are depicted in Figures~\ref{fig:simul1} and \ref{fig:simul2}, with each figure corresponding to the 32 scenarios for a given $n_\bfnum{0}$. For each figure, we present boxplots for 8 scenarios associated with each $(d,t)$ in each row, grouped by $\Delta_f$ and further split by $\Delta_p$ to facilitate comparison.

In Figure~\ref{fig:simul1} and \ref{fig:simul2}, the LL estimator outperforms the NW estimator, while the TL estimator generally outperforms LL, in terms of mean prediction error. The results also highlight the distinct effects of $\Delta_p$ and $\Delta_f$. An increase in $\Delta_p$ generally worsens the performance of the transfer learning estimator, consistent with the theoretical findings. Likewise, in line with the theory, the performance decreases as $\Delta_f$ increases. However, when $t=0.1$, corresponding to weak dependence among the covariates, local linear estimation performs sufficiently well that TL exhibits similar or even inferior performance compared to LL when $\Delta_f=3$. This phenomenon may be interpreted as an instance of negative transfer learning (\cite{perkins1992transfer}).

\subsection{Real data application}

\subsubsection{Data description}

Rapid advances in high-throughput profiling have enabled the construction of genomic predictors of drug response using large panels of cancer cell lines (\cite{barretina2012cancer,ferreira2013importance,garnett2012systematic}). As documented in \cite{barretina2012cancer,garnett2012systematic}, the CCLE provides a comprehensive resource linking gene expression to anti-cancer drug responses across cell lines. In the version analyzed here, the dataset reports responses to 24 drugs in 288 cancer cell lines, with each line characterized by expression levels for 18,988 genes. The complete list of drugs is given in Table~\ref{table:all_drug}. These data are widely employed in drug discovery for candidate screening (\cite{juan2018rationalizing}) and in studies of cancer biology and therapeutic efficacy (\cite{sharma2010cell}), owing to their cost-effectiveness and effectively unlimited replicative capacity (\cite{ferreira2013importance}).  

In our analysis, following \cite{Lee2024}, we take IC50 value as the response. For each drug, IC50 is the concentration that yields 50\% growth inhibition (\cite{barretina2012cancer}), and it serves as a summary measure of drug sensitivity across cell lines. Building on this setup, we extend the empirical analysis of \cite{Lee2024} to evaluate transfer-learned estimators for the five drugs listed in their Table~7. Among these (AZD6244, PD-0325901, Topotecan, 17-AAG, Irinotecan), we focus on the latter three: Topotecan, 17-AAG, and Irinotecan.

To implement transfer learning, we standardize the response across drugs so that IC50 values lie on a comparable scale. The goal is to align the regression functions and thereby facilitate the transfer of functional similarity. Empirically, this heuristic normalization performs well; accordingly, we adopt it throughout, rescaling the response within each drug to have sample standard deviation 2.5. For each of the three drugs, we first selected 3000 genes with the largest variances across the 288 cell lines and then chose 450 genes with the largest correlation coefficients with IC50. Thus, we considered $n_\bfnum{0}=288$ cell lines and $d=450$ features, scaling each covariate to lie between 0 and 1.  

\begin{table}[ht]
\centering
\caption{List of all drugs considered in the analysis, sorted alphabetically. Drugs in boldface indicate those used for our empirical study.}
\label{table:all_drug}
\begin{tabular}{cccc}
\hline
\textbf{17-AAG} & AEW541 & AZD0530 & AZD6244 \\
Erlotinib & \textbf{Irinotecan} & L-685458 & Lapatinib \\
LBW242 & Nilotinib & Nutlin-3 & Paclitaxel \\
Panobinostat & PD-0325901 & PD-0332991 & PF2341066 \\
PHA-665752 & PLX4720 & RAF265 & Sorafenib \\
TAE684 & TKI258 & \textbf{Topotecan} & ZD-6474 \\
\hline
\end{tabular}
\end{table}

\subsubsection{Transferable source detection}  

For notational convenience, for each target drug (Topotecan, 17-AAG, Irinotecan), let $\{(\Xv_\bfnum{b}^{i}, Y_\bfnum{b}^{i})\}_{i=1}^{n_\bfnum{b}}$, $\bfnum{b}\in\{1,2,\ldots,23\}$, denote the samples corresponding to the 23 drugs other than the given target drug. Auxiliary drugs were selected using the transferable source detection algorithm introduced in Section~\ref{subsec:apdx-source-detection}. Specifically, we randomly selected 200 samples from the full dataset and, for each $\bfnum{b}\in\{1,\ldots,23\}$, computed the score $\frac{1}{2}\sum_{r=1}^2 \widehat L_\bfnum{0}^{\langle r\rangle}(\hatfv_{\{\bfnum{0},\bfnum{b}\}}^{\rmtp, \langle r \rangle})$. This procedure was repeated twice, and the average of the two scores was used to rank the candidates. The top $|\cA_{\rm add}|$ drugs, corresponding to the $|\cA_{\rm add}|$ smallest scores, were then chosen as auxiliary drugs. The auxiliary drugs were determined after fixing the $d=450$ covariates with respect to the target drug, so that the target and auxiliary samples share the same covariates but differ in their responses. The top three auxiliary drugs selected by this procedure are summarized in Table~\ref{table:tdsource}.  

\begin{table}[ht]
\centering
\caption{Auxiliary drugs selected by the transferable source detection algorithm of Section~\ref{subsec:apdx-source-detection} for each target drug.}
\label{table:tdsource}
\begin{tabular}{cc}
\hline
Target drug & Auxiliary drugs (top 3) \\
\hline
Topotecan & LBW242, AZD0530, Erlotinib \\
Irinotecan & Erlotinib, 17-AAG, Paclitaxel \\
17-AAG & LBW242, Paclitaxel, Nutlin-3 \\
\hline
\end{tabular}
\end{table}

\subsubsection{Benchmark methods}
  
We compare our locally linear and transfer-learned estimators with the NW estimator of \cite{Lee2024} and the transfer-learning estimator for high-dimensional linear regression of \cite{tian2023transfer}. For the linear transfer-learning algorithm, we implemented their transferable source detection procedure. Specifically, we computed their score twice using the same random subsample of 200 observations from the full dataset, averaged the two scores, and then selected the top $|\cA_{\rm lin}|$ drugs accordingly. The top three auxiliary drugs identified by this procedure are reported in Table~\ref{table:tdsourcelinear}.  
Notably, the drugs selected by the linear detection algorithm significantly differ from those obtained by our procedure in Table~\ref{table:tdsource}. This may indicate that our method more effectively captures nonlinear functional similarity than the algorithm of \cite{tian2023transfer}.  

\begin{table}[ht]
\centering
\caption{Auxiliary drugs selected by the transferable source detection algorithm of \cite{tian2023transfer} for each target drug.}
\label{table:tdsourcelinear}
\begin{tabular}{cc}
\hline
Target drug & Auxiliary drugs (top 3) \\
\hline
Topotecan & Irinotecan, Paclitaxel, PF2341066 \\
Irinotecan & Topotecan, Panobinostat, Paclitaxel \\
17-AAG & RAF265, TAE684, Erlotinib \\
\hline
\end{tabular}
\end{table}

\subsubsection{Empirical results}

\begin{figure}[ht]
\centering
\includegraphics[width=\textwidth]{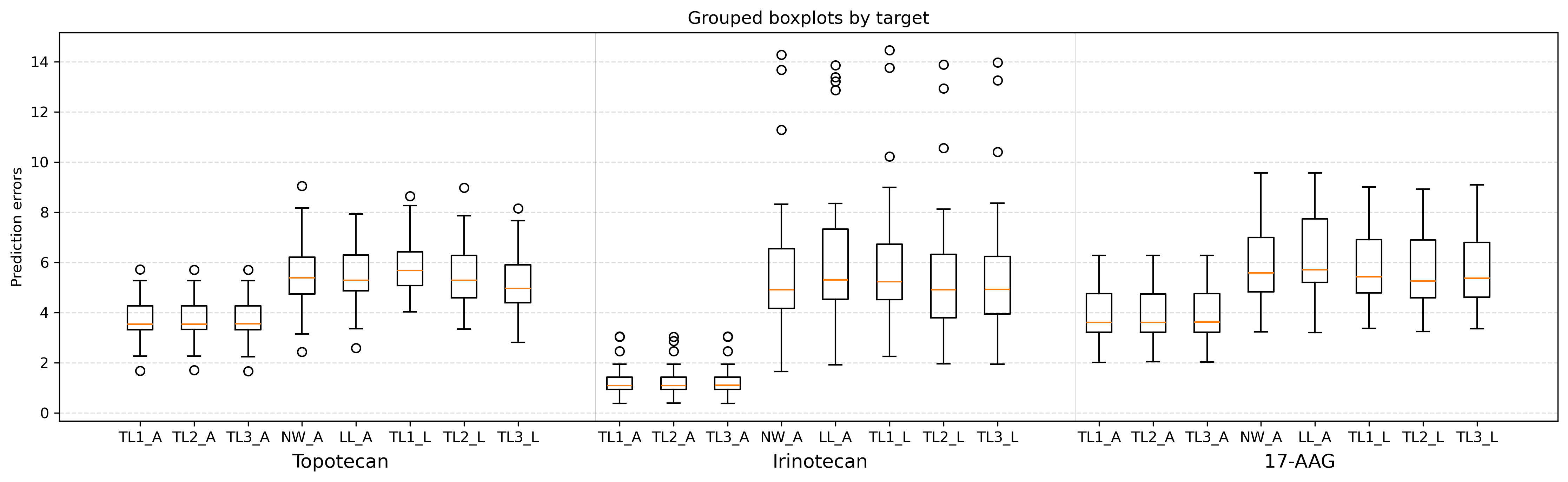}
\caption{Boxplots of prediction errors over 50 replications for each method.}
\label{fig:result}
\end{figure}

As for the transferable source detection algorithm, we randomly split the data into a training set of size 200 and a test set of size 88, and repeated this procedure $M=50$ times. For each replication, we computed the prediction error of a generic regression function estimator $\widetilde f_\bfnum{0}$, defined as  
\begin{align*}
{\rm PE}(\widetilde f_\bfnum{0}) = \frac{1}{n_{\rm test}}\sum_{i=1}^{n_{\rm test}} \left(Y_\bfnum{0}^{i} - \widetilde f_\bfnum{0}(\Xv_\bfnum{0}^{i})\right)^2.
\end{align*}
Boxplots of the 50 prediction errors for each method are displayed in Figure~\ref{fig:result}. In the notation, subscripts ``A'' indicate results from additive models, while subscripts ``L'' refer to the linear method of \cite{tian2023transfer}. The labels ``NW'' and ``LL'' denote the Nadaraya–Watson and locally linear estimators, respectively. In particular, TL$\ell\_{\rm A}$ and TL$\ell\_{\rm L}$ for $\ell\in\{1,2,3\}$ denote our proposed additive transfer-learned estimator and the linear transfer-learned estimator, respectively, with the top $\ell$ auxiliary samples selected by the source detection algorithm. The results show that TL$1\_{\rm A}$, TL$2\_{\rm A}$, and TL$3\_{\rm A}$ uniformly outperform the other methods. Moreover, our algorithm exhibits robustness, with its performance remaining stable regardless of the number of auxiliary drugs.  
For 17-AAG, although the linear transfer-learned estimators already improve upon the NW and locally linear estimators, the superior performance of our transfer-learned estimators is especially evident.

\section*{Acknowlegements}
The first author gratefully acknowledges Woonyoung Chang for the series of helpful discussions. 

\newpage

\section*{Appendix}

\renewcommand{\theequation}{A.\arabic{equation}}
\setcounter{equation}{0}

\renewcommand{\thesubsection}{A.\arabic{subsection}}
\setcounter{subsection}{0}

\renewcommand{\thelemma}{A.\arabic{lemma}}
\setcounter{lemma}{0}

\renewcommand{\thetheorem}{A.\arabic{theorem}}
\setcounter{theorem}{0}

\renewcommand{\theproposition}{A.\arabic{proposition}}
\setcounter{proposition}{0}

\renewcommand{\thecorollary}{A.\arabic{corollary}}
\setcounter{corollary}{0}

\renewcommand{\theremark}{A.\arabic{remark}}
\setcounter{remark}{0}

\renewcommand{\thefigure}{A.\arabic{figure}}
\setcounter{figure}{0}

\subsection{Transferable source detection} \label{subsec:apdx-source-detection}

To complete our theoretical development, we propose a transferable source detection algorithm along with its theoretical guarantee. We begin by introducing the algorithm and then present a theorem establishing that, under some conditions, the proposed method successfully identifies the true informative set $\cA$.

Suppose we observe datasets $\{(\Xv_\bfnum{b}^{i}, Y_\bfnum{b}^{i})\}_{i=1}^{n_\bfnum{b}}$ for $\bfnum{b} \in\Bc$. We assume that each dataset shares a common additive structure of the form 
\begin{align*} 
\Eb[Y_\bfnum{b} \mid \Xv_\bfnum{b}] = \Eb[Y_\bfnum{b}] + \sumj f_\bfsnum{b}{j}(X_\bfsnum{b}{j}), 
\end{align*} 
where $f_\bfsnum{b}{j}$ denotes the $j$th additive component in the $\bfnum{b}$th population.
The goal is to identify a subset $\cA \subset \Bc$ such that the transfer learning procedure described in Section~\ref{sec:tl} can be effectively applied using the selected sources. We basically follow the source detection algorithm introduced in \cite{tian2023transfer}, which is tailored for our nonparametric setting. 

Let the target sample \(\{(\Xv_\bfnum{0}^{i}, Y_\bfnum{0}^{i})\}_{i=1}^{n_\bfnum{0}}\) be randomly and equally divided into two disjoint subsamples, denoted by \(\{(\Xv_\bfnum{0}^{{i},\langle r\rangle}, Y_\bfnum{0}^{{i},\langle r\rangle})\}_{i=1}^{n_\bfnum{0}/2}\) for \(r=1,2\). For each \(r = 1,2\), we first construct the estimator \(\hatfv_\bfnum{0}^{\rmtp,\langle r\rangle}\) via the locally linear fLasso algorithm described in Section~\ref{subsec:ll-estimation}, using the subsample \(\{(\Xv_\bfnum{0}^{i}, Y_\bfnum{0}^{i})\}_{i=1}^{n_\bfnum{0}} \setminus \{(\Xv_\bfnum{0}^{{i},\langle r\rangle}, Y_\bfnum{0}^{{i},\langle r\rangle})\}_{i=1}^{n_\bfnum{0}/2}\) and the penalty parameter \(\lambda_\bfnum{0}^{\langle r\rangle}\). In this stage, the bandwidths are chosen to be uniformly asymptotic to \(n_\bfnum{0}^{-1/5}\).
Additionally, for each \(r = 1,2\), we construct the first-stage transfer-learned estimator \(\hatfv_{\{\bfnum{0}, \bfnum{b}\}}^{\rmtp, \langle r\rangle}\) as introduced in Section~\ref{subsec:tl-estimation}. In this procedure, the same subsample \(\{(\Xv_\bfnum{0}^{i}, Y_\bfnum{0}^{i})\}_{i=1}^{n_\bfnum{0}} \setminus \{(X_\bfnum{0}^{{i},\langle r\rangle}, Y_\bfnum{0}^{{i},\langle r\rangle})\}_{i=1}^{n_\bfnum{0}/2}\) is used as the target sample, and the full sample \(\{(\Xv_\bfnum{b}^{i}, Y_\bfnum{b}^{i})\}_{i=1}^{n_\bfnum{0}}\) is used as the auxiliary source. The bandwidths in this stage are set to be uniformly asymptotic to \((n_\bfnum{0} + 2n_\bfnum{b})^{-1/5}\), and the penalty parameter \(\lambda_{\{\bfnum{0},\bfnum{b}\}}^{\rmtl1,\langle r\rangle}\) is applied for the estimation.

Define 
\[
\widehat L_\bfnum{0}^{\langle r\rangle}(\gv^\rmtp) := \frac{2}{n_\bfnum{0}}\sum_{i=1}^{n_\bfnum{0}/2}\Big|g(\Xv_\bfnum{0}^{{i},\angl{3-r}}) - \hatf_\bfnum{0}^\angl{r}(\Xv_\bfnum{0}^{{i},\angl{3-r}})\Big|.
\]
In this algorithm, we compare the deviations between the target-only estimator and the transfer-learned estimator by evaluating the loss differences between $\widehat L_\bfnum{0}^{\langle r\rangle}(\hatfv_{\{\bfnum{0}, \bfnum{b}\}}^{\rmtp, \langle r\rangle})$ and $\widehat L_\bfnum{0}^{\langle r\rangle}(\hatfv_\bfnum{0}^{\rmtp,\langle r\rangle})$.
The $\bfnum{b}$th sample is rejected as an auxiliary (informative) source if
\[
\frac{1}{2}\sum_{r=1}^2 \widehat L_\bfnum{0}^{\langle r\rangle}(\hatfv_{\{\bfnum{0}, \bfnum{b}\}}^{\rmtp, \langle r\rangle})\ge \frac{c_{\rm SD}}{4}
\]
where \( c_{\rm SD} > 0 \) is a constant specified later in Theorem~\ref{thm:apdx-source-detection}. Notably, this method does not require a specific choice of the bandwidth parameter $\eta_\delta$.

We now present a simple theoretical guarantee for the above procedure. Let \( \widehat{\mathcal{A}} \) denote the set of sources identified as informative by the source detection algorithm. For theoretical simplicity, we assume that all datasets \(\{(\Xv_\bfnum{b}^{i}, Y_\bfnum{b}^{i})\}_{i=1}^{n_\bfnum{b}}\), including the target sample, are drawn independently from mutually distinct populations. Although strong, this assumption is also implicitly adopted in \cite{tian2023transfer} to establish theoretical guarantees for their version of the source detection algorithm. Let \( \fv_\bfnum{\{0,b\}}^\rmtp \) denote the true objective corresponding to the estimator \( \hatfv_\bfnum{\{0,b\}} \). The proof of Theorem~\ref{thm:apdx-source-detection} is deferred to the supplementary material.

\begin{theorem} \label{thm:apdx-source-detection}
Assume the conditions in Corollary~\ref{cor:ll-bound-error-pop} and \ref{cor:tl-total}. Also, assume that
\begin{align*}
\Eb\Big[\Big|f_\bfnum{\{0,b\}}(\Xv_\bfnum{0}) - f_\bfnum{0}(\Xv_\bfnum{0})\Big|\Big] &\ge c_{\rm SD}, \quad \bfnum{b}\not\in\cA,
\end{align*}
for some absolute constant $c_{\rm SD}>0$. Then, for any \( \xi > 0 \), there exists \( N = N(\xi) > 0 \) such that if \( \min_{\bfnum{b}\in\{\bz\}\cup \cA}n_\bfnum{b} > N(\xi) \), it holds that
$\Pb(\widehat \cA = \cA) \ge 1 - \xi$.
\end{theorem}
\newpage

\renewcommand{\theequation}{S.\arabic{equation}}
\setcounter{equation}{0}

\renewcommand{\thesubsection}{S.\arabic{subsection}}
\setcounter{subsection}{0}

\renewcommand{\thelemma}{S.\arabic{lemma}}
\setcounter{lemma}{0}

\renewcommand{\thetheorem}{S.\arabic{theorem}}
\setcounter{theorem}{0}

\renewcommand{\theproposition}{S.\arabic{proposition}}
\setcounter{proposition}{0}

\renewcommand{\thecorollary}{S.\arabic{corollary}}
\setcounter{corollary}{0}

\renewcommand{\theremark}{S.\arabic{remark}}
\setcounter{remark}{0}

\renewcommand{\thefigure}{S.\arabic{figure}}
\setcounter{figure}{0}

\section*{Supplementary materials}

\subsection{A sufficient condition for norm compatibility} \label{subsec:apdx-norm-compt}

The following proposition establishes an explicit norm-compatibility condition between the additive space $\sH_\mathrm{add}^\rmtp$ and the product space $\sH_\mathrm{prod}^\rmtp$. While the argument parallels earlier results for the Nadaraya Watson setting \cite{Lee2024}, the locally linear setting necessitates a direct modification of the classical approach. Hence, we only sketch the proof of the following proposition.

\begin{proposition} \label{prop:suff-cond-norm-cmpt}
Assume that conditions (P1)--(P2) hold for the target population. Also, for some fixed $\alpha>0$, condition (B-$\alpha$) holds with the reference bandwidth of $h_\bfsnum{0}{j}$ denoted by $h_\bfnum{0}$. Also suppose there exist absolute constants $\varphi > 0$ and $0 < \psi < (\frac{C_{p,L}^\rmuniv\mu_2}{C_{p,L}^\rmuniv\mu_2 + 4\sqrt{\varphi}})^2$ such that after an appropriate permutation of indices $1,2,\ldots,d$ the following holds:
\begin{align*}
\int_{[0,1]^2} \left( p_\bfsnum{0}{jk}(x_j,x_k) - p_\bfsnum{0}{j}(x_j)p_\bfsnum{0}{k}(x_k) \right)^2 \dxj \dxk \le \varphi \cdot \psi^{|j-k|},
\end{align*}
for all $(j,k) \in [d]^2$. Then there exists an absolute constant $0 < C_\bfnum{0} < \infty$ such that if $\gv^\rmtp = (g_j^\rmtp : j\in[d])$ satisfies the constraints $\int_0^1 g_j^\rmv(x_j)^\top \tp_\bfsnum{0}{j}(x_j)\dxj=0$ for $j\in[d]$, and
\begin{align*}
\sum_{j \not\in \cS_\bfnum{0}} \norm{g_j^\rmtp}_\tMnum{0} \le C \sum_{j \in \cS_\bfnum{0}} \norm{g_j^\rmtp}_\tMnum{0},
\end{align*}
for some $0 < C < \infty$, then it holds that
\begin{align*}
\left\Vert \sumj g_j^\rmtp \right\Vert_\tMnum{0}^2 \ge \left( \frac{C_{p,L}^\rmuniv\mu_2 - \sqrt{\psi}(C_{p,L}^\rmuniv\mu_2 + 4\sqrt{\varphi})}{(1 - \sqrt{\psi})C_{p,L}^\rmuniv\mu_2} - C_\bfnum{0}(1 + C)^2 \cdot \sqrt{h_\bfnum{0}}|\cS_\bfnum{0}| \right) \sumj \norm{g_j^\rmtp}_\tMnum{0}^2.
\end{align*}
\end{proposition}

\subsection{A concentration bound for degenerate $U$-statistics}

In this section, we present a concentration inequality for degenerate $U$-statistics of a specific form. Although a related result and its proof appear as Theorem~1 in \cite{chakrabortty2018tail}, we restate them here with modifications for completeness and clarity, using our own notation and assumptions. A key modification involves the definition of the term $\Omega_{n,1}$ in Theorem~\ref{thm:u-stat}. For more detailed discussion, see Remark~\ref{remark:thms1}. We adopt more general notation to facilitate the broader applicability of our results.

Let $\Wb$ be a symmetric measurable function and define $Z_i = (\Xv^i, \ve^i)$ for $1 \le i \le n$. We assume that $\ve^i$ satisfy condition (R-$\alpha$) for some fixed $\alpha > 0$. Note that
\begin{align*}
\Eb[|\ve^i|^2 \mid \Xv^i] 
&= \int_0^1 \Pb(|\ve^i| \ge \sqrt{t} | \Xv^i)\, \rmd t \le \frac{4}{\alpha} \Gamma\left(\frac{2}{\alpha}\right) C_\ve^2,
\end{align*}
almost surely for all $1 \le i \le n$. 
Consider the degenerate $U$-statistic
\begin{align*}
\Ub_n := \sumiiprime \Wb_n(Z_i, Z_{i'}).
\end{align*}
We say that $\Ub_n$ is degenerate if
\begin{align*}
\Eb[\Wb_n(Z_i, Z_{i'}) | Z_i] = \Eb[\Wb_n(Z_i, Z_{i'}) | Z_{i'}] = 0, \quad \text{for all } 1 \le i \ne i' \le n.
\end{align*}
Suppose further that $\Wb_n$ takes the specific form
\begin{align*}
\Wb_n(Z_i, Z_{i'}) = \ve^i W_n(\Xv^i, \Xv^{i'}) \ve^{i'},
\end{align*}
for some symmetric measurable function $W_n$ satisfying $\sup_{\xv, \xv' \in [0,1]^d} |W_n(\xv, \xv')|=: B_{n,W}<\infty$. Put $\alpha^*:=\alpha\wedge 1$. To describe the concentration inequality, we define the additional quantities. Let $\Omega_{n,1} := B_{n,W} (\log n)^{\frac{1}{\alpha^*}+\frac{2}{\alpha}}$. Moreover, define
\begin{align*}
\Omega_{n,2} &:= \left( \sumiiprime \Eb\left(W_n(\Xv^i, \Xv^{i'})^2\right) \right)^{\frac{1}{2}}, \\
\Omega_{n,3} &:= \sup\left\{ \sumiiprime \Eb\left(\eta_i(\Xv^i) W_n(\Xv^i, \Xv^{i'}) \zeta_{i'}(\Xv^{i'})\right) : \sum_{i=1}^n \Eb(\eta_i(\Xv^i)^2) \le 1,\, \sum_{i=1}^n \Eb(\zeta_{i}(\Xv^{i})^2) \le 1 \right\}, \\
\Omega_{n,4} &:= (\log n)^{\frac{1}{\alpha}} \sup_{\xv \in [0,1]^d} \left( \sum_{i=1}^n \Eb\left(W_n(\Xv^i,\xv)^2\right)\right)^{\frac{1}{2}}, \\
\Omega_{n,5} &:= (\log n)^{\frac{1}{2}} \Omega_{n,4} + (\log n) \Omega_{n,1}.
\end{align*}
The terms $\Omega_{n,\ell}$ for $1\le \ell\le 5$ also appear in Theorem~3.2 of \cite{Gine2000exp}. Now, we state the theorem. The proof is deffered to Section~\ref{sec:supple-pf-thm-u-stat}

\begin{theorem} \label{thm:u-stat}
There exists a constant $C_\alpha$ depending only on $\alpha>0$, such that 
\begin{align*}
\Pb\left(|\Ub_n|\ge C_\alpha\left(t^{\frac{2}{\alpha^*}}\Omega_{n,1}+ t^{\frac{1}{2}}\Omega_{n,2} + t\Omega_{n,3} + t^{\frac{1}{2}+\frac{1}{\alpha^*}}\Omega_{n,4} + t^{\frac{1}{\alpha^*}}\Omega_{n,5} \right)\right) \le 2\exp(-t), 
\end{align*}
where $\alpha^* = \alpha\wedge 1$. 
\end{theorem}

\subsection{Proof of Theorem~\ref{thm:u-stat}} \label{sec:supple-pf-thm-u-stat}
Before presenting the proof, we introduce five lemmas that will be used in establishing the main result. The proofs of Lemma~\ref{lem:start-u-stat} and \ref{lem:bound-Mve} are deferred to Section~\ref{subsubsec:supple-pf-lem-u-stat} and \ref{subsubsec:supple-pf-lem-bound-Mve}, while the proofs of the remaining lemmas are omitted, as they follow directly from results in the existing literature. The corresponding references are indicated in each lemma. In this proof, we use the notation $C_\alpha$ to denote a constant that depends only on $\alpha$, which may take different values in different instances.

For a random variable $V$, we define its $\ell$-norm by
\begin{align*}
\norm{V}_\ell := \Eb( |V|^\ell )^{\frac{1}{\ell}}.
\end{align*}
Additionally, for $\Phi_\alpha(x) := \exp(x^\alpha) - 1$, we define the Orlicz norm of $U$ with respect to $\Phi_\alpha$ as
\begin{align*}
\norm{V}_{\Phi_\alpha} := \inf\left\{ C > 0 : \Eb\left( \Phi_\alpha\left( \frac{|V|}{C} \right) \right) \le 1 \right\}.
\end{align*}

\begin{lemma}[Theorem 3.2 in \cite{Gine2000exp}] \label{lem:thm3.2gine}
Let $h$ be a bounded bivariate function, and let $(V_i : i \in [n])$ and $(V_i' : i \in [n])$ be two independent sequences of identically distributed random variables, where $V_i \overset{d}{=} V_i'$ for all $i\in[n]$. Consider the decoupled $U$-statistic $\sumiiprime h(V_i, V_{i'}')$, and assume it is degenerate of order 2. Define $h_{i,i'} := h(V_i, V_{i'}')$. Then, there exists an absolute constant $0 < C < \infty$ such that for any $\ell \ge 2$,
\begin{align*}
\normbig{\sumiiprime h_{i,i'}}_\ell &\le C\Biggr(\ell^{\frac{1}{2}}\left(\sumiiprime \Eb(h_{i,i'}^2)\right)^{\frac{1}{2}} + \ell\norm{(h_{i,i'})}_{L^2\to L^2} \\
&\quad + \ell^{\frac{3}{2}}\left\{\Eb\left(\max_{i\in[n]}\Eb\left(\sum_{i'=1}^n h_{i,i'}^2\Biggr|V_i\right)^{\frac{1}{2}}\right) + \Eb\left(\max_{i'\in[n]}\Eb\left(\sum_{i=1}^n h_{i,i'}^2\Biggr|V_{i'}'\right)^{\frac{1}{2}}\right)\right\} \\
&\quad + \ell^2 \Eb\left(\max_{1\le i\ne i'\le n}|h_{i,i'}|^\ell\right)^{\frac{1}{\ell}}\Biggr),
\end{align*}
where
\begin{align*}
\norm{(h_{i,i'})}_{L^2 \to L^2} := \sup\left\{ \sumiiprime \Eb\left( \eta_i(V_i) h_{i,i'} \zeta_{i'}(V_{i'}') \right) : \sumin \Eb \eta_i(V_i)^2 \le 1, \, \sumin \Eb \zeta_{i}(V_{i}')^2 \le 1 \right\}.
\end{align*}
\end{lemma}

For Lemmas~\ref{lem:prop6.8ledoux} and \ref{lem:prop6.21ledoux}, we define the $\ell$-norm and the Orlicz norm for a random element $V$ taking values in a Banach space $(\Bs, \norm{\cdot}_\Bs)$ as follows:
\[
\norm{V}_\ell := \Eb(\norm{V}_\Bs^\ell)^{\frac{1}{\ell}}, \quad  \norm{V}_{\Phi_\alpha} := \inf\left\{ C > 0 : \Eb\left( \Phi_\alpha\left( \frac{\norm{V}_\Bs}{C} \right) \right) \le 1 \right\}.
\]

\begin{lemma}[Proposition 6.8 in \cite{ledoux2011prob}] \label{lem:prop6.8ledoux}
Let $0 < \ell < \infty$ and let $(V_i : i \in [n])$ be independent random elements taking values in an $L_p$ space over a Banach space $(\Bs, \norm{\cdot}_\Bs)$. Define the partial sums $S_k := \sum_{i=1}^k V_i$ for $k \le n$. Then, for
\[
t_0 := \inf\left\{ t > 0 : \Pb\left( \max_{k \le n} \norm{S_k}_\Bs > t \right) \le (2 \cdot 4^\ell)^{-1} \right\},
\]
it holds that
\[
\Eb\left( \max_{k \le n} \norm{S_k}_\Bs^\ell \right) \le 2 \cdot 4^\ell \Eb\left( \max_{i \in [n]} \norm{V_i}_\Bs^\ell \right) + 2(t_0)^\ell.
\]
\end{lemma}

\begin{lemma}[Proposition 6.21 in \cite{ledoux2011prob}] \label{lem:prop6.21ledoux}
There exists a constant $C_\alpha > 0$, depending only on $\alpha$, such that for any finite sequence $(V_i : i \in [n])$ of independent mean-zero random elements taking values in the Orlicz space with respect to $\Phi_\alpha$ over a Banach space $(\Bs, \norm{\cdot}_\Bs)$, the following bounds hold. If $0 < \alpha \le 1$, then
\[
\left\Vert \sum_{i=1}^n V_i \right\Vert_{\Phi_\alpha} \le C_\alpha \left( \left\Vert \sum_{i=1}^n V_i \right\Vert_1 + \left\Vert \max_{i \in [n]} \norm{V_i}_\Bs \right\Vert_{\Phi_\alpha} \right).
\]
If $1 < \alpha \le 2$, then
\[
\left\Vert \sum_{i=1}^n V_i \right\Vert_{\Phi_\alpha} \le C_\alpha \left( \left\Vert \sum_{i=1}^n V_i \right\Vert_1 + \left( \sum_{i=1}^n \norm{V_i}_{\Phi_\alpha}^\beta \right)^{1/\beta} \right),
\]
where $\frac{1}{\alpha} + \frac{1}{\beta} = 1$.
\end{lemma}

\begin{lemma}[Symmetrization] \label{lem:start-u-stat}
For any $\ell \ge 1$, it holds that
\begin{align*}
\norm{\Ub_n}_\ell \le 48 \left\Vert \sumiiprime w_i \Wb(Z_i, Z_{i'}') w_{i'}' \right\Vert_\ell,
\end{align*}
where $(w_i, w_i' : i \in [n])$ are Rademacher random variables that are independent of $(Z_i, Z_i' : i\in[n])$. Here, $(w_i : i \in [n])$ is independent of $(w_i' : i \in [n])$ and $Z_1' = ({\Xv'}^1, {\ve'}^1), \ldots, Z_n' = ({\Xv'}^n, {\ve'}^n)$ are $n$ independent copies of $(\Xv, \ve)$ and are also independent of $Z_1, \ldots, Z_n$.
\end{lemma}

\begin{lemma}[Maximal inequality] \label{lem:bound-Mve}
Let $\Xb_n:=\{\Xv^1, \ldots, \Xv^n\}$. It holds almost surely that
\begin{align*}
\Eb\left(\max_{i \in [n]} |\ve^i| \,\big|\, \Xb_n\right) \le C_\alpha (\log n)^{\frac{1}{\alpha}}.
\end{align*}
Moreover,
\begin{align*}
\left\Vert \max_{i \in [n]} |\ve^i| \right\Vert_{\Phi_\alpha \mid \Xb_n} \le C_\alpha (\log n)^{\frac{1}{\alpha}}, \quad \text{a.s.},
\end{align*}
where $\norm{\cdot}_{\Phi_\alpha \mid \Xb_n}$ denotes the Orlicz norm with respect to $\Phi_\alpha$, conditional on $\Xb_n$.
\end{lemma}

\smallskip\noindent
{\it Proof of Theorem~\ref{thm:u-stat}.}
\smallskip

We claim that
\begin{align} \label{claim-lem:start-u-stat}
\norm{\Ub_n}_\ell  \le C_\alpha \left( \ell^{\frac{2}{\alpha^*}} \Omega_{n,1} + \ell^{\frac{1}{2}} \Omega_{n,2} + \ell \Omega_{n,3} + \ell^{\frac{1}{2} + \frac{1}{\alpha^*}} \Omega_{n,4} + \ell^{\frac{1}{\alpha^*}} \Omega_{n,5} \right), \quad \ell\ge 2.
\end{align}
Applying Markov's inequality to the claim in \eqref{claim-lem:start-u-stat} yields the desired result.
 
From Lemma~\ref{lem:start-u-stat}, it suffices to show that
\begin{align}
\left\Vert \sumiiprime w_i \Wb_n(Z_i, Z_{i'}') w_{i'}' \right\Vert_\ell \le C_\alpha \left( \ell^{\frac{2}{\alpha^*}} \Omega_{n,1} + \ell^{\frac{1}{2}} \Omega_{n,2} + \ell \Omega_{n,3} + \ell^{\frac{1}{2} + \frac{1}{\alpha^*}} \Omega_{n,4} + \ell^{\frac{1}{\alpha^*}} \Omega_{n,5} \right), \quad \ell\ge 2.
\end{align}
Fix $\ell\ge 2$. To this end, we employ a truncation technique. Let $\Xb_n := \{\Xv^1, \ldots, \Xv^n\}$ and ${\Xb}_n'=\{{\Xv'}^1, \ldots, {\Xv'}^n\}$, and define
\[
M_\ve := 8\Eb\left( \max_{i \in [n]} |\ve^i| \,\big|\, \Xb_n \right).
\]
Define the truncated variables
\begin{align*}
T_{i,1} &:= \ve^i \cdot I(|\ve^i| \le M_\ve), \quad T_{i,2} := \ve^i \cdot I(|\ve^i| > M_\ve), \\
T_{i,1}' &:= {\ve'}^{i} \cdot I(|{\ve'}^{i}| \le M_\ve), \quad T_{i,2}' := {\ve'}^{i} \cdot I(|{\ve'}^{i}| > M_\ve).
\end{align*}
Observe that
\begin{align*}
\Wb_n(Z_i, Z_{i'}') &= \ve^i W_n(\Xv^i, {\Xv'}^{i'}) \ve^{i'} \\
&= T_{i,1} W_n(\Xv^i, {\Xv'}^{i'})  T_{i',1}' + T_{i,1}W_n(\Xv^i, {\Xv'}^{i'})  T_{i',2}' \\
&\hspace{3.65cm} + T_{i,2} W_n(\Xv^i, {\Xv'}^{i'})  T_{i',1}' + T_{i,2} W_n(\Xv^i, {\Xv'}^{i'})  T_{i',2}'.
\end{align*}
This decomposition yields
\[
\sum_{1 \le i < i' \le n} w_i \Wb_n(Z_i, Z_{i'}') w_{i'}' = \Uc_{n,1} + \Uc_{n,2} + \Uc_{n,3} + \Uc_{n,4},
\]
where
\begin{align*}
\Uc_{n,1} &:= \sumiiprime w_i T_{i,1} W_n(\Xv^i, {\Xv'}^{i'})  T_{i',1}' w_{i'}', \\
\Uc_{n,2} &:= \sumiiprime w_i T_{i,2} W_n(\Xv^i, {\Xv'}^{i'})  T_{i',1}' w_{i'}', \\
\Uc_{n,3} &:= \sumiiprime w_i T_{i,1} W_n(\Xv^i, {\Xv'}^{i'})  T_{i',2}' w_{i'}', \\
\Uc_{n,4} &:= \sumiiprime w_i T_{i,2} W_n(\Xv^i, {\Xv'}^{i'})  T_{i',2}' w_{i'}'.
\end{align*}
It is worth noting that each of $\Uc_{n,1}, \Uc_{n,2}, \Uc_{n,3}, \Uc_{n,4}$ is a degenerate and decoupled $U$-statistic.

First, we bound $\norm{\Uc_{n,1}}_\ell$. Let $\Vb_n:=\{V_1, \ldots, V_n\}$ and $\Vb_n'=\{V_1',\ldots, V_n'\}$ with $V_i=(w_i, \Xv^i, \ve^i)$ and $V_i'=(w_i', {\Xv'}^i, {\ve'}^i)$. From Lemma~\ref{lem:thm3.2gine}, we observe that 
\begin{align*}
\norm{\Uc_{n,1}}_\ell \le C_0\left(\ell^{\frac{1}{2}}\cdot \Uc_{n,1}^{(1)} + \ell\cdot \Uc_{n,1}^{(2)} + \ell^{\frac{3}{2}}\cdot \Uc_{n,1}^{(3)} + \ell^2\cdot \Uc_{n,1}^{(4)}\right),
\end{align*}
where $0<C_0<\infty$ is an absolute constant and  
\begin{align*}
\Uc_{n,1}^{(1)} &:= \left(\sumiiprime \Eb\left((T_{i,1})^2 W_n(\Xv^i, {\Xv'}^{i'}) ^2(T_{i',1}')^2\right)\right)^{\frac{1}{2}}, \\
\Uc_{n,1}^{(2)} &:= \sup\Biggr\{\sumiiprime \Eb(\eta_i(V_i)w_iT_{i,1}W_n(\Xv^i, {\Xv'}^{i'}) T_{i',1}'w_{i'}'\zeta_{i'}(V_{i'}')): \\
&\hspace{6cm}\sumin \Eb(\eta_i(V_i)^2)\le1, \, \sumin \Eb(\zeta_{i}(V_i')^2)\le 1\Biggr\}, \\
\Uc_{n,1}^{(3)} &:= \Eb\left(\max_{i'\in[n]}\Eb\left(\sumin (T_{i,1})^2 W_n(\Xv^i, {\Xv'}^{i'})^2(T_{i',1}')^2\Big|\Vb_n'\right)^{\frac{1}{2}}\right) \\
\Uc_{n,1}^{(4)} &:= \Eb\left(\max_{1\le i\ne i'\le n}|T_{i,1}W_n(\Xv^i, {\Xv'}^{i'}) T_{i',1}'|^\ell\right)^{\frac{1}{\ell}}.
\end{align*}
Note that 
\begin{align*}
\Eb\left((T_{i,1})^2 W_n(\Xv^i, {\Xv'}^{i'})^2(T_{i',1}')^2\right) &\le \Eb\left(\Eb(|\ve_i|^2|\Xv_i)\Eb\left( W_n(\Xv^i, {\Xv'}^{i'})^2\right)\Eb(|\ve_i'|^2|\Xv_i')\right) \\
&\le C_\alpha \Eb(W_n(\Xv^i, {\Xv'}^{i'})^2).
\end{align*}
This entails that
\begin{align} \label{eq1-1}
\Uc_{n,1}^{(1)} \le C_\alpha \cdot \Omega_{n,2}. 
\end{align}
For the term $\Uc_{n,1}^{(2)}$, we claim that
\begin{align} \label{eq1-2}
\begin{aligned}
\Uc_{n,1}^{(2)} &\le C_\alpha \cdot \Omega_{n,3}. 
\end{aligned}
\end{align}
A proof of this claim is deferred to Section~\ref{subsubsec:supple-claim-eq1-2}.
From Lemma~\ref{lem:bound-Mve}, we obtain
\begin{align} \label{eq1-3}
\begin{aligned}
\Uc_{n,1}^{(3)} &\le C_\alpha M_\ve \Eb\left(\max_{i'\in[n]}\Eb\left(\sumin W_n(\Xv^i, {\Xv'}^{i'})^2 \Big|\Vb_n'\right)^{\frac{1}{2}}\right) \\
&\le C_\alpha M_\ve \sup_{\xv\in[0,1]^d}\left(\sumin\Eb\left( W_n(\Xv^i, \xv)^2\right)\right)^{\frac{1}{2}} \\
&\le C_\alpha \cdot \Omega_{n,4}.
\end{aligned}
\end{align}
Here, we have used $\Eb((\ve^i)^2 |\Xv^i)\le C_\alpha$. We may derive that
\begin{align} \label{eq1-4}
\begin{aligned}
\Uc_{n,1}^{(4)} &\le C_\alpha B_{n,W} (\log n)^{\frac{2}{\alpha}} = C_\alpha\cdot (\log n)^{-\frac{1}{\alpha^*}}\Omega_{n,1}. 
\end{aligned}
\end{align}
Combining \eqref{eq1-1}, \eqref{eq1-2}, \eqref{eq1-3}, and \eqref{eq1-4}, we conclude that
\begin{align} \label{eq1}
\norm{\Uc_{n,1}}_\ell \le C_\alpha\left(\ell^2 (\log n)^{-\frac{1}{\alpha^*}}\Omega_{n,1} + \ell^{\frac{1}{2}}\Omega_{n,2}+ \ell \Omega_{n,3} + \ell^{\frac{3}{2}}\Omega_{n,4} \right).
\end{align}

Next, we analyze the term $\Uc_{n,2}$. Define
\begin{align*}
g_i(\Xv^i, \Vb_n') := \sum_{i'=1, \ne i}^n W_n(\Xv^i, {\Xv'}^{i'}) T_{i',1}' w_{i'}',
\end{align*}
so that we have $\Uc_{n,2} = \sumin w_i T_{i,2} g_i(\Xv^i, \Vb_n')$. Since
\begin{align} \label{prob-bound}
\Pb\left(\max_{k \le n} \left| \sum_{i=1}^k w_i T_{i,2} g_i(\Xv^i, \Vb_n') \right| > 0 \Big| \Xb_n, \Vb_n'\right) \le \Pb\left(\max_{i \in [n]} |\ve^i| > M_\ve \Big| \Xb_n\right) \le \frac{1}{8},
\end{align}
an application of Lemma~\ref{lem:prop6.8ledoux} yields
\begin{align*}
\Eb(|\Uc_{n,2}| | \Xb_n, \Vb_n') &\le 8 \Eb\left(\max_{i \in [n]} |w_i T_{i,2} g_i(\Xv^i, \Vb_n')| \Big| \Xb_n, \Vb_n'\right) \\
&\le 8 \Eb\left(\max_{i \in [n]} |\ve_i| \Big| \Xb_n\right) \max_{i \in [n]} |g_i(\Xv^i, \Vb_n')| \\
&\le M_\ve \max_{i \in [n]} |g_i(\Xv^i, \Vb_n')|.
\end{align*}
Hence, by Lemma~\ref{lem:prop6.21ledoux}, it follows that for $0 < \alpha \le 1$,
\begin{align*}
\norm{\Uc_{n,2}}_{\Phi_\alpha | \Xb_n, \Vb_n'} &\le C_\alpha \left( M_\ve \max_{i \in [n]} |g_i(\Xv^i, \Vb_n')| + \normbig{\max_{i \in [n]} |w_i T_{i,2} g_i(\Xv^i, \Vb_n')|}_{\Phi_\alpha | \Xb_n, \Vb_n'} \right) \\
&\le C_\alpha \left( M_\ve \max_{i \in [n]} |g_i(\Xv^i, \Vb_n')| + \normbig{\max_{i \in [n]} |\ve_i|}_{\Phi_\alpha | \Xb_n} \max_{i \in [n]} |g_i(\Xv^i, \Vb_n')| \right) \\
&\le C_\alpha (\log n)^{\frac{1}{\alpha}} \max_{i \in [n]} |g_i(\Xv^i, \Vb_n')|,
\end{align*}
where the last inequality uses Lemma~\ref{lem:bound-Mve}. Also, for $\alpha > 1$, we claim
\begin{align} \label{claim-eq2}
\norm{\Uc_{n,2}}_{\Phi_{\alpha^*} | \Xb_n, \Vb_n'} \le C_\alpha (\log n)^{\frac{1}{\alpha}} \max_{i \in [n]} |g_i(\Xv^i, \Vb_n')|,
\end{align}
where $\alpha^* = \alpha\wedge 1$. The proof of claim is deffered to Section~\ref{subsubsec:claim-eq2}. Then, a straightforward calculation gives
\begin{align*}
\Eb\left(|\Uc_{n,2}|^\ell | \Xb_n, \Vb_n'\right) \le C_\alpha^\ell  \ell^{\frac{\ell}{\alpha^*}} (\log n)^{\frac{\ell}{\alpha}} \max_{i \in [n]} |g_i(\Xv^i, \Vb_n')|^\ell, \quad \ell \ge 2,
\end{align*}
and thus
\begin{align} \label{eq2-1}
\Eb\left(|\Uc_{n,2}|^\ell\right) \le C_\alpha^\ell \ell^{\frac{\ell}{\alpha^*}} (\log n)^{\frac{\ell}{\alpha}} \Eb\left(\max_{i \in [n]} |g_i(\Xv^i, \Vb_n')|^\ell\right), \quad \ell \ge 2.
\end{align}
It remains to bound $\Eb(\max_{i \in [n]} |g_i(\Xv^i, \Vb_n')|^\ell)$. To this end, note that $g_i(\Xv^i, \Vb_n')$ is a sum of independent, mean-zero random variables with uniform bound $B_{n,W} M_\ve$, and variance bound given by
\begin{align*}
\Var(g_i(\Xv^i, \Vb_n')) &= \sum_{i'=1, \ne i}^n \Eb\left(W_n(\Xv^i, {\Xv'}^{i'})^2 (T_{i',1}')^2\right) \\
&\le \left(\sup_{\xv\in[0,1]^d}\Var(\ve|\Xv=\xv)\right) \cdot  \sup_{\xv \in [0,1]^d} \left( \sum_{i'=1, \ne i}^n \Eb\left(W_n(\Xv^i, {\Xv'}^{i'})^2\right) \right).
\end{align*}
Note that $\sup_{\xv\in[0,1]^d}\Var(\ve|\Xv=\xv)\le C_\alpha$. Since the right-hand side does not depend on $i$ and uniformly bounded, define
\begin{align*}
\Wc_n := \left(\sup_{\xv\in[0,1]^d}\Var(\ve|\Xv=\xv)\right)^{\frac{1}{2}} \cdot   \left(\sup_{\xv \in [0,1]^d} \left( \sum_{i'=1, \ne i}^n \Eb\left(W_n(\xv, {\Xv'}^{i'})^2\right) \right)^{\frac{1}{2}}\right).
\end{align*}
From Bernstein’s inequality, we obtain
\begin{align*}
\Pb\left(|g_i(\Xv^i, \Vb_n')| \ge \frac{2 B_{n,W} M_\ve}{3} t + \Wc_n \sqrt{t}\right) \le 2 \exp(-t).
\end{align*}
For $L > 0$, define
\begin{align*}
\Psi_L(x) := \exp\left\{\left(\frac{\sqrt{1 + 2Lx} - 1}{L}\right)^2\right\} - 1,
\end{align*}
and let $\|\cdot\|_{\Psi_L}$ denote the associated Bernstein-Orlicz norm. For more details on Bernstein-Orlicz norm, refer to \cite{geer2013berns}. Then, by Lemma~2 of \cite{geer2013berns}, it follows that
\begin{align*}
\max_{1\le i\le n}\|g_i(\Xv^i, \Vb_n')\|_{\Psi_{\sqrt{3}L_n}} \le \sqrt{3} \Wc_n ,
\end{align*}
where 
\begin{align*}
L_n = \frac{4B_{n,W}M_\ve}{3\Wc_n}. 
\end{align*}
From Lemma~4 in \cite{geer2013berns}, we deduce that
\begin{align*}
&\Pb\left(\max_{i \in [n]} |g_i(\Xv^i, \Vb_n')| - \Wc_n \sqrt{3 \log(n+1)} - 2 B_{n,W} M_\ve \log(n+1) \ge \Wc_n \sqrt{3t} + 2 B_{n,W} M_\ve t \right) \\
&\le 2 \exp(-t), \quad t > 0.
\end{align*}
Using this inequality, it follows that
\begin{align*}
\Eb\left(\max_{i \in [n]} |g_i(\Xv^i, \Vb_n')|^\ell\right)
&= \int_0^\infty \Pb\left(\max_{i \in [n]} |g_i(\Xv^i, \Vb_n')| \ge t^{\frac{1}{\ell}} \right) \, \rmd t \\
&\le C_\alpha^\ell \left(\Wc_n^\ell (\log n)^{\frac{\ell}{2}} + (B_{n,W} M_\ve)^\ell (\log n)^\ell + \ell^{\frac{\ell}{2}} \Wc_n^\ell + \ell^\ell (B_{n,W} M_\ve)^\ell \right).
\end{align*}
Substituting this bound into \eqref{eq2-1} and recalling that
\begin{align*}
\Wc_n (\log n)^{1/\alpha} \le C_\alpha \cdot \Omega_{n,4} \quad \text{and} \quad M_\ve \le C_\alpha \cdot (\log n)^{\frac{1}{\alpha}},
\end{align*}
we conclude that
\begin{align} \label{eq2}
\norm{\Uc_{n,2}}_\ell \le C_\alpha \left( \ell^{1 + \frac{1}{\alpha^*}} (\log n)^{-\frac{1}{\alpha^*}} \Omega_{n,1} + \ell^{\frac{1}{2} + \frac{1}{\alpha^*}} \Omega_{n,4} + \ell^{\frac{1}{\alpha^*}} \Omega_{n,5} \right), \quad \ell\ge 2. 
\end{align}

We note that the bound for $\norm{\Uc_{n,3}}_\ell$ coincides with that of $\norm{\Uc_{n,2}}_\ell$ due to symmetry. Let
\begin{align*}
g_i^\star(\Xv^i, \Vb_n) := \sum_{i'=1,\ne i}^n W_n(\Xv^i, {\Xv'}^{i'})T_{i',2}'w_{i'}'. 
\end{align*}
For the term $\Uc_{n,4}$, using an argument analogous to that leading to \eqref{eq2-1}, we observe that
\begin{align*}
\Eb\left(|\Uc_{n,4}|^\ell\right) \le C_\alpha^\ell \ell^{\frac{\ell}{\alpha^*}}(\log n)^{\frac{\ell}{\alpha}}\Eb\left(\max_{i\in[n]}|g_i^\star(\Xv^i, \Vb_n')|^\ell\right), \quad \ell\ge 2. 
\end{align*}
Therefore, it suffices to analyze the term $\Eb(\max_{i\in[n]}|g_i^\star(\Xv^i, \Vb_n')|^\ell)$. Since
\begin{align*}
\Pb\left(\max_{k\le n}\left|\sum_{i'=1}^k W_n(\Xv^i, {\Xv'}^{i'})T_{i',2}'w_{i'}'\right|>0 \Big|\Xb_n, \Xb_n'\right) \le \frac{1}{8}
\end{align*}
as in \eqref{prob-bound}, where we put $W_n(\Xv^i, {\Xv'}^i)=0$ in the above inequality, an application of Lemma~\ref{lem:prop6.8ledoux} yields
\begin{align*}
\Eb\left(|g_i^\star(\Xv^i, \Vb_n')| | \Xb_n, \Xb_n'\right) &\le 8\Eb\left(\max_{i'\in[n]}\left|W_n(\Xv^i, {\Xv'}^{i'})T_{i',2}'\right|\Big| \Xb_n, \Xb_n'\right) \\
&\le B_{n,W}M_\ve. 
\end{align*}
Combining this with Lemma~\ref{lem:prop6.21ledoux}, we may obtain
\begin{align*}
\norm{g_i^\star(\Xv^i, \Vb_n')}_{\Phi_{\alpha^*}|\Xb_n, \Xb_n'} &\le C_\alpha\left(B_{n,W}M_\ve + B_{n,W}(\log n)^{\frac{1}{\alpha}}\right) \\
&\le C_\alpha B_{n,W}(\log n)^{\frac{1}{\alpha}}.
\end{align*}
By the arguments regarding maximal inequality as in Lemma~\ref{lem:bound-Mve}, we get
\begin{align*}
\normbig{\max_{i\in[n]}|g_i^\star(\Xv^i, \Vb_n')|}_{\Phi_{\alpha^*}|\Xb_n, \Xb_n'} &\le C_\alpha B_{n,W}(\log n)^{\frac{1}{\alpha^*}+\frac{1}{\alpha}}.
\end{align*}
Using the preceding bound, we deduce that
\begin{align*}
\Eb\left(\max_{i\in[n]}|g_i^\star(\Xv^i, \Vb_n')|^\ell\right) \le C_\alpha^\ell \ell^{\frac{\ell}{\alpha^*}}B_{n,W}^\ell (\log n)^{\frac{\ell}{\alpha^*}+\frac{\ell}{\alpha}}.
\end{align*}
Consequently, we conclude that
\begin{align} \label{eq4}
\norm{\Uc_{n,4}}_\ell &\le C_\alpha\cdot \ell^{\frac{2}{\alpha^*}} B_{n,W}(\log n)^{\frac{1}{\alpha^*}+\frac{2\ell}{\alpha}} \le \ell^{\frac{2}{\alpha^*}}\Omega_{n,1}.
\end{align}
Combining the bounds in \eqref{eq1}, \eqref{eq2}, and \eqref{eq4}, the theorem follows.

\begin{remark} \label{remark:thms1}
The main distinction between our Theorem~\ref{thm:u-stat} and Theorem~1 in \cite{chakrabortty2018tail} lies in the treatment of the term $\Uc_{n,4}$. For this analysis, \cite{chakrabortty2018tail} invoked Theorems 6.8 and 6.21 from \cite{ledoux2011prob} simultaneously. However, we observe that their argument contains a logical gap. Upon correcting this issue, we obtain a slightly looser bound than that in \cite{chakrabortty2018tail}, though it remains optimal up to a logarithmic factor.
\end{remark}

\subsubsection{Proof of \eqref{eq1-2}} \label{subsubsec:supple-claim-eq1-2}

Given a sequence of bounded measurable functions $(\mathfrak{g}_i : i \in [n])$, we have
\begin{align} \label{duality-argument}
\sup\left\{\sumin \Eb(\eta_i(V_i)\mathfrak{g}_i(V_i)) : \sumin \Eb(\eta_i(V_i))^2 \le 1 \right\} = \Eb\left(\sumin \mathfrak{g}_i(V_i)^2\right)^{\frac{1}{2}}.
\end{align}
If $\Eb(\sumin \mathfrak{g}_i(V_i)^2) = 0$, then the claim holds trivially. Otherwise, applying Hölder’s inequality yields
\begin{align*}
\sumin \Eb(\eta_i(V_i)\mathfrak{g}_i(V_i)) \le \Eb\left(\sumin \eta_i(V_i)^2\right)^{\frac{1}{2}} \Eb\left(\sumin \mathfrak{g}_i(V_i)^2\right)^{\frac{1}{2}} \le \Eb\left(\sumin \mathfrak{g}_i(V_i)^2\right)^{\frac{1}{2}}.
\end{align*}
For the reverse inequality, we may set
\begin{align*}
\eta_i(V_i) = \mathfrak{g}_i(V_i) \cdot \Eb\left(\sumin \mathfrak{g}_i(V_i)^2\right)^{-\frac{1}{2}}.
\end{align*}
We establish \eqref{eq1-2} by using the duality argument, where \textit{duality} often refers to the identity given in \eqref{duality-argument}.

Define
\begin{align*}
G_i(V_i; \Vb_n') := \sum_{i'=1, \ne i}^n w_i T_{i,1} W_n(\Xv^i, {\Xv'}^{i'}) T_{i',1}' w_{i'}'.
\end{align*}
Then, for any sequences $(\eta_i : i \in [n])$ and $(\zeta_i : i \in [n])$ satisfying $\sumin \Eb(\eta_i(V_i)^2) \le 1$ and $\sumin \Eb(\zeta_i(V_i')^2) \le 1$, it holds that
\begin{align*}
\sumiiprime \Eb\left(\eta_i(V_i) w_i T_{i,1} W_n(\Xv^i, {\Xv'}^{i'}) T_{i',1}' w_{i'}' \zeta_{i'}(V_{i'})\right) &\le \sumin \Eb\left(\eta_i(V_i) \Eb\left(G_i(V_i; \Vb_n')^2 \mid \Vb_n\right)^{\frac{1}{2}}\right) \\
&\le \Eb\left(\sumin \Eb\left(G_i(V_i; \Vb_n')^2 \mid \Vb_n\right)\right)^{\frac{1}{2}},
\end{align*}
where each inequality follows by an application of H\"older inequality. Combined with a corresponding reverse inequality argument, as in \eqref{duality-argument}, we obtain
\begin{align*}
\Uc_{n,1}^{(2)} = \Eb\left(\sumin \Eb\left(G_i(V_i; \Vb_n')^2 \mid \Vb_n\right)\right)^{\frac{1}{2}} \le C_\alpha \Eb\left(\sumin \Eb\left(W_n(\Xv^i, {\Xv'}^{i'})^2 \mid \Vb_n\right)\right)^{\frac{1}{2}} = C_\alpha \cdot \Omega_{n,3}.
\end{align*}
For the last equality, we once again used the duality argument.

\subsubsection{Proof of \eqref{claim-eq2}} \label{subsubsec:claim-eq2}

Fix $\alpha > 1$. Applying Lemma~\ref{lem:prop6.21ledoux} with $\alpha^* = \alpha \wedge 1 = 1$, we obtain
\begin{align*}
\norm{\Uc_{n,2}}_{\Phi_{\alpha^*}} \le C_1 \left( M_\ve \max_{i \in [n]} |g_i(\Xv^i, \Vb_n')| + \normbig{\max_{i \in [n]} |\ve^i|}_{\Phi_1 | \Xb_n} \max_{i \in [n]} |g_i(\Xv^i, \Vb_n')| \right),
\end{align*}
for some absolute constant $0 < C_1 < \infty$. Observe that, for any $0 < C < \infty$,
\begin{align*}
\Eb\left(\exp\left(\frac{\max_{i \in [n]} |\ve^i|}{C}\right) \Big| \Xb_n\right)
&\le \Eb\left(\exp\left(\frac{\max_{i \in [n]} |\ve^i|}{C}\right) I\left(\max_{i \in [n]} |\ve^i| \le C\right) \Big| \Xb_n\right) \\
&\quad + \Eb\left(\exp\left(\frac{\max_{i \in [n]} |\ve^i|^\alpha}{C^\alpha}\right) I\left(\max_{i \in [n]} |\ve^i| > C\right) \Big| \Xb_n\right) \\
&\le \exp\left(1\right) + \Eb\left(\exp\left(\frac{\max_{i \in [n]} |\ve^i|^\alpha}{C^\alpha} \right) \Big| \Xb_n\right),
\end{align*}
which implies that
\begin{align*}
\normbig{\max_{i \in [n]} |\ve^i|}_{\Phi_1 | \Xb_n} \le C_\alpha \normbig{\max_{i \in [n]} |\ve^i|}_{\Phi_\alpha | \Xb_n}.
\end{align*}
Combining this relation with the argument previously used for $0 < \alpha \le 1$, we conclude the proof of \eqref{claim-eq2}.

\subsubsection{Proof of Lemma~\ref{thm:u-stat}} \label{subsubsec:supple-pf-lem-u-stat}
We sketch the proof. 
Applying Theorem~3.1.1 in \cite{Pena1999decoupling}, we obtain that for all $\ell \ge 1$,
\begin{align*}
\Eb\left(\left|\sumiiprime \Wb(Z_i, Z_{i'})\right|^\ell\right)^{\frac{1}{\ell}} \le 12\Eb\left(\left|\sumiiprime \Wb(Z_i, Z_{i'}')\right|^\ell\right)^{\frac{1}{\ell}},
\end{align*}
where $(Z_i' : i \in [n])$ are i.i.d. copies of $Z = (\Xv,\ve)$ that are independent of $(Z_i : i \in [n])$. For any $\ell \ge 1$, we observe that
\begin{align*}
\Eb(|\ve|^\ell | \Xv) &= \int_0^1 \Pb(|\ve| \ge t^{1/\ell} | \Xv)\, \rmd t \le \frac{2\ell}{\alpha} C_\ve^{-\ell} \Gamma\left(\frac{\ell}{\alpha}\right) < \infty, \quad \text{a.s}.
\end{align*}
Moreover, since $\Wb$ is symmetric, the argument in the proof of Theorem~3.5.2 in \cite{Pena1999decoupling} yields
\begin{align*}
\Eb\left(\left|\sumiiprime \Wb(Z_i, Z_{i'}')\right|^\ell\right)^{\frac{1}{\ell}} = 4 \Eb\left(\left|\sumiiprime w_i \Wb(Z_i, Z_{i'}') w_{i'}'\right|^\ell\right)^{\frac{1}{\ell}}.
\end{align*}
This completes the proof.

\subsubsection{Proof of Lemma~\ref{lem:bound-Mve}} \label{subsubsec:supple-pf-lem-bound-Mve}

Define the function $\Phi_\alpha^*(x) := \exp(x^\alpha/C_\ve^\alpha) - 1$. When $\alpha \ge 1$, the function $\Phi_\alpha^*$ is convex. Hence, by Jensen’s inequality, we have
\begin{align*}
\Phi_\alpha^*\left(\Eb\left(\max_{i\in[n]}|\ve^i| \Big|\Xb_n\right)\right) &\le \Eb\left(\Phi_\alpha^*\left(\max_{i\in[n]}|\ve^i|\right)\Big|\Xb_n\right) \\
&\le \Eb\left(\sumin \Phi_\alpha^*(|\ve^i|)\Big|\Xb_n\right) \\
&\le 2n.
\end{align*}
Since $(\Phi_\alpha^*)^{-1}(x) = C_\ve (\log(x+1))^{\frac{1}{\alpha}}$, it follows that
\begin{align*}
\Eb\left(\max_{i\in[n]}|\ve^i| \Big|\Xb_n\right) \le C_\ve (\log 2n)^{\frac{1}{\alpha}},
\end{align*}
which completes the proof of the first assertion of the lemma when $\alpha \ge 1$.

If $0 < \alpha \le 1$, the function $\Phi_\alpha^*$ is no longer convex. In this case, applying Theorem~3.1 of \cite{arun2022moving} in conjunction with the argument in the proof of Lemma~3 of \cite{geer2013berns}, we obtain
\begin{align*}
\Eb\left(\max_{i\in[n]}|\ve^i| \Big|\Xb_n\right) \le C_\alpha\left(\sqrt{\log(n+1)} + (\log(n+1))^{\frac{1}{\alpha}}\right) \le C_\alpha(\log n)^{\frac{1}{\alpha}},
\end{align*}
where last inequality follows as $\alpha<1$. 

We prove a more general version of the second assertion in the lemma. For i.i.d. random variables $\{V_i\}_{i=1}^n$ with $\norm{V_i}_{\Phi_\alpha}=C$ for some $0<C<\infty$, we have  
\begin{align*}
\Eb\left(\exp\left(\frac{\max_{i\in[n]}|V_i|^\alpha}{C^\alpha}\right)\right) \le \Eb\left(\sumin \exp\left(\frac{|V_i|^\alpha}{C^\alpha}\right)\right) \le 2n.
\end{align*}
Let $C':=(\frac{\log 2n}{2})^{\frac{1}{\alpha}}C$. Then, by Jensen's inequality,  
\begin{align*}
\Eb\left(\exp\left(\frac{\max_{i\in[n]}|V_i|^\alpha}{(C')^\alpha}\right)\right) &= \Eb\left(\exp\left(\frac{\max_{i\in[n]}|V_i|^\alpha}{C^\alpha \cdot \frac{\log 2n}{\log 2}}\right)\right) \\
&= \Eb\left(\exp\left(\frac{\max_{i\in[n]}|V_i|^\alpha}{C^\alpha}\right)\right)^{\frac{\log 2}{\log 2n}} \\
&= 2,
\end{align*}
which implies that $\norm{\max_{i\in[n]}|V_i|}_{\Phi_\alpha} \le C_\alpha (\log n)^{1/\alpha}$. This completes the proof of the second assertion in the lemma.

\subsection{Technical Proofs for Section~\ref{sec:ll}}
 
This section presents the technical details supporting the results in Section~\ref{sec:ll}. Throughout the proofs, all (in)equalities are understood to hold either almost surely or with probability tending to one. We often use the notations $C_\ell$ for $\ell\in\Nb$ to denote (absolute) constants, whose values may change from line to line.

\subsubsection{Proof of Lemma~\ref{lem:ll-bound-Delta}}

From Lemma~\ref{lem:tech1}, we may verify that 
\begin{align} \label{lower-eigen-lem1}
\min_{j\in[d]}\inf_{x_j\in[0,1]}\lambda_\rmmin(\hM_\bfsnum{0}{jj}(x_j)) > 0
\end{align}
holds with probability tending to one. In what follows, we frequently make use of \eqref{lower-eigen-lem1} without explicitly mentioning it in the proofs of the claims.
In addition, applying the same lemma, we deduce that there exists an absolute constant $0<C_1<\infty$ such that 
\begin{align*}
\norm{\Delta_\bfsnum{0}{j}^\rmtp}_\hMnum{0}^2 \le C_1 \norm{\Delta_\bfsnum{0}{j}^\rmtp}_\Mnum{0}^2
\end{align*}
holds with probability tending to one. Since the constant $C_1$ does not depend on the index $j$, it suffices to establish that 
\begin{align} \label{what-to-show-lem1}
\max_{j\in[d]}\norm{\Delta_\bfsnum{0}{j}^\rmtp}_\Mnum{0}^2 \lesssim |\cS_\bfnum{0}|^2h_\bfnum{0}^4 + \frac{1}{n_\bfnum{0}h_\bfnum{0}} + A(n_\bfnum{0}, h_\bfnum{0}, d;\alpha). 
\end{align}
To this end, observe that
\begin{align*}
U_j\cdot \Delta_\bfsnum{0}{j}^\rmtp(x_j) &= U_j \cdot \left(\hatm_\bfsnum{0}{j}^\rmtp(x_j) - \hPi_\bfsnum{0}{j}(f_\bfnum{0}^\rmtp)(x_j)\right) \\
&= \hM_\bfsnum{0}{jj}(x_j)^{-1}  \Biggr\{ \frac{1}{n_\bfnum{0}}\suminum{0} Z_\bfsnum{0}{j}^i(x_j) K_{h_\bfsnum{0}{j}}(x_j, X_\bfsnum{0}{j}^i)\ve_\bfsnum{0}{i} \\
&\quad + \frac{1}{n_\bfnum{0}}\suminum{0} Z_\bfsnum{0}{j}^i(x_j) K_{h_\bfsnum{0}{j}}(x_j, X_\bfsnum{0}{j}^i)\left(f_\bfsnum{0}{j}(X_\bfsnum{0}{j}^i) - Z_\bfsnum{0}{j}^i(x_j)^\top f_\bfsnum{0}{j}(x_j)\right) \\
&\quad + \frac{1}{n_\bfnum{0}}\suminum{0} Z_\bfsnum{0}{j}^i(x_j) K_{h_\bfsnum{0}{j}}(x_j, X_\bfsnum{0}{j}^i) \\
&\hspace{2cm} \times \left(\int_0^1 \left(f_\bfsnum{0}{k}(X_\bfsnum{0}{k}^i) - Z_\bfsnum{0}{k}^i(x_k)^\top f_\bfsnum{0}{k}(x_k)\right)K_{h_\bfsnum{0}{k}}(x_k, X_\bfsnum{0}{k}^i)\dxk\right) \Biggr\} \\
&\overset{\rm let}{=:} \hatm_j^{A,\rmv}(x_j) + \hatm_j^{B,\rmv}(x_j) + \hatm_j^{C,\rmv}(x_j).
\end{align*}
We claim the following stochastic bounds:
\begin{align}
\max_{j\in[d]}\norm{\hatm_j^{A,\rmv}}_\Mnum{0}^2 &\lesssim \frac{1}{n_\bfnum{0}h_\bfnum{0}} + A(n_\bfnum{0},h_\bfnum{0},d;\alpha), \label{claim-2-lem1} \\
\max_{j\in[d]}\norm{\hatm_j^{B,\rmv}}_\Mnum{0}^2 &\lesssim h_\bfnum{0}^4, \label{claim-3-lem1} \\
\max_{j\in[d]}\norm{\hatm_j^{C,\rmv}}_\Mnum{0}^2 &\lesssim |\cS_\bfnum{0}|^2h_\bfnum{0}^4. \label{claim-4-lem1} 
\end{align}
It is evident that claims \eqref{claim-2-lem1}--\eqref{claim-4-lem1} together imply the lemma.

We note that \eqref{claim-2-lem1} is a direct consequence of Lemma~\ref{lem:u-stat1}, since \eqref{lower-eigen-lem1} holds with probability tending to one. We now outline the proof of \eqref{claim-3-lem1}. To establish \eqref{claim-3-lem1}, we observe that 
\begin{align*}
&\frac{1}{n_\bfnum{0}}\suminum{0} Z_\bfsnum{0}{j}^i(x_j) K_{h_\bfsnum{0}{j}}(x_j, X_\bfsnum{0}{j}^i)\left(f_\bfsnum{0}{j}(X_\bfsnum{0}{j}^i) - Z_\bfsnum{0}{j}^i(x_j)^\top f_\bfsnum{0}{j}^\rmv(x_j)\right) \\
&= \frac{h_\bfsnum{0}{j}^2}{2}\frac{1}{n_\bfnum{0}}\suminum{0} Z_\bfsnum{0}{j}^i(x_j) K_{h_\bfsnum{0}{j}}(x_j, X_\bfsnum{0}{j}^i)f_\bfsnum{0}{j}''(x_j) + \frac{1}{n_\bfnum{0}}\suminum{0} Z_\bfsnum{0}{j}^i(x_j) K_{h_\bfsnum{0}{j}}(x_j, X_\bfsnum{0}{j}^i)r_j(x_j),
\end{align*}
for some stochastic function $r_j : [0,1] \to \Rb$ satisfying $\max_{j\in[d]}\sup_{x_j\in[0,1]}|r_j(x_j)| = o_p(h_\bfnum{0}^2)$. Combining this with standard results from kernel smoothing theory yields \eqref{claim-3-lem1}.
The proof of \eqref{claim-4-lem1} is essentially identical to that of \eqref{claim-3-lem1}, and is therefore omitted.

\subsubsection{Proof of Theorem~\ref{thm:ll-bound-error-emp}}

We first argue that the deviance term arising from $\bar Y_\bfnum{0} - \Eb(Y_\bfnum{0})$ is negligible compared to the other terms. That is,
\begin{align}
\norm{(\bar Y_\bfnum{0}-\Eb(Y_\bfnum{0}),0_d^\top)^\top}_\Mnum{0}^2 \lesssim |\cS_\bfnum{0}|^2\frac{\log n_\bfnum{0}}{n_\bfnum{0}}\lesssim |\cS_\bfnum{0}|^2h_\bfnum{0}^4\lesssim \frac{1}{n_\bfnum{0}h_\bfnum{0}} + A(n_\bfnum{0},h_\bfnum{0},d;\alpha), \label{claim-1-lem1}
\end{align}
where the last inequality follows from the order condition imposed on $|\cS_\bfnum{0}|$. We note that although the upper bound in \eqref{claim-1-lem1} can be improved, the stated form suffices for our purpose. Specifically, we may substitute $\log n_\bfnum{0}$ in the above bound with a function of $n_\bfnum{0}$ that diverges to infinity as $n_\bfnum{0} \to \infty$. 
To see this, observe that
\begin{align*}
\Pb\left(|\bar Y_\bfnum{0} - \Eb(Y_\bfnum{0})| \ge C_1(|\cS_\bfnum{0}|+1)\sqrt{\frac{\log n_\bfnum{0}}{n_\bfnum{0}}}\right) &\le \Pb\left(\left|\frac{1}{n_\bfnum{0}}\suminum{0} \ve_\bfnum{0}^{i}\right| \ge C_1\sqrt{\frac{\log n_\bfnum{0}}{n_\bfnum{0}}}\right) \\
&\quad + \sumjs\Pb\left(\left|\frac{1}{n_\bfnum{0}}\sumin f_\bfsnum{0}{j}(X_\bfsnum{0}{j}^i)\right| \ge C_1\sqrt{\frac{\log n_\bfnum{0}}{n_\bfnum{0}}}\right).
\end{align*}
By Markov's inequality, we obtain
\begin{align*}
\Pb\left(\left|\frac{1}{n_\bfnum{0}}\suminum{0} \ve_\bfnum{0}^{i}\right| \ge C_1\sqrt{\frac{\log n_\bfnum{0}}{n_\bfnum{0}}}\right) \le \frac{\Var(\ve_\bfnum{0}^{1})}{C_1^2\log n_\bfnum{0}} \lesssim (\log n_\bfnum{0})^{-1} = o(1),
\end{align*}
where the last equality follows from the order condition on $h_\bfnum{0}$ specified in condition (B-$\alpha$). Here, we have used the fact that
\begin{align*}
\Var(\ve_\bfnum{0}{1}) = \Eb((\ve_\bfnum{0}^{1})^2) = \int_0^1 \Pb(|\ve_\bfnum{0}^{1}|\ge t^{\frac{1}{2}})\, \rmd t \le \frac{4}{\alpha}\Gamma\left(\frac{2}{\alpha}\right)C_\ve^2,
\end{align*}
which follows from condition (R-$\alpha$) imposed on the error term $\ve_\bfnum{0}$.
Since $|f_\bfsnum{0}{j}(X_\bfsnum{0}{j}^i)|\le C_{f,0}$ almost surely, applying Bernstein's inequality, we further obtain
\begin{align*}
\Pb\left(\left|\frac{1}{n_\bfnum{0}}\sumin f_\bfsnum{0}{j}(X_\bfsnum{0}{j}^i)\right| \ge C_1\sqrt{\frac{\log n_\bfnum{0}}{n_\bfnum{0}}}\right) \le 2\exp\left(-\frac{C_1^2 \log n_\bfnum{0}}{2C_{f,0}^2 + \frac{2}{3}C_{f,0}C_1}\right),
\end{align*}
provided that $n_\bfnum{0}$ is sufficiently large such that $\frac{\log n_\bfnum{0}}{n_\bfnum{0}}\le 1$. This implies
\begin{align*}
\sumjs\Pb\left(\left|\frac{1}{n_\bfnum{0}}\sumin f_\bfsnum{0}{j}(X_\bfsnum{0}{j}^i)\right| \ge C_1h_\bfnum{0}^2\right) \le 2\exp\left(\log \left(\frac{|\cS_\bfnum{0}|}{2}\right)-\frac{C_1^2 \log n_\bfnum{0}}{2C_{f,0}^2 + \frac{2}{3}C_{f,0}C_1}\right) = o(1),
\end{align*}
since $|\cS_\bfnum{0}|\ll n_\bfnum{0}$, as stated in the assumptions of the theorem. This completes the proof of \eqref{claim-1-lem1}.
Based on this observation, without loss of generality, we henceforth treat $\bar Y_\bfnum{0}$ as $\Eb(Y_\bfnum{0})$.

Let $\alpha_\bfsnum{0}{j}^\rmtp := \hatf_\bfsnum{0}{j}^\rmtp - f_\bfsnum{0}{j}^\rmtp$ and $\alpha_\bfnum{0}^\rmtp := \hatf_\bfnum{0}^\rmtp - f_\bfnum{0}^\rmtp$. Recall that the penalized loss functional $\hL_\bfnum{0}^\rmpen$ is defined as
\begin{align} \label{pen-loss-thm1}
\hL_\bfnum{0}^\rmpen(\gv^\rmtp) = \hL_\bfnum{0}(\gv^\rmtp) + \lambda_\bfnum{0}\sumj \norm{g_j^\rmtp}_\hMnum{0},
\end{align}
where $\hL_\bfnum{0}$ denotes the standard squared loss functional associated with kernel smoothing. Since $\hatfv_\bfnum{0}^\rmtp = (\hatf_\bfsnum{0}{j}^\rmtp : j \in [d])$ minimizes $\hL_\bfnum{0}^\rmpen$, it follows from \eqref{pen-loss-thm1} that
\begin{align*}
\hPi(\hatf_\bfnum{0}^\rmtp) = \hatm_\bfsnum{0}{j}^\rmtp - \lambda_\bfnum{0}\nu_\bfsnum{0}{j}^\rmtp,
\end{align*}
so that
\begin{align}
\hPi_\bfsnum{0}{j}(\alpha_\bfnum{0}^\rmtp) = \Delta_\bfsnum{0}{j}^\rmtp - \lambda_\bfnum{0}\nu_\bfsnum{0}{j}^\rmtp,
\end{align}
where $\nu_\bfsnum{0}{j}^\rmtp$ denotes a subgradient of $\norm{\cdot}_\hMnum{0}$ at $\hatf_\bfsnum{0}{j}^\rmtp$. The subgradient $\nu_\bfsnum{0}{j}^\rmtp$ is further characterized as
\begin{align*}
\nu_\bfsnum{0}{j}^\rmtp = \begin{cases}
\hatf_\bfsnum{0}{j}^\rmtp/\norm{\hatf_\bfsnum{0}{j}^\rmtp}_\hMnum{0}, &\quad \text{if }\norm{\hatf_\bfsnum{0}{j}^\rmtp}_\hMnum{0} \ne 0, \\
\text{any } v_j^\rmtp \in \sH_j^\rmtp \text{ with } \norm{v_j^\rmtp}_\hMnum{0} \le 1, &\quad \text{otherwise},
\end{cases}
\end{align*}
and satisfies
\begin{align} \label{prop-subgrad}
\ip{\nu_\bfsnum{0}{j}^\rmtp, g_j^\rmtp}_\hMnum{0} \ge \norm{\hatf_\bfsnum{0}{j}^\rmtp}_\hMnum{0} - \norm{\hatf_\bfsnum{0}{j}^\rmtp - g_j^\rmtp}_\hMnum{0}, \quad g_j \in \sH_j^\rmtp.
\end{align}
From \eqref{prop-subgrad}, we may derive that 
\begin{align} \label{subgrad-ineq}
\ip{\nu_\bfsnum{0}{j}^\rmtp, \alpha_\bfsnum{0}{j}^\rmtp}_\hMnum{0} \ge \norm{\hatf_\bfsnum{0}{j}^\rmtp}_\hMnum{0} - \norm{f_\bfsnum{0}{j}^\rmtp}_\hMnum{0} \begin{cases}
\ge -\norm{\alpha_\bfsnum{0}{j}^\rmtp}, &\quad \text{if }j\in\cS_\bfnum{0}, \\
=\norm{\alpha_\bfsnum{0}{j}^\rmtp}, &\quad \text{if }j\not\in\cS_\bfnum{0}. 
\end{cases}
\end{align}

Recall that $\Delta_\bfnum{0} = \max_{j\in[d]}\norm{\Delta_\bfsnum{0}{j}^\rmtp}_\hMnum{0}$. Applying \eqref{subgrad-ineq}, we observe that
\begin{align*}
\norm{\alpha_\bfnum{0}^\rmtp}_\hMnum{0}^2 &= \sumj \ip{\alpha_\bfnum{0}^\rmtp, \alpha_\bfsnum{0}{j}^\rmtp}_\hMnum{0} \\
&= \sumj \ip{\hPi_\bfsnum{0}{j}(\alpha_\bfnum{0}^\rmtp), \alpha_\bfsnum{0}{j}^\rmtp}_\hMnum{0} \\
&= \sumj \ip{\Delta_\bfsnum{0}{j}^\rmtp - \lambda_\bfnum{0}\nu_\bfsnum{0}{j}^\rmtp, \alpha_\bfsnum{0}{j}^\rmtp}_\hMnum{0} \\
&\le \Delta_\bfnum{0} \sumj \norm{\alpha_\bfsnum{0}{j}^\rmtp}_\hMnum{0} - \lambda_\bfnum{0} \left\{ \sumjnots \norm{\alpha_\bfsnum{0}{j}^\rmtp}_\hMnum{0} - \sumjs \norm{\alpha_\bfsnum{0}{j}^\rmtp}_\hMnum{0} \right\} \\
&\le (\lambda_\bfnum{0} + \Delta_\bfnum{0}) \sumjs \norm{\alpha_\bfsnum{0}{j}^\rmtp}_\hMnum{0} - (\lambda_\bfnum{0} - \Delta_\bfnum{0}) \sumjnots \norm{\alpha_\bfsnum{0}{j}^\rmtp}_\hMnum{0}.
\end{align*}
Since there exists a constant $\mathfrak{C}_\bfnum{0} > 1$ such that $\lambda_\bfnum{0} \ge \mathfrak{C}_\bfnum{0} \Delta_\bfnum{0}$, it follows that
\begin{align*}
\lambda_\bfnum{0}^{-1}\norm{\alpha_\bfnum{0}^\rmtp}_\hMnum{0}^2 \le \frac{\mathfrak{C}_\bfnum{0}+1}{\mathfrak{C}_\bfnum{0}} \sumjs \norm{\alpha_\bfsnum{0}{j}^\rmtp}_\hMnum{0} - \frac{\mathfrak{C}_\bfnum{0}-1}{\mathfrak{C}_\bfnum{0}} \sumjnots \norm{\alpha_\bfsnum{0}{j}^\rmtp}_\hMnum{0}.
\end{align*}
Therefore, we obtain
\begin{align} \label{main-cond1-thm1}
\sumj \norm{\alpha_\bfsnum{0}{j}^\rmtp}_\hMnum{0} \le \frac{2\mathfrak{C}_\bfnum{0}}{\mathfrak{C}_\bfnum{0}-1} \sumjs \norm{\alpha_\bfsnum{0}{j}^\rmtp}_\hMnum{0},
\end{align}
and
\begin{align} \label{main-cond2-thm1}
\lambda_\bfnum{0}^{-1}\norm{\alpha_\bfnum{0}^\rmtp}_\hMnum{0}^2 \le \frac{\mathfrak{C}_\bfnum{0}+1}{\mathfrak{C}_\bfnum{0}} \sumjs \norm{\alpha_\bfsnum{0}{j}^\rmtp}_\hMnum{0}.
\end{align}

We prove only the first assertion of the theorem using the relation in \eqref{main-cond1-thm1}. Once the first assertion is established, the second follows directly from \eqref{main-cond2-thm1}. Let $\sD_\bfnum{0} := \sumjs \norm{\alpha_\bfsnum{0}{j}^\rmtp}_\hMnum{0}$. Recall that the matrix $\tM_\bfnum{0}(\cdot)$ is defined by $\tM_\bfnum{0}(\cdot) := \Eb(\hM_\bfnum{0}(\cdot))$, and define the projection operator $\tPi_\bfsnum{0}{0}$ analogously to $\hPi_\bfsnum{0}{0}$, which projects onto $\Rb^\rmtp$ with respect to the inner product $\ip{\cdot, \cdot}_\hMnum{0}$, by replacing $\hM_\bfnum{0}$ with $\tM_\bfnum{0}$ in the definition. 
Let $\alpha_\bfsnum{0}{j}^{\rmtp, \tc} := \alpha_\bfsnum{0}{j}^\rmtp - \tPi_\bfsnum{0}{0}(\alpha_\bfsnum{0}{j}^\rmtp)$ and $\alpha_\bfnum{0}^{\rmtp, \tc}:= \sumj \alpha_\bfsnum{0}{j}^{\rmtp, \tc}$, and define $\cD_\bfnum{0} := \max_{j\in\cS_\bfnum{0}} (\max(\norm{\alpha_\bfsnum{0}{j}^\rmtp}_\hMnum{0} - \norm{\alpha_\bfsnum{0}{j}^{\rmtp, \tc}}_\hMnum{0}),0 )$. We claim that
\begin{align} \label{main-claim-thm1}
\cD_\bfnum{0} \lesssim h_\bfnum{0}^2 + \sqrt{\frac{\log (|\cS_\bfnum{0}|\vee n_\bfnum{0})}{n_\bfnum{0}}}.
\end{align}
The proof of the claim in \eqref{main-claim-thm1} is deferred to the end of the proof.
Suppose now that the claim in \eqref{main-claim-thm1} holds. Then observe that
\begin{align*}
\sD_\bfnum{0} \le \sumjs \norm{\alpha_\bfsnum{0}{j}^{\rmtp, \tc}}_\hMnum{0} + |\cS_\bfnum{0}|\cD_\bfnum{0}.
\end{align*}
We consider two cases separately: (i) $\sumjs \norm{\alpha_\bfsnum{0}{j}^{\rmtp, \tc}}_\hMnum{0} \le |\cS_\bfnum{0}|\cD_\bfnum{0}$; and (ii) $\sumjs \norm{\alpha_\bfsnum{0}{j}^{\rmtp, \tc}}_\hMnum{0} > |\cS_\bfnum{0}|\cD_\bfnum{0}$. In case (i), we obtain $\sD_\bfnum{0} \le 2|\cS_\bfnum{0}|\cD_\bfnum{0}$, which, together with the claim in \eqref{main-claim-thm1}, yields the desired conclusion.

For case (ii), observe that
\begin{align*}
\sD_\bfnum{0} \le 2 \sumjs \norm{\alpha_\bfsnum{0}{j}^{\rmtp, \tc}}_\hMnum{0}.
\end{align*}
Let $\xi_\bfnum{0} > 0$ be a sufficiently small constant such that
\begin{align} \label{xicond-thm1}
2\cdot \frac{\mathfrak{C}_\bfnum{0}+1}{\mathfrak{C}_\bfnum{0}-1} \le 2\cdot \sqrt{\frac{1+\xi_\bfnum{0}}{1-\xi_\bfnum{0}}} \cdot \frac{\mathfrak{C}_\bfnum{0}+1}{\mathfrak{C}_\bfnum{0}-1} \le C_\bfnum{0},
\end{align}
where $C_\bfnum{0}$ is the constant specified in the statement of the theorem.
Then, by Lemma~\ref{lem:tech1}, we have
\begin{align*}
1-\xi_\bfnum{0} &\le \min_{j\in[d]}\inf_{x_j\in[0,1]}\lambda_\rmmin\left(\tM_\bfsnum{0}{jj}(x_j)^{-\frac{1}{2}}\hM_\bfsnum{0}{jj}(x_j) \tM_\bfsnum{0}{jj}(x_j)^{-\frac{1}{2}}\right) \\
&\le \max_{j\in[d]}\sup_{x_j\in[0,1]}\lambda_\rmmax\left(\tM_\bfsnum{0}{jj}(x_j)^{-\frac{1}{2}}\hM_\bfsnum{0}{jj}(x_j) \tM_\bfsnum{0}{jj}(x_j)^{-\frac{1}{2}}\right) \le 1+\xi_\bfnum{0}.
\end{align*}
Using this together with \eqref{xicond-thm1} and the fact that
\begin{align*}
\norm{g_j^\rmtp}_\tMnum{0}^2 = \norm{g_j^\rmtp - \tPi_\bfsnum{0}{0}(g_j^\rmtp)}_\tMnum{0}^2 + \norm{\tPi_\bfsnum{0}{0}(g_j^\rmtp)}_\tMnum{0}^2, \quad g_j^\rmtp \in \sH_j^\rmtp,
\end{align*}
we may verify that
\begin{align*}
\sumjnots \norm{\alpha_\bfsnum{0}{j}^{\rmtp, \tc}}_\tMnum{0}
&\le \sumjnots \norm{\alpha_\bfsnum{0}{j}^\rmtp}_\tMnum{0} \\
&\le \sqrt{\frac{1}{1-\xi_\bfnum{0}}} \cdot\sumjnots \norm{\alpha_\bfsnum{0}{j}^\rmtp}_\hMnum{0} \\
&\le \sqrt{\frac{1}{1-\xi_\bfnum{0}}} \cdot \frac{\mathfrak{C}_\bfnum{0}+1}{\mathfrak{C}_\bfnum{0}-1} \cdot \sumjs \norm{\alpha_\bfsnum{0}{j}^\rmtp}_\hMnum{0} \\
&\le \sqrt{\frac{1}{1-\xi_\bfnum{0}}} \cdot \frac{\mathfrak{C}_\bfnum{0}+1}{\mathfrak{C}_\bfnum{0}-1} \cdot\left( \sumjs \norm{\alpha_\bfsnum{0}{j}^{\rmtp, \tc}}_\hMnum{0} + |\cS_\bfnum{0}|\cD_\bfnum{0} \right) \\
&\le 2 \sqrt{\frac{1}{1-\xi_\bfnum{0}}} \cdot \frac{\mathfrak{C}_\bfnum{0}+1}{\mathfrak{C}_\bfnum{0}-1} \cdot\sumjs \norm{\alpha_\bfsnum{0}{j}^{\rmtp, \tc}}_\hMnum{0} \\
&\le 2 \sqrt{\frac{1+\xi_\bfnum{0}}{1-\xi_\bfnum{0}}} \cdot \frac{\mathfrak{C}_\bfnum{0}+1}{\mathfrak{C}_\bfnum{0}-1} \cdot\sumjs \norm{\alpha_\bfsnum{0}{j}^{\rmtp, \tc}}_\tMnum{0} \\
&\le C_\bfnum{0} \sumjs \norm{\alpha_\bfsnum{0}{j}^{\rmtp, \tc}}_\tMnum{0}.
\end{align*}
From the definition of $\phi_\bfnum{0}$, it follows that
\begin{align} \label{norm-compat-thm1}
\norm{\alpha_\bfnum{0}^{\rmtp, \tc}}_\tMnum{0}^2 \ge \phi_\bfnum{0}(C_\bfnum{0}) \sumjs \norm{\alpha_\bfsnum{0}{j}^{\rmtp, \tc}}_\tMnum{0}^2.
\end{align}

From \eqref{norm-compat-thm1}, we may derive that
\begin{align} \label{sDsquare-thm1}
\begin{aligned}
\sD_\bfnum{0}^2 &\le |\cS_\bfnum{0}| \sumjs \norm{\alpha_\bfsnum{0}{j}^\rmtp}_\hMnum{0}^2 \\
&\le 2|\cS_\bfnum{0}| \left( \sumjs \norm{\alpha_\bfsnum{0}{j}^{\rmtp, \tc}}_\hMnum{0}^2 + |\cS_\bfnum{0}| \cD_\bfnum{0}^2 \right) \\
&\le 2|\cS_\bfnum{0}| (1+\xi_\bfnum{0}) \sumjs \norm{\alpha_\bfsnum{0}{j}^{\rmtp, \tc}}_\tMnum{0}^2 + 2|\cS_\bfnum{0}|^2 \cD_\bfnum{0}^2 \\
&\le 2(1+\xi_\bfnum{0}) \frac{|\cS_\bfnum{0}|}{\phi_\bfnum{0}(C_\bfnum{0})} \norm{\alpha_\bfnum{0}^{\rmtp, \tc}}_\tMnum{0}^2 + 2|\cS_\bfnum{0}|^2 \cD_\bfnum{0}^2.
\end{aligned}
\end{align}

We claim that there exists an absolute constant $0 < \mathscr{C}_\bfnum{0} < \infty$ such that
\begin{align} \label{main-claim2-thm1}
\norm{\alpha_\bfnum{0}^{\rmtp,\tc}}_\tMnum{0}^2 \le \norm{\alpha_\bfnum{0}^\rmtp}_\hMnum{0}^2 + \mathscr{C}_\bfnum{0} \left( \frac{1}{n_\bfnum{0}h_\bfnum{0}^2} + B(n_\bfnum{0}, h_\bfnum{0}^2, d) \right)^{\frac{1}{2}} \sD_\bfnum{0}^2.
\end{align}
The proof of this claim is deferred to the end of the argument.
Suppose now that the claim holds. Since $\phi_\bfnum{0}(C_\bfnum{0})$ is bounded away from zero and
\begin{align*}
|\cS_\bfnum{0}| \left( \frac{1}{n_\bfnum{0}h_\bfnum{0}^2} + B(n_\bfnum{0}, h_\bfnum{0}^2, d) \right)^{\frac{1}{2}} \ll 1,
\end{align*}
we may, without loss of generality, assume that
\begin{align} \label{clue-thm1}
2\mathscr{C}_\bfnum{0}(1+\xi_\bfnum{0}) \frac{|\cS_\bfnum{0}|}{\phi_\bfnum{0}(C_\bfnum{0})} \left( \frac{1}{n_\bfnum{0}h_\bfnum{0}^2} + B(n_\bfnum{0}, h_\bfnum{0}^2, d) \right)^{\frac{1}{2}} \le \xi_\bfnum{0}.
\end{align}
Combining \eqref{main-cond2-thm1}, \eqref{main-claim2-thm1}, and \eqref{clue-thm1} with \eqref{sDsquare-thm1}, we obtain
\begin{align*}
\sD_\bfnum{0}^2 &\le 2\frac{1+\xi_\bfnum{0}}{1-\xi_\bfnum{0}} \cdot \frac{|\cS_\bfnum{0}|}{\phi_\bfnum{0}(C_\bfnum{0})} \norm{\alpha_\bfnum{0}^\rmtp}_\hMnum{0}^2 + \frac{2}{1-\xi_\bfnum{0}}|\cS_\bfnum{0}|^2 \cD_\bfnum{0}^2 \\
&\le 2|\cS_\bfnum{0}| \cdot \frac{1+\xi_\bfnum{0}}{1-\xi_\bfnum{0}} \cdot \left( \frac{\mathfrak{C}_\bfnum{0}+1}{\mathfrak{C}_\bfnum{0}} \right) \cdot \frac{\lambda_\bfnum{0}}{\phi_\bfnum{0}(C_\bfnum{0})} \sD_\bfnum{0} + \frac{2}{1-\xi_\bfnum{0}}|\cS_\bfnum{0}|^2 \cD_\bfnum{0}^2.
\end{align*}
Finally, this implies that
\begin{align*}
\sD_\bfnum{0} \lesssim |\cS_\bfnum{0}| \left( \frac{\lambda_\bfnum{0}}{\phi_\bfnum{0}} + \cD_\bfnum{0} \right),
\end{align*}
which, together with the order condition on $\lambda_\bfnum{0}$ and the claim in \eqref{main-claim2-thm1}, completes the proof of the theorem.

It remains to prove claims~\eqref{main-claim-thm1} and~\eqref{main-claim2-thm1}, whose proofs are provided below.
 
\smallskip\noindent
{\it Proof of \eqref{main-claim-thm1}.}
\smallskip

Observe that
\begin{align*}
\norm{\alpha_\bfsnum{0}{j}^{\rmtp, \tc}}_\hMnum{0} &= \norm{\hatf_\bfsnum{0}{j}^\rmtp - f_\bfsnum{0}{j}^\rmtp - \tPi_\bfsnum{0}{0}^\rmtp(\hatf_\bfsnum{0}{j}^\rmtp - f_\bfsnum{0}{j}^\rmtp)}_\hMnum{0} \\
&= \norm{\hatf_\bfsnum{0}{j}^\rmtp - f_\bfsnum{0}{j}^\rmtp + \hPi_\bfsnum{0}{0}(f_\bfsnum{0}{j}^\rmtp) - \hPi_\bfsnum{0}{0}(f_\bfsnum{0}{j}^\rmtp) - \tPi_\bfsnum{0}{0}(\hatf_\bfsnum{0}{j}^\rmtp - f_\bfsnum{0}{j}^\rmtp)}_\hMnum{0} \\
&\ge \norm{\hatf_\bfsnum{0}{j}^\rmtp - f_\bfsnum{0}{j}^\rmtp + \hPi_\bfsnum{0}{0}(f_\bfsnum{0}{j}^\rmtp)}_\hMnum{0} \\
&\ge \norm{\alpha_\bfsnum{0}{j}^\rmtp}_\hMnum{0} - \norm{\hPi_\bfsnum{0}{0}(f_\bfsnum{0}{j}^\rmtp)}_\hMnum{0}.
\end{align*}
From this, we obtain
\begin{align*}
\norm{\alpha_\bfsnum{0}{j}^\rmtp}_\hMnum{0} - \norm{\alpha_\bfsnum{0}{j}^{\rmtp, \tc}}_\hMnum{0} \le \norm{\hPi_\bfsnum{0}{0}(f_\bfsnum{0}{j}^\rmtp)}_\hMnum{0} \le \norm{\tPi_\bfsnum{0}{0}(f_\bfsnum{0}{j}^\rmtp)} + \norm{\hPi_\bfsnum{0}{0}(f_\bfsnum{0}{j}^\rmtp) - \tPi_\bfsnum{0}{0}(f_\bfsnum{0}{j}^\rmtp)}_\hMnum{0}.
\end{align*}
We now establish the following two bounds:
\begin{align}
\max_{j\in\cS_\bfnum{0}} \norm{\tPi_\bfsnum{0}{0}(f_\bfsnum{0}{j}^\rmtp)}_\hMnum{0} &\lesssim h_\bfnum{0}^2, \label{claim-1-claim1} \\
\max_{j\in\cS_\bfnum{0}} \norm{\hPi_\bfsnum{0}{0}(f_\bfsnum{0}{j}^\rmtp) - \tPi_\bfsnum{0}{0}(f_\bfsnum{0}{j}^\rmtp)}_\hMnum{0} &\lesssim \sqrt{\frac{\log (|\cS_\bfnum{0}|\vee n_\bfnum{0})}{n_\bfnum{0}}}. \label{claim-2-claim1}
\end{align}
Clearly, combining \eqref{claim-1-claim1} and \eqref{claim-2-claim1} yields \eqref{main-claim-thm1}.

To prove \eqref{claim-1-claim1}, we note that
\begin{align*}
\norm{\tPi_\bfsnum{0}{0}(f_\bfsnum{0}{j}^\rmtp)}_\hMnum{0} &= \left| \int_0^1 f_\bfsnum{0}{j}^\rmv(x_j)^\top \tp_\bfsnum{0}{j}^\rmv(x_j) \dxj \right| \\
&= \left| \int_{[0,1]^2} \left( f_\bfsnum{0}{j}(x_j) + (u_j - x_j) f_\bfsnum{0}{j}'(x_j) \right) K_{h_\bfsnum{0}{j}}(x_j, u_j) p_\bfsnum{0}{j}(u_j) \duj \dxj \right| \\
&\le \frac{h_\bfsnum{0}{j}^2}{2} \sup_{x_j \in [0,1]} |f_\bfsnum{0}{j}''(x_j)| \\
&\le \frac{C_{f,2} h_\bfnum{0}^2}{2 C_{h,L}}.
\end{align*}
Since the right-hand side is uniform in $j$, this establishes \eqref{claim-1-claim1}.

We note that \eqref{claim-2-claim1} is not a direct consequence of Lemma~\ref{lem:u-stat2}. Observe that
\begin{align*}
\norm{\hPi_\bfsnum{0}{0}(f_\bfsnum{0}{j}^\rmtp) - \tPi_\bfsnum{0}{0}(f_\bfsnum{0}{j}^\rmtp)}_\hMnum{0} = \left|\int_0^1 f_\bfsnum{0}{j}^\rmv(x_j)^\top \left(\hp_\bfsnum{0}{j}^\rmv(x_j) - \tp_\bfsnum{0}{j}^\rmv(x_j)\right)\dxj\right|.
\end{align*}
For $1 \le i \le n_\bfnum{0}$ and $j \in \cS_\bfnum{0}$, define
\begin{align*}
T_\bfsnum{0}{j}^i := \int_0^1 \left(f_\bfsnum{0}{j}(x_j) + (X_\bfsnum{0}{j}^i - x_j)f_\bfsnum{0}{j}'(x_j)\right) K_{h_\bfsnum{0}{j}}(x_j, X_\bfsnum{0}{j}^i) \dxj.
\end{align*}
Then, we have
\begin{align*}
\int_0^1 f_\bfsnum{0}{j}^\rmv(x_j)^\top \left(\hp_\bfsnum{0}{j}^\rmv(x_j) - \tp_\bfsnum{0}{j}^\rmv(x_j)\right)\dxj = \frac{1}{n_\bfnum{0}} \suminum{0} \left( T_\bfsnum{0}{j}^i - \Eb(T_\bfsnum{0}{1j}) \right).
\end{align*}
Let $\widetilde T_\bfsnum{0}{j}^i := T_\bfsnum{0}{j}^i - \Eb(T_\bfsnum{0}{j}^i)$. Since there exists an absolute constant $0 < C_T < \infty$ such that $\max_{j \in \cS_\bfnum{0}} \max_{1 \le i \le n_\bfnum{0}} |T_\bfsnum{0}{j}^i| \le C_T$, applying Bernstein's inequality yields
\begin{align*}
\Pb\left(\left|\frac{1}{n_\bfnum{0}} \suminum{0} \widetilde T_\bfsnum{0}{j}^i \right| \ge t \right) \le 2 \exp\left( -\frac{n_\bfnum{0} t^2}{8C_T^2 + \frac{4}{3} C_T t} \right).
\end{align*}
Therefore, for sufficiently large $n_\bfnum{0}$ such that $\frac{\sqrt{\log (|\cS_\bfnum{0}|\vee n_\bfnum{0})}}{n_\bfnum{0}} \le 1$, we obtain
\begin{align} \label{prob-bound-claim1-thm1}
\begin{aligned}
\Pb\left( \max_{j \in \cS_\bfnum{0}} \left| \frac{1}{n_\bfnum{0}} \suminum{0} \widetilde T_\bfsnum{0}{j}^i \right| \ge C \sqrt{ \frac{\log (|\cS_\bfnum{0}| \vee n_\bfnum{0})}{n_\bfnum{0}} } \right)
&\le 2|\cS_\bfnum{0}| \exp\left( -\frac{ \log (|\cS_\bfnum{0}| \vee n_\bfnum{0}) C^2 }{8C_T^2 + \frac{4}{3} C_T C} \right) \\
&\le \exp\left( \log(|\cS_\bfnum{0}|) - \frac{ \log (|\cS_\bfnum{0}| \vee n_\bfnum{0}) C^2 }{8C_T^2 + \frac{4}{3} C_T C} \right).
\end{aligned}
\end{align}
By choosing $C$ sufficiently large in \eqref{prob-bound-claim1-thm1}, the desired result follows.

\smallskip\noindent
{\it Proof of \eqref{main-claim2-thm1}.}
\smallskip

Lemma~\ref{lem:u-stat2} and Lemma~\ref{lem:u-stat3} imply that there exists an absolute constant $0 < \mathscr{C}_\bfnum{0}^* < \infty$ such that for any $g_j^\rmtp \in \sH_j^\rmtp$ and $g_k^\rmtp \in \sH_k^\rmtp$,
\begin{align} \label{important-claim2-thm1}
\begin{aligned}
\left\Vert U_j^\top \cdot (\hM_\bfsnum{0}{jj} - \tM_\bfsnum{0}{jj})g_j^\rmv \right\Vert_\Mnum{0} &\le \mathscr{C}_\bfnum{0}^*\left(\frac{1}{n_\bfnum{0}h_\bfnum{0}} + B(n_\bfnum{0}, h_\bfnum{0}, d)\right)^{\frac{1}{2}}\norm{g_j^\rmtp}_\Mnum{0}, \\
\left\Vert U_j^\top \cdot \int_0^1 (\hM_\bfsnum{0}{jk}(\cdot,x_k) - \tM_\bfsnum{0}{jk}(\cdot, x_k))g_k^\rmv(x_k)\dxk \right\Vert_\Mnum{0} &\le \mathscr{C}_\bfnum{0}^*\left(\frac{1}{n_\bfnum{0}h_\bfnum{0}^2} + B(n_\bfnum{0}, h_\bfnum{0}^2, d)\right)^{\frac{1}{2}}\norm{g_k^\rmtp}_\Mnum{0},
\end{aligned}
\end{align}
with probability tending to one. Observe that
\begin{align*}
&\norm{\alpha_\bfnum{0}^\rmtp}_\tMnum{0}^2 - \norm{\alpha_\bfnum{0}^\rmtp}_\hMnum{0}^2 \\
&= \int_{[0,1]^d}\left(\sumj \alpha_\bfsnum{0}{j}^\rmtp(x_j)\right)^\top \left(\tM_\bfnum{0}(\xv) - \hM_\bfnum{0}(\xv)\right)\left(\sumj \alpha_\bfsnum{0}{j}^\rmtp(x_j)\right) \dxv \\
&= \sumj \int_0^1 \alpha_\bfsnum{0}{j}^\rmv(x_j)^\top \left(\tM_\bfsnum{0}{jj}(x_j) - \hM_\bfsnum{0}{jj}(x_j)\right)\alpha_\bfsnum{0}{j}^\rmv(x_j)\dxj \\
&\quad + 2\sumjk \int_{[0,1]^2} \alpha_\bfsnum{0}{j}^\rmv(x_j)^\top \left(\tM_\bfsnum{0}{jk}(x_j,x_k) - \hM_\bfsnum{0}{jk}(x_j,x_k)\right)\alpha_\bfsnum{0}{k}^\rmv(x_k)\dxj\dxk.
\end{align*}
Since
\begin{align*}
\min_{j \in [d]} \inf_{x_j \in [0,1]} \lambda_\rmmin(M_\bfsnum{0}{jj}(x_j)) \ge C_{p,L}^\rmuniv\mu_2,
\end{align*}
the first term can be bounded by
\begin{align} \label{clue1-claim2-thm1}
\begin{aligned}
&\sumj \left| \int_0^1 \alpha_\bfsnum{0}{j}^\rmv(x_j)^\top \left(\tM_\bfsnum{0}{jj}(x_j) - \hM_\bfsnum{0}{jj}(x_j)\right)\alpha_\bfsnum{0}{j}^\rmv(x_j)\dxj \right| \\
&\le \frac{1}{C_{p,L}^\rmuniv\mu_2} \sumj \norm{\alpha_\bfsnum{0}{j}^\rmtp}_\Mnum{0} \cdot \left\Vert U_j^\top \cdot (\hM_\bfsnum{0}{jj} - \tM_\bfsnum{0}{jj}) \alpha_\bfsnum{0}{j}^\rmv \right\Vert_\Mnum{0} \\
&\le \frac{\mathscr{C}_\bfnum{0}^*}{C_{p,L}^\rmuniv\mu_2} \left(\frac{1}{n_\bfnum{0}h_\bfnum{0}} + B(n_\bfnum{0}, h_\bfnum{0}, d)\right)^{\frac{1}{2}} \sumj \norm{\alpha_\bfsnum{0}{j}^\rmtp}_\Mnum{0}^2,
\end{aligned}
\end{align}
where the last inequality follows from the first part of \eqref{important-claim2-thm1}. Similarly, we bound the second term as
\begin{align} \label{clue2-claim2-thm1}
\begin{aligned}
&\sumjk \left| \int_{[0,1]^2} \alpha_\bfsnum{0}{j}^\rmv(x_j)^\top \left(\tM_\bfsnum{0}{jk}(x_j,x_k) - \hM_\bfsnum{0}{jk}(x_j,x_k)\right) \alpha_\bfsnum{0}{k}^\rmv(x_k) \dxj\dxk \right| \\
&\le \frac{1}{C_{p,L}^\rmuniv\mu_2} \sumjk \norm{\alpha_\bfsnum{0}{j}^\rmtp}_\Mnum{0} \cdot \left\Vert U_j^\top \cdot \int_0^1 (\hM_\bfsnum{0}{jk}(\cdot,x_k) - \tM_\bfsnum{0}{jk}(\cdot,x_k)) \alpha_\bfsnum{0}{k}^\rmv(x_k) \dxk \right\Vert_\Mnum{0} \\
&\le \frac{\mathscr{C}_\bfnum{0}^*}{C_{p,L}^\rmuniv\mu_2} \left(\frac{1}{n h^2} + B(n, h^2, d)\right)^{\frac{1}{2}} \sumjk \norm{\alpha_\bfsnum{0}{j}^\rmtp}_\Mnum{0} \cdot \norm{\alpha_\bfsnum{0}{k}^\rmtp}_\Mnum{0},
\end{aligned}
\end{align}
where we applied the second part of \eqref{important-claim2-thm1}. Combining \eqref{clue1-claim2-thm1} and \eqref{clue2-claim2-thm1}, and using the fact
\begin{align*}
\frac{1}{n_\bfnum{0}h_\bfnum{0}} + B(n_\bfnum{0}, h_\bfnum{0}, d) \le \frac{1}{n_\bfnum{0}h_\bfnum{0}^2} + B(n_\bfnum{0}, h_\bfnum{0}^2, d),
\end{align*}
we obtain
\begin{align*}
\left| \norm{\alpha_\bfnum{0}^\rmtp}_\tMnum{0}^2 - \norm{\alpha_\bfnum{0}^\rmtp}_\hMnum{0}^2 \right| \le \frac{\mathscr{C}_\bfnum{0}^*}{C_{p,L}^\rmuniv\mu_2} \left(\frac{1}{n h^2} + B(n, h^2, d)\right)^{\frac{1}{2}} \left(\sumj \norm{\alpha_\bfsnum{0}{j}^\rmtp}_\Mnum{0} \right)^2.
\end{align*}
From Lemma~\ref{lem:tech1}, we have
\begin{align*}
\frac{C_{p,L}^\rmuniv \mu_2}{3} \le \min_{j \in [d]} \inf_{x_j \in [0,1]} \lambda_{\rmmin}\left( \hM_{jj}(x_j) \right) \le \max_{j \in [d]} \sup_{x_j \in [0,1]} \lambda_{\rmmax}\left( \hM_{jj}(x_j) \right) \le 3 C_{p,U}^{\rmuniv}
\end{align*}
with probability tending to one. Hence, for all $j\in[d]$,
\begin{align*}
\norm{g_j^\rmtp}_\Mnum{0}^2 \le \frac{3 C_{p,U}^\rmuniv}{C_{p,L}^\rmuniv \mu_2} \norm{g_j^\rmtp}_\hMnum{0}^2, \quad \text{for all } g_j^\rmtp \in \sH_j^\rmtp.
\end{align*}
Applying this yields
\begin{align*}
\left| \norm{\alpha_\bfnum{0}^\rmtp}_\tMnum{0}^2 - \norm{\alpha_\bfnum{0}^\rmtp}_\hMnum{0}^2 \right| &\le \frac{3 \mathscr{C}_\bfnum{0}^* C_{p,U}^\rmuniv}{(C_{p,L}^\rmuniv \mu_2)^2} \cdot \left(\frac{1}{n h^2} + B(n, h^2, d)\right)^{\frac{1}{2}} \left(\sumj \norm{\alpha_\bfsnum{0}{j}^\rmtp}_\hMnum{0} \right)^2 \\
&\le \frac{12 \mathscr{C}_\bfnum{0}^* C_{p,U}^\rmuniv}{(C_{p,L}^\rmuniv \mu_2)^2} \cdot \left( \frac{\mathfrak{C}_\bfnum{0}}{\mathfrak{C}_\bfnum{0} - 1} \right)^2 \cdot \left( \frac{1}{n h^2} + B(n, h^2, d) \right)^{\frac{1}{2}} \sD_\bfnum{0}^2,
\end{align*}
where we have used \eqref{main-cond1-thm1}. By setting
\[
\mathscr{C}_\bfnum{0} = \frac{12 \mathscr{C}_\bfnum{0}^* C_{p,U}^\rmuniv}{(C_{p,L}^\rmuniv \mu_2)^2} \cdot \left( \frac{\mathfrak{C}_\bfnum{0}}{\mathfrak{C}_\bfnum{0} - 1} \right)^2,
\]
the desired result follows since
\begin{align*}
\norm{\alpha_\bfnum{0}^{\rmtp,\tc}}_\tMnum{0}^2 \le \norm{\alpha_\bfnum{0}^\rmtp}_\tMnum{0}^2. 
\end{align*}

\subsubsection{Proof of Corollary~\ref{cor:ll-bound-error-pop}}

We sketch the proof. Recall the definitions of $\alpha_\bfsnum{0}{j}^\rmtp$, $\alpha_\bfsnum{0}{j}^{\rmtp, \tc}$, $\alpha_\bfnum{0}^\rmtp$, and $\alpha_\bfnum{0}^{\rmtp, \tc}$ from the proof of Theorem~\ref{thm:ll-bound-error-emp}. Additionally, define $\alpha_\bfsnum{0}{j}^{\rmtp, \rmc}:= \alpha_\bfsnum{0}{j}^\rmtp - \Pi_\bfsnum{0}{0}(\alpha_\bfsnum{0}{j}^\rmtp)$ and let $\alpha_\bfnum{0}^{\rmtp, \rmc} := \sumj \alpha_\bfsnum{0}{j}^{\rmtp, \rmc}$. Along the lines of the proof of \eqref{claim-propa1}, one may show that there exist absolute constants $0<a<b<\infty$ such that
\begin{align} \label{pop-version-thm2} 
a\sumj \norm{\alpha_\bfsnum{0}{j}^{\rmtp, \rmc}}_\Mnum{0}^2 \le \norm{\alpha_\bfnum{0}^{\rmtp, \rmc}}_\Mnum{0}^2 \le b\sumj \norm{\alpha_\bfsnum{0}{j}^{\rmtp, \rmc}}_\Mnum{0}^2.
\end{align}
Similarly, Proposition~\ref{prop:suff-cond-norm-cmpt} implies the existence of absolute constants $0<\widetilde a<\widetilde b<\infty$ such that
\begin{align} \label{tilde-version-thm2}
\widetilde a(1-\sqrt{h_\bfnum{0}}|\cS_\bfnum{0}|)\sumj \norm{\alpha_\bfsnum{0}{j}^{\rmtp, \tc}}_\tMnum{0}^2 \le \norm{\alpha_\bfnum{0}^{\rmtp, \tc}}_\tMnum{0}^2 \le \widetilde b(1-\sqrt{h_\bfnum{0}}|\cS_\bfnum{0}|)\sumj \norm{\alpha_\bfsnum{0}{j}^{\rmtp, \tc}}_\tMnum{0}^2.
\end{align}
Furthermore, from standard kernel smoothing theory, it can be shown that there exist absolute constants $0<c_1<c_2<\infty$ such that
\begin{align*}
\norm{g_j^\rmtp}_\Mnum{0} \le c_1 \norm{g_j^\rmtp}_\tMnum{0} \le c_2 \norm{g_j^\rmtp}_\hMnum{0}, \quad g_j^\rmtp\in\sH_j^\rmtp,
\end{align*}
uniformly over $j\in[d]$, with probability tending to one. Combining this with~\eqref{pop-version-thm2} and~\eqref{tilde-version-thm2}, we derive
\begin{align*}
\norm{\alpha_\bfnum{0}^\rmtp}_\Mnum{0}^2 &\le 2\norm{\alpha_\bfnum{0}^{\rmtp, \rmc}}_\Mnum{0}^2 + 2\norm{\Pi_\bfsnum{0}{0}(\alpha_\bfnum{0}^\rmtp)}_\Mnum{0}^2 \\
&\le 2b\sumj \norm{\alpha_\bfsnum{0}{j}^{\rmtp, \rmc}}_\Mnum{0}^2 + 2\norm{\Pi_\bfsnum{0}{0}(\alpha_\bfnum{0}^\rmtp)}_\Mnum{0}^2 \\
&\le 2b\sumj \norm{\alpha_\bfsnum{0}{j}^{\rmtp, \tc}}_\Mnum{0}^2 + 2\norm{\Pi_\bfsnum{0}{0}(\alpha_\bfnum{0}^\rmtp)}_\Mnum{0}^2 \\
&\le 2c_1b \sumj \norm{\alpha_\bfsnum{0}{j}^{\rmtp, \tc}}_\tMnum{0}^2 + 2\norm{\Pi_\bfsnum{0}{0}(\alpha_\bfnum{0}^\rmtp)}_\Mnum{0}^2 \\
&\le \frac{2c_1b}{\widetilde a(1-\sqrt{h_\bfnum{0}}|\cS_\bfnum{0}|)}\norm{\alpha_\bfnum{0}^{\rmtp, \tc}}_\tMnum{0}^2 + 2\norm{\Pi_\bfsnum{0}{0}(\alpha_\bfnum{0}^\rmtp)}_\Mnum{0}^2 \\
&\le \frac{2c_1b}{\widetilde a(1-\sqrt{h_\bfnum{0}}|\cS_\bfnum{0}|)}\left(\norm{\alpha_\bfnum{0}^\rmtp}_\hMnum{0}^2 + \mathscr{C}_\bfnum{0} \left( \frac{1}{n_\bfnum{0}h_\bfnum{0}^2} + B(n_\bfnum{0}, h_\bfnum{0}^2, d) \right)^{\frac{1}{2}} \sD_\bfnum{0}^2\right) + 2\norm{\Pi_\bfsnum{0}{0}(\alpha_\bfnum{0}^\rmtp)}_\Mnum{0}^2,
\end{align*}
where the last inequality follows from~\eqref{main-claim2-thm1}. Since
\begin{align*}
\sqrt{h_\bfnum{0}}|\cS_\bfnum{0}|,\quad |\cS_\bfnum{0}|\left( \frac{1}{n_\bfnum{0}h_\bfnum{0}^2} + B(n_\bfnum{0}, h_\bfnum{0}^2, d) \right)^{\frac{1}{2}} \ll 1,
\end{align*}
it suffices to show that
\begin{align} \label{what-to-show-thm2}
\norm{\Pi_\bfsnum{0}{0}(\alpha_\bfnum{0}^\rmtp)}_\Mnum{0}^2 \lesssim \norm{\alpha_\bfnum{0}^\rmtp}_\hMnum{0}^2.
\end{align}

We note that, for any $g^\rmtp\in\sH_\rmadd^\rmtp$, the projection $\Pi_\bfsnum{0}{0}(g^\rmtp)$ takes the form $(c_j^\rmtp, 0_d^\top)^\top$. Based on this observation, denote by $c_\bfsnum{0}{j}^\rmtp$ the first element of $\Pi_\bfsnum{0}{0}(\alpha_\bfsnum{0}{j}^\rmtp)$. Recall that $p_j^\rmv = (p_j, 0)^\top$. Then, it holds that
\begin{align*}
c_\bfsnum{0}{j}^\rmtp &= \int_0^1 \left(\hatf_\bfsnum{0}{j}(x_j) - f_\bfsnum{0}{j}(x_j)\right) p_\bfsnum{0}{j}(x_j)\dxj \\
&= \int_0^1 \alpha_\bfsnum{0}{j}^\rmv(x_j)^\top \left(p_\bfsnum{0}{j}^\rmv(x_j) - \hp_\bfsnum{0}{j}^\rmv(x_j)\right)\dxj - \int_0^1 f_\bfsnum{0}{j}^\rmv(x_j)^\top \hp_\bfsnum{0}{j}^\rmv(x_j)\dxj.
\end{align*}
We claim that there exist absolute constants $0 < C_1, C_2 < \infty$ such that
\begin{align} \label{claim1-thm2}
\left|\int_0^1 \alpha_\bfsnum{0}{j}^\rmv(x_j)^\top \left(p_\bfsnum{0}{j}^\rmv(x_j) - \hp_\bfsnum{0}{j}^\rmv(x_j)\right)\dxj \right| \le C_1\sqrt{h_\bfnum{0}}\norm{\alpha_\bfsnum{0}{j}^\rmtp}_\hMnum{0}, \quad j\in[d],
\end{align}
and
\begin{align} \label{claim2-thm2}
\left|\int_0^1 f_\bfsnum{0}{j}^\rmv(x_j)^\top \hp_\bfsnum{0}{j}^\rmv(x_j)\dxj\right| \begin{cases}
\le C_2 h_\bfnum{0}^2, &\quad j\in\cS_\bfnum{0}, \\
= 0, &\quad j\notin\cS_\bfnum{0},
\end{cases}
\end{align}
with probability tending to one. The bounds in~\eqref{claim1-thm2} and~\eqref{claim2-thm2} together imply~\eqref{what-to-show-thm2}. To see this, let
\begin{align*}
D_\bfsnum{0}{1j} &:= \int_0^1 \alpha_\bfsnum{0}{j}^\rmv(x_j)^\top \left(p_\bfsnum{0}{j}^\rmv(x_j) - \hp_\bfsnum{0}{j}^\rmv(x_j)\right)\dxj, \\
D_\bfsnum{0}{2j} &:= \int_0^1 f_\bfsnum{0}{j}^\rmv(x_j)^\top \hp_\bfsnum{0}{j}^\rmv(x_j)\dxj.
\end{align*}
Then it follows that
\begin{align*}
\norm{\Pi_\bfsnum{0}{0}(\alpha_\bfnum{0}^\rmtp)}_\hMnum{0}^2 &= \left|\sumj D_\bfsnum{0}{1j} + \sumjs D_\bfsnum{0}{2j}\right|^2 \\
&\le 2\left(\sumj |D_\bfsnum{0}{1j}|\right)^2 + 2\left(\sumjs |D_\bfsnum{0}{2j}|\right)^2 \\
&\lesssim h_\bfnum{0} \left(\sumj \norm{\alpha_\bfsnum{0}{j}^\rmtp}_\hMnum{0}\right)^2 + |\cS_\bfnum{0}|^2 h_\bfnum{0}^4 \\
&\lesssim \norm{\alpha_\bfnum{0}^\rmtp}_\hMnum{0}^2.
\end{align*}
Here, we use the condition that 
\begin{align*}
|\cS_\bfnum{0}|h_\bfnum{0}^2 \lesssim \left(\frac{1}{n_\bfnum{0}h_\bfnum{0}} + A(n_\bfnum{0}, h_\bfnum{0}, d;\alpha)\right)^{\frac{1}{2}}.
\end{align*}
It remains to verify claims~\eqref{claim1-thm2} and~\eqref{claim2-thm2}. As both follow from standard kernel smoothing theory, the details are omitted.

\subsubsection{Proof of Theorem~\ref{thm:ll-minimax}}

It is without loss of generality to assume that each covariate $X_\bfsnum{0}{j}$ is uniformly distributed on $[0,1]$ when proving the theorem. To justify this reduction, suppose that 
\begin{align*}
\inf_{\tilde{f}} \sup_{f_\bfnum{0} \in \sF_\bfsnum{0}{\rmadd}^s(\beta,L)} \Pb_{f,\mathrm{unif}}\left( \norm{\tilde{f} - f_\bfnum{0}}_{p_\bfnum{0}}^2 \gtrsim s\left(n^{-\frac{2\beta}{2\beta+1}} + \frac{\log(d/s)}{n}\right) \right) \ge \frac{1}{2},
\end{align*}
where $\Pb_{f,\mathrm{unif}}$ denotes the probability measure under the assumption that the true regression function is $f_\bfnum{0}$ and that each $X_\bfsnum{0}{j}$ follows the uniform distribution on $[0,1]$. The infimum is taken over all measurable functions of the target sample $\{(\Xv_\bfsnum{0}{i}, Y_\bfsnum{0}{i})\}_{i=1}^{n_\bfnum{0}}$. 
Let $F_\bfsnum{0}{j}$ be the cumulative distribution function of $X_\bfsnum{0}{j}$. Under assumption (P1), $F_\bfsnum{0}{j}$ is strictly increasing, and thus $X_\bfsnum{0}{j}$ has one-to-one correspondence with uniformly distributed variable via $U_\bfsnum{0}{j} := F_\bfsnum{0}{j}(X_\bfsnum{0}{j})$. This change of variables preserves measurability, so the collection of estimators—measurable functions of the observed data—remains the same under both the general and uniform designs.
On the other hand, the set of distributions over which the supremum is taken becomes smaller under the uniform design, since the probability measure space is restricted to covariates with uniform marginals. That is,
\begin{align*}
\sup_{f_\bfnum{0} \in \sF_\bfsnum{0}{\mathrm{add}}^s(\beta,L)} \Pb_{f,\mathrm{unif}}\left(E_{(\Xv_\bfnum{0}, Y_\bfnum{0})}\right) \le \sup_{f_\bfnum{0} \in \sF_\bfsnum{0}{\mathrm{add}}^s(\beta,L)} \Pb_{f}\left(E_{(\Xv_\bfnum{0}, Y_\bfnum{0})}\right)
\end{align*}
for any measurable event $E_{(\Xv_\bfnum{0}, Y_\bfnum{0})}$ of $\{(\Xv_\bfsnum{0}{i}, Y_\bfsnum{0}{i})\}_{i=1}^{n_\bfnum{0}}$. Therefore, assuming the uniformity of the covariates leads to a smaller or equal minimax risk, and thus provides a valid lower bound for the general case. Throughout the following, we assume without further mention that each covariate $X_\bfsnum{0}{j}$ is uniformly distributed on $[0,1]$. The function class $\sF_\bfsnum{0}{j}(\beta, L)$ is understood to be the collection of all functions $g_j$ satisfying 
\begin{align*}
g_j \in \Sigma(\beta, L) \quad \text{and} \quad \int_0^1 g_j(x_j) \dxj = 0.
\end{align*}

To prove the theorem, we construct a set of functions 
\[
\sG := \left\{0,g^1, \ldots, g^M\right\} \subset \sF_\bfsnum{0}{\rmadd}^s(\beta, L),
\]
that are sufficiently separated from one another. In order to ensure that each $g^\ell$ belongs to $\sF_\bfsnum{0}{\rmadd}^s(\beta, L)$, we construct component functions $g_j^\ell \in \sF_\bfsnum{0}{j}(\beta, L)$ forming $g^\ell$, such that
\begin{align*}
\int_0^1 g_j^\ell(x_j)\dxj = 0.
\end{align*}
To this end, we choose a nonzero function $\kappa: \Rb \to \Rb$ satisfying the following conditions:
\begin{itemize}[itemsep=0pt, topsep=0pt, parsep=0pt, partopsep=0pt]
\item[($\kappa1$)] $\kappa\in\Sigma(\beta, 1)\cap C^\infty(\Rb)$;
\item[($\kappa2$)] ${\rm supp}(\kappa)=(-\frac{1}{2}, \frac{1}{2})$;
\item[($\kappa3$)] $\kappa_\infty:=\sup_{u\in\Rb}|\kappa(u)|<\infty$ and $\kappa_2 := \int_\Rb \kappa(u)^2\,\rmd u>0$;
\item[($\kappa4$)] $\int_{-1/2}^{1/2} \kappa(u)\, \rmd u=0$.
\end{itemize}
We emphasize that condition ($\kappa4$) ensures that $g_j^\ell \in \sF_\bfsnum{0}{j}(\beta, L)$ under a suitable construction, which constitutes a key difference from existing approaches. The existence of such a function $\kappa$ is guaranteed, as one may take $\kappa = \kappa_0$, where
\[
\kappa_0(u) := c_\kappa \cdot u \exp\left(-\frac{1}{1 - 4u^2}\right) I\left(-\frac{1}{2} \le u \le \frac{1}{2}\right),
\]
for some normalization constant $c_\kappa > 0$.
Let $N$ be a natural number whose value will be specified later. Put $\xi_l = (l - \frac{1}{2})/N$, and define
\begin{align*}
\eta_{jl}(u_j) := \frac{L}{2} \cdot b^\beta \cdot \kappa\left(\frac{u_j - \xi_l}{b}\right),
\end{align*}
where $b = 1/N$. Since $\eta_{jl}$ and $\eta_{jl'}$ have disjoint supports whenever $l \ne l'$, and $\eta_{jl} \in \sF_\bfsnum{0}{j}(\beta, L)$, the following construction satisfies the required conditions. For any matrix $A \in \{-1, 0, 1\}^{d \times N}$ with exactly $s$ nonzero rows, define
\begin{align*}
g_{A,j}(x_j) &:= \sum_{l=1}^N a_{jl} \eta_{jl}(x_j), \\
g_A(x_1, \ldots, x_d) &:= \sumj g_{A,j}(x_j),
\end{align*}
where $a_{jl}$ denotes the $(j,l)$-entry of $A$. Clearly, $g_A \in \sF_\bfsnum{0}{\rmadd}^s(\beta, L)$.

To fully characterize the set $\sG$, it remains to construct a collection of matrices with $s$ nonzero rows. We follow the construction of \cite{yuan2016minimax}, incorporating the Varshamov–Gilbert lemma as presented in \cite{massart2007concen}. For the sake of completeness, we reproduce the essential details here.
Applying the Varshamov–Gilbert lemma, we can construct a set $\{\theta_1, \ldots, \theta_{M_1}\} \subset \{0,1\}^d$ such that
\begin{itemize}
\item[(a)] $\norm{\theta_l}_{\ell_1} = s$ for all $1 \le l \le M_1$;
\item[(b)] for any $l \ne l'$, $\norm{\theta_l - \theta_{l'}}_1 \ge \frac{s}{2}$;
\item[(c)] $\log M_1 \ge \frac{s}{4} \log (d/s)$.
\end{itemize}
Here, $\norm{\cdot}_{\ell_1}$ denotes the $\ell_1$-norm of a vector. Each $\theta_l$ specifies the indices of the nonzero rows in a matrix. 
Next, we characterize the values in those nonzero rows by filling them with $\pm1$ entries. To this end, we again invoke the Varshamov–Gilbert lemma to construct a set $\{\Gamma_1, \ldots, \Gamma_{M_2}\} \subset \{-1,1\}^{s \times N}$ satisfying
\begin{itemize}
\item[(a$'$)] for any $l \ne l'$, $\norm{\Gamma_l - \Gamma_{l'}}_F^2 \ge \frac{Ns}{2}$;
\item[(b$'$)] $\log M_2 \ge \frac{Ns}{8}$.
\end{itemize}
Here, $\norm{\cdot}_F$ denotes the Frobenius norm of a matrix. Each pair $(\theta_l, \Gamma_{l'})$ uniquely determines a matrix, denoted by $A(\theta_l, \Gamma_{l'})$. Finally, we define a set $\sG$ by $\sG:=\{0\}\cup \widetilde \sG$ where 
\begin{align*}
\widetilde \sG := \left\{ g_{A(\theta_l, \Gamma_{l'})} : 1 \le l \le M_1, \; 1 \le l' \le M_2 \right\}.
\end{align*}
Simply write $\widetilde \sG=\{g_{A_\ell}:1\le \ell\le M\}$ where $M=M_1M_2$. Note that (c) together with (b$'$) implies that $\log M \ge \frac{s}{4}\log (d/s)+\frac{Ns}{8}$.

Let $\Mc := \{A_\ell : 1 \le \ell \le M\}$ denote the collection of constructed matrices. Note that 
\begin{align*}
\int_0^1 \eta_{jl}(x_j)^2 \dxj &= \frac{L^2}{4}b^{2\beta+1}\int_0^1 \kappa(x_j)^2 \dxj = \frac{L^2\kappa_2}{4}b^{2\beta+1}.
\end{align*}
This, together with the inequality in \eqref{ll-norm-ineq-minimx}, implies that
\begin{align*}
\norm{g_A - g_B}_\pnum{0}^2 &\ge C_{\sF, L} \sumj \norm{g_{A,j} - g_{B,j}}_\pnum{0}^2 \\
&= C_{\sF, L} \sumj \int_0^1 \left\{ \sum_{l=1}^N (a_{jl} - b_{jl}) \eta_{jl}(x_j) \right\}^2 \dxj \\
&= C_{\sF, L} \sumj \sum_{l=1}^N (a_{jl} - b_{jl})^2 \int_0^1 \eta_{jl}(x_j)^2 \dxj \\
&= \frac{C_{\sF, L} L^2 \kappa_2}{4} b^{2\beta+1} \sumj \sum_{l=1}^N (a_{jl} - b_{jl})^2 \\
&= \frac{C_{\sF, L} L^2 \kappa_2}{4} b^{2\beta+1} \norm{A - B}_F^2,
\end{align*}
for any $A, B \in \Mc$, where $a_{jl}$ and $b_{jl}$ denote the $(j, l)$-entries of $A$ and $B$, respectively. Here, we used the fact that $\eta_{jl}$ and $\eta_{jl'}$ have disjoint supports for $l \ne l'$ in the third equality. Using (a$'$), we further obtain
\begin{align} \label{lower-bound-thm3}
\norm{g_A - g_B}_\pnum{0}^2 \ge \frac{C_{\sF, L} L^2 \kappa_2}{4} b^{2\beta+1} \norm{A - B}_F^2 \ge \frac{C_{\sF, L} L^2 \kappa_2}{8} N^{-2\beta} s.
\end{align}
Similarly, for any $A\in\Mc$, we can derive that
\begin{align} \label{upper-bound-thm3}
\begin{aligned}
\norm{g_A}_\pnum{0}^2 &\le C_{\sF,U}\sumj \norm{g_{A,j}}_\pnum{0}^2 \\
&=C_{\sF,U}\sumj \sum_{l=1}^N a_{jl}^2\int_0^1 \eta_{jl}(x_j)^2\dxj \\
&=\frac{C_{\sF,U}L^2\kappa_2}{4}b^{2\beta+1}\sumj\sum_{l=1}^N a_{jl}^2 \\
&= \frac{C_{\sF,U}L^2 \kappa_2}{4}N^{-2\beta}s. 
\end{aligned}
\end{align}

We obtain the minimax lower bound via Fano's lemma. Let $P_\bfsnum{0}{\ell}$, for $1 \le \ell \le M$, denote the joint distribution of $\{(\Xv_\bfsnum{0}{i}, Y_\bfsnum{0}{i})\}_{i=1}^{n_\bfnum{0}}$ when the true regression function is $g_{A_\ell}$, and let $P_\bfsnum{0}{0}$ denote the joint distribution when the regression function is identically zero. Let $\KL{\cdot}{\cdot}$ denote the Kullback--Leibler divergence. Then, we have
\begin{align} \label{kl-bound-thm3}
\begin{aligned}
&\KL{P_\bfsnum{0}{\ell}}{P_\bfsnum{0}{0}} \\
&= \suminum{0} \int_{[0,1]^d}p_\bfnum{0}(\xv_\bfnum{0}^{i})\int_\Rb p_{\ve_\bfnum{0}|\xv_\bfnum{0}}(y_\bfsnum{0}{i})\log \left(\frac{p_{\ve_\bfnum{0}|\xv_\bfnum{0}}(y_\bfnum{0}^{i})}{p_{\ve_\bfnum{0}|\xv_\bfnum{0}}(y_\bfnum{0}^{i} + g_{A_\ell}(\xv_\bfnum{0}^{i}))}\right)\, \rmd y_\bfnum{0}^{i} \, \rmd \xv_\bfnum{0}^{i} \\
&\le c_\ve \suminum{0}\norm{g_{A_\ell}}_\pnum{0}^2 \\
&\le \frac{c_\ve C_{\sF, U}L^2 \kappa_2}{4}n_\bfnum{0}N^{-2\beta}s,
\end{aligned}
\end{align}
whenever
\begin{align} \label{condition-kl-thm3}
\sup_{\xv\in[0,1]^d}|g_{A_\ell}(\xv)|\le \frac{L\kappa_\infty}{2}N^{-\beta}s \le v_\ve.
\end{align}
Applying Corollary~2.6 of \cite{tsybakov2009nonpara} together with \eqref{kl-bound-thm3}, we obtain
\begin{align} \label{res-fano-lemma-thm3}
\begin{aligned}
&\inf_{\widetilde f}\sup_{f_\bfnum{0}\in\sF_\bfsnum{0}{\rmadd}^s(\beta, L)}\Pb_f\left(\norm{\widetilde f - f_\bfnum{0}}_\pnum{0}^2 \ge \frac{1}{4}\min_{A\ne B\in\Mc}\norm{g_A - g_B}_\pnum{0}^2\right) \\
&\hspace{6cm}\ge 1-\frac{{c_\ve C_{\sF, U}L^2 \kappa_2}n_\bfnum{0}N^{-2\beta}s + 4\log 2}{4\log M} \\
&\hspace{6cm}\ge 1-\frac{2{c_\ve C_{\sF, U}L^2 \kappa_2}n_\bfnum{0}N^{-2\beta}s + 8\log 2}{2s\log (d/s) + Ns}.
\end{aligned}
\end{align}
Here, we used the fact that $\log M = \log M_1 + \log M_2 \ge \frac{s\log (d/s)}{4} + \frac{Ns}{8}$.

By choosing $N = C_{N,1}n_\bfnum{0}^{\frac{1}{2\beta+1}}$ for sufficiently large constant $C_{N,1}>0$, \eqref{res-fano-lemma-thm3} yields
\begin{align} \label{final-res1-thm3}
\inf_{\widetilde f}\sup_{f_\bfnum{0}\in\sF_\bfsnum{0}{\rmadd}^s(\beta, L)}\Pb_f\left(\norm{\widetilde f - f_\bfnum{0}}_\pnum{0}^2 \gtrsim sn_\bfnum{0}^{-\frac{2\beta}{2\beta+1}}\right) \ge \frac{3}{4}.
\end{align}
Here, we have used the notation $\gtrsim$ in probability arguments to indicate that the inequality holds up to a multiplicative constant $0<C<\infty$, depending only on $C_{\sF,L}, C_{\sF,U}, \beta$ and $L$. Alternatively, choosing $N = C_{N,2}(\frac{n_\bfnum{0}}{\log (d/s)})^{\frac{1}{2\beta}}$ for sufficiently large $C_{N,2}>0$, we obtain from \eqref{res-fano-lemma-thm3}
\begin{align} \label{final-res2-thm3}
\inf_{\widetilde f}\sup_{f_\bfnum{0}\in\sF_\bfsnum{0}{\rmadd}^s(\beta, L)}\Pb_f\left(\norm{\widetilde f - f_\bfnum{0}}_\pnum{0}^2 \gtrsim s\frac{\log (d/s)}{n_\bfnum{0}}\right) \ge \frac{3}{4}.
\end{align}
Clearly, \eqref{final-res1-thm3} and \eqref{final-res2-thm3} together imply the claim of the theorem. It remains to verify that the above choices of $N$ satisfy \eqref{condition-kl-thm3}. This follows from condition~\eqref{ll-restriction-s}, and the details are therefore omitted.

\subsection{Technical Proofs for Section~\ref{sec:tl}}

This section presents the technical details supporting the results in Section~\ref{sec:tl}. Throughout the proofs, all (in)equalities are understood to hold either almost surely or with probability tending to one. We often use the notations $C_\ell$ for $\ell\in\Nb$ to denote (absolute) constants, whose values may change from line to line.

\subsubsection{Proof of Proposition~\ref{prop:invertibility-PicA}}

First, we prove the invertibility of the operator $\rmI^\rmtp + \Pi_\bfnum{a}^{\ominus,\rmtp}$ for all $\ab \in \{\bz\} \cup \cA$. Fix $\ab \in \{\bz\} \cup \cA$. By definition, $\Pi_\bfnum{a}^{\ominus,\rmtp}$ can be represented as a $d \times d$ matrix of kernel integral operators. Specifically, $\Pi_\bfnum{a}^{\ominus,\rmtp}$ is defined as a matrix-valued kernel operator whose $(j,k)$-entry, denoted by $\pi_\bfsnum{a}{jk} : \sH_k^\rmtp \to \sH_j^\rmtp$, is given by
\begin{align*}
\pi_\bfsnum{a}{jk}(g_k^\rmtp) = \Pi_\bfsnum{a}{j}(g_k^\rmtp), \quad g_k^\rmtp \in \sH_k^\rmtp.
\end{align*}
Each operator $\pi_\bfsnum{a}{jk}$ is Hilbert–Schmidt, and thus compact. Since $d < \infty$ and every compact operator is the norm-limit of finite-rank operators, it follows that $\Pi_\bfnum{a}^{\ominus,\rmtp}$ is itself compact. Let $\sigma_p(\Qc)$ denote the point spectrum of a bounded linear operator $\Qc:\sH_\rmprod^\rmtp \to\sH_\rmprod^\rmtp$. By Theorem~6.8 of \cite{brezis2011functional} and Corollary~4.15 of \cite{conway1990functional}, the operator $\rmI^\rmtp + \Pi_\bfnum{a}^{\ominus,\rmtp}$ is invertible if and only if $-1 \notin \sigma_p(\Pi_\bfnum{a}^{\ominus,\rmtp})$.

We proceed by contradiction. Suppose that $-1 \in \sigma_p(\Pi_\bfnum{a}^{\ominus,\rmtp})$, so that there exists a nonzero function tuple $\etav^\rmtp = (\eta_j^\rmtp : j \in [d]) \in \sH_\rmprod^\rmtp$, where $\eta_j^\rmtp = U_j^\top \cdot (\eta_j, \eta_j^{(1)})^\top$, satisfying
\begin{align} \label{start-prop1}
(\rmI^\rmtp + \Pi_\bfnum{a}^{\ominus,\rmtp})(\etav^\rmtp) = -\etav^\rmtp.
\end{align}
For each $j \in [d]$, define the centered function $\eta_j^\rmc = \eta_j - \Eb(\eta_j(X_\bfsnum{a}{j}))$. From \eqref{start-prop1}, we obtain
\begin{align*} 
\begin{aligned}
-\norm{\etav^\rmtp}_\Mnum{a}^2 &= \ip{(\rmI^\rmtp + \Pi_\bfnum{a}^{\ominus,\rmtp})(\etav^\rmtp), \etav^\rmtp}_\Mnum{a} \\
&= \Eb\left( \left( \sumj \eta_j^\rmc(X_\bfsnum{a}{j}) \right)^2 \right) + \sumj \Eb\left( \eta_j(X_\bfsnum{a}{j}) \right)^2 + \mu_2\sumj \Eb\left( \eta_j^{(1)}(X_\bfsnum{a}{j})^2 \right).
\end{aligned}
\end{align*}
Since 
\begin{align*}
\norm{\etav^\rmtp}_\Mnum{a}^2 = \sumj \Eb(\eta_j(X_\bfsnum{a}{j}))^2 + \mu_2 \sumj \Eb(\eta_j^{(1)}(X_\bfsnum{a}{j}))^2,
\end{align*}
it reduces to 
\begin{align} \label{exp-prop1}
 \Eb\left( \left( \sumj \eta_j^\rmc(X_\bfsnum{a}{j}) \right)^2 \right) + 2\sumj \Eb\left( \eta_j(X_\bfsnum{a}{j}) \right)^2 + 2\mu_2\sumj \Eb\left( \eta_j^{(1)}(X_\bfsnum{a}{j})^2 \right)=0.
\end{align}
Since condition (T1) holds, it follows from \eqref{exp-prop1} that the tuple $\etav^{\rmtp,\rmc} = (\eta_j^{\rmtp,\rmc} : j \in [d])$, with $\eta_j^{\rmtp,\rmc} = U_j^\top \cdot (\eta_j^\rmc, \eta_j^{(1)})^\top$, must be identically zero. 
This with \eqref{exp-prop1} implies that $\etav^\rmtp$ is also the zero function tuple. This contradicts the assumption that $\etav^\rmtp$ is nonzero, and therefore establishes that $\rmI^\rmtp + \Pi_\bfnum{a}^{\ominus,\rmtp}$ is invertible.

Next, we prove the invertibility of the operator $\rmI^\rmtp + \Pi_\bfnum{\cA}^{\ominus,\rmtp}$. Since conditions (P1)--(P2) imposed on each auxiliary population imply that the aggregated marginal and pairwise densities $p_\bfsnum{\cA}{j}$ and $p_\bfsnum{\cA}{jk}$ also satisfy the same conditions, it suffices to verify that $-1 \notin \sigma_p(\Pi_\bfnum{\cA}^{\ominus,\rmtp})$. Suppose, by way of contradiction, that there exists a nonzero function tuple $\etav^\rmtp \in \sH_\rmprod^\rmtp$ such that
\begin{align*}
(\rmI^\rmtp + \Pi_\bfnum{\cA}^{\ominus,\rmtp})(\etav^\rmtp) = -\etav^\rmtp.
\end{align*}
Then, by the same argument as before, we obtain
\begin{align} \label{ca-version-prop1}
\ip{(\rmI^\rmtp + \Pi_\bfnum{\cA}^{\ominus,\rmtp})(\etav^\rmtp), \etav^\rmtp}_\Mnum{\cA} = -\norm{\etav^\rmtp}_\Mnum{\cA}^2.
\end{align}
Using the identity
\begin{align*}
\Mc_\cA^\rmtp (\rmI^\rmtp + \Pi_\bfnum{\cA}^{\ominus,\rmtp}) = \sumaac w_\bfnum{a} \Mc_\bfnum{a}^\rmtp (\rmI^\rmtp + \Pi_\bfnum{a}^{\ominus,\rmtp}),
\end{align*}
we deduce from \eqref{ca-version-prop1} that
\begin{align*}
-\sumaac w_\bfnum{a} \norm{\etav^\rmtp}_\Mnum{a}^2 = \sumaac w_\bfnum{a} \ip{(\rmI^\rmtp + \Pi_\bfnum{a}^{\ominus,\rmtp})(\etav^\rmtp), \etav^\rmtp}_\Mnum{a}.
\end{align*}
Since each operator $\rmI^\rmtp + \Pi_\bfnum{a}^{\ominus,\rmtp}$ is invertible by the argument established previously, it follows that the right-hand side is nonnegative only when $\etav^\rmtp$ is the zero function tuple, yielding a contradiction. This completes the proof.

\subsubsection{Proof of Proposition~\ref{prop:bound-PicA}}

For notational convenience, let $\Tc_\bfnum{a}^\rmtp := \Mc_\bfnum{a}^\rmtp(\rmI^\rmtp + \Pi_\bfnum{a}^{\ominus,\rmtp})$ for $\ab \in \{\bz\} \cup \cA$, and define $\Tc_\bfnum{\cA}^\rmtp := \Mc_\bfnum{\cA}^\rmtp(\rmI^\rmtp + \Pi_\bfnum{\cA}^{\ominus,\rmtp})$. Recall from Proposition~\ref{prop:invertibility-PicA} that the operators $\rmI^\rmtp + \Pi_\bfnum{a}^{\ominus,\rmtp}$ for $\ab \in \{\bz\} \cup \cA$, as well as $\rmI^\rmtp + \Pi_\bfnum{\cA}^{\ominus,\rmtp}$, are invertible. This implies that $\Tc_\bfnum{a}^\rmtp$ for all $\ab \in \{\bz\} \cup \cA$ and $\Tc_\bfnum{\cA}^\rmtp$ are also invertible.
We claim that
\begin{align} \label{claim-prop2}
\max\left\{\norm{(\Tc_\bfnum{0}^\rmtp)^{-1}}_\opnumnum{0}{1}, \norm{(\Tc_\bfnum{\cA}^\rmtp)^{-1}}_\opnumnum{0}{1}\right\} < \infty.
\end{align}
We emphasize that the previous invertibility result does not guarantee \eqref{claim-prop2}, since invertibility alone only ensures that
\begin{align*}
\max\left\{\norm{(\Tc_\bfnum{0}^\rmtp)^{-1}}_\opnumnum{0}{2}, \norm{(\Tc_\bfnum{\cA}^\rmtp)^{-1}}_\opnumnum{0}{2}\right\} < \infty.
\end{align*}

Suppose the claim in \eqref{claim-prop2} holds. Observe that
\begin{align*}
(\Tc_\bfnum{\cA}^\rmtp)^{-1} 
&= \left(\Tc_\bfnum{\cA}^\rmtp - \Tc_\bfnum{0}^\rmtp + \Tc_\bfnum{0}^\rmtp\right)^{-1} \\
&= \left(\sumaac w_\bfnum{a} (\Tc_\bfnum{a}^\rmtp - \Tc_\bfnum{0}^\rmtp) + \Tc_\bfnum{0}^\rmtp\right)^{-1} \\
&= (\Tc_\bfnum{0}^\rmtp)^{-1} - (\Tc_\bfnum{0}^\rmtp)^{-1} \left(\sumaac w_\bfnum{a} (\Tc_\bfnum{a}^\rmtp - \Tc_\bfnum{0}^\rmtp)\right) (\Tc_\bfnum{\cA}^\rmtp)^{-1}.
\end{align*}
Taking the $\norm{\cdot}_\opnumnum{0}{1}$ on both sides and recalling the definition of $\eta_{p,1}$, we obtain
\begin{align*}
\norm{(\Tc_\bfnum{\cA}^\rmtp)^{-1}}_\opnumnum{0}{1} 
\le \sk + \sk \eta_{p,1} \norm{(\Tc_\bfnum{\cA}^\rmtp)^{-1}}_\opnumnum{1}.
\end{align*}
Since $\sk \eta_{p,1} \le \gamma < 1$ by condition (T2), it follows that
\begin{align*}
\norm{(\Tc_\cA^\rmtp)^{-1}}_\opnumnum{0}{1} \le \frac{\sk}{1 - \sk \eta_{p,1}}.
\end{align*}

It remains to prove \eqref{claim-prop2}. We only verify that $\norm{(\Tc_\bfnum{0}^\rmtp)^{-1}}_\opnumnum{0}{1}<\infty$, as the bound for $\norm{(\Tc_\bfnum{\cA}^\rmtp)^{-1}}_\opnumnum{\cA}{1}$ follows analogously. For any function tuple $\etav^\rmtp \in \sH_\rmprod^\rmtp$, the Hölder inequality yields
\begin{align*}
\sumj \norm{\eta_j^\rmtp}_\Mnum{0} \le d \left(\sumj \norm{\eta_j^\rmtp}_\Mnum{0}^2\right)^{\frac{1}{2}}.
\end{align*}
Combining this with the fact that
\begin{align*}
\left\{\gv^\rmtp \in \sH_\rmprod^\rmtp : \sumj \norm{g_j^\rmtp}_\Mnum{0} \le 1\right\} \subset \left\{\gv^\rmtp \in \sH_\rmprod^\rmtp : \sumj \norm{g_j^\rmtp}_\Mnum{0}^2 \le 1\right\},
\end{align*}
we obtain
\begin{align*}
\norm{(\Tc_\bfnum{0}^\rmtp)^{-1}}_\opnumnum{0}{1} \le d \norm{(\Tc_\bfnum{0}^\rmtp)^{-1}}_\opnumnum{0}{2} < \infty.
\end{align*}

\subsubsection{Proof of Proposition~\ref{prop:functional-similarity}}

Recall the definitions $\Tc_\bfnum{a}^\rmtp := \Mc_\bfnum{a}^\rmtp(\rmI^\rmtp + \Pi_\bfnum{a}^{\ominus,\rmtp})$ for $\ab \in \{\bz\} \cup \cA$, and define $\Tc_\bfnum{\cA}^\rmtp := \Mc_\bfnum{\cA}^\rmtp(\rmI^\rmtp + \Pi_\bfnum{\cA}^{\ominus,\rmtp})$. From \eqref{fundamental-identity-delta}, we have
\begin{align*}
\deltav_\cA^\rmtp &= \sumaac w_\bfnum{a}\deltav_\bfnum{a}^\rmtp + (\Tc_\cA^\rmtp)^{-1}\left\{\sumaac w_\bfnum{a}\left(\Tc_\bfnum{a}^\rmtp(\deltav_\bfnum{a}^\rmtp) - \Tc_\bfnum{\cA}^\rmtp(\deltav_\bfnum{a}^\rmtp)\right)\right\} \\
&= \sumaac w_\bfnum{a} \deltav_\bfnum{a}^\rmtp + (\Tc_\cA^\rmtp)^{-1} \left\{ \sumaac w_\bfnum{a} \left( \Tc_\bfnum{a}^\rmtp(\deltav_\bfnum{a}^\rmtp) - \Tc_\bfnum{0}^\rmtp(\deltav_\bfnum{a}^\rmtp) + \Tc_\bfnum{0}^\rmtp(\deltav_\bfnum{a}^\rmtp) - \Tc_\bfnum{\cA}^\rmtp(\deltav_\bfnum{a}^\rmtp) \right) \right\}.
\end{align*}
We observe that
\begin{align} \label{ineq-prop3}
\norm{\Tc_\cA^\rmtp - \Tc_\bfnum{0}^\rmtp}_\opnumnum{0}{1} \le \sumaac w_\bfnum{a} \norm{\Tc_\bfnum{a}^\rmtp - \Tc_\bfnum{0}^\rmtp}_\opnumnum{0}{1} \le \eta_{p,1},
\end{align}
where we used the definition of $\eta_{p,1}$. Taking $\norm{\cdot}_\opnumnum{0}{1}$ on both sides and applying \eqref{ineq-prop3}, we derive
\begin{align*}
\sumj \normbig{\delta_\bfsnum{\cA}{j}^\rmtp - \sumaac w_\bfnum{a} \delta_\bfsnum{a}{j}^\rmtp}_\Mnum{0}
&\le \frac{\sk}{1 - \sk \eta_{p,1}} \cdot 2 \eta_{p,1} \cdot \left( \sumaac w_\bfnum{a} \sumj \norm{\delta_\bfsnum{a}{j}^\rmtp}_\Mnum{0} \right) \\
&= \frac{2 \sk \eta_{p,1}}{1 - \sk \eta_{p,1}} \eta_\delta,
\end{align*}
which is the desired result.

\subsubsection{Proof of Proposition~\ref{prop:tl-norm-compt}}

Suppose that $\gv^\rmtp = (g_j^\rmtp : j \in [d])$ is a function tuple satisfying the conditions of the proposition. Define $g_\bfsnum{a}{0j}^\rmtp := \tPi_\bfsnum{a}{0}(g_j^\rmtp)$, where the projection operator $\tPi_\bfsnum{a}{0}$ is defined analogously to $\tPi_\bfsnum{0}{0}$, with the density $\tp_\bfnum{0}$ replaced by $\tp_\bfnum{a}$. We claim that there exists an absolute constant $0 < C_1 < \infty$ such that
\begin{align}
\norm{g_\bfsnum{a}{0j}^\rmtp}_\tMnum{\cA} &\le C_1 \sqrt{\eta_{p,2} + h_\cA^2} \norm{g_j^\rmtp}_\tMnum{\cA}, \label{claim1-prop4} \\
\sumjnots \norm{g_j^\rmtp - g_\bfsnum{a}{0j}^\rmtp}_\tMnum{a} &\le \frac{4C_{p,U}^\rmuniv C}{C_{p,L}^\rmuniv \mu_2} \sumjs \norm{g_j^\rmtp - g_\bfsnum{a}{0j}^\rmtp}_\tMnum{a}. \label{claim2-prop4}
\end{align}
Note that the norms in \eqref{claim2-prop4} are evaluated with respect to $\tMnum{a}$, and $C$ is the constant from the proposition satisfying
\begin{align*}
\sumjnots \norm{g_j^\rmtp}_\tMnum{\cA} \le C \sumjs \norm{g_j^\rmtp}_\tMnum{\cA}.
\end{align*}
The proofs of these claims are deferred to the end of the proof.

We now observe that
\begin{align*}
&\left|\int_0^1 \int_0^1 g_j^\rmv(x_j)^\top \tM_\bfsnum{\cA}{jk}(x_j,x_k) g_k^\rmv(x_k)\dxj\dxk\right| \\
&\le \sumaac w_\bfnum{a} \left|\int_0^1 \int_0^1 (g_j^\rmv(x_j) - g_\bfsnum{a}{0j}^\rmv)^\top \tM_\bfsnum{a}{jk}(x_j,x_k) (g_k^\rmv(x_k) - g_\bfsnum{a}{0k}^\rmv)\dxj\dxk\right| \\
&\quad + \sumaac w_\bfnum{a} \left| (g_\bfsnum{a}{0j}^\rmv)^\top \int_0^1 \int_0^1 \tM_\bfsnum{a}{jk}(x_j, x_k) (g_k^\rmv(x_k) - g_\bfsnum{a}{0k}^\rmv)\dxj\dxk \right| \\
&\quad + \sumaac w_\bfnum{a} \left| \int_0^1 \int_0^1 (g_j^\rmv(x_j) - g_\bfsnum{a}{0j}^\rmv)^\top \tM_\bfsnum{a}{jk}(x_j, x_k)\dxj\dxk \cdot g_\bfsnum{a}{0k}^\rmv \right| \\
&\quad + \sumaac w_\bfnum{a} \left| (g_\bfsnum{a}{0j}^\rmv)^\top \int_0^1 \int_0^1 \tM_\bfsnum{a}{jk}(x_j, x_k)\dxj\dxk \cdot g_\bfsnum{a}{0k}^\rmv \right| \\
&=: \sumaac w_\bfnum{a} \left( \Gc_\bfsnum{a}{jk}^{(1)} + \Gc_\bfsnum{a}{jk}^{(2)} + \Gc_\bfsnum{a}{jk}^{(3)} + \Gc_\bfsnum{a}{jk}^{(4)} \right).
\end{align*}
From standard kernel smoothing theory, we may show that there exists an absolute constant $0 < C_2<\infty$ such that
\begin{align*}
\left(\int_0^1 \int_0^1 \norm{\tM_\bfsnum{a}{jk}(x_j,x_k) - M_\bfsnum{a}{jk}(x_j, x_k)}_F^2\dxj\dxk\right)^{\frac{1}{2}} &\le \frac{ C_2}{2}\sqrt{h_\cA}, \\
\left(\int_0^1 \int_0^1 \norm{\tp_\bfsnum{a}{j}^\rmv(x_j) \tp_\bfsnum{a}{k}^\rmv(x_k)^\top - p_\bfsnum{a}{j}^\rmv(x_j) p_\bfsnum{a}{k}^\rmv(x_k)^\top}_F^2\dxj\dxk\right)^{\frac{1}{2}} &\le \frac{C_2}{2}\sqrt{h_\cA}.
\end{align*}
Then, using \eqref{claim2-prop4} and the arguments from the proof of Proposition~\ref{prop:suff-cond-norm-cmpt}, we obtain that
\begin{align*}
2\sumjk \Gc_\bfsnum{a}{jk}^{(1)} 
&\le 2\sqrt{\varphi}\frac{\sqrt{\psi}}{1-\sqrt{\psi}} \sumj \norm{g_j^\rmtp - g_\bfsnum{a}{0j}^\rmtp}_{I_{d+1}}^2 + C_2 \sqrt{h_\bfnum{\cA}} \left(\sumj \norm{g_j^\rmtp - g_\bfsnum{a}{0j}^\rmtp}_{I_{d+1}}\right)^2 \\
&\le \sqrt{\varphi}\frac{\sqrt{\psi}}{1-\sqrt{\psi}} \cdot \frac{4}{C_{p,L}^\rmuniv \mu_2} \left(\sumj \norm{g_j^\rmtp - g_\bfsnum{a}{0j}^\rmtp}_\tMnum{a}^2\right) \\
&\quad + \frac{2C_2}{C_{p,L}^\rmuniv \mu_2} \sqrt{h_\bfnum{\cA}} \left( \sumj \norm{g_j^\rmtp - g_\bfsnum{a}{0j}^\rmtp}_\tMnum{a} \right)^2 \\
&\le \sqrt{\varphi}\frac{\sqrt{\psi}}{1-\sqrt{\psi}} \cdot \frac{4}{C_{p,L}^\rmuniv \mu_2} \left(\sumj \norm{g_j^\rmtp - g_\bfsnum{a}{0j}^\rmtp}_\tMnum{a}^2\right) \\
&\quad + \frac{2\widetilde C_\cA^{(2)}}{C_{p,L}^\rmuniv \mu_2} \sqrt{h_\bfnum{\cA}} \left(1 + \frac{4C_{p,U}^\rmuniv C}{C_{p,L}^\rmuniv \mu_2} \right)^2 \left( \sumjs \norm{g_j^\rmtp - g_\bfsnum{a}{0j}^\rmtp}_\tMnum{a} \right)^2 \\
&\le \sqrt{\varphi}\frac{\sqrt{\psi}}{1-\sqrt{\psi}} \cdot \frac{8 C_{p,U}^\rmuniv}{(C_{p,L}^\rmuniv \mu_2)^2} \left(\sumj \norm{g_j^\rmtp}_\tMnum{\cA}^2\right) \\
&\quad + \frac{4C_2 C_{p,U}^\rmuniv}{(C_{p,L}^\rmuniv \mu_2)^2} \left(1 + \frac{4C_{p,U}^\rmuniv C}{C_{p,L}^\rmuniv \mu_2} \right)^2 \sqrt{h_\bfnum{\cA}} |\cS_\bfnum{0}| \left( \sumjs \norm{g_j^\rmtp}_\tMnum{\cA}^2 \right),
\end{align*}
where the last inequality follows from the fact that $\norm{g_j^\rmtp - g_\bfsnum{a}{0j}^\rmtp}_\tMnum{a} \le \norm{g_j^\rmtp}_\tMnum{a}$. Similarly, we may derive that 
\begin{align*}
2\sumjk \Gc_\bfsnum{a}{jk}^{(2)},2\sumjk \Gc_\bfsnum{a}{jk}^{(3)} &\le \sqrt{\varphi}\frac{\sqrt{\psi}}{1-\sqrt{\psi}} \frac{4\sqrt{C_{p,U}^\rmuniv}C_1}{C_{p,L}^\rmuniv\mu_2} \sqrt{\eta_{p,2}+h_\cA}\left(\sumj \norm{g_j^\rmtp}_\tMnum{\cA}^2\right)\\
&\quad + \frac{2\sqrt{C_{p,U}^\rmuniv}C_1C_2}{C_{p,L}^\rmuniv \mu_2}\sqrt{\eta_{p,2}+h_\cA}\sqrt{h_\bfnum{\cA}}|\cS_\bfnum{0}|(1+C)^2 \left(\sumjs \norm{g_j^\rmtp}_\tMnum{\cA}^2\right)
\end{align*}
and
\begin{align*}
2\sumjk \Gc_\bfsnum{a}{jk}^{(4)} &\le \sqrt{\varphi}\frac{\sqrt{\psi}}{1-\sqrt{\psi}} 2C_1^2 (\eta_{p,2}+h_\cA)\left(\sumj \norm{g_j^\rmtp}_\tMnum{\cA}^2\right)  \\
&\quad + C_1^2C_2(1+C)^2 (\eta_{p,2}+h_\cA)\sqrt{h_\cA}|\cS_\bfnum{0}|\left(\sumjs \norm{g_j^\rmtp}_\tMnum{\cA}^2\right).
\end{align*}
From this with the fact that $\eta_{p,2}=o(1)$, for all sufficiently large $n_\bfnum{0}$, we have
\begin{align*}
2\sumjk\left(\Gc_\bfsnum{a}{jk}^{(2)} + \Gc_\bfsnum{a}{jk}^{(3)} + \Gc_\bfsnum{a}{jk}^{(4)}\right) \le \frac{1}{8}\cdot \left(2\sumjk \Gc_\bfsnum{a}{jk}^{(1)}\right).
\end{align*}
Then, the proposition follows since
\begin{align*}
\normbig{\sumj g_j^\rmtp}_\tMnum{\cA}^2 &\ge \sumj \norm{g_j^\rmtp}_\tMnum{\cA}^2 - 2\sumjk \left|\int_0^1 \int_0^1 g_j^\rmv(x_j)^\top \tM_\bfsnum{\cA}{jk}(x_j,x_k) g_k^\rmv(x_k)\dxj\dxk\right| \\
&\ge \sumj \norm{g_j^\rmtp}_\tMnum{\cA}^2 - 2\sumaac w_\bfnum{a} \, \sumjk\left(\Gc_\bfsnum{a}{jk}^{(1)} + \Gc_\bfsnum{a}{jk}^{(2)} + \Gc_\bfsnum{a}{jk}^{(3)} + \Gc_\bfsnum{a}{jk}^{(4)}\right) \\
&\ge \sumj \norm{g_j^\rmtp}_\tMnum{\cA}^2 - \frac{9}{8}\left(2\sumaac w_\bfnum{a} \, \sumjk \Gc_\bfsnum{a}{jk}^{(1)}\right).
\end{align*}

It remains to prove \eqref{claim1-prop4} and \eqref{claim2-prop4}. For \eqref{claim1-prop4}, we observe that
\begin{align*}
g_\bfsnum{a}{0j}^\rmtp &= \int_0^1 g_j^\rmv(x_j)^\top \tp_\bfsnum{a}{j}^\rmv(x_j)\dxj \\
&= \int_0^1 g_j^\rmv(x_j)^\top \left\{\tp_\bfsnum{a}{j}^\rmv(x_j) - \tp_\bfsnum{\cA}{j}^\rmv(x_j)\right\}\dxj \\
&\le \norm{g_j^\rmtp}_{I_{d+1}} \norm{\tp_\bfsnum{a}{j}^\rmtp - \tp_\bfsnum{\cA}{j}^\rmtp}_{I_{d+1}},
\end{align*}
where $\tp_\bfsnum{a}{j}^\rmtp = U_j^\top \cdot \tp_\bfsnum{a}{j}^\rmv$ and $\tp_\bfsnum{\cA}{j}^\rmtp = U_j^\top \cdot \tp_\bfsnum{\cA}{j}^\rmv$. Define $p_\bfsnum{a}{j}^\rmtp := U_j^\top \cdot p_\bfsnum{a}{j}^\rmv$ and $p_\bfsnum{\cA}{j}^\rmtp := U_j^\top \cdot p_\bfsnum{\cA}{j}^\rmv$. Then it follows that
\begin{align*}
\norm{\tp_\bfsnum{a}{j}^\rmtp - \tp_\bfsnum{\cA}{j}^\rmtp}_{I_{d+1}} &\le \norm{\tp_\bfsnum{a}{j}^\rmtp - p_\bfsnum{a}{j}^\rmtp}_{I_{d+1}} + \norm{p_\bfsnum{a}{j}^\rmtp - p_\bfsnum{\cA}{j}^\rmtp}_{I_{d+1}} + \norm{\tp_\bfsnum{\cA}{j}^\rmtp - p_\bfsnum{\cA}{j}^\rmtp}_{I_{d+1}} \\
&\le C_3 \sqrt{h_\bfnum{\cA} + \eta_{p,2}},
\end{align*}
for some absolute constant $0 < C_3< \infty$. This with the fact that
\begin{align*}
\norm{g_j^\rmtp}_{I_{d+1}}\le \sqrt{\frac{2}{C_{p,L}^\rmuniv\mu_2}}\norm{g_j^\rmtp}_\tMnum{\cA}
\end{align*}
completes the proof of \eqref{claim1-prop4}.
To establish \eqref{claim2-prop4}, note that
\begin{align*}
\sumjnots \norm{g_j^\rmtp - g_\bfsnum{a}{0j}^\rmtp}_\tMnum{a}
&\le \sqrt{\frac{4C_{p,U}^\rmuniv}{C_{p,L}^\rmuniv \mu_2}} \sumjnots \norm{g_j^\rmtp - g_\bfsnum{a}{0j}^\rmtp}_\tMnum{\cA} \\
&\le \sqrt{\frac{4C_{p,U}^\rmuniv}{C_{p,L}^\rmuniv \mu_2}} \sumjnots \norm{g_j^\rmtp}_\tMnum{\cA} \\
&\le C \sqrt{\frac{4C_{p,U}^\rmuniv}{C_{p,L}^\rmuniv \mu_2}} \sumjs \norm{g_j^\rmtp - g_\bfsnum{a}{0j}^\rmtp}_\tMnum{\cA} \\
&\le \frac{4C_{p,U}^\rmuniv C}{C_{p,L}^\rmuniv \mu_2} \sumjs \norm{g_j^\rmtp - g_\bfsnum{a}{0j}^\rmtp}_\tMnum{a}.
\end{align*}

\subsubsection{Proof of Lemma~\ref{lem:tl-nonasymp-bound-Delta}}

Observe that
\begin{align*}
\Delta_\bfsnum{\cA}{j}^\rmv(x_j) &= \hM_\bfsnum{\cA}{jj}(x_j)^{-1}\Biggr[\sumaac w_\bfnum{a} \Biggr\{\frac{1}{n_\bfnum{a}}\suminum{a} Z_\bfsnum{a}{j}^i(x_j) K_{h_\bfsnum{\cA}{j}}(x_j,X_\bfsnum{a}{j}^i)\Biggr(Y_\bfnum{a}^{i} - \bar Y_\bfnum{a} - Z_\bfsnum{a}{j}^i(x_j)^\top f_\bfsnum{a}{j}^\rmv(x_j) \\
&\qquad - \sumknotj \int_0^1 Z_\bfsnum{a}{k}^i(x_k)^\top f_\bfsnum{a}{k}^\rmv(x_k)\Biggr) + \hM_\bfsnum{a}{jj}(x_j)\left(\delta_\bfsnum{a}{j}^\rmv(x_j) - \delta_\bfsnum{\cA}{j}^\rmv(x_j)\right) \\
&\quad + \sumknotj \int_0^1 \hM_\bfsnum{a}{jk}(x_j,x_k)\left(\delta_\bfsnum{a}{k}^\rmv(x_k) - \delta_\bfsnum{\cA}{k}^\rmv(x_k)\right)\dxk\Biggr\}\Biggr],
\end{align*}
where we have used the identity $f_\bfsnum{a}{j}^\rmv - f_\bfsnum{\cA}{j}^\rmv = \delta_\bfsnum{a}{j}^\rmv - \delta_\bfsnum{\cA}{j}^\rmv$. Define
\begin{align*}
\Delta_\bfsnum{a}{j}^{\rmv, (1)}(x_j) &:=\frac{1}{n_\bfnum{a}}\suminum{a} Z_\bfsnum{a}{j}^i(x_j) K_{h_\bfsnum{\cA}{j}}(x_j,X_\bfsnum{a}{j}^i)\\
&\hspace{3cm} \times \Biggr(Y_\bfnum{a}^{i} - \bar Y_\bfnum{a} - Z_\bfsnum{a}{j}^i(x_j)^\top f_\bfsnum{a}{j}^\rmv(x_j) - \sumknotj \int_0^1 Z_\bfsnum{a}{k}^i(x_k)^\top f_\bfsnum{a}{k}^\rmv(x_k)\Biggr), \\
\Delta_\bfsnum{a}{j}^{\rmv, (2)}(x_j) &:= \hM_\bfsnum{a}{jj}(x_j)\left(\delta_\bfsnum{a}{j}^\rmv(x_j) - \delta_\bfsnum{\cA}{j}^\rmv(x_j)\right), \\
\Delta_\bfsnum{a}{j}^{\rmv,(3)}(x_j) &:= \sumknotj \int_0^1 \hM_\bfsnum{a}{jk}(x_j,x_k)\left(\delta_\bfsnum{a}{k}^\rmv(x_k) - \delta_\bfsnum{\cA}{k}^\rmv(x_k)\right)\dxk.
\end{align*}
Since the eigenvalues of $\hM_\bfsnum{\cA}{jj}(x_j)$ are uniformly bounded away from zero over $x_j \in [0,1]$ and $j \in [d]$, it suffices to bound the norms of $\sumaac w_\bfnum{a}\Delta_\bfsnum{a}{j}^{\rmtp, (\ell)} = U_j^\top \cdot \sumaac w_\bfnum{a}\Delta_\bfsnum{a}{j}^{\rmv, (\ell)}$ for $1 \le \ell \le 3$.

Along the lines of the proof of Lemma~\ref{lem:ll-bound-Delta}, we may show that 
\begin{align*}
\max_{j\in[d]}\norm{\Delta_\bfsnum{a}{j}^{\rmtp, (1)}}_\Mnum{0} \le C_1\left(|\cS_\bfnum{a}|\left(\sqrt{\frac{\log n_\bfnum{a}}{n_\bfnum{a}}}+ h_\bfnum{\cA}^2\right) + \sqrt{\frac{1}{n_\bfnum{a}h_\bfnum{\cA}}} + A(n_\bfnum{a}, h_\bfnum{\cA}, d;\alpha)^{\frac{1}{2}}\right)
\end{align*}
for some absolute constant $0<C_1<\infty$ with probability tending to one. Since a standard probabilistic argument yields that
\begin{align*}
&\Pb\left(\max_{j\in[d]}\normbig{\sumaac w_\bfnum{a} \Delta_\bfsnum{a}{j}^{\rmtp, (1)}}_\Mnum{0} \ge C_1|\cA|\left(|\cS_\bfnum{\cA}|\left(\sqrt{\frac{\log n_\bfnum{\cA}}{n_\bfnum{\cA}}}+ h_\bfnum{\cA}^2\right) + \sqrt{\frac{1}{n_\bfnum{\cA}h_\bfnum{\cA}}} + A(n_\bfnum{\cA}, h_\bfnum{\cA}, d;\alpha)^{\frac{1}{2}}\right) \right) \\
&\le \sumaac \Pb\left(\max_{j\in[d]}\norm{\Delta_\bfsnum{a}{j}^{\rmtp, (1)}}_\Mnum{0} \ge C_1\left(|\cS_\bfnum{a}|\left(\sqrt{\frac{\log n_\bfnum{a}}{n_\bfnum{a}}}+ h_\bfnum{\cA}^2\right) + \sqrt{\frac{1}{n_\bfnum{a}h_\bfnum{\cA}}} + A(n_\bfnum{a}, h_\bfnum{\cA}, d;\alpha)^{\frac{1}{2}}\right)\right),
\end{align*}
together with the conditions $|\cA|<\infty$ and $\frac{\log n_\cA}{n_\cA h_\cA^4} = o(1)$, we conclude that 
\begin{align} \label{order-1-lem2}
\max_{j\in[d]}\normbig{\sumaac w_\bfnum{a} \Delta_\bfsnum{a}{j}^{\rmtp, (1)}}_\Mnum{0} \lesssim |\cS_\bfnum{\cA}|h_\bfnum{\cA}^2 + \sqrt{\frac{1}{n_\bfnum{\cA}h_\bfnum{\cA}}} + A(n_\bfnum{\cA}, h_\bfnum{\cA}, d;\alpha)^{\frac{1}{2}}.
\end{align}

For the second term involving $\Delta_\bfsnum{a}{j}^{\rmv, (2)}$, we observe that
\begin{align*}
\sumaac w_\bfnum{a}\Delta_\bfsnum{a}{j}^{\rmv, (2)}(x_j) &= \sumaac w_\bfnum{a} \left(\hM_\bfsnum{a}{jj}(x_j) - \hM_\bfsnum{\cA}{jj}(x_j)\right)\delta_\bfsnum{a}{j}^\rmv(x_j) \\
&\hspace{4cm}+ \hM_\bfsnum{\cA}{jj}(x_j)\left(\sumaac w_\bfnum{a}\delta_\bfsnum{a}{j}^\rmv(x_j) - \delta_\bfsnum{\cA}{j}^\rmv(x_j)\right) \\
&\overset{\rm let}{:=} \Delta_\bfsnum{\cA}{j}^{\rmv, (2-1)}(x_j) + \Delta_\bfsnum{\cA}{j}^{\rmv, (2-2)}(x_j).
\end{align*}
Define
\begin{align*}
N_\bfsnum{\cA}{j}(x_j) := \left(\begin{array}{cc}
\mu_\bfsnum{\cA}{j,0}(x_j) & \frac{\mu_\bfsnum{\cA}{j,1}(x_j)}{\mu_2} \\
\mu_\bfsnum{\cA}{j,1}(x_j) & \frac{\mu_\bfsnum{\cA}{j,2}(x_j)}{\mu_2}
\end{array}\right), \quad j \in [d].
\end{align*}
To control the norm of $\Delta_\bfsnum{\cA}{j}^{\rmv, (2-1)}$, we claim
\begin{align}
&\max_{j\in[d]}\left[\sumaac w_\bfnum{a} \left(\int_0^1 \normbig{\hM_\bfsnum{a}{jj}(x_j) - \tM_\bfsnum{a}{jj}(x_j)}_F^2 \dxj\right)\right] \lesssim \frac{1}{n_\cA h_\cA} + B(n_\cA, h_\cA, d), \label{claim-1-1-lem2} \\
&\max_{j\in[d]}\left(\int_0^1 \normbig{\tM_\bfsnum{a}{jj}(x_j) - \tM_\bfsnum{\cA}{jj}(x_j) - N_\bfsnum{\cA}{j}(x_j)(M_\bfsnum{a}{jj}(x_j) - M_\bfsnum{\cA}{jj}(x_j))}_F^2 \dxj\right) \lesssim h_\cA^2\eta_{p,3}^2. \label{claim-1-2-lem2}
\end{align}
We prove these claims at the end of the proof. Note that \eqref{claim-1-1-lem2}, together with Jensen's inequality, implies
\begin{align}
\max_{j\in[d]}\left(\int_0^1 \normbig{\hM_\bfsnum{\cA}{jj}(x_j) - \tM_\bfsnum{\cA}{jj}(x_j)}_F^2 \dxj\right) \lesssim \frac{1}{n_\cA h_\cA} + B(n_\cA, h_\cA, d). \label{claim-1-3-lem2}
\end{align}
Observe that
\begin{align*}
\Delta_\bfsnum{\cA}{j}^{\rmv, (2-1)}(x_j) &= \sumaac w_\bfnum{a}\Biggr\{\left(\hM_\bfsnum{a}{jj}(x_j) - \tM_\bfsnum{a}{jj}(x_j)\right) - \left(\hM_\bfsnum{\cA}{jj}(x_j) - \tM_\bfsnum{\cA}{jj}(x_j)\right) \\
&\qquad + \left(\tM_\bfsnum{a}{jj}(x_j) - \tM_\bfsnum{\cA}{jj}(x_j)\right) \Biggr\}\delta_\bfsnum{a}{j}^\rmv(x_j).
\end{align*}
From \eqref{claim-1-1-lem2}, \eqref{claim-1-2-lem2}, and \eqref{claim-1-3-lem2}, we deduce that
\begin{align} \label{clue1-second-lem2}
\begin{aligned}
\Delta_\bfsnum{\cA}{j}^{\rmv, (2-1)}(x_j) = N_\bfsnum{\cA}{j}(x_j)(M_\bfsnum{a}{jj}(x_j) - M_\bfsnum{\cA}{jj}(x_j))\delta_\bfsnum{a}{j}^\rmv(x_j) + R_\bfsnum{\cA}{j}^{\rmv, (2)}(x_j;\delta_\bfsnum{a}{j}^\rmtp),
\end{aligned}
\end{align}
where $R_\bfsnum{\cA}{j}^{\rmv, (2)}(\cdot;\delta_\bfsnum{a}{j}^\rmtp)$ denotes a generic function satisfying
\begin{align*}
\norm{R_\bfsnum{\cA}{j}^{\rmtp, (2)}(\cdot; \delta_\bfsnum{a}{j}^\rmtp)}_\Mnum{0} \le C_2\left(\sqrt{\frac{1}{n_\cA h_\cA}} + B(n_\cA, h_\cA, d)^{\frac{1}{2}} + h_\cA\eta_{p,3}\right)\norm{\delta_\bfsnum{a}{j}^\rmtp}_\Mnum{0},
\end{align*}
for some absolute constant $0<C_2<\infty$. Moreover, it is straightforward to obtain
\begin{align} \label{clue2-second-lem2}
\norm{\Delta_\bfsnum{\cA}{j}^{\rmtp, (2-2)}}_\Mnum{0} \le C_3\normbig{\sumaac w_\bfnum{a}\delta_\bfsnum{a}{j}^\rmtp - \delta_\bfsnum{\cA}{j}^\rmtp}_\Mnum{0}.
\end{align}
for some absolute constant $0<C_3<\infty$.

The analysis of the last term $\sumaac w_\bfnum{a}\Delta_\bfsnum{a}{j}^{\rmv, (3)}$ proceeds analogously to that of $\sumaac w_\bfnum{a}\Delta_\bfsnum{a}{j}^{\rmv, (2)}$. Define
\begin{align*}
L_\bfsnum{\cA}{j}(x_j):= \left(\begin{array}{cc}
\mu_\bfsnum{\cA}{j,0}(x_j) & \mu_\bfsnum{\cA}{j,1}(x_j) \\
0 & 0
\end{array}\right), \quad j\in[d].
\end{align*}
In this part, we additionally establish the following bounds:
\begin{align}
&\max_{(j,k)\in[d]^2}\Biggr[ \sumaac w_\bfnum{a} \left(\int_{[0,1]^2}\normbig{\hM_\bfsnum{a}{jk}(x_j, x_k) - \tM_\bfsnum{a}{jk}(x_j, x_k)}_F^2\dxj\dxk\right)\Biggr] \lesssim \frac{1}{n_\cA h_\cA^2} + B(n_\cA, h_\cA^2, d), \label{claim-2-1-lem2}\\
&\max_{(j,k)\in[d]^2} \Biggr( \int_{[0,1]^2} \Big\Vert\tM_\bfsnum{a}{jk}(x_j,x_k) - \tM_\bfsnum{\cA}{jk}(x_j,x_k) \nonumber\\ 
&\hspace{2cm} - N_\bfsnum{\cA}{j}(x_j)L_\bfsnum{\cA}{k}(x_k)(p_\bfsnum{a}{jk}(x_j,x_k) - p_\bfsnum{\cA}{jk}(x_j,x_k))\Big\Vert_F^2\dxj\dxk \Biggr) \lesssim h_\cA^2\eta_{p,3}^2. \label{claim-2-2-lem2}
\end{align}
We prove the claims at the end of the proof. Applying similar arguments as in the derivation of \eqref{clue1-second-lem2} and \eqref{clue2-second-lem2}, and invoking \eqref{claim-2-1-lem2} and \eqref{claim-2-2-lem2}, we obtain
\begin{align*} 
&\sumaac w_\bfnum{a}\Delta_\bfsnum{a}{j}^{\rmv, (3)}(x_j) \\
&= N_\bfsnum{\cA}{j}(x_j)\sumknotj\int_0^1 (M_\bfsnum{a}{jk}(x_j,x_k) - M_\bfsnum{\cA}{jk}(x_j, x_k))\delta_\bfsnum{a}{k}^\rmv(x_k)\dxk  \\
&\quad + N_\bfsnum{\cA}{j}(x_j)\sumknotj\int_0^1 (L_\bfsnum{\cA}{k}(x_k)-I_2)(M_\bfsnum{a}{jk}(x_j,x_k) - M_\bfsnum{\cA}{jk}(x_j, x_k))\delta_\bfsnum{a}{k}^\rmv(x_k)\dxk \\
&\quad + R_\bfsnum{\cA}{j}^{\rmv,(3)}(x_j;\{\delta_\bfsnum{a}{k}^\rmtp:k\ne j\}),
\end{align*}
where $R_\bfsnum{\cA}{j}^{\rmv, (3)}(\cdot;\{\delta_\bfsnum{a}{j}^\rmtp:k\ne j\})$ denotes a generic term satisfying
\begin{align*}
\norm{R_\bfsnum{\cA}{j}^{\rmv, (3)}(\cdot;\{\delta_\bfsnum{a}{j}^\rmtp:k\ne j\})}_\Mnum{0} &\le C_4\Biggr\{\left(\sqrt{\frac{1}{n_\cA h_\cA^2}} + B(n_\cA, h_\cA^2, d)^{\frac{1}{2}} + h_\cA\eta_{p,3}\right)\left(\sumknotj \norm{\delta_\bfsnum{a}{k}^\rmtp}_\Mnum{0}\right) \\
&\quad + \sumknotj \normbig{\sumaac w_\bfnum{a}\delta_\bfsnum{a}{k}^\rmtp - \delta_\bfsnum{\cA}{k}^\rmtp}_\Mnum{0}\Biggr\},
\end{align*}
for some absolute constant $0<C_4<\infty$. 
Observe that
\begin{align*}
&\int_0^1 (L_\bfsnum{\cA}{k}(x_k)-I_2)(M_\bfsnum{a}{jk}(x_j,x_k) - M_\bfsnum{\cA}{jk}(x_j, x_k))\delta_\bfsnum{a}{k}^\rmv(x_k)\dxk \\
&= \int_0^1 \left(\begin{array}{cc}
\mu_\bfsnum{\cA}{k,0}(x_k)-1 & 0 \\
0 & 0
\end{array}\right) (p_\bfsnum{a}{jk}(x_j,x_k) - p_\bfsnum{\cA}{jk}(x_j, x_k))\delta_\bfsnum{a}{k}^\rmv(x_k)\dxk.
\end{align*}
Since, for $j\in[d]$, $\mu_\bfsnum{\cA}{j,0}(x_j) = 1$ for all $x_j \in [2h_\bfsnum{\cA}{j}, 1-2h_\bfsnum{\cA}{j}]$ and is uniformly bounded otherwise, we conclude
\begin{align*}
&\normbig{U_j^\top \cdot N_\bfsnum{\cA}{j}(x_j)\int_0^1 (L_\bfsnum{\cA}{k}(x_k)-I_2)(M_\bfsnum{a}{jk}(x_j,x_k) - M_\bfsnum{\cA}{jk}(x_j, x_k))\delta_\bfsnum{a}{k}^\rmv(x_k)\dxk}_\Mnum{0} \\
&\le C_5h_\cA \eta_{p,3}\norm{\delta_\bfsnum{a}{k}^\rmtp}_\Mnum{0},
\end{align*}
for some absolute constant $0<C_5<\infty$. It is therefore valid to write
\begin{align} \label{order-3-lem2}
\begin{aligned}
\sumaac w_\bfnum{a}\Delta_\bfsnum{a}{j}^{\rmv, (3)}(x_j) &= N_\bfsnum{\cA}{j}(x_j)\sumknotj\int_0^1 (M_\bfsnum{a}{jk}(x_j,x_k) - M_\bfsnum{\cA}{jk}(x_j, x_k))\delta_\bfsnum{a}{k}^\rmv(x_k)\dxk \\
&\quad + R_\bfsnum{\cA}{j}^{\rmv,(3)}(x_j;\{\delta_\bfsnum{a}{k}^\rmtp:k\ne j\}).
\end{aligned}
\end{align}
Let $\Tc_\bfnum{a}^\rmtp := \Mc_\bfnum{a}^\rmtp(\rmI^\rmtp + \Pi_\bfnum{a}^{\ominus,\rmtp})$ for $\ab \in \cA$. We observe that
\begin{align*}
U_j^\top \cdot &\Biggr((M_\bfsnum{a}{jj} - M_\bfsnum{\cA}{jj})\delta_\bfsnum{a}{j}^\rmv + \sumknotj \int_0^1\left(M_\bfsnum{a}{jk}(\cdot, x_k) - M_\bfsnum{\cA}{jk}(\cdot, x_k)\right)\delta_\bfsnum{a}{k}^\rmv(x_k)\dxk \\
&\quad - \int_0^1 \diag(1,0)(p_\bfsnum{a}{k}(x_k) - p_\bfsnum{\cA}{k}(x_k))\delta_\bfsnum{a}{k}^\rmv(x_k)\dxk \Biggr)
\end{align*}
corresponds to the $j$-th component of $(\Tc_\bfnum{a}^\rmtp - \Tc_\bfnum{\cA})\deltav_\bfnum{a}^\rmtp$. Therefore, we obtain
\begin{align} \label{operator-lem2}
\begin{aligned}
&\max_{j\in[d]}\normbig{U_j^\top \cdot \left((M_\bfsnum{a}{jj} - M_\bfsnum{\cA}{jj})\delta_\bfsnum{a}{j}^\rmv + \sumknotj \int_0^1\left(M_\bfsnum{a}{jk}(\cdot, x_k) - M_\bfsnum{\cA}{jk}(\cdot, x_k)\right)\delta_\bfsnum{a}{k}^\rmv(x_k)\dxk\right)} \\
&\le (\norm{\Tc_\bfnum{a}^\rmtp - \Tc_\bfnum{\cA}^\rmtp}_\opnumnum{0}{1} + \eta_{p,2})\eta_\delta \le (\eta_{p,1} + \eta_{p,2})\eta_\delta.
\end{aligned}
\end{align}
Since
\begin{align*}
\sup_{x_j\in[0,1]}\max_{j\in[d]}\lambda_\rmmax\left(N_\bfsnum{\cA}{j}(x_j)\right)\le C_5,
\end{align*}
for some absolute constant $0<C_5<\infty$, it follows from \eqref{operator-lem2}, \eqref{clue1-second-lem2}, \eqref{clue2-second-lem2}, and \eqref{order-3-lem2} that
\begin{align*}
\max_{j\in[d]}\normbig{\sumaac w_\bfnum{a}\left(\Delta_\bfsnum{a}{j}^{\rmtp,(2)} + \Delta_\bfsnum{a}{j}^{\rmtp, (3)}\right)}_\Mnum{0} &\lesssim \left(\sqrt{\frac{1}{n_\cA h_\cA^2}} + B(n_\cA, h_\cA^2, d)^{\frac{1}{2}} + h_\cA\eta_{p,3} + \eta_{p,1} + \eta_{p,2}\right) \eta_\delta \\
&\quad + \sumj \normbig{\sumaac w_\bfnum{a}\delta_\bfsnum{a}{j}^\rmtp - \delta_\bfsnum{\cA}{j}^\rmtp}_\Mnum{0} \\
&\lesssim \left(\sqrt{\frac{1}{n_\cA h_\cA^2}} + B(n_\cA, h_\cA^2, d)^{\frac{1}{2}} + h_\cA\eta_{p,3} + \eta_{p,1} + \eta_{p,2}\right) \eta_\delta + \eta_{p,\delta}.
\end{align*}
Together with \eqref{order-1-lem2}, this completes the proof.

It remains to verify the claims \eqref{claim-1-1-lem2}, \eqref{claim-1-2-lem2}, \eqref{claim-2-1-lem2}, and \eqref{claim-2-2-lem2}. The bounds in \eqref{claim-1-2-lem2} and \eqref{claim-2-2-lem2} follow from Lemma~\ref{lem:u-stat2} and Lemma~\ref{lem:u-stat3}, respectively, together with standard probabilistic arguments. Hence, it suffices to prove \eqref{claim-1-2-lem2} and \eqref{claim-2-2-lem2}.
To prove \eqref{claim-1-2-lem2}, we show that for $1 \le \ell, \ell' \le 2$,
\begin{align*}
\max_{j\in[d]}\sup_{x_j\in[0,1]}\left|\left(\tM_\bfsnum{a}{jj}(x_j) - \tM_\bfsnum{\cA}{jj}(x_j)\right)_{\ell, \ell'} - \left(N_\bfsnum{\cA}{j}(x_j)\left(M_\bfsnum{a}{jj}(x_j) - M_\bfsnum{\cA}{jj}(x_j)\right)\right)_{\ell,\ell'}\right| \lesssim h_\cA\eta_{p,3}.
\end{align*}
To see this, observe that
\begin{align*}
\left(\tM_\bfsnum{a}{jj}(x_j) - \tM_\bfsnum{\cA}{jj}(x_j)\right)_{\ell, \ell'} &= \int_0^1 \left(\frac{u_j-x_j}{h_\bfsnum{\cA}{j}}\right)^{\ell+\ell'-2}K_{h_\bfsnum{\cA}{j}}(x_j, u_j) (p_\bfsnum{a}{j}(u_j) - p_\bfsnum{\cA}{j}(u_j))\duj.
\end{align*}
By Taylor's theorem, we have
\begin{align*}
p_\bfsnum{a}{j}(u_j) - p_\bfsnum{\cA}{j}(u_j) = p_\bfsnum{a}{j}(x_j) - p_\bfsnum{\cA}{j}(x_j) + \int_{x_j}^{u_j}\frac{\partial (p_\bfsnum{a}{j} - p_\bfsnum{\cA}{j})(t)}{\partial t}\, \rmd t.
\end{align*}
Combining this with the identity
\begin{align*}
\left(N_\bfsnum{\cA}{j}(x_j)(M_\bfsnum{a}{jj}(x_j) - M_\bfsnum{\cA}{jj}(x_j))\right)_{\ell, \ell'} = \mu_\bfsnum{\cA}{j,\ell+\ell'-2}(x_j)(p_\bfsnum{a}{j}(x_j) - p_\bfsnum{\cA}{j}(x_j)),
\end{align*}
we deduce that
\begin{align*}
&\left|\left(\tM_\bfsnum{a}{jj}(x_j) - \tM_\bfsnum{\cA}{jj}(x_j)\right)_{\ell, \ell'} - \left(N_\bfsnum{\cA}{j}(x_j)\left(M_\bfsnum{a}{jj}(x_j) - M_\bfsnum{\cA}{jj}(x_j)\right)\right)_{\ell,\ell'}\right| \\
&\le \left|\int_0^1 \left(\frac{u_j-x_j}{h_\bfsnum{\cA}{j}}\right)^{\ell+\ell'-2}K_{h_\bfsnum{\cA}{j}}(x_j, u_j)\int_{x_j}^{u_j}\frac{\partial (p_\bfsnum{a}{j} - p_\bfsnum{\cA}{j})(t)}{\partial t}\, \rmd t\duj\right| \\
&\le 2h_\bfsnum{\cA}{j}\eta_{p,3} \\
&\le \frac{2}{C_{h,L}}h_\cA\eta_{p,3}.
\end{align*}

The proof of \eqref{claim-2-2-lem2} follows similarly, so we only sketch the argument. By Taylor's theorem, we write
\begin{align*}
p_\bfsnum{a}{jk}(u_j, u_k) - p_\bfsnum{\cA}{jk}(u_j, u_k) &= p_\bfsnum{a}{jk}(x_j, x_k) - p_\bfsnum{\cA}{jk}(x_j,x_k) \\
&\quad  + \int_{x_k}^{u_k}\frac{\partial (p_\bfsnum{a}{jk}(x_j, \cdot) - p_\bfsnum{\cA}{jk}(x_j, \cdot))(t)}{\partial t}\, \rmd t \\
&\quad + \int_{x_j}^{u_j}\frac{\partial (p_\bfsnum{a}{jk}(\cdot,x_k) - p_\bfsnum{\cA}{jk}(\cdot, x_k))(t)}{\partial t}\, \rmd t.
\end{align*}
Moreover,
\begin{align*}
&\left(N_\bfsnum{\cA}{j}(x_j)L_\bfsnum{\cA}{k}(x_k)(M_\bfsnum{a}{jk}(x_j,x_k) - M_\bfsnum{\cA}{jk}(x_j,x_k))\right)_{\ell, \ell'} \\
&= \mu_\bfsnum{\cA}{j,\ell-1}(x_j)\mu_\bfsnum{\cA}{k,\ell'-1}(x_k)(p_\bfsnum{a}{jk}(x_j,x_k) - p_\bfsnum{\cA}{jk}(x_j, x_k)).
\end{align*}
It then follows that
\begin{align*}
&\left|\left(\tM_\bfsnum{a}{jk}(x_j,x_k) - \tM_\bfsnum{\cA}{jk}(x_j,x_k)\right)_{\ell, \ell'} - \left(N_\bfsnum{\cA}{j}(x_j)L_\bfsnum{\cA}{k}(x_k)(M_\bfsnum{a}{jk}(x_j,x_k) - M_\bfsnum{\cA}{jk}(x_j,x_k))\right)_{\ell, \ell'}\right| \\
&\le \Big|\int_{[0,1]^2} \left(\frac{x_j-u_j}{h_\bfsnum{\cA}{j}}\right)^{\ell-1} \left(\frac{x_k-u_k}{h_\bfsnum{\cA}{k}}\right)^{\ell'-1}K_{h_\bfsnum{\cA}{j}}(x_j,u_j)K_{h_\bfsnum{\cA}{k}}(x_k, u_k)  \\
&\qquad \times \int_{x_k}^{u_k}\frac{\partial (p_\bfsnum{a}{jk}(x_j, \cdot) - p_\bfsnum{\cA}{jk}(x_j, \cdot))(t)}{\partial t}\, \rmd t \duj\duk\Big| \\
&\quad +  \Big|\int_{[0,1]^2} \left(\frac{x_j-u_j}{h_\bfsnum{\cA}{j}}\right)^{\ell-1} \left(\frac{x_k-u_k}{h_\bfsnum{\cA}{k}}\right)^{\ell'-1}K_{h_\bfsnum{\cA}{j}}(x_j,u_j)K_{h_\bfsnum{\cA}{k}}(x_k, u_k) \\
&\qquad \times \int_{x_j}^{u_j}\frac{\partial (p_\bfsnum{a}{jk}(\cdot,x_k) - p_\bfsnum{\cA}{jk}(\cdot, x_k))(t)}{\partial t}\, \rmd t \duj\duk\Big| \\
&\le 2(h_\bfsnum{\cA}{j}+h_\bfsnum{\cA}{k})\eta_{p,3} \\
&\le \frac{4}{C_{h,L}}h_\cA \eta_{p,3}.
\end{align*}
Clearly, this shows \eqref{claim-2-2-lem2}.

\subsubsection{Proof of Theorem~\ref{thm:tl-aggr}}

For $j \in [d]$, define $\beta_\bfsnum{\cA}{j}^\rmtp := \hatf_\bfsnum{\cA}{j}^\rmtp - f_\bfsnum{\cA}{j}^\rmtp$ and let $\beta_\bfnum{\cA}^\rmtp := \sumj \beta_\bfsnum{\cA}{j}$. As in the proof of Theorem~\ref{thm:ll-bound-error-emp}, we begin by observing that
\begin{align*}
\hPi_\bfsnum{\cA}{j}(\beta_\cA^\rmtp )= \Delta_\bfsnum{\cA}{j}^\rmtp - \lambda_\cA^{\rmtl1}\nu_\bfsnum{\cA}{j}^\rmtp, 
\end{align*}
where $\nu_\bfsnum{\cA}{j}^\rmtp$ denotes a subgradient of $\norm{\cdot}_\hMnum{\cA}$ at $\hatf_\bfsnum{\cA}{j}^\rmtp$. This subgradient satisfies
\begin{align*}
\ip{\nu_\bfsnum{\cA}{j}^\rmtp, g_j^\rmtp}_\hMnum{\cA} \ge \norm{\hatf_\bfsnum{\cA}{j}^\rmtp}_\hMnum{\cA} - \norm{\hatf_\bfsnum{\cA}{j}^\rmtp - g_j^\rmtp}_\hMnum{\cA}, \quad g_j^\rmtp \in \sH_j^\rmtp.
\end{align*}
It follows that:
\begin{itemize}
\item When $j \in \cS_\bfnum{0}$,
\begin{align*}
\ip{\nu_\bfsnum{\cA}{j}^\rmtp, \beta_\bfsnum{\cA}{j}^\rmtp}_\hMnum{\cA} \ge \norm{\hatf_\bfsnum{\cA}{j}^\rmtp}_\hMnum{\cA} - \norm{f_\bfsnum{\cA}{j}^\rmtp}_\hMnum{\cA} \ge -\norm{\beta_\bfsnum{\cA}{j}^\rmtp}_\hMnum{\cA};
\end{align*}
\item When $j \notin \cS_\bfnum{0}$,
\begin{align*}
\ip{\nu_\bfsnum{\cA}{j}^\rmtp, \beta_\bfsnum{\cA}{j}^\rmtp}_\hMnum{\cA} \ge \norm{\hatf_\bfsnum{\cA}{j}^\rmtp}_\hMnum{\cA} - \norm{\delta_\bfsnum{\cA}{j}^\rmtp}_\hMnum{\cA} \ge \norm{\beta_\bfsnum{\cA}{j}^\rmtp}_\hMnum{\cA} - 2\norm{\delta_\bfsnum{\cA}{j}^\rmtp}_\hMnum{\cA}.
\end{align*}
\end{itemize}
Combining these yields
\begin{align} \label{first-thm4}
\begin{aligned}
\norm{\beta_\bfnum{\cA}^\rmtp}_\hMnum{\cA}^2 &= \sumj \ip{\Delta_\bfsnum{\cA}{j}^\rmtp- \lambda_\cA^{\rmtl1}\nu_\bfsnum{\cA}{j}^\rmtp, \beta_\bfsnum{\cA}{j}^\rmtp}_\hMnum{\cA} \\
&\le (\Delta_\cA+\lambda_\cA^{\rmtl1})\sumjs \norm{\beta_\bfsnum{\cA}{j}^\rmtp}_\hMnum{\cA} + (\Delta_\cA - \lambda_\cA^{\rmtl1})\sumjnots \norm{\beta_\bfsnum{\cA}{j}^\rmtp}_\hMnum{\cA} \\
&\hspace{5cm}+ \sqrt{\frac{12C_{p,U}^\rmuniv}{C_{p,L}^\rmuniv\mu_2}}\lambda_\cA^{\rmtl1}(\eta_{\delta, \cS_\bfnum{0}^c} + \eta_{p,\delta,\cS_\bfnum{0}^c}) \\
&\le \frac{\mathfrak{C}_\cA+1}{\mathfrak{C}_\cA}\lambda_\cA^{\rmtl1}\sumjs \norm{\beta_\bfsnum{\cA}{j}^\rmtp}_\hMnum{\cA}- \frac{\mathfrak{C}_\cA-1}{\mathfrak{C}_\cA}\lambda_\cA^{\rmtl1}\sumjnots \norm{\beta_\bfsnum{\cA}{j}^\rmtp}_\hMnum{\cA} \\
&\hspace{5cm} +\sqrt{\frac{12C_{p,U}^\rmuniv}{C_{p,L}^\rmuniv\mu_2}}\lambda_\cA^{\rmtl1}(\eta_{\delta, \cS_\bfnum{0}^c} + \eta_{p,\delta,\cS_\bfnum{0}^c}).
\end{aligned}
\end{align}
Here, we have used the fact that the inequality
\begin{align*}
\norm{g_j^\rmtp}_\hMnum{\cA} \le \sqrt{\frac{3C_{p,U}^\rmuniv}{C_{p,L}^\rmuniv\mu_2}}\norm{g_j^\rmtp}_\Mnum{0}, \quad g_j^\rmtp \in \sH_j^\rmtp
\end{align*}
holds with probability tending to one.

Next, we consider two cases separately. The first case is when 
\begin{align} \label{case-1-1-thm4}
\sumjs \norm{\beta_\bfsnum{\cA}{j}^\rmtp}_\hMnum{\cA} \le \mathfrak{C}_\cA\sqrt{\frac{12C_{p,L}^\rmuniv}{C_{p,L}^\rmuniv\mu_2 }}(\eta_{\delta, \cS_\bfnum{0}^c}+\eta_{p,\delta,\cS_\bfnum{0}^c}).
\end{align}
Under the condition in \eqref{case-1-1-thm4}, it follows that 
\begin{align*}
\norm{\beta_\cA^\rmtp}_\hMnum{\cA}^2 + \frac{\mathfrak{C}_\cA-1}{\mathfrak{C}_\cA}\lambda_\cA^{\rmtl1}\sumjnots \norm{\beta_\bfsnum{\cA}{j}^\rmtp}_\hMnum{\cA} \le  (\mathfrak{C}_\cA+2)\sqrt{\frac{12C_{p,L}^\rmuniv}{C_{p,L}^\rmuniv\mu_2 }}\lambda_\cA^{\rmtl1}(\eta_{\delta, \cS_\bfnum{0}^c}+\eta_{p,\delta,\cS_\bfnum{0}^c}).
\end{align*}
This implies that
\begin{align} \label{case-1-order-1-thm4}
\norm{\beta_\cA^\rmtp}_\hMnum{\cA}^2 \lesssim \lambda_\cA^{\rmtl1}(\eta_{\delta, \cS_\bfnum{0}^c}+\eta_{p,\delta,\cS_\bfnum{0}^c}).
\end{align}
Moreover, since
\begin{align*}
\sumjnots \norm{\beta_\bfsnum{\cA}{j}^\rmtp}_\hMnum{\cA} \lesssim \eta_{\delta, \cS_\bfnum{0}^c}+\eta_{p,\delta,\cS_\bfnum{0}^c}, 
\end{align*}
together with \eqref{case-1-order-1-thm4}, we also obtain
\begin{align} \label{case-1-order-2-thm4}
\norm{\beta_\cA^\rmtp}_\hMnum{\cA}^2 \le \left(\sumj \norm{\beta_\bfsnum{\cA}{j}^\rmtp}_\hMnum{\cA}\right)^2 \lesssim (\eta_{\delta, \cS_\bfnum{0}^c}+\eta_{p,\delta,\cS_\bfnum{0}^c})^2. 
\end{align}
Combining \eqref{case-1-order-1-thm4} and \eqref{case-1-order-2-thm4}, we conclude that
\begin{align*}
\norm{\beta_\cA^\rmtp}_\hMnum{\cA}^2 \lesssim \lambda_\cA^{\rmtl1}(\eta_{\delta, \cS_\bfnum{0}^c}+\eta_{p,\delta,\cS_\bfnum{0}^c}) \wedge (\eta_{\delta, \cS_\bfnum{0}^c}+\eta_{p,\delta,\cS_\bfnum{0}^c})^2. 
\end{align*}
This establishes the desired result in the case of \eqref{case-1-1-thm4}.

Secondly, we consider the complementary case where
\begin{align} \label{case-1-2-thm4}
\sumjs \norm{\beta_\bfsnum{\cA}{j}^\rmtp}_\hMnum{\cA} > \mathfrak{C}_\cA\sqrt{\frac{12C_{p,L}^\rmuniv}{C_{p,L}^\rmuniv\mu_2}}(\eta_{\delta, \cS_\bfnum{0}^c}+\eta_{p,\delta,\cS_\bfnum{0}^c}).
\end{align}
In this case, we observe that
\begin{align*}
\norm{\beta_\cA^\rmtp}_\hMnum{\cA}^2 \le \frac{\mathfrak{C}_\cA+2}{\mathfrak{C}_\cA}\lambda_\cA^{\rmtl1} \sumjs \norm{\beta_\bfsnum{\cA}{j}^\rmtp}_\hMnum{\cA} - \frac{\mathfrak{C}_\cA-1}{\mathfrak{C}_\cA} \sumjnots \norm{\beta_\bfsnum{\cA}{j}^\rmtp}_\hMnum{\cA}. 
\end{align*}
This implies that 
\begin{align} \label{clue-1-thm4}
\sumjs \norm{\beta_\bfsnum{\cA}{j}^\rmtp}_\hMnum{\cA} \le \frac{\mathfrak{C}_\cA-1}{\mathfrak{C}_\cA+2} \sumjnots \norm{\beta_\bfsnum{\cA}{j}^\rmtp}_\hMnum{\cA},
\end{align}
and 
\begin{align} \label{clue-2-thm4}
\norm{\beta_\cA^\rmtp}_\hMnum{\cA}^2 \le \frac{\mathfrak{C}_\cA+2}{\mathfrak{C}_\cA} \lambda_\cA^{\rmtl1} \sumjs \norm{\beta_\bfsnum{\cA}{j}^\rmtp}_\hMnum{\cA}.
\end{align}

For convenience, let $\sD_\cA := \sumjs \norm{\beta_\bfsnum{\cA}{j}^\rmtp}_\hMnum{\cA}$. We now establish the theorem under the condition in \eqref{case-1-2-thm4}, utilizing the compatibility condition stated in terms of the norm $\norm{\cdot}_\tMnum{\cA}$. For each $j\in[d]$, define 
\[
\cD_\bfsnum{\cA}{j} := \max(\norm{\beta_\bfsnum{\cA}{j}^\rmtp}_\hMnum{\cA} - \norm{\beta_\bfsnum{\cA}{j}^{\rmtp, \tc}}_\hMnum{\cA},0),
\]
where $\beta_\bfsnum{\cA}{j}^{\rmtp, \tc} := \beta_\bfsnum{\cA}{j}^\rmtp - \tPi_\bfsnum{\cA}{0}(\beta_\bfsnum{\cA}{j}^\rmtp)$. We claim that
\begin{align} \label{claim-thm4}
\sumjs \cD_\bfsnum{\cA}{j} \lesssim |\cS_\bfnum{0}|\left(h_\cA^2 + \sqrt{\frac{\log (|\cS_\bfnum{0}|\vee n_\cA)}{n_\cA}}\right) + \eta_{p,\delta, \cS_\bfnum{0}} + \eta_{p,2}\eta_{\delta, \cS_\bfnum{0}}.
\end{align}
The proof of this claim is deferred to the end of the argument.
Since
\begin{align*}
\sD_\cA \le \sumjs \norm{\beta_\bfsnum{\cA}{j}^{\rmtp, \tc}}_\hMnum{\cA} + \sumjs \cD_\bfsnum{\cA}{j},
\end{align*}
the theorem follows directly from the claim \eqref{claim-thm4} whenever $\sumjs \norm{\beta_\bfsnum{\cA}{j}^{\rmtp, \tc}}_\hMnum{\cA} \le \sumjs \cD_\bfsnum{\cA}{j}$. Therefore, in the following, we restrict our attention to the case where $\sumjs \norm{\beta_\bfsnum{\cA}{j}^{\rmtp, \tc}}_\hMnum{\cA} > \sumjs \cD_\bfsnum{\cA}{j}$. Under this condition, we have
\begin{align} \label{sD-condition-thm4}
\sD_\cA \le \sumjs \norm{\beta_\bfsnum{\cA}{j}^{\rmtp, \tc}}_\hMnum{\cA}.
\end{align}
Let $\xi_\cA > 0$ be a sufficiently small constant such that
\begin{align*}
2\frac{\mathfrak{C}_\cA+2}{\mathfrak{C}_\cA-1} \le 2\sqrt{\frac{1+\xi_\cA}{1-\xi_\cA}}\frac{\mathfrak{C}_\cA+2}{\mathfrak{C}_\cA-1} \le C_\cA,
\end{align*}
where $C_\cA$ is the constant defined in the statement of the theorem. By an argument analogous to that used in the proof of Lemma~\ref{lem:tech1}, we may establish that
\begin{align} \label{eigenvalue-thm4}
\begin{aligned}
1 - \xi_\cA &\le \min_{j \in [d]} \inf_{x_j \in [0,1]} \lambda_\rmmin\left(\tM_\bfsnum{\cA}{jj}(x_j)^{-\frac{1}{2}} \hM_\bfsnum{\cA}{jj}(x_j) \tM_\bfsnum{\cA}{jj}(x_j)^{-\frac{1}{2}}\right) \\
&\le \max_{j \in [d]} \sup_{x_j \in [0,1]} \lambda_\rmmax\left(\tM_\bfsnum{\cA}{jj}(x_j)^{-\frac{1}{2}} \hM_\bfsnum{\cA}{jj}(x_j) \tM_\bfsnum{\cA}{jj}(x_j)^{-\frac{1}{2}}\right) \le 1 + \xi_\cA.
\end{aligned}
\end{align}
Combining \eqref{clue-1-thm4}, \eqref{sD-condition-thm4}, and \eqref{eigenvalue-thm4} with the definition of $\xi_\cA$, we obtain
\begin{align*}
\sumjnots \norm{\beta_\bfsnum{\cA}{j}^{\rmtp, \tc}}_\tMnum{\cA}
&\le \sumjnots \norm{\beta_\bfsnum{\cA}{j}^\rmtp}_\tMnum{\cA} \\
&\le \sqrt{\frac{1}{1 - \xi_\cA}} \sumjnots \norm{\beta_\bfsnum{\cA}{j}^\rmtp}_\hMnum{\cA} \\
&\le \sqrt{\frac{1}{1 - \xi_\cA}} \frac{\mathfrak{C}_\cA+2}{\mathfrak{C}_\cA-1} \sumjs \norm{\beta_\bfsnum{\cA}{j}^\rmtp}_\hMnum{\cA} \\
&\le 2 \sqrt{\frac{1}{1 - \xi_\cA}} \frac{\mathfrak{C}_\cA+2}{\mathfrak{C}_\cA-1} \sumjs \norm{\beta_\bfsnum{\cA}{j}^{\rmtp, \tc}}_\hMnum{\cA} \\
&\le 2 \sqrt{\frac{1+\xi_\cA}{1 - \xi_\cA}} \frac{\mathfrak{C}_\cA+2}{\mathfrak{C}_\cA-1} \sumjs \norm{\beta_\bfsnum{\cA}{j}^{\rmtp, \tc}}_\tMnum{\cA} \\
&\le C_\cA \sumjs \norm{\beta_\bfsnum{\cA}{j}^{\rmtp, \tc}}_\tMnum{\cA}.
\end{align*}
Let $\beta_\cA^{\rmtp, \tc} := \sumj \beta_\bfsnum{\cA}{j}^{\rmtp, \tc}$. By the definition of the compatibility constant $\phi_\cA(\cdot)$, we conclude that
\begin{align} \label{compat-thm4}
\norm{\beta_\cA^{\rmtp, \tc}}_\tMnum{\cA}^2 \ge \phi_\cA(C_\cA) \sumjs \norm{\beta_\cA^{\rmtp, \tc}}_\tMnum{\cA}^2.
\end{align}

From the compatibility inequality in \eqref{compat-thm4}, we obtain
\begin{align} \label{res-compat-thm4}
\begin{aligned}
\sD_\cA^2 &= \left(\sumjs \norm{\beta_\bfsnum{\cA}{j}^\rmtp}_\hMnum{\cA}\right)^2 \\
&\le \left(\sumjs \norm{\beta_\bfsnum{\cA}{j}^{\rmtp, \tc}}_\hMnum{\cA} + \sumjs \cD_\bfsnum{\cA}{j}\right)^2 \\
&\le 2|\cS_\bfnum{0}|\sumjs \norm{\beta_\bfsnum{\cA}{j}^{\rmtp, \tc}}_\hMnum{\cA}^2 + 2\left(\sumjs \cD_\bfsnum{\cA}{j}\right)^2 \\
&\le 2(1+\xi_\bfnum{0})|\cS_\bfnum{0}|\sumjs \norm{\beta_\bfsnum{\cA}{j}^{\rmtp, \tc}}_\tMnum{\cA}^2 + 2\left(\sumjs \cD_\bfsnum{\cA}{j}\right)^2 \\
&\le 2(1+\xi_\cA)\frac{|\cS_\bfnum{0}|}{\phi_\cA(C_\cA)}\norm{\beta_\cA^{\rmtp, \tc}}_\tMnum{\cA}^2 + 2\left(\sumjs \cD_\bfsnum{\cA}{j}\right)^2.
\end{aligned}
\end{align}
Using arguments similar to those leading to \eqref{main-claim-thm1} in the proof of Theorem~\ref{thm:ll-bound-error-emp}, we may show that there exists an absolute constant $0 < \Cs_\cA < \infty$ such that
\begin{align} \label{approx-thm4}
\norm{\beta_\cA^{\rmtp, \tc}}_\tMnum{\cA}^2 \le \norm{\beta_\cA^\rmtp}_\hMnum{\cA}^2 + \Cs_\cA\left(\frac{1}{n_\cA h_\cA^2} + B(n_\cA, h_\cA^2, d)\right)^{\frac{1}{2}} \sD_\cA^2.
\end{align}
Recalling the order condition imposed on $|\cS_\bfnum{0}|$, we may ensure that for sufficiently large $n_\bfnum{0}$, the inequality
\begin{align} \label{last-clue-thm4}
2\Cs_\cA(1+\xi_\cA)\frac{|\cS_\bfnum{0}|}{\phi_\cA(C_\cA)}\left(\frac{1}{n_\cA h_\cA^2} + B(n_\cA, h_\cA^2, d)\right)^{\frac{1}{2}} \le \xi_\cA
\end{align}
holds. Combining \eqref{clue-2-thm4}, \eqref{res-compat-thm4}, \eqref{approx-thm4}, and \eqref{last-clue-thm4}, we obtain
\begin{align*}
\sD_\cA^2 &\le 2\frac{1+\xi_\cA}{1-\xi_\cA}\frac{|\cS_\bfnum{0}|}{\phi_\cA(C_\cA)}\norm{\beta_\cA^\rmtp}_\hMnum{\cA}^2 + \frac{2}{1-\xi_\cA}\left(\sumjs \cD_\bfsnum{\cA}{j}\right)^2 \\
&\le |\cS_\bfnum{0}|\frac{1+\xi_\cA}{1-\xi_\cA}\frac{\mathfrak{C}_\cA+2}{\mathfrak{C}_\cA}\frac{\lambda_\cA^{\rmtl1}}{\phi_\cA(C_\cA)}\sD_\cA + \frac{2}{1-\xi_\cA}\left(\sumjs \cD_\bfsnum{\cA}{j}\right)^2,
\end{align*}
which, in conjunction with the claim in \eqref{claim-thm4}, completes the proof of the theorem.

It remains to prove the claim \eqref{claim-thm4}. We note that this step constitutes the most distinctive part of the present proof, in contrast to the argument used in Theorem~\ref{thm:ll-bound-error-emp}.

\smallskip\noindent
{\it Proof of \eqref{claim-thm4}.}
\smallskip

Observe that 
\begin{align*}
\norm{\beta_\bfsnum{\cA}{j}^{\rmtp, \tc}}_\hMnum{\cA} &= \norm{\beta_\bfsnum{\cA}{j}^\rmtp - \tPi_\bfsnum{\cA}{j}(\beta_\bfsnum{\cA}{j}^\rmtp)}_\hMnum{\cA} \\
&=\norm{\beta_\bfsnum{\cA}{j}^\rmtp - \hPi_\bfsnum{\cA}{0}(\beta_\bfsnum{\cA}{j}^\rmtp) + \hPi_\bfsnum{\cA}{0}(\beta_\bfsnum{\cA}{j}^\rmtp) - \tPi_\bfsnum{\cA}{j}(\beta_\bfsnum{\cA}{j})}_\hMnum{\cA} \\
&\ge\norm{\beta_\bfsnum{\cA}{j}^\rmtp - \hPi_\bfsnum{\cA}{0}(f_\bfsnum{\cA}{j}^\rmtp)}_\hMnum{\cA} \\
&\ge \norm{\beta_\bfsnum{\cA}{j}^\rmtp}_\hMnum{\cA} - \norm{\hPi_\bfsnum{\cA}{0}(f_\bfsnum{\cA}{j}^\rmtp)}_\hMnum{\cA}.
\end{align*}
This implies that 
\begin{align} \label{clue-all-cl}
\begin{aligned}
\cD_\bfsnum{\cA}{j}  = \norm{\beta_\bfsnum{\cA}{j}^{\rmtp, \tc}}_\hMnum{\cA} - \norm{\beta_\bfsnum{\cA}{j}^\rmtp}_\hMnum{\cA} &\le \norm{\hPi_\bfsnum{\cA}{0}(f_\bfsnum{\cA}{j}^\rmtp)}_\hMnum{\cA} \\
&\le \norm{\tPi_\bfsnum{\cA}{0}(f_\bfsnum{\cA}{j}^\rmtp)}_\hMnum{\cA} + \norm{(\hPi_\bfsnum{\cA}{0}-\tPi_\bfsnum{\cA}{0})(f_\bfsnum{\cA}{j}^\rmtp)}_\hMnum{\cA}.
\end{aligned}
\end{align}
We now bound each term on the right-hand side in \eqref{clue-all-cl}. For the first term, we have
\begin{align*}
\norm{\tPi_\bfsnum{\cA}{0}(f_\bfsnum{\cA}{j}^\rmtp)}_\hMnum{\cA} &= \left|\int_0^1 f_\bfsnum{\cA}{j}^\rmv(x_j)^\top \tp_\bfsnum{\cA}{j}^\rmv(x_j)\dxj\right| \\
&\le \left|\sumaac w_\ab \int_0^1 f_\bfsnum{a}{j}^\rmv(x_j)^\top \tp_\bfsnum{\cA}{j}^\rmv(x_j)\dxj\right| \\
&\hspace{3cm} + \left|\int_0^1 \left(f_\bfsnum{\cA}{j}^\rmv(x_j) - \sumaac w_\ab f_\bfsnum{a}{j}^\rmv(x_j)\right)^\top\tp_\bfsnum{\cA}{j}^\rmv(x_j)\dxj\right|.
\end{align*} 
Note that
\begin{align} \label{clue-1-0-cl}
\begin{aligned}
&\sumaac w_\ab \int_0^1 f_\bfsnum{a}{j}^\rmv(x_j)^\top \tp_\bfsnum{\cA}{j}^\rmv(x_j)\dxj \\
&= \sumaac w_\ab\int_0^1 f_\bfsnum{a}{j}^\rmv(x_j)^\top\left(\tp_\bfsnum{\cA}{j}^\rmv(x_j) - \tp_\bfsnum{a}{j}^\rmv(x_j)\right)\dxj  + O(h_\cA^2) \\
&= \sumaac w_\ab\int_0^1 \delta_\bfsnum{a}{j}^\rmv(x_j)^\top \left(\tp_\bfsnum{\cA}{j}^\rmv(x_j) - \tp_\bfsnum{a}{j}^\rmv(x_j)\right)\dxj + O(h_\cA^2). 
\end{aligned}
\end{align}
uniformly over $j\in[d]$ and $\ab\in\cA$. Here, we used the identity $\sumaac w_\ab \tp_\bfsnum{a}{j}^\rmv = \tp_\bfsnum{\cA}{j}^\rmv$ for the last equality. Also, it holds that
\begin{align} \label{clue-1-1-cl}
\begin{aligned}
&\int_0^1 \int_0^1 \left(\delta_\bfsnum{a}{j}(x_j) + (u_j-x_j)f_\bfsnum{a}{j}'(x_j)\right) K_{h_\bfsnum{\cA}{j}}(x_j, u_j) \left(p_\bfsnum{\cA}{j}(u_j)-p_\bfsnum{a}{j}(u_j)\right)\dxj\duj \\
&=\int_0^1 \int_0^1 \delta_\bfsnum{a}{j}(u_j) K_{h_\bfsnum{\cA}{j}}(x_j, u_j) \left(p_\bfsnum{\cA}{j}(u_j)-p_\bfsnum{a}{j}(u_j)\right)\dxj\duj  + O(h_\cA^2) \\
&= \int_0^1  \delta_\bfsnum{a}{j}(u_j) \left(p_\bfsnum{\cA}{j}(u_j)-p_\bfsnum{a}{j}(u_j)\right)\duj + O(h_\cA^2), 
\end{aligned}
\end{align}
uniformly over $j\in[d]$ and $\ab\in\cA$. From \eqref{clue-1-1-cl} together with \eqref{clue-1-0-cl}, it follows that
\begin{align} \label{result-1-1-cl}
\sumjs  \left|\sumaac w_\ab \int_0^1 f_\bfsnum{a}{j}^\rmv(x_j)^\top \tp_\bfsnum{\cA}{j}^\rmv(x_j)\dxj\right| \lesssim |\cS_\bfnum{0}|h_\cA^2 + \eta_{p,2}\eta_{\delta, \cS_\bfnum{0}}. 
\end{align}
Moreover, standard kernel smoothing theory implies that each entry of $\tp_\bfsnum{\cA}{j}^\rmv$ is uniformly bounded. Thus, applying Hölder's inequality yields
\begin{align} \label{result-1-2-cl}
\sumjs \left|\int_0^1 \left(f_\bfsnum{\cA}{j}^\rmv(x_j) - \sumaac w_\ab f_\bfsnum{a}{j}^\rmv(x_j)\right)^\top\tp_\bfsnum{\cA}{j}^\rmv(x_j)\dxj\right| \lesssim \eta_{p, \delta,\cS_\bfnum{0}}. 
\end{align}
Combining \eqref{result-1-1-cl} and \eqref{result-1-2-cl}, we obtain
\begin{align} \label{result-1-cl}
\sumjs \norm{\tPi_\bfsnum{\cA}{0}(f_\bfsnum{\cA}{j}^\rmtp)}_\hMnum{\cA} \lesssim |\cS_\bfnum{0}|h_\cA^2 + \eta_{p,2}\eta_{\delta, \cS_\bfnum{0}} + \eta_{p, \delta,\cS_\bfnum{0}}.
\end{align}

For the second term in \eqref{clue-all-cl}, we observe that 
\begin{align*}
\norm{(\hPi_\bfsnum{\cA}{0}-\tPi_\bfsnum{\cA}{0})(f_\bfsnum{\cA}{j}^\rmtp)}_\hMnum{\cA} &= \left|\int_0^1 f_\bfsnum{\cA}{j}^\rmv(x_j)^\top \left(\hp_\bfsnum{\cA}{j}(x_j) - \tp_\bfsnum{\cA}{j}^\rmv(x_j)\right)\dxj \right| \\
&\le \left|\sumaac w_\ab\int_0^1 f_\bfsnum{a}{j}^\rmv(x_j)^\top \left(\hp_\bfsnum{\cA}{j}^\rmv(x_j) - \tp_\bfsnum{\cA}{j}^\rmv(x_j)\right)\dxj\right| \\
&\quad + \left|\int_0^1 \left(f_\bfsnum{\cA}{j}^\rmv(x_j) - \sumaac w_\ab f_\bfsnum{a}{j}(x_j)\right)^\top \left(\hp_\bfsnum{\cA}{j}^\rmv(x_j) - \tp_\bfsnum{\cA}{j}^\rmv(x_j)\right)\dxj \right|.
\end{align*}
For each $\ab\in\cA$, it can be shown—along similar lines as the proof of Theorem~\ref{thm:ll-bound-error-emp}—that there exists an absolute constant $0<C_1<\infty$ such that 
\begin{align*}
\max_{j\in\cS_\bfnum{0}}\left|\int_0^1 f_\bfsnum{a}{j}^\rmv(x_j)^\top \left(\hp_\bfsnum{b}{j}^\rmv(x_j) - \tp_\bfsnum{b}{j}^\rmv(x_j)\right)\dxj\right| \le C_1\sqrt{\frac{\log (|\cS_\bfnum{0}|\vee n_\bfnum{b})}{n_\bfnum{b}}}\le C_1\sqrt{\frac{\log (|\cS_\bfnum{0}|\vee n_\bfnum{\cA})}{n_\bfnum{b}}}
\end{align*}
with probability tending to one for all $\bfnum{b}\in\cA$. Since $|\cA|<\infty$, it follows that 
\begin{align*}
&\Pb\left(\max_{j\in\cS_\bfnum{0}}\left|\int_0^1 f_\bfsnum{a}{j}^\rmv(x_j)^\top \left(\hp_\bfsnum{\cA}{j}^\rmv(x_j) - \tp_\bfsnum{\cA}{j}^\rmv(x_j)\right)\dxj\right| \ge |\cA|C_1\sqrt{\frac{\log (|\cS_\bfnum{0}|\vee n_\bfnum{\cA})}{n_\bfnum{\cA}}}\right) \\
&\le \Pb\left(\sum_{\bfnum{b}\in\cA} w_\bfnum{b}\cdot \max_{j\in\cS_\bfnum{0}}\left|\int_0^1 f_\bfsnum{a}{j}^\rmv(x_j)^\top \left(\hp_\bfsnum{b}{j}^\rmv(x_j) - \tp_\bfsnum{b}{j}^\rmv(x_j)\right)\dxj\right| \ge |\cA|C_1\sqrt{\frac{\log (|\cS_\bfnum{0}|\vee n_\bfnum{\cA})}{n_\bfnum{\cA}}}\right) \\
&\le \sum_{\bfnum{b}\in\cA}\Pb\left(w_\bfnum{b}\cdot \max_{j\in\cS_\bfnum{0}}\left|\int_0^1 f_\bfsnum{a}{j}^\rmv(x_j)^\top \left(\hp_\bfsnum{b}{j}^\rmv(x_j) - \tp_\bfsnum{b}{j}^\rmv(x_j)\right)\dxj\right| \ge C_1\sqrt{\frac{\log (|\cS_\bfnum{0}|\vee n_\bfnum{\cA})}{n_\bfnum{\cA}}}\right) \\
&\le \sum_{\bfnum{b}\in\cA}\Pb\left(\max_{j\in\cS_\bfnum{0}}\left|\int_0^1 f_\bfsnum{a}{j}^\rmv(x_j)^\top \left(\hp_\bfsnum{b}{j}^\rmv(x_j) - \tp_\bfsnum{b}{j}^\rmv(x_j)\right)\dxj\right| \ge C_1\sqrt{\frac{\log (|\cS_\bfnum{0}|\vee n_\bfnum{\cA})}{n_\bfnum{b}}}\right) \\
&= o(1).
\end{align*}
Therefore, we obtain 
\begin{align} \label{result-2-1-cl}
\max_{j\in\cS_\bfnum{0}}\left|\sumaac w_\ab\int_0^1 f_\bfsnum{a}{j}^\rmv(x_j)^\top \left(\hp_\bfsnum{\cA}{j}^\rmv(x_j) - \tp_\bfsnum{\cA}{j}^\rmv(x_j)\right)\dxj\right| &\lesssim \sqrt{\frac{\log (|\cS_\bfnum{0}|\vee n_\cA)}{n_\cA}}. 
\end{align}
Next, using arguments analogous to those in the proof of Lemma~\ref{lem:u-stat2}, we may show that 
\begin{align} \label{clue-2-0-cl}
\max_{j\in\cS_\bfnum{0}}\norm{U_j^\top \cdot (\hp_\bfsnum{\cA}{j}^\rmv - \tp_\bfsnum{\cA}{j}^\rmv)}_{I_{d+1}}\lesssim \left(\frac{1}{n_\cA h_\cA} + B(n_\cA, h_\cA, |\cS_\bfnum{0}|)\right)^{\frac{1}{2}}. 
\end{align}
Also, we have
\begin{align} \label{clue-2-1-cl}
\sumjs \normbig{f_\bfsnum{\cA}{j}^\rmv - \sumaac w_\ab f_\bfsnum{a}{j}} = \sumjs \normbig{\delta_\bfsnum{\cA}{j}^\rmv - \sumaac w_\ab \delta_\bfsnum{a}{j}}_{I_{d+1}} \lesssim \eta_{p,\delta, \cS_\bfnum{0}}. 
\end{align}
From \eqref{clue-2-0-cl} together with \eqref{clue-2-1-cl}, we get
\begin{align} \label{result-2-2-cl}
\begin{aligned}
&\sumjs\left|\int_0^1 \left(f_\bfsnum{\cA}{j}^\rmv(x_j) - \sumaac w_\ab f_\bfsnum{a}{j}(x_j)\right)^\top \left(\hp_\bfsnum{\cA}{j}^\rmv(x_j) - \tp_\bfsnum{\cA}{j}^\rmv(x_j)\right)\dxj \right| \\
&\lesssim \left(\frac{1}{n_\cA h_\cA} + B(n_\cA, h_\cA, |\cS_\bfnum{0}|)\right)^{\frac{1}{2}}\eta_{p,\delta, \cS_\bfnum{0}}. 
\end{aligned}
\end{align}
Combining \eqref{result-2-1-cl} and \eqref{result-2-2-cl}, we obtain
\begin{align} \label{result-2-cl}
\sumjs \norm{(\hPi_\bfsnum{\cA}{0}-\tPi_\bfsnum{\cA}{0})(f_\bfsnum{\cA}{j}^\rmtp)}_\hMnum{\cA} \lesssim |\cS_\bfnum{0}|\sqrt{\frac{\log (|\cS_\bfnum{0}|\vee n_\cA)}{n_\cA}} + \left(\frac{1}{n_\cA h_\cA} + B(n_\cA, h_\cA, |\cS_\bfnum{0}|)\right)^{\frac{1}{2}}\eta_{p,\delta, \cS_\bfnum{0}}.
\end{align}
Finally, results in \eqref{result-1-cl} and \eqref{result-2-cl} complete the proof of \eqref{claim-thm4} as 
\begin{align*}
\frac{1}{n_\cA h_\cA} + B(n_\cA, h_\cA, |\cS_\bfnum{0}|) \ll 1. 
\end{align*}

\subsubsection{Proof of Theorem~\ref{thm:tl-debias}}

Recall the definitions of $\Delta_\bfsnum{0}{j}^\rmtp$ and $\Delta_\bfnum{0}$ introduced in Theorem~\ref{thm:ll-bound-error-emp}. Define $\gamma_\bfsnum{\cA}{j}^\rmtp := \hdelta_\bfsnum{\cA}{j}^\rmtp - \delta_\bfsnum{\cA}{j}^{\rmtp}$ and let $\gamma_\bfnum{\cA}^\rmtp := \sumj \gamma_\bfsnum{\cA}{j}^\rmtp$. Let $\tnu_\bfsnum{\cA}{j}^\rmtp$ denote the sub-gradient of $\norm{\cdot}_\hMnum{0}$ evaluated at $\hdelta_\bfsnum{\cA}{j}^\rmtp$. We observe that
\begin{align} \label{sub-grad-thm5}
\ip{\tnu_\bfsnum{\cA}{j}^\rmtp, \gamma_\bfsnum{\cA}{j}^\rmtp}_\hMnum{0}\ge \norm{\hdelta_\bfsnum{\cA}{j}^\rmtp}_\hMnum{0} - \norm{\delta_\bfsnum{\cA}{j}^\rmtp}_\hMnum{0} \le \norm{\gamma_\bfsnum{\cA}{j}^\rmtp}_\hMnum{0} - 2\norm{\delta_\bfsnum{\cA}{j}^\rmtp}_\hMnum{0}, \quad j\in[d].
\end{align}
Recall that $\hf_\bfsnum{\cA}{j}^{\rmtp, \hc} := \hf_\bfsnum{\cA}{j}^\rmtp - \hPi_\bfsnum{0}{0}(\hf_\bfsnum{\cA}{j}^\rmtp)$ and define $\hf_\bfnum{\cA}^{\rmtp, \hc} := \sumj \hf_\bfsnum{\cA}{j}^{\rmtp, \hc}$. Let $\beta_\cA^{\rmtp, \hc} := \beta_\cA^\rmtp - \hPi_\bfsnum{0}{0}(\beta_\cA^\rmtp)$. Since
\begin{align*}
\hatm_\bfsnum{0}{j}^\rmtp = \hPi_\bfsnum{0}{j}(\hf_\bfnum{\cA}^{\rmtp, \hc} + \hdelta_\bfnum{\cA}^{\rmtp}) + \lambda_\cA^{\rmtl2}\tnu_\bfsnum{\cA}{j}^\rmtp,
\end{align*}
we deduce from \eqref{sub-grad-thm5} that
\begin{align} \label{main-eq-thm5}
\begin{aligned}
\norm{\gamma_\cA^\rmtp}_\hMnum{0}^2 &= \sumj \ip{\hPi_\bfsnum{0}{j}(\gamma_\cA^\rmtp), \gamma_\bfsnum{\cA}{j}^\rmtp}_\hMnum{0} \\
&=\sumj \ip{\Delta_\bfsnum{0}{j}^\rmtp - \hPi_\bfsnum{0}{j}(\beta_\cA^{\rmtp, \hc}) - \lambda_\cA^{\rmtl2}\tnu_\bfsnum{\cA}{j}^\rmtp, \gamma_\bfsnum{\cA}{j}^\rmtp}_\hMnum{0} + \ip{\hPi_\bfsnum{0}{0}(f_\cA^\rmtp), \hPi_\bfsnum{0}{0}(\delta_\cA^\rmtp)}_\hMnum{0} \\
&\le -\left(\frac{\mathfrak{C}_\bfnum{0}'-1}{\mathfrak{C}_\bfnum{0}'}\right)\lambda_\cA^{\rmtl2}\sumj \norm{\gamma_\bfsnum{\cA}{j}^\rmtp}_\hMnum{0} + \norm{\beta_\cA^{\rmtp,\hc}}_\hMnum{0}\norm{\gamma_\cA^\rmtp}_\hMnum{0} \\
&\quad + 2\lambda_\cA^{\rmtl2}\sumj \norm{\delta_\bfsnum{\cA}{j}^\rmtp}_\hMnum{0}+  \ip{\hPi_\bfsnum{0}{0}(f_\cA^\rmtp), \hPi_\bfsnum{0}{0}(\delta_\cA^\rmtp)}_\hMnum{0}.
\end{aligned}
\end{align}
Here, we have used the fact that $\hdelta_\bfsnum{\cA}{j}^\rmtp$ is orthogonal to $\Rb^\rmtp$ under the inner product $\ip{\cdot,\cdot}_\hMnum{0}$.

We claim that
\begin{align} \label{claim-thm5}
\left|\ip{\hPi_\bfsnum{0}{0}(f_\cA^\rmtp), \hPi_\bfsnum{0}{0}(\delta_\cA^\rmtp)}_\hMnum{0}\right| \lesssim \lambda_\cA^{\rmtl2}\eta_\delta + (\eta_{p,\delta} + |\cS_\bfnum{0}|\eta_{p,2})\cdot (|\cS_\bfnum{0}|\lambda_\cA^{\rmtl2}\vee (\eta_{p,\delta} + |\cS_\bfnum{0}|\eta_{p,2})).
\end{align}
The proof of \eqref{claim-thm5} is deferred to the end of the theorem. Define
\begin{align*}
\eta_{p,\delta}^* := \eta_{p,\delta} + \frac{1}{\lambda_\cA^{\rmtl2}}\cdot (\eta_{p,\delta} + |\cS_\bfnum{0}|\eta_{p,2})\cdot (|\cS_\bfnum{0}|\lambda_\cA^{\rmtl2}\vee (\eta_{p,\delta} + |\cS_\bfnum{0}|\eta_{p,2})).
\end{align*}
Assuming \eqref{claim-thm5} holds, we obtain from \eqref{main-eq-thm5} that
\begin{align*}
&\left(\norm{\gamma_\cA^\rmtp}_\hMnum{0} - \frac{1}{2}\norm{\beta_\cA^{\rmtp}}_\hMnum{0}\right)^2 + \left(\frac{\mathfrak{C}_\bfnum{0}'-1}{\mathfrak{C}_\bfnum{0}'}\right)\lambda_\cA^{\rmtl2}\sumj \norm{\gamma_\bfsnum{\cA}{j}^\rmtp}_\hMnum{0} \\
&\le \frac{1}{4}\norm{\beta_\cA^{\rmtp,\hc}}_\hMnum{0}^2 + 2\lambda_\cA^{\rmtl2}\sumj \norm{\delta_\bfsnum{\cA}{j}^\rmtp}_\hMnum{0} + \lambda_\cA^{\rmtl2}\eta_{p,\delta}^* \\
&\lesssim \norm{\beta_\cA^{\rmtp,\hc}}_\hMnum{0}^2 + \lambda_\cA^{\rmtl2}(\eta_\delta + \eta_{p,\delta}^*),
\end{align*}
where we used the fact that
\begin{align*}
\sumj \norm{\delta_\bfsnum{\cA}{j}^\rmtp}_\hMnum{0} \lesssim \eta_\delta + \eta_{\delta,p}.
\end{align*}

We divide the proof of the theorem into two separate cases. If
\begin{align*}
\norm{\beta_\cA^{\rmtp,\hc}}_\hMnum{0}^2 \le \lambda_\cA^{\rmtl2}(\eta_\delta + \eta_{p,\delta}^*),
\end{align*}
then
\begin{align*}
&\left(\norm{\gamma_\cA^\rmtp}_\hMnum{0} - \frac{1}{2}\norm{\beta_\cA^{\rmtp}}_\hMnum{0}\right)^2 + \left(\frac{\mathfrak{C}_\bfnum{0}'-1}{\mathfrak{C}_\bfnum{0}'}\right)\lambda_\cA^{\rmtl2}\sumj \norm{\gamma_\bfsnum{\cA}{j}^\rmtp}_\hMnum{0} \lesssim \lambda_\cA^{\rmtl2}(\eta_\delta + \eta_{p,\delta}^*),
\end{align*}
which yields
\begin{align}
\norm{\gamma_\cA^\rmtp}_\hMnum{0}^2 &\lesssim \lambda_\cA^{\rmtl2}(\eta_\delta + \eta_{p,\delta}^*), \label{clue-1-1-thm5} \\
\sumj \norm{\gamma_\bfsnum{\cA}{j}^\rmtp}_\hMnum{0} &\lesssim \eta_\delta + \eta_{p,\delta}^*. \label{clue-1-2-thm5}
\end{align}
Since $\norm{\gamma_\cA^\rmtp}_\hMnum{0} \le \sumj \norm{\gamma_\bfsnum{\cA}{j}^\rmtp}_\hMnum{0}$, inequalities \eqref{clue-1-1-thm5} and \eqref{clue-1-2-thm5} imply that
\begin{align*}
\norm{\gamma_\cA^\rmtp}_\hMnum{0} \lesssim \left(\lambda_\cA^{\rmtl2}(\eta_\delta + \eta_{p,\delta}^*)\right) \wedge \left(\eta_\delta + \eta_{p,\delta}^*\right)^2,
\end{align*}
which, together with \eqref{clue-1-2-thm5}, establishes the theorem.
Otherwise, when
\begin{align*}
\norm{\beta_\cA^{\rmtp,\hc}}_\hMnum{0}^2 > \lambda_\cA^{\rmtl2}(\eta_\delta + \eta_{p,\delta}^*),
\end{align*}
we can similarly show that
\begin{align*}
\norm{\gamma_\cA^\rmtp}_\hMnum{0} &\lesssim \norm{\beta_\cA^\rmtp}_\hMnum{0}^2, \\
\sumj \norm{\gamma_\bfsnum{\cA}{j}^\rmtp}_\hMnum{0} &\lesssim \frac{1}{\lambda_\cA^{\rmtl2}}\norm{\beta_\cA^\rmtp}_\hMnum{0}^2,
\end{align*}
which completes the proof.

It remains to prove the claim in \eqref{claim-thm5}, for which we provide a sketch. Observe that
\begin{align*}
\hPi_\bfsnum{0}{0}(f_\cA^\rmtp) = \hPi_\bfsnum{0}{0}(\delta_\cA^\rmtp) + (\hPi_\bfsnum{0}{0} - \Pi_\bfsnum{0}{0})(f_\bfnum{0}^\rmtp).
\end{align*}
This yields
\begin{align*}
\left|\ip{\hPi_\bfsnum{0}{0}(f_\cA^\rmtp), \hPi_\bfsnum{0}{0}(\delta_\cA^\rmtp)}_\hMnum{0}\right| &\le \norm{\hPi_\bfsnum{0}{0}(\delta_\cA^\rmtp)}_\hMnum{0}^2 + \norm{(\hPi_\bfsnum{0}{0} - \Pi_\bfsnum{0}{0})(f_\bfnum{0}^\rmtp)}_\hMnum{0}\norm{\hPi_\bfsnum{0}{0}(\delta_\cA^\rmtp)}_\hMnum{0}.
\end{align*}
Note that
\begin{align*}
\hPi_\bfsnum{0}{0}(\delta_\cA^\rmtp) &= \hPi_\bfsnum{0}{0}\left(\delta_\cA^\rmtp - \sumaac w_\ab \delta_\ab^\rmtp\right) + \sumaac w_\ab\hPi_\bfsnum{0}{0}(\delta_\ab^\rmtp) \\
&=\hPi_\bfsnum{0}{0}\left(\delta_\cA^\rmtp - \sumaac w_\ab \delta_\ab^\rmtp\right) + \sumaac w_\ab(\hPi_\bfsnum{0}{0} - \Pi_\bfsnum{0}{0})(\delta_\ab^\rmtp) + \sumaac w_\ab \Pi_\bfsnum{0}{0}(\delta_\ab^\rmtp).
\end{align*}
Standard arguments from the proofs of Lemma~\ref{lem:u-stat2} and Lemma~\ref{lem:tech2} yield
\begin{align*}
\norm{(\hPi_\bfsnum{0}{0} - \tPi_\bfsnum{0}{0})(\delta_\ab^\rmtp)}_\hMnum{0} &\lesssim \lambda_\cA^{\rmtl2}\eta_\delta, \\
\norm{(\tPi_\bfsnum{0}{0} - \Pi_\bfsnum{0}{0})(\delta_\ab^\rmtp)}_\hMnum{0} &\lesssim \sqrt{h_\bfnum{0}}\eta_\delta \wedge (|\cS_\bfnum{0}|\vee|\cS_\bfnum{\cA}|)h_\bfnum{0}^2.
\end{align*}
These imply
\begin{align} \label{clue1-cl-thm5}
\norm{(\hPi_\bfsnum{0}{0} - \Pi_\bfsnum{0}{0})(\delta_\ab^\rmtp)}_\hMnum{0} \lesssim \lambda_\cA^{\rmtl2}\eta_\delta + \sqrt{h_\bfnum{0}}\eta_\delta \wedge (|\cS_\bfnum{0}|\vee|\cS_\bfnum{\cA}|)h_\bfnum{0}^2.
\end{align}
Furthermore, from the identity $\Pi_\bfsnum{0}{0}(\delta_\ab^\rmtp) = (\Pi_\bfsnum{0}{0} - \Pi_\bfsnum{a}{0})(\delta_\ab^\rmtp) + (\Pi_\bfsnum{0}{0} - \Pi_\bfsnum{a}{0})(f_\bfnum{0}^\rmtp)$, it follows that
\begin{align*}
\norm{\Pi_\bfsnum{0}{0}(\delta_\ab^\rmtp)}_\hMnum{0} \lesssim (\eta_\delta + |\cS_\bfnum{0}|)\eta_{p,2} \lesssim |\cS_\bfnum{0}|\eta_{p,2}.
\end{align*}
Combining this with \eqref{clue1-cl-thm5} yields
\begin{align} \label{result-0-cl-thm5}
\norm{\delta_\cA^\rmtp}_\hMnum{0} \lesssim \eta_{p,\delta} + |\cS_\bfnum{0}|\eta_{p,2} + \sqrt{h_\bfnum{0}}\eta_\delta \wedge(|\cS_\bfnum{0}|\vee|\cS_\bfnum{\cA}|)h_\bfnum{0}^2.
\end{align}
This immediately implies
\begin{align} \label{result-1-cl-thm5}
\begin{aligned}
\norm{\delta_\cA^\rmtp}_\hMnum{0}^2 &\lesssim (\eta_{p,\delta} + |\cS_\bfnum{0}|\eta_{p,2})^2 + h_\bfnum{0}\eta_\delta^2 \wedge (|\cS_\bfnum{0}|\vee|\cS_\bfnum{\cA}|)^2h_\bfnum{0}^4 \\
&\lesssim (\eta_{p,\delta} + |\cS_\bfnum{0}|\eta_{p,2})^2 + \lambda_\cA^{\rmtl2}\eta_\delta, 
\end{aligned}
\end{align}
where the last inequality uses the condition in \eqref{add-condition-thm5}.

From standard arguments, we may also show that
\begin{align*}
\norm{(\hPi_\bfsnum{0}{0} - \Pi_\bfsnum{0}{0})(f_\bfnum{0}^\rmtp)}_\hMnum{0} &\lesssim |\cS_\bfnum{0}|\left(h_\bfnum{0}^4 + \frac{1}{n_\bfnum{0}h_\bfnum{0}} + B(n_\bfnum{0}, h_\bfnum{0}, |\cS_\bfnum{0}|)\right)^{\frac{1}{2}} \\ 
&\lesssim |\cS_\bfnum{0}|\lambda_\cA^{\rmtl2}.
\end{align*}
Combining this with \eqref{result-0-cl-thm5}, we obtain
\begin{align*}
\norm{(\hPi_\bfsnum{0}{0} - \Pi_\bfsnum{0}{0})(f_\bfnum{0}^\rmtp)}_\hMnum{0}\norm{\hPi_\bfsnum{0}{0}(\delta_\cA^\rmtp)}_\hMnum{0} &\lesssim |\cS_\bfnum{0}|\lambda_\cA^{\rmtl2}(\eta_{p,\delta} + |\cS_\bfnum{0}|\eta_{p,2}) + |\cS_\bfnum{0}|\lambda_\cA^{\rmtl2}(\lambda_\cA^{\rmtl2} + \sqrt{h_\bfnum{0}})\eta_\delta \\
&\lesssim |\cS_\bfnum{0}|\lambda_\cA^{\rmtl2}(\eta_{p,\delta} + |\cS_\bfnum{0}|\eta_{p,2}) + \lambda_\cA^{\rmtl2}\eta_\delta.
\end{align*}
Here, we used the condition $|\cS_\bfnum{0}|(\lambda_\cA^{\rmtl2} + \sqrt{h_\bfnum{0}})\lesssim 1$. This bound, together with \eqref{result-1-cl-thm5}, establishes \eqref{claim-thm5}.

\subsubsection{Proof of Corollary~\ref{cor:tl-total}}

We note that even under the heterogeneous regime, a similar line of analysis can be applied. In the homogeneous regime, where $p_\bfsnum{0}{jk} \equiv p_\bfsnum{a}{jk}$ for all $(j,k)\in[d]^2$ and $\ab\in\cA$, we have
\begin{align*}
\lambda_\cA^{\rmtl1}&\sim h_\cA^2 + \sqrt{\frac{1}{n_\cA h_\cA}} + A(n_\cA, h_\cA, d;\alpha)^{\frac{1}{2}}, \\
\lambda_\cA^{\rmtl2}&\sim h_\bfnum{0}^2 + \sqrt{\frac{1}{n_\bfnum{0} h_\bfnum{0}}} + A(n_\bfnum{0}, h_\bfnum{0}, d;\alpha)^{\frac{1}{2}}.
\end{align*}
Recall the definitions of $\beta_\bfsnum{\cA}{j}^\rmtp$, $\beta_\bfnum{\cA}^\rmtp$, $\gamma_\bfsnum{\cA}{j}^\rmtp$, and $\gamma_\bfnum{\cA}^\rmtp$ from the proofs of Theorems~\ref{thm:tl-aggr} and~\ref{thm:tl-debias}. Also, define $\beta_\bfsnum{\cA}{j}^{\rmtp, \hc} := \beta_\bfsnum{\cA}{j}^\rmtp - \hPi_\bfsnum{0}{0}(\beta_\bfsnum{\cA}{j}^\rmtp)$ and $\beta_\cA^{\rmtp, \hc} := \sumj \beta_\bfsnum{\cA}{j}^{\rmtp, \hc}$. Under these notations, the conclusions of Theorems~\ref{thm:tl-aggr} and~\ref{thm:tl-debias} reduce to
\begin{align} \label{befres-step-cor2} 
\begin{aligned}
\sumj \norm{\beta_\bfsnum{\cA}{j}^\rmtp}_\hMnum{\cA} &\lesssim |\cS_\bfnum{0}|\lambda_\cA^{\rmtl1} + \eta_\delta, \\
\norm{\beta_\cA^{\rmtp}}_\hMnum{\cA}^2 &\lesssim |\cS_\bfnum{0}|(\lambda_\cA^{\rmtl1})^2 + \lambda_\cA^{\rmtl1}\eta_\delta \wedge \eta_\delta^2,
\end{aligned}
\end{align}
and
\begin{align*}
\sumj \norm{\gamma_\bfsnum{\cA}{j}^\rmtp}_\hMnum{0} &\lesssim \frac{1}{\lambda_\cA^{\rmtl2}}\norm{\beta_\cA^{\rmtp, \hc}}_\hMnum{0}^2 + \eta_\delta, \\
\norm{\gamma_\cA^{\rmtp}}_\hMnum{0}^2 &\lesssim \norm{\beta_\cA^{\rmtp, \hc}}_\hMnum{0}^2 + \lambda_\cA^{\rmtl2}\eta_\delta \wedge \eta_\delta^2.
\end{align*}

We now outline the proof. The argument proceeds in three steps. In the first step, we establish that $\norm{\beta_\cA^\rmtp}_\Mnum{\cA}^2$ admits the same upper bound as $\norm{\beta_\cA^\rmtp}_\hMnum{\cA}^2$. In the second step, we show that 
\begin{align*}
\norm{\beta_\cA^{\rmtp, \hc}}_\hMnum{0}^2 \lesssim |\cS_\bfnum{0}|(\lambda_\cA^{\rmtl1})^2 + \lambda_\cA^{\rmtl2}\eta_\delta\wedge \eta_\delta^2. 
\end{align*}
Since $\Mnum{\cA} \equiv \Mnum{0}$ under the homogeneous regime, these two steps together imply that
\begin{align} \label{res-step-cor2}
\begin{aligned}
\sumj \norm{\gamma_\bfsnum{\cA}{j}^\rmtp}_\hMnum{0} &\lesssim |\cS_\bfnum{0}|\frac{(\lambda_\cA^{\rmtl1})^2}{\lambda_\cA^{\rmtl2}} + \eta_\delta, \\
\norm{\gamma_\cA^\rmtp}_\hMnum{0}^2 &\lesssim |\cS_\bfnum{0}|(\lambda_\cA^{\rmtl1})^2 + \lambda_\cA^{\rmtl2}\eta_\delta \wedge \eta_\delta^2.
\end{aligned}
\end{align}
In the final step, we show that $\norm{\gamma_\cA^\rmtp}_\Mnum{0}^2$ also satisfies the same upper bound as $\norm{\gamma_\cA^\rmtp}_\hMnum{0}^2$. Combining these estimates gives  
\begin{align*}
\norm{\hatf_\bfnum{0}^{\rmtp, \rmtl} - f_\bfnum{0}^\rmtp}_\Mnum{0}^2 \lesssim \norm{\beta_\cA^\rmtp}_\Mnum{0}^2 + \norm{\gamma_\cA^\rmtp}_\Mnum{0}^2 \lesssim |\cS_\bfnum{0}|(\lambda_\cA^{\rmtl1})^2 + (\lambda_\cA^{\rmtl2}\eta_\delta \wedge \eta_\delta^2),
\end{align*}
where we have used the identity $M_\cA \equiv \Mnum{0}$. This completes the proof of the corollary.

\smallskip\noindent
{\it Proof of the first step.}
\smallskip

Using the arguments from the proof of Corollary~\ref{cor:ll-bound-error-pop}, we obtain
\begin{align*}
\norm{\beta_\cA^\rmtp}_\Mnum{\cA}^2 \lesssim \norm{\beta_\cA^\rmtp}_\hMnum{\cA}^2 + \left(\frac{1}{n_\cA h_\cA^2} + B(n_\cA, h_\cA^2, d)\right)^{\frac{1}{2}} \left( \sumj \norm{\beta_\bfsnum{\cA}{j}^\rmtp}_\hMnum{\cA} \right)^2 + \norm{\Pi_\bfsnum{\cA}{0}(\beta_\cA^\rmtp)}_\Mnum{\cA}^2.
\end{align*}
By applying \eqref{befres-step-cor2} and assuming that
\begin{align*}
|\cS_\bfnum{0}|\left(\frac{1}{n_\cA h_\cA^2} + B(n_\cA, h_\cA^2, d)\right)^{\frac{1}{2}} &\lesssim 1, \\
\left(\frac{1}{n_\cA h_\cA^2} + B(n_\cA, h_\cA^2, d)\right)^{\frac{1}{2}} \eta_\delta^2 &\lesssim \lambda_\cA^{\rmtl1} \eta_\delta,
\end{align*}
we deduce that
\begin{align*}
\left(\frac{1}{n_\cA h_\cA^2} + B(n_\cA, h_\cA^2, d)\right)^{\frac{1}{2}} \left( \sumj \norm{\beta_\bfsnum{\cA}{j}^\rmtp}_\hMnum{\cA} \right)^2 \lesssim |\cS_\bfnum{0}|(\lambda_\cA^{\rmtl1})^2 + \lambda_\cA^{\rmtl1}\eta_\delta \wedge \eta_\delta^2.
\end{align*}
Thus, it remains to bound $\norm{\Pi_\bfsnum{\cA}{0}(\beta_\cA^\rmtp)}_\Mnum{\cA}^2$ by the same quantity. Under the homogeneous regime, we have $f_\bfsnum{\cA}{j}^\rmtp = \sumaac w_\ab f_\bfsnum{a}{j}^\rmtp$, and hence we observe that
\begin{align*}
\norm{\Pi_\bfsnum{\cA}{0}(\beta_\cA^\rmtp)}_\Mnum{\cA} &\le \sumj \left| \int_0^1 \beta_\bfsnum{\cA}{j}^\rmv(x_j)^\top\left( \hp_\bfsnum{\cA}{j}^\rmv(x_j) - p_\bfsnum{\cA}{j}^\rmv(x_j) \right)\dxj \right|  \\
&\quad + \sumaac w_\ab \sumj \left| \int_0^1 f_\bfsnum{a}{j}^\rmv(x_j)^\top \left(\hp_\bfsnum{\cA}{j}^\rmv(x_j)  - p_\bfsnum{\cA}{j}^\rmv(x_j)\right)\dxj \right|.
\end{align*}
After a series of standard but tedious calculations based on kernel smoothing theory, we obtain
\begin{align*}
\norm{\Pi_\bfsnum{\cA}{0}(\beta_\cA^\rmtp)}_\Mnum{\cA}^2 &\lesssim h_\cA \left(\sumj \norm{\beta_\bfsnum{\cA}{j}^\rmtp}_\hMnum{\cA}\right)^2 + \left|\cup_{\ab\in\cA} \cS_\ab\right|^2 h_\cA^4 \\
&\lesssim h_\cA \left(\sumj \norm{\beta_\bfsnum{\cA}{j}^\rmtp}_\hMnum{\cA}\right)^2 + |\cS_\cA|^2 h_\cA^4 \\
&\lesssim |\cS_\bfnum{0}| (\lambda_\cA^{\rmtl1})^2,
\end{align*}
where we have used the conditions $h_\cA |\cS_\bfnum{0}| \ll 1$ and
\begin{align*}
h_\cA\eta_\delta^2 \lesssim \lambda_\cA^{\rmtl1}\eta_\delta, \quad \text{and} \quad |\cS_\cA| h_\cA^2 \lesssim \lambda_\cA^{\rmtl1}.
\end{align*}
This completes the argument for the first step.

\smallskip\noindent
{\it Proof of the second step.}
\smallskip

We observe that 
\begin{align*}
\norm{\beta_\cA^{\rmtp, \hc}}_\hMnum{0}^2 &\lesssim \norm{\beta_\cA^{\rmtp, \hc}}_\tMnum{0}^2 + \left(\frac{1}{n_\bfnum{0}h_\bfnum{0}^2} + B(n_\bfnum{0}, h_\bfnum{0}^2, d)\right)^{\frac{1}{2}}\left(\sumj \norm{\beta_\bfsnum{\cA}{j}^{\rmtp, \hc}}_\hMnum{0}\right)^2 \\
&\lesssim \norm{\beta_\cA^{\rmtp, \tc}}_\tMnum{0}^2 + \norm{(\tPi_\bfsnum{0}{0} - \hPi_\bfsnum{0}{0})(\beta_\cA^\rmtp)}_\tMnum{0}^2 + \left(\frac{1}{n_\bfnum{0}h_\bfnum{0}^2} + B(n_\bfnum{0}, h_\bfnum{0}^2, d)\right)^{\frac{1}{2}}\left(\sumj \norm{\beta_\bfsnum{\cA}{j}^{\rmtp}}_\hMnum{0}\right)^2 \\
&\lesssim \sumj \norm{\beta_\bfsnum{\cA}{j}^{\rmtp, \tc}}_\tMnum{0}^2 + \left(\frac{1}{n_\bfnum{0}h_\bfnum{0}^2} + B(n_\bfnum{0}, h_\bfnum{0}^2, d)\right)^{\frac{1}{2}}\left(\sumj \norm{\beta_\bfsnum{\cA}{j}^{\rmtp}}_\hMnum{0}\right)^2 \\
&\lesssim \sumj \norm{\beta_\bfsnum{\cA}{j}^{\rmtp, \rmc}}_\Mnum{0}^2 + \norm{(\tPi_\bfsnum{0}{0}-\Pi_\bfsnum{0}{0})(\beta_\cA^{\rmtp})}_\Mnum{0}^2 + \left(\frac{1}{n_\bfnum{0}h_\bfnum{0}^2} + B(n_\bfnum{0}, h_\bfnum{0}^2, d)\right)^{\frac{1}{2}}\left(\sumj \norm{\beta_\bfsnum{\cA}{j}^{\rmtp}}_\hMnum{0}\right)^2 \\
&\lesssim \norm{\beta_\bfnum{\cA}^{\rmtp}}_\Mnum{0}^2 + \left(h_\bfnum{0}\vee\left(\frac{1}{n_\bfnum{0}h_\bfnum{0}^2} + B(n_\bfnum{0}, h_\bfnum{0}^2, d)\right)^{\frac{1}{2}}\right)\left(\sumj \norm{\beta_\bfsnum{\cA}{j}^{\rmtp}}_\hMnum{0}\right)^2
\end{align*}
Since 
\begin{align*}
\left(h_\bfnum{0}\vee\left(\frac{1}{n_\bfnum{0}h_\bfnum{0}^2} + B(n_\bfnum{0}, h_\bfnum{0}^2, d)\right)^{\frac{1}{2}}\right)\eta_\delta^2&\lesssim \lambda_\cA^{\rmtl2}\eta_\delta, \\ 
\left(h_\bfnum{0}\vee\left(\frac{1}{n_\bfnum{0}h_\bfnum{0}^2} + B(n_\bfnum{0}, h_\bfnum{0}^2, d)\right)^{\frac{1}{2}}\right)|\cS_\bfnum{0}|&\lesssim 1,  
\end{align*}
it follows from the first bound in~\eqref{befres-step-cor2} that the desired result holds.

\smallskip\noindent
{\it Proof of the third step.}
\smallskip

Following the steps of the proof of Corollary~\ref{cor:ll-bound-error-pop}, we obtain
\begin{align*}
\norm{\gamma_\cA^\rmtp}_\Mnum{0}^2 \lesssim \norm{\gamma_\cA^\rmtp}_\hMnum{0}^2 + \left(\frac{1}{n_\bfnum{0}h_\bfnum{0}^2} + B(n_\bfnum{0}, h_\bfnum{0}^2, d)\right)^{\frac{1}{2}}\left(\sumj \norm{\gamma_\bfsnum{\cA}{j}^\rmtp}_\hMnum{0}\right)^2 + \norm{\Pi_\bfsnum{0}{0}(\gamma_\cA^\rmtp)}_\Mnum{0}^2. 
\end{align*}
From~\eqref{res-step-cor2}, under the condition $\lambda_\cA^{\rmtl1}\lesssim \lambda_\cA^{\rmtl2}$, it follows that
\begin{align*}
\left(\frac{1}{n_\bfnum{0}h_\bfnum{0}^2} + B(n_\bfnum{0}, h_\bfnum{0}^2, d)\right)^{\frac{1}{2}}\left(\sumj \norm{\gamma_\bfsnum{\cA}{j}^\rmtp}_\hMnum{0}\right)^2 \lesssim |\cS_\bfnum{0}|(\lambda_\cA^{\rmtl1})^2 + \lambda_\cA^{\rmtl2}\eta_\delta\wedge \eta_\delta^2. 
\end{align*}
Moreover, by arguments similar to those used in the proof of the first step, we may show that
\begin{align*}
\norm{\Pi_\bfsnum{0}{0}(\gamma_\cA^\rmtp)}_\Mnum{0} &\le \sumj\left|\int_0^1 \gamma_\bfsnum{\cA}{j}^\rmv(x_j)^\top\left(\hp_\bfsnum{0}{j}^\rmv(x_j) - p_\bfsnum{0}{j}^\rmv(x_j)\right)\dxj\right| \\
&\quad + \sumaac w_\bfnum{a} \sumj \left|\int_0^1 \delta_\bfsnum{a}{j}^\rmv(x_j)^\top \left(\hp_\bfsnum{0}{j}^\rmv(x_j) - p_\bfsnum{0}{j}^\rmv(x_j)\right)\dxj \right| \\
&\le \sqrt{h_\bfnum{0}}\left(\sumj \norm{\gamma_\bfsnum{\cA}{j}^\rmtp}_\hMnum{0} + \eta_\delta\right). 
\end{align*}
Hence, under the conditions used in the previous steps, we could obtain
\begin{align*}
\norm{\Pi_\bfsnum{0}{0}(\gamma_\cA^\rmtp)}_\Mnum{0}^2 \lesssim |\cS_\bfnum{0}|(\lambda_\cA^{\rmtl1})^2 + \lambda_\cA^{\rmtl2}\eta_\delta\wedge \eta_\delta^2.
\end{align*}
This completes the proof.

\subsubsection{Proof of Theorem~\ref{thm:tl-minimax}}

In this proof, we use the notation $\gtrsim$ in probability arguments to indicate that the inequality holds up to a multiplicative constant $0<C<\infty$, depending only on $C_{\sF,L}, C_{\sF,U}, \beta$ and $L$. We first consider the following two cases:
\begin{itemize}
\item[(i)] All auxiliary populations share the same functional structure as the target population; that is, $f_\bfsnum{a}{j} \equiv f_\bfsnum{0}{j}$ for all $j \in [d]$ and $\ab \in \cA$. Moreover, the target and auxiliary populations are mutually independent;
\item[(ii)] All auxiliary populations are non-informative; that is, $f_\bfsnum{a}{j} \equiv 0$ for all $j \in [d]$ and $\ab \in \cA$.
\end{itemize}
In case~(i), following the arguments used in the proof of Theorem~\ref{thm:ll-minimax}, we obtain the lower bound
\begin{align} \label{case-1-thm-minimax}
\inf_\tf \sup_{(f_\bfnum{0}, (f_\bfnum{a}:\ab\in\cA))\in\sF_\bfsnum{0}{\rmadd}^{s, \rmtl}(\beta, L)}\Pb_f\left(\norm{\tf-f_\bfnum{0}}_{p_\bfnum{0}}^2\gtrsim sC(n_\cA, s,d;\beta)\right)\ge \frac{3}{4}.
\end{align}

In case~(ii), we note that $\sumj \norm{f_\bfsnum{0}{j}}_\pnum{0}\le \eta_\delta$. In terms of the notations in Theorem~\ref{thm:ll-minimax}, this condition reduces to
\begin{align*}
LN^{-\beta}s\lesssim \eta_\delta. 
\end{align*}
If $\eta_\delta$ is sufficiently small such that
\begin{align*}
\eta_\delta\sqrt{\frac{n_\bfnum{0}}{\log (d/s)}}<1,
\end{align*}
then we set $s'=1$, $N=1$, and $L'=C_L\eta_\delta$ for some constant $C_L>0$. It is legitimate to assume that $L' < L$, since $\eta_\delta \ll 1$. It follows that $L'N^{-2\beta}s'\lesssim \eta_\delta$. The arguments leading to \eqref{res-fano-lemma-thm3} then yield
\begin{align*}
&\inf_\tf \sup_{(f_\bfnum{0}, (f_\bfnum{a}:\ab\in\cA))\in\sF_\bfsnum{0}{\rmadd}^{s, \rmtl}(\beta, L)}\Pb_f\left(\norm{\tf-f_\bfnum{0}}_\pnum{0} \gtrsim \eta_\delta^2\right) \\
&\ge \inf_\tf \sup_{(f_\bfnum{0}, (f_\bfnum{a}:\ab\in\cA))\in\sF_\bfsnum{0}{\rmadd}^{s', \rmtl}(\beta, L')}\Pb_f\left(\norm{\tf-f_\bfnum{0}}_\pnum{0} \gtrsim \eta_\delta^2\right) \\
&\ge 1-\frac{2c_\ve C_{\sF,U}C_L^2 \kappa_2 n_\bfnum{0}\eta_\delta^2 + 8\log 2}{2\log d + 1} \\
&\ge \frac{3}{4},
\end{align*}
by choosing $C_L$ sufficiently small. 
On the other hand, when
\begin{align*}
\eta_\delta n_\bfnum{0}^{\frac{\beta}{2\beta+1}}<1,
\end{align*}
we let $s'=1$, $N=C_N\cdot n_\bfnum{0}^{\frac{1}{2\beta+1}}$ for some constant $C_N>0$, and $L'=\eta_\delta n_\bfnum{0}^{\frac{\beta}{2\beta+1}}\cdot L < L$. It holds that $L'N^{-2\beta}s'\lesssim \eta_\delta$. Then, we may verify that
\begin{align} \label{extreme-case-thm-minimax}
\begin{aligned}
&\inf_\tf \sup_{(f_\bfnum{0}, (f_\bfnum{a}:\ab\in\cA))\in\sF_\bfsnum{0}{\rmadd}^{s, \rmtl}(\beta, L)}\Pb_f\left(\norm{\tf-f_\bfnum{0}}_\pnum{0} \gtrsim \eta_\delta^2\right) \\
&\ge \inf_\tf \sup_{(f_\bfnum{0}, (f_\bfnum{a}:\ab\in\cA))\in\sF_\bfsnum{0}{\rmadd}^{s', \rmtl}(\beta, L')}\Pb_f\left(\norm{\tf-f_\bfnum{0}}_\pnum{0} \gtrsim \eta_\delta^2\right) \\
&\ge 1-\frac{2 c_\ve C_{\sF,U}L^2 \kappa_2 C_N^{-2\beta} n_\bfnum{0}\eta_\delta^2 + 8\log 2}{2\log d + C_N n_\bfnum{0}^{\frac{1}{2\beta+1}}} \\
&\ge 1-\frac{2 c_\ve C_{\sF,U}L^2 \kappa_2 C_N^{-2\beta} \eta_\delta^2 + \frac{8\log 2}{n_\bfnum{0}}}{\frac{2\log d}{n_\bfnum{0}} + C_N n_\bfnum{0}^{-\frac{2\beta}{2\beta+1}}} \\
&\ge \frac{3}{4},
\end{aligned}
\end{align}
by choosing $C_N$ sufficiently large. Here, we have used the fact that $\eta_\delta^2 \le n_\bfnum{0}^{-\frac{2\beta}{2\beta+1}}$.
Hence, in the following proof, we may assume without loss of generality that
\begin{align} \label{cond-eta-delta-thm-minimax}
\eta_\delta \left(\sqrt{\frac{n_\bfnum{0}}{\log (d/s)}} \wedge n_\bfnum{0}^{\frac{\beta}{2\beta+1}}\right) \ge 1.
\end{align}

Next, we obtain the lower bound by dividing case (ii) into the following four subcases:
\begin{itemize}
\item[(ii-1)] $\eta_\delta \ge s n_\bfnum{0}^{-\frac{\beta}{2\beta+1}}$ and $\eta_\delta \ge s \sqrt{\frac{\log (d/s)}{n_\bfnum{0}}}$;
\item[(ii-2)] $s \sqrt{\frac{\log (d/s)}{n_\bfnum{0}}} \le \eta_\delta \le s n_\bfnum{0}^{-\frac{\beta}{2\beta+1}}$;
\item[(ii-3)] $s n_\bfnum{0}^{-\frac{\beta}{2\beta+1}} \le \eta_\delta \le s \sqrt{\frac{\log (d/s)}{n_\bfnum{0}}}$;
\item[(ii-4)] $\eta_\delta \le s n_\bfnum{0}^{-\frac{\beta}{2\beta+1}}$ and $\eta_\delta \le s \sqrt{\frac{\log (d/s)}{n_\bfnum{0}}}$.
\end{itemize}
In case (ii-1), the standard choices of $L$, $N$, and $s$ as in the proof of Theorem~\ref{thm:ll-minimax} remain valid. Therefore, we have
\begin{align*}
\inf_\tf \sup_{(f_\bfnum{0}, (f_\bfnum{a}:\ab \in \cA)) \in \sF_\bfsnum{0}{\rmadd}^{s, \rmtl}(\beta, L)} \Pb_f\left(\norm{\tf - f_\bfnum{0}}_\pnum{0} \gtrsim s C(n_\bfnum{0}, s, d; \beta)\right) \ge \frac{3}{4}.
\end{align*}

In case (ii-4), assume first that $\eta_\delta \le s n_\bfnum{0}^{-\frac{\beta}{2\beta+1}}$. Let $s' = \lfloor \eta_\delta n_\bfnum{0}^{\frac{\beta}{2\beta+1}} \rfloor \le s$. This is valid since \eqref{cond-eta-delta-thm-minimax} holds. Choosing $N = C_N n_\bfnum{0}^{\frac{\beta}{2\beta+1}}$ for some constant $C_N > 0$, it follows from \eqref{res-fano-lemma-thm3} that
\begin{align*}
&\inf_\tf \sup_{(f_\bfnum{0}, (f_\bfnum{a}: \ab \in \cA)) \in \sF_\bfsnum{0}{\rmadd}^{s, \rmtl}(\beta, L)} \Pb_f\left(\norm{\tf - f_\bfnum{0}}_\pnum{0} \gtrsim \eta_\delta n_\bfnum{0}^{-\frac{\beta}{2\beta+1}}\right) \\
&\ge \inf_\tf \sup_{(f_\bfnum{0}, (f_\bfnum{a}: \ab \in \cA)) \in \sF_\bfsnum{0}{\rmadd}^{s', \rmtl}(\beta, L)} \Pb_f\left(\norm{\tf - f_\bfnum{0}}_\pnum{0} \gtrsim \eta_\delta n_\bfnum{0}^{-\frac{\beta}{2\beta+1}}\right) \\
&\ge 1 - \frac{2 c_\ve C_{\sF,U} L^2 \kappa_2 C_N^{-2\beta} n_\bfnum{0}^{\frac{\beta}{2\beta+1}} \eta_\delta + 8 \log 2}{2 s' \log (d/s') + C_N n_\bfnum{0}^{\frac{\beta+1}{2\beta+1}} \eta_\delta} \\
&\ge \frac{7}{8},
\end{align*}
for sufficiently large $C_N$. 

Alternatively, if $\eta_\delta \le s \sqrt{\frac{\log (d/s)}{n_\bfnum{0}}}$, let $s' = \lfloor \eta_\delta \sqrt{\frac{n_\bfnum{0}}{\log (d/s)}} \rfloor \le s$, and set $N = C_N (\frac{n_\bfnum{0}}{\log (d/s)})^{\frac{1}{2\beta}}$. Then we obtain
\begin{align*}
&\inf_\tf \sup_{(f_\bfnum{0}, (f_\bfnum{a}: \ab \in \cA)) \in \sF_\bfsnum{0}{\rmadd}^{s, \rmtl}(\beta, L)} \Pb_f\left(\norm{\tf - f_\bfnum{0}}_\pnum{0} \gtrsim \eta_\delta \sqrt{\frac{\log (d/s)}{n_\bfnum{0}}}\right) \\
&\ge \inf_\tf \sup_{(f_\bfnum{0}, (f_\bfnum{a}: \ab \in \cA)) \in \sF_\bfsnum{0}{\rmadd}^{s', \rmtl}(\beta, L)} \Pb_f\left(\norm{\tf - f_\bfnum{0}}_\pnum{0} \gtrsim \eta_\delta \sqrt{\frac{\log (d/s)}{n_\bfnum{0}}}\right) \\
&\ge 1 - \frac{2 c_\ve C_{\sF,U} L^2 \kappa_2 C_N^{-2\beta} \eta_\delta \sqrt{\frac{n_\bfnum{0}}{\log (d/s)}} \cdot \log (d/s) + 8 \log 2}{2 \eta_\delta \sqrt{\frac{n_\bfnum{0}}{\log (d/s)}} \cdot \log (d/s) + C_N \left(\frac{n_\bfnum{0}}{\log (d/s)}\right)^{\frac{1}{2\beta}} s'} \\
&\ge \frac{7}{8},
\end{align*}
for sufficiently large $C_N$. Thus, for case (ii-4), we have
\begin{align*}
\inf_\tf \sup_{(f_\bfnum{0}, (f_\bfnum{a}: \ab \in \cA)) \in \sF_\bfsnum{0}{\rmadd}^{s, \rmtl}(\beta, L)} \Pb_f\left(\norm{\tf - f_\bfnum{0}}_\pnum{0} \gtrsim \eta_\delta C(n_\bfnum{0}, s, d; \beta)^{\frac{1}{2}}\right) \ge \frac{3}{4}.
\end{align*}

For the remaining cases (ii-2) and (ii-3), the same lower bound as in case (ii-4) can be established. To illustrate, we focus on case (ii-2), as the argument for case (ii-3) is analogous. Since $\eta_\delta \le s n_\bfnum{0}^{-\frac{\beta}{2\beta+1}}$, the argument used in case (ii-4) leads to
\begin{align}\label{case2-thm-minimax}
\inf_\tf \sup_{(f_\bfnum{0}, (f_\bfnum{a}: \ab \in \cA)) \in \sF_\bfsnum{0}{\rmadd}^{s, \rmtl}(\beta, L)} \Pb_f\left(\norm{\tf - f_\bfnum{0}}_\pnum{0} \gtrsim \eta_\delta n_\bfnum{0}^{-\frac{\beta}{2\beta+1}}\right) \ge \frac{3}{4}.
\end{align}
Note that in case (ii-2),
\begin{align*}
\frac{\log (d/s)}{n_\bfnum{0}} \le n_\bfnum{0}^{-\frac{2\beta}{2\beta+1}}.
\end{align*}
Combining this with \eqref{case2-thm-minimax}, we obtain
\begin{align*}
\inf_\tf \sup_{(f_\bfnum{0}, (f_\bfnum{a}: \ab \in \cA)) \in \sF_\bfsnum{0}{\rmadd}^{s, \rmtl}(\beta, L)} \Pb_f\left(\norm{\tf - f_\bfnum{0}}_\pnum{0} \gtrsim \eta_\delta C(n_\bfnum{0}, s, d; \beta)^{\frac{1}{2}}\right) \ge \frac{3}{4}.
\end{align*}
Combining the lower bounds from all cases (i), (ii-1)--(ii-4), as well as from \eqref{extreme-case-thm-minimax}, yields the desired result.

\subsection{Technical proofs for Appendix}

This section presents the technical details supporting the result in Appendix. Throughout the proofs, all (in)equalities are understood to hold either almost surely or with probability tending to one. We use the notation $C$ to denote an absolute constant, whose value may change from line to line.

\subsubsection{Proof of Theorem~\ref{thm:apdx-source-detection}}

We sketch the proof. Consider the event under which the following bounds hold:
\begin{align} \label{results-thmappx}
\begin{aligned}
\norm{\hatf_\bfnum{0}^\rmtp - f_\bfnum{0}^\rmtp}_\Mnum{0}^2 &\lesssim |\cS_\bfnum{0}|\left(h_\bfnum{0}^4 + \frac{1}{n_\bfnum{0}h_\bfnum{0}} + A(n_\bfnum{0}, h_\bfnum{0},d;\alpha)\right), \\
\norm{\hatf_\bfnum{\{0,b\}}^\rmtp - f_\bfnum{\{0,b\}}^\rmtp}_\Mnum{0}^2 &\lesssim |\cS_\bfnum{0}|\left(h_{\{0,b\}}^4 + \frac{1}{(n_\bfnum{0}+2n_\bfnum{b})h_{\{0,b\}}} + A(n_\bfnum{0} + 2n_\bfnum{b}, h_{\{0,b\}},d;\alpha)\right) \\
&\quad  + \left(h_\bfnum{0}^4 + \frac{1}{n_\bfnum{0}h_\bfnum{0}} + A(n_\bfnum{0}, h_\bfnum{0},d;\alpha)\right)^{\frac{1}{2}}\eta_\delta\wedge \eta_\delta^2
\end{aligned}
\end{align}
for all $\bfnum{b} \in \Bc$, $h_\bfnum{0} \sim n_\bfnum{0}^{-1/5}$, and $h_{\{\bfnum{0}, \bfnum{b}\}} \sim (n_\bfnum{0} + 2n_\bfnum{b})^{-1/5}$. This event holds with probability tending to one.

Let $L_0$ denote the expected loss,
\[
L_0(\gv^\rmtp) := \Eb\big[\big|g(\Xv_\bfnum{0}) - f_\bfnum{0}(\Xv_\bfnum{0})\big|\big].
\]
Note that $L_0(\fv_\bfnum{0}^\rmtp)=0=\widehat L_0^\angl{r}(\hatfv_\bfnum{0}^{\rmtp, \angl{r}})$. Observe that
\begin{align*}
\widehat L_0^\angl{r}(\hatfv_{\{\bfnum{0}, \bfnum{b}\}}^{\rmtp, \angl{r}}) 
&\ge \frac{2}{n_\bfnum{0}} \sum_{i=1}^{n_\bfnum{0}/2} \big|f_{\{\bfnum{0}, \bfnum{b}\}}(\Xv_\bfnum{0}^{{i},\angl{3-r}}) - f_\bfnum{0}(\Xv_\bfnum{0}^{{i},\angl{3-r}})\big| \\
&\quad - \frac{2}{n_\bfnum{0}} \sum_{i=1}^{n_\bfnum{0}/2} \big|\hatf_{\{\bfnum{0}, \bfnum{b}\}}^\angl{r}(\Xv_\bfnum{0}^{{i},\angl{3-r}}) - f_{\{0,b\}}(\Xv_\bfnum{0}^{{i},\angl{3-r}})\big| 
- \frac{2}{n_\bfnum{0}} \sum_{i=1}^{n_\bfnum{0}/2} \big|\hatf_\bfnum{0}^\angl{r}(\Xv_\bfnum{0}^{{i},\angl{3-r}}) - f_\bfnum{0}(\Xv_\bfnum{0}^{{i},\angl{3-r}})\big|
\end{align*}
and
\begin{align*}
\widehat L_0^\angl{r}(\hatfv_{\{\bfnum{0}, \bfnum{b}\}}^{\rmtp, \angl{r}}) 
&\le \frac{2}{n_\bfnum{0}} \sum_{i=1}^{n_\bfnum{0}/2} \big|f_{\{\bfnum{0}, \bfnum{b}\}}(\Xv_\bfnum{0}^{{i},\angl{3-r}}) - f_\bfnum{0}(\Xv_\bfnum{0}^{{i},\angl{3-r}})\big| \\
&\quad + \frac{2}{n_\bfnum{0}} \sum_{i=1}^{n_\bfnum{0}/2} \big|\hatf_{\{\bfnum{0}, \bfnum{b}\}}^\angl{r}(\Xv_\bfnum{0}^{{i},\angl{3-r}}) - f_{\{\bfnum{0}, \bfnum{b}\}}(\Xv_\bfnum{0}^{{i},\angl{3-r}})\big| 
+ \frac{2}{n_\bfnum{0}} \sum_{i=1}^{n_\bfnum{0}/2} \big|\hatf_\bfnum{0}^\angl{r}(\Xv_\bfnum{0}^{{i},\angl{3-r}}) - f_\bfnum{0}(\Xv_\bfnum{0}^{{i},\angl{3-r}})\big|.
\end{align*}
We prove that
\begin{align*}
\widehat L_0^\angl{r}(\hatfv_{\{\bfnum{0}, \bfnum{b}\}}^{\rmtp, \angl{r}}) &\ge \frac{c_{\rm SD}}{4}, \quad \bfnum{b}\in\Bc\setminus\cA, \\
\widehat L_0^\angl{r}(\hatfv_{\{\bfnum{0}, \bfnum{b}\}}^{\rmtp, \angl{r}}) &\le \frac{c_{\rm SD}}{8}, \quad \bfnum{b}\in\cA,
\end{align*}
hold with probability tending to one for $r=1,2$. Clearly, this implies the theorem.  

It suffices to show that for $r=1,2$, with probability tending to one,
\begin{align*} 
\begin{aligned}
\frac{2}{n_\bfnum{0}} \sum_{i=1}^{n_\bfnum{0}/2} \big|f_{\{\bfnum{0}, \bfnum{b}\}}(\Xv_\bfnum{0}^{{i},\angl{3-r}}) - f_\bfnum{0}(\Xv_\bfnum{0}^{{i},\angl{3-r}})\big| &\ge \frac{3c_{\rm SD}}{8}, \quad \bfnum{b} \in \Bc \setminus \cA, \\
\frac{2}{n_\bfnum{0}} \sum_{i=1}^{n_\bfnum{0}/2} \big|f_{\{\bfnum{0}, \bfnum{b}\}}(\Xv_\bfnum{0}^{{i},\angl{3-r}}) - f_\bfnum{0}(\Xv_\bfnum{0}^{{i},\angl{3-r}})\big| &\le \frac{c_{\rm SD}}{8}, \quad \bfnum{b} \in \cA, \\
\frac{2}{n_\bfnum{0}} \sum_{i=1}^{n_\bfnum{0}/2} \big|\hatf_{\{\bfnum{0}, \bfnum{b}\}}^\angl{r}(\Xv_\bfnum{0}^{{i},\angl{3-r}}) - f_{\{\bfnum{0}, \bfnum{b}\}}(\Xv_\bfnum{0}^{{i},\angl{3-r}})\big| &\le \frac{c_{\rm SD}}{16}, \\
\frac{2}{n_\bfnum{0}} \sum_{i=1}^{n_\bfnum{0}/2} \big|\hatf_\bfnum{0}^\angl{r}(\Xv_\bfnum{0}^{{i},\angl{3-r}}) - f_\bfnum{0}(\Xv_\bfnum{0}^{{i},\angl{3-r}})\big| &\le \frac{c_{\rm SD}}{16}.
\end{aligned}
\end{align*}
These inequalities follow from Chebyshev’s inequality together with the $L^2$ bounds established in Theorems~\ref{cor:ll-bound-error-pop} and~\ref{cor:tl-total} as in \eqref{results-thmappx}, noting that $L^1$ errors are controlled by their $L^2$ counterparts.

\subsection{Technical lemmas} \label{subsec:technical-lemmas}

We now state three lemmas that will be used in the proofs of our main theoretical results. These lemmas follow from $U$-statistic theory, such as Theorem~\ref{thm:u-stat}. All proofs are deferred to Section~\ref{subsec:proofs-tech-lemmas}. To the best of our knowledge, this is the first result of its kind established using $U$-statistic theory. In both the statements and proofs, we employ general notation. For example, in what follows, the matrix-valued function $M(\cdot)$ is understood to represent $M_{\bfnum{0}}(\cdot)$ with $\Xv_{\bfnum{0}}$ replaced by a generic random vector $\Xv$. Define $\Bb(1)$ to be the unit ball in $\sH_\rmadd^\rmtp$, i.e.,
\begin{align*}
\Bb(1) := \left\{ g^\rmtp \in \sH_\rmadd^\rmtp : \norm{g^\rmtp}_M \le 1 \right\}.
\end{align*}
Recall the definition of $B(n,h,d)$.

\begin{lemma} \label{lem:u-stat1}
Assume that (P1), (R-$\alpha$) and (B-$\alpha$) hold with given $\alpha>0$. Then, it follows that
\begin{align*}
\max_{j\in[d]} \left\Vert U_j^\top \cdot \frac{1}{n}\sumin Z_{j}^i(x_j) K_{h_j}(x_j, X_{j}^i)\ve^i \right\Vert_M^2 \lesssim \frac{1}{nh} + A(n, h,d;\alpha). 
\end{align*}
\end{lemma}

\begin{lemma} \label{lem:u-stat2}
Assume that (P1) and (B-$\alpha$) hold with given $\alpha>0$. Then, it follows that
\begin{align*}
\max_{j\in[d]}\sup_{g_j^\rmtp\in\sH_j^\rmtp\cap \Bb(1)}\left\Vert U_j^\top \cdot (\hM_{jj} - \tM_{jj})g_j^\rmv \right\Vert_M^2 \lesssim \frac{1}{nh} +  B(n,h,d). 
\end{align*}
In particular, when $g_j^\rmtp= U_j^\top \cdot (1,0)^\top$, we further obtain
\begin{align*}
\max_{j\in[d]}\left\Vert U_j^\top \cdot(\hp_j^\rmv - \tp_j^\rmv)\right\Vert_M^2 \lesssim \frac{1}{nh} + B(n,h,d). 
\end{align*}
\end{lemma}

\begin{lemma} \label{lem:u-stat3}
Assume that (P1)--(P2) and (B-$\alpha$) hold with given $\alpha>0$. Then, it follows that
\begin{align*}
\max_{(j,k)\in[d]^2}\sup_{g_k^\rmtp\in\sH_k^\rmtp\cap\Bb(1)}\left\Vert U_j^\top \cdot \int_0^1 (\hM_{jk}(\cdot,x_k) - \tM_{jk}(\cdot, x_k))g_k^\rmv(x_k)\dxk\right\Vert_M^2 \lesssim \red{\frac{1}{nh^2} + B(n,h^2,d)}. 
\end{align*}
\end{lemma}

Next, we introduce two additional lemmas. Since their proofs follow from standard kernel smoothing theory combined with exponential inequalities, as in \cite{Lee2024}, we omit the proofs. Define the incomplete moments
\begin{align*}
\mu_{j,\ell}(x_j) := \int_0^1 \left( \frac{u_j - x_j}{h_j} \right)^\ell K_{h_j}(x_j, u_j) \, \duj, \quad \ell=0,1,2.
\end{align*}
We also define the matrix-valued function
\begin{align*}
N_{jj}(x_j) := \begin{pmatrix}
\mu_{j,0}(x_j) & \mu_{j,1}(x_j)/\mu_2 \\
\mu_{j,1}(x_j) & \mu_{j,2}(x_j)/\mu_2
\end{pmatrix}.
\end{align*}
Note that 
\begin{align*}
\mu_2 = \int_{-1}^1 v^2 K(v)\, \rmd v \le \int_{-1}^1 K(v) =1. 
\end{align*}

\begin{lemma} \label{lem:tech1}
Assume that (P1) and (B-$\alpha$) hold with given $\alpha > 0$. Then, it follows that 
\begin{align*}
\frac{C_{p,L}^\rmuniv \mu_2}{2} \le \min_{j\in[d]}\inf_{x_j\in[0,1]}\lambda_{\rmmin}\left(\tM_{jj}(x_j)\right) \le \max_{j\in[d]}\sup_{x_j\in[0,1]}\lambda_{\rmmax}\left(\tM_{jj}(x_j)\right) \le 2C_{p,U}^{\rmuniv}
\end{align*}
for all sufficiently large $n$. Furthermore, for any small constant $\xi > 0$, we have
\begin{align*}
1-\xi &\le \min_{j\in[d]}\inf_{x_j\in[0,1]}\lambda_\rmmin\left(\tM_{jj}(x_j)^{-\frac{1}{2}}\hM_{jj}(x_j) \tM_{jj}(x_j)^{-\frac{1}{2}}\right) \\
&\le \max_{j\in[d]}\sup_{x_j\in[0,1]}\lambda_\rmmax\left(\tM_{jj}(x_j)^{-\frac{1}{2}}\hM_{jj}(x_j) \tM_{jj}(x_j)^{-\frac{1}{2}}\right) \le 1+\xi
\end{align*}
with probability tending to one.
\end{lemma}

\begin{lemma} \label{lem:tech2}
Assume that (P1)--(P2) and (B-$\alpha$) hold with given $\alpha>0$. Then, it follows that 
\begin{align*}
\max_{j\in[d]}\sup_{g_j^\rmtp\in\sH_j^\rmtp\cap \Bb(1)}\left\Vert U_j^\top \cdot \left(\tM_{jj} - N_{jj}M_{jj}\right) g_j^\rmv \right\Vert_M &\lesssim \sqrt{h}, \\
\max_{(j,k)\in[d]^2}\sup_{g_k^\rmtp\in\sH_k^\rmtp\cap \Bb(1)}\left\Vert U_j^\top \cdot \int_0^1 \left(\tM_{jk}(\cdot, x_k) - N_{jj}(\cdot)M_{jk}(\cdot, x_k)\right) g_k^\rmv(x_k)\dxk \right\Vert_M &\lesssim \sqrt{h}. 
\end{align*}
\end{lemma}

\subsection{Additional technical proofs} \label{subsec:proofs-tech-lemmas}

In this section, we use the notation $C_\alpha$ to denote a constant that depends only on $\alpha$, which may take different values in different instances. 

\subsubsection{Proof of Proposition~\ref{prop:suff-cond-norm-cmpt}} \label{subsubsec:proof-propa1}

Since we adopt the strategy in \cite{Lee2024} used in the proof of their Proposition~1, we outline the argument here. It suffices to show that
\begin{align} \label{claim-propa1}
\begin{aligned}
&2\sumjk \left|\int_0^1 \int_0^1 g_j^\rmv(x_j)^\top \tM_\bfsnum{0}{jk}(x_j,x_k) g_k^\rmv(x_k)\dxj\dxk\right| \\
&\le \sqrt{\varphi}\frac{\sqrt{\psi}}{1-\sqrt{\psi}}\frac{4}{C_{p,L}^\rmuniv\mu_2} \sumj \norm{g_j^\rmtp}_\tMnum{0}^2 + C_\bfnum{0}(1+C)^2\sqrt{h_\bfnum{0}}|\cS_\bfnum{0}|\sumjs \norm{g_j^\rmtp}_\tMnum{0}^2 \\
&\le \sqrt{\varphi}\frac{\sqrt{\psi}}{1-\sqrt{\psi}}\frac{4}{C_{p,L}^\rmuniv\mu_2} \sumj \norm{g_j^\rmtp}_\tMnum{0}^2 + C_\bfnum{0}(1+C)^2\sqrt{h_\bfnum{0}}|\cS_\bfnum{0}|\sumj \norm{g_j^\rmtp}_\tMnum{0}^2,
\end{aligned}
\end{align}
for some constant $0<C_\bfnum{0}<\infty$, since the remaining parts follow from the inequality
\begin{align*}
\normbig{\sumj g_j^\rmtp}_\tMnum{0}^2 \ge \sumj \norm{g_j^\rmtp}_\tMnum{0}^2 - 2\sumjk \left|\int_0^1 \int_0^1 g_j^\rmv(x_j)^\top \tM_\bfsnum{0}{jk}(x_j,x_k) g_k^\rmv(x_k)\dxj\dxk\right|.
\end{align*}

To this end, we claim that there exists an absolute constant $0<\widetilde C_1<\infty$ such that
\begin{align} \label{claim-of-claim-propa1}
\begin{aligned}
\max_{(j,k)\in[d]^2}\int_0^1 \int_0^1 \normbig{\tM_\bfsnum{0}{jk}(x_j,x_k) - M_\bfsnum{0}{jk}(x_j,x_k)}_F^2\dxj\dxk &\le \frac{C_1^2}{4}h_\bfnum{0}, \\
\max_{(j,k)\in[d]^2}\int_0^1 \int_0^1 \normbig{\tp_\bfsnum{0}{j}^\rmv(x_j)\tp_\bfsnum{0}{k}^\rmv(x_k)^\top - p_\bfsnum{0}{j}^\rmv(x_j)p_\bfsnum{0}{k}^\rmv(x_k)^\top}_F^2 \dxj\dxk &\le \frac{C_1^2}{4}h_\bfnum{0},
\end{aligned}
\end{align}
where $\norm{\cdot}_F$ denotes the Frobenius norm. These bounds follow from standard results in kernel smoothing theory and are omitted for brevity. Using \eqref{claim-of-claim-propa1}, we derive
\begin{align*}
&2\sumjk \left|\int_0^1 \int_0^1 g_j^\rmv(x_j)^\top \tM_\bfsnum{0}{jk}(x_j,x_k) g_k^\rmv(x_k)\dxj\dxk\right| \\
&=2\sumjk \left|\int_0^1 \int_0^1 g_j^\rmv(x_j)^\top \left(\tM_\bfsnum{0}{jk}(x_j,x_k) - \tp_\bfsnum{0}{j}^\rmv(x_j)\tp_\bfsnum{0}{k}^\rmv(x_k)^\top\right)g_k^\rmv(x_k)\dxj\dxk\right| \\
&\le 2\sumjk \norm{g_j^\rmtp}_{I_{d+1}}\norm{g_k^\rmtp}_{I_{d+1}}\cdot \left(\int_0^1\int_0^1\normbig{\tM_\bfsnum{0}{jk}(x_j,x_k) - \tp_\bfsnum{0}{j}^\rmv(x_j)\tp_\bfsnum{0}{k}^\rmv(x_k)^\top}_F^2\dxj\dxk\right)^{\frac{1}{2}} \\
&\le 2\sumjk \norm{g_j^\rmtp}_{I_{d+1}}\norm{g_k^\rmtp}_{I_{d+1}}\sqrt{\varphi}\psi^{|j-k|/2} + C_1\sqrt{h_\bfnum{0}}\cdot 2\sumjk \norm{g_j^\rmtp}_{I_{d+1}}\norm{g_k^\rmtp}_{I_{d+1}} \\
&\le \sumjk \left(\norm{g_j^\rmtp}_{I_{d+1}}^2+\norm{g_k^\rmtp}_{I_{d+1}}^2\right)\sqrt{\varphi}\psi^{|j-k|/2} + C_1\sqrt{h_\bfnum{0}}\cdot 2\sumjk \norm{g_j^\rmtp}_{I_{d+1}}\norm{g_k^\rmtp}_{I_{d+1}} \\
&\le 2\sqrt{\varphi}\frac{\sqrt{\psi}}{1-\sqrt{\psi}}\sumj \norm{g_j^\rmtp}_{I_{d+1}}^2 + C_1\sqrt{h_\bfnum{0}}\left(\sumj \norm{g_j^\rmtp}_{I_{d+1}}\right)^2.
\end{align*}
From Lemma~\ref{lem:tech1}, we have for all $j\in[d]$ that
\begin{align*}
\norm{g_j^\rmtp}_{I_{d+1}} \le \sqrt{\frac{2}{C_{p,L}^\rmuniv\mu_2}}\norm{g_j^\rmtp}_\tMnum{0}.
\end{align*}
Substituting this and defining 
\begin{align*}
C_\bfnum{0} :=\frac{2C_1}{C_{p,L}^\rmuniv\mu_2},
\end{align*}
we obtain the desired \eqref{claim-propa1}.

\subsubsection{Proof of Lemma~\ref{lem:u-stat1}}

We observe that
\begin{align} \label{decompose-univ}
\begin{aligned}
&\max_{j\in[d]}\left\Vert U_j^\top \cdot \frac{1}{n} \sum_{i=1}^n Z_{j}^i(x_j) K_{h_j}(x_j, X_{j}^i) \ve^i \right\Vert_M^2 \\
&\le \max_{j\in[d]}\left(\frac{1}{n^2} \sum_{i=1}^n \int_0^1 Z_{j}^i(x_j)^\top M_{jj}(x_j) Z_{j}^i(x_j) K_{h_j}(x_j, X_{j}^i)^2 \, \dxj \cdot (\ve^i)^2\right) \\
&\quad + \max_{j\in[d]}\left(\frac{1}{n^2} \sumiiprime \int_0^1 Z_{j}^i(x_j)^\top M_{jj}(x_j) Z_{j}^{i'}(x_j) K_{h_j}(x_j, X_{j}^i) K_{h_j}(x_j, X_{j}^{i'}) \, \dxj \cdot \ve^i \ve^{i'}\right).
\end{aligned}
\end{align}
Note that
\begin{align*}
Z_{j}^i(x_j)^\top M_{jj}(x_j) Z_{j}^i(x_j) K_{h_j}(x_j, X_{j}^i)^2 \le 4 K_{h_j}(x_j, X_{j}^i)^2.
\end{align*}
Using this bound, we obtain
\begin{align} \label{first-term-bound-univ}
\begin{aligned}
&\max_{j\in[d]}\left(\frac{1}{n^2} \sum_{i=1}^n \int_0^1 Z_{j}^i(x_j)^\top M_{jj}(x_j) Z_{j}^i(x_j) K_{h_j}(x_j, X_{j}^i)^2 \dxj \cdot (\ve^i)^2\right) \\
&\le \max_{j\in[d]}\left(\frac{4}{nh_j} \sum_{i=1}^n \int_0^1 (K^2)_{h_j}(x_j, X_{j}^i) \dxj \cdot \left(\sumin (\ve^i)^2\right)\right) \\
&\lesssim \frac{1}{nh}, 
\end{aligned}
\end{align}
where $(K^2)_{h}(u,v) := \frac{1}{h} K_{h}(u,v)^2$. We have used the fact that 
\begin{align} \label{kernel-property}
\max_{j\in[d]}\sup_{v\in[0,1]}\left(\int_0^1 (K^2)_{h_j}(u,v)\, \rmd u\right)<\infty
\end{align}
This yields the bound for the first term in \eqref{decompose-univ}.

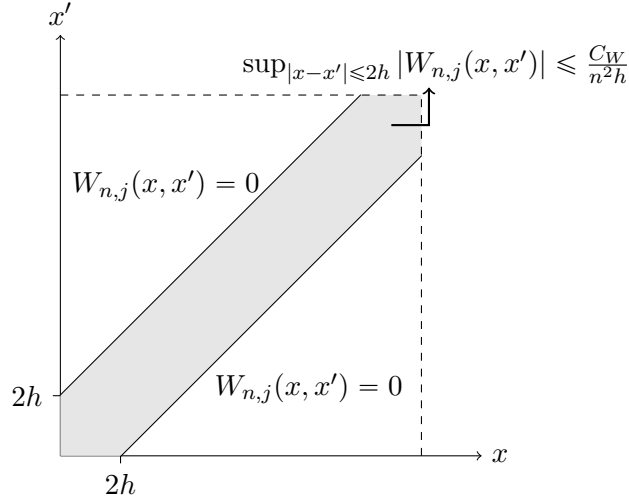
\begin{figure}
\centering
\begin{tikzpicture}[scale=2]

\draw[->] (0,0) -- (2.8,0) node[right] {$x$};
\draw[->] (0,0) -- (0,2.8) node[above] {$x'$};

\draw[thick, domain=0.4:2.4] plot (\x,{\x-0.4});
\draw[thick, domain=0:2] plot (\x,{\x+0.4});

\begin{scope}
  \clip (0,0) rectangle (2.8,2.8); 
  \fill[gray!20]
    (0.4,0) -- (2.4,2) -- (2.4, 2.4) -- (2,2.4) -- (0,0.4) -- (0,0) -- cycle;
\end{scope}

\draw[dashed] (2.4,0) -- (2.4,2.4);
\draw[dashed] (0,2.4) -- (2.4,2.4);

\node at (1.65,0.45) {$W_{n,j}(x,x') = 0$};
\node at (0.7,1.8) {$W_{n,j}(x,x') = 0$};
\node at (2.5,2.6) {$\sup_{|x-x'|\le 2h}|W_{n,j}(x,x')| \leq \frac{C_W}{n^2h}$};

\draw[thick, ->] (2.2,2.2) -- (2.45,2.2) -- (2.45,2.45);

\draw (0.4,0pt) -- (0.4,-0.05) node[below] {$2h$};
\draw (0pt,0.4) -- (-0.05,0.4) node[left] {$2h$};

\end{tikzpicture}
\caption{Illustration of the support and magnitude of $W_{n,j}(x, x')$ on $[0,1]^2$. The function $W_n(x, x')$ is nonzero only when $|x - x'| \le 2h$, and is uniformly bounded by $\frac{C_W}{n^2h}$ for an absolute constant $C_W$ within gray band.}
\label{fig:Wn-prop}
\end{figure}

For the second term in \eqref{decompose-univ}, we apply Theorem~\ref{thm:u-stat}. Denote this term by $\Ub_{n,j}$. Then, it can be written as
\begin{align*}
\Ub_{n,j} = \sumiiprime \ve_i \, W_{n,j}(X_{j}^i, X_{j}^{i'}) \, \ve^{i'},
\end{align*}
where
\begin{align*}
W_{n,j}(X_{j}^i, X_{j}^{i'}) := \frac{1}{n^2} \int_0^1 Z_{j}^i(x_j)^\top M_{jj}(x_j) Z_{j}^{i'}(x_j) K_{h_j}(x_j, X_{j}^i) K_{h_j}(x_j, X_{j}^{i'}) \, \dxj.
\end{align*}
We note that $W_{n,j}$ is a symmetric and measurable function on $[0,1]^2$. Moreover, $W_{n,j}(x, x')$ vanishes whenever $|x - x'| \ge 2h$, due to the compact support of the kernel function. This structure allows us to visualize $W_{n,j}$ as depicted in Figure~\ref{fig:Wn-prop}. In the figure, $W_n$ is uniformly bounded by $C_W / (n^2 h)$ for some absolute constant $C_W > 0$, and its support is contained in the gray region, which has Lebesgue measure proportional to $h$, and identically zero outside this region. 

Next, we derive bounds for the terms $\Omega_{n,\ell}^{(j)}$, which corresponds to $\Omega_{n,\ell}$ in Theorem~\ref{thm:u-stat}. First, it is clear that 
\begin{align} \label{omega1-univ}
\Omega_{n,1}^{(j)} \le \frac{C_W (\log n)^{\frac{1}{\alpha^*} + \frac{2}{\alpha}}}{n^2h}.
\end{align}
Since 
\begin{align*}
\Eb(W_{n,j}(X_{j}^i, X_{j}^{i'})^2) \le  \frac{C_W^2}{n^4h^2}\cdot h = \frac{C_W^2}{n^4h}, 
\end{align*}
it follows that 
\begin{align} \label{omega2-univ}
\Omega_{n,2}^{(j)}\le \left(n(n-1) \cdot \frac{C_W^2}{n^4h}\right)^{\frac{1}{2}} \le \frac{2C_W}{nh^{\frac{1}{2}}}.
\end{align}
For the term $\Omega_{n,3}^{(j)}$, we first note that $\sup_{x\in[0,1]}\Eb(|W_{n,j}(x, X_{j}^{i'})|) \le \frac{C_W}{n^2}$. This entails that, for $\{\eta_i\}_{i=1}^n$ and $\{\zeta_i\}_{i=1}^n$ such that 
\begin{align*}
\sumin \Eb(\eta_i(X_{j}^i)^2)\le 1, \quad \sumin \Eb(\zeta_i(X_{j}^i)^2)\le 1, 
\end{align*}
it follows that 
\begin{align*}
&\sumiiprime \Eb(\eta_i(X_{j}^i)|W_{n,j}(X_{j}^i, X_{j}^{i'})|\zeta_{i'}(X_{j}^{i'}))  \\
&\le \frac{1}{2}\sumiiprime \left\{\Eb(\eta_i(X_{j}^i)^2|W_{n,j}(X_{j}^i, X_{j}^{i'})|) + \Eb(\zeta_{i'}(X_{j}^{i'})^2|W_{n,j}(X_{j}^i, X_{j}^{i'})|)\right\} \\
&\le \frac{C_W}{2n^2}\sumiiprime \left\{\Eb(\eta_i(X_{j}^i)^2) + \Eb(\zeta_{i'}(X_{j}^{i'})^2)\right\} \\
&\le \frac{C_W}{n}. 
\end{align*}
Here, we used Young’s inequality for the first inequality. This gives 
\begin{align} \label{omega3-univ}
\Omega_{n,3}^{(j)} \le \frac{C_W}{n}.
\end{align}
A similar approach leading to \eqref{omega2-univ} yields 
\begin{align} \label{omega4-univ}
\Omega_{n,4}^{(j)} \le (\log n)^{\frac{1}{\alpha}}\left(\frac{C_W^2}{n^4h^2}\cdot nh\right)^{\frac{1}{2}} \le \frac{C_W(\log n)^{\frac{1}{\alpha}}}{n^{\frac{3}{2}}h^{\frac{1}{2}}}. 
\end{align}
Recalling that $\Omega_{n,5}^{(j)} = (\log n)^{\frac{1}{2}}\Omega_{n,1}^{(j)} + (\log n)\Omega_{n,4}^{(j)}$ and the following result from Theorem~\ref{thm:u-stat}:
\begin{align*}
\Pb\left(|\Ub_{n,j}| \ge C_\alpha\left(t^{\frac{2}{\alpha^*}}\Omega_{n,1}^{(j)} + t^{\frac{1}{2}}\Omega_{n,2}^{(j)} + t\Omega_{n,3}^{(j)} + t^{\frac{1}{2} + \frac{1}{\alpha^*}}\Omega_{n,4}^{(j)} + t^{\frac{1}{\alpha^*}}\Omega_{n,5}^{(j)}\right)\right) \le 2\exp(-t). 
\end{align*}
Combining the results in \eqref{omega1-univ}, \eqref{omega2-univ}, \eqref{omega3-univ} and \eqref{omega4-univ}, and plugging in $t=C_1\log d$ for some absolute constant $0<C_1<\infty$, we further obtain that 
\begin{align*}
\Pb\left(\max_{j\in[d]}|\Ub_{n,j}| \ge C_\alpha \cdot A(n,h,d;\alpha)\right) \lesssim d^{-1}
\end{align*}
which together with \eqref{first-term-bound-univ} completes the proof.

\subsubsection{Proof of Lemma~\ref{lem:u-stat2} and \ref{lem:u-stat3}}

We provide the proof of Lemma~\ref{lem:u-stat3} only, as the proof of Lemma~\ref{lem:u-stat2} is similar and simpler. For notational convenience, we often write
\begin{align*}
b_{ij}(x_j) := \left(\frac{X_{j}^i - x_j}{h_j}\right), \quad \kappa_{ij}(x_j) := K_{h_j}(x_j,X_{j}^i), \quad j \in [d].
\end{align*}
Observe that, for any $g_k^\rmtp \in \sH_k^\rmtp\in\Bb(1)$,
\begin{align*}
&\left\Vert U_j^\top\int_0^1 \left(\hM_{jk}(\cdot, x_k) - \tM_{jk}(\cdot,x_k)\right)g_k^\rmv(x_k)\dxk\right\Vert_M^2 \le \int_0^1 \int_0^1 \normbig{\hM_{jk}(x_j,x_k) - \tM_{jk}(x_j,x_k)}_F^2 \dxj\dxk ,
\end{align*}
where $\norm{\cdot}_F$ denotes the Frobenius norm of a matrix. Here, we have used the inequality
\begin{align*}
\norm{A b} \le \norm{A}_{\rmop} \cdot \norm{b} \le \norm{A}_F \cdot \norm{b}, \quad A \in \Rb^{\ell \times \ell}_{\rm sym}, \, b \in \Rb^\ell,
\end{align*}
where $\Rb^{\ell \times \ell}_{\rm sym}$ denotes the space of symmetric matrices, $\norm{\cdot}$ denotes the Euclidean norm, and $\norm{\cdot}_{\rmop}$ denotes the operator norm.
We note that the $(\ell, \ell')$-th element of $\hM_{jk}(x_j,x_k) - \tM_{jk}(x_j,x_k)$ is given by 
\begin{align*}
\frac{1}{n}\sumin \left\{b_{ij}(x_j)^{\ell-1}b_{ik}(x_k)^{\ell'-1}\kappa_{ij}(x_j)\kappa_{ik}(x_k) - \Eb\left(b_{1j}(x_j)^{\ell-1}b_{1k}(x_k)^{\ell'-1}\kappa_{1j}(x_j)\kappa_{1k}(x_k)\right)\right\},
\end{align*}
for $1\le \ell,\ell'\le 2$. We denote this quantity by $\Ms_{n,jk,\ell,\ell'}(x_j,x_k)$. We claim that 
\begin{align} \label{claim-biv}
\max_{(j,k)\in[d]^2} \left(\int_0^1\int_0^1 \Ms_{n,jk,\ell,\ell'}(x_j,x_k)^2\dxj\dxk \right) \lesssim \frac{1}{nh^2} + B(n,h^2,d), \quad 1\le \ell,\ell'\le 2. 
\end{align}

Below, we provide the proof of the claim in \eqref{claim-biv} for the case $\ell = \ell' = 1$, as the other cases can be treated analogously. Observe that 
\begin{align*}
&\int_0^1 \int_0^1 \left\{\frac{1}{n}\sumin \kappa_{ij}(x_j)\kappa_{ik}(x_k) - \Eb\left(\kappa_{1j}(x_j)\kappa_{1k}(x_k)\right) \right\}^2\dxj\dxk \\
&=\frac{1}{n^2}\sumin \int_0^1 \int_0^1 \left\{ \kappa_{ij}(x_j)\kappa_{ik}(x_k) - \Eb\left(\kappa_{1j}(x_j)\kappa_{1k}(x_k)\right) \right\}^2\dxj\dxk \\
&\quad + \frac{1}{n^2}\sumiiprime \int_0^1 \int_0^1 \left\{ \kappa_{ij}(x_j)\kappa_{ik}(x_k) - \Eb\left(\kappa_{1j}(x_j)\kappa_{1k}(x_k)\right) \right\} \\
&\hspace{5cm}\times \left\{ \kappa_{i'j}(x_j)\kappa_{i'k}(x_k) - \Eb\left(\kappa_{1j}(x_j)\kappa_{1k}(x_k)\right) \right\}\dxj\dxk \\
&\overset{\rm let}{=:} U_{n,jk}^{(1)} + U_{n,jk}^{(2)}. 
\end{align*}
We note that
\begin{align*}
\frac{1}{n^2} \sumin \int_0^1 \int_0^1 \kappa_{ij}(x_j)^2 \kappa_{ik}(x_k)^2 \dxj\dxk &= \frac{1}{n^2h_jh_k} \sumin \int_0^1 \int_0^1 (K^2)_{h_j}(x_j,X_{j}^i)(K^2)_{h_k}(x_k, X_{k}^i) \dxj\dxk.
\end{align*}
Together with \eqref{kernel-property} in the proof of Lemma~\ref{lem:u-stat1}, this implies
\begin{align} \label{first-result-sjk1}
\max_{(j,k)\in[d]^2} \left(\int_0^1 \int_0^1 \left\{\frac{1}{n}\sumin \kappa_{ij}(x_j)\kappa_{ik}(x_k) - \Eb\left(\kappa_{1j}(x_j)\kappa_{1k}(x_k)\right) \right\}^2\dxj\dxk\right) \lesssim \frac{1}{nh^2}. 
\end{align}
Moreover, since
\begin{align*}
\Eb(\kappa_{1j}(x_j)\kappa_{1k}(x_k)) &= \int_0^1 \int_0^1 K_{h_j}(x_j,u_j) K_{h_k}(x_k, u_k) p_{j,k}(u_j,u_k) \duj\duk \\
&\le C_{p,U}^{\rmbiv,1}\int_0^1 \int_0^1 K_{h_j}(x_j,u_j) K_{h_k}(x_k, u_k) \duj\duk \\
&\le 4C_{p,U}^{\rmbiv,1},
\end{align*}
it can be shown that
\begin{align} \label{second-result-sjk1}
\max_{(j,k)\in[d]^2}\sup_{x_j,x_k\in[0,1]} \left|\Eb(\kappa_{1j}(x_j)\kappa_{1k}(x_k))\right|\le C_1
\end{align}
for some absolute constant $0 < C_1 < \infty$. Combining \eqref{first-result-sjk1} and \eqref{second-result-sjk1}, and applying Young's inequality, we obtain
\begin{align} \label{bound-sjk1}
\max_{(j,k)\in[d]^2} |U_{n,jk}^{(1)}| \lesssim \frac{1}{nh^2}. 
\end{align}

Next, we bound the second term $U_{n,jk}^{(2)}$.
Define a symmetric function $W_{n,jk}$ by
\begin{align*}
&W_{n,jk}((X_{j}^i, X_{k}^i), (X_{j}^{i'}, X_{k}^{i'})) := \frac{1}{n^2} \int_0^1 \int_0^1 \left\{ \kappa_{ij}(x_j)\kappa_{ik}(x_k) - \Eb\left(\kappa_{1j}(x_j)\kappa_{1k}(x_k)\right) \right\} \\
&\hspace{7cm}\times \left\{ \kappa_{i'j}(x_j)\kappa_{i'k}(x_k) - \Eb\left(\kappa_{1j}(x_j)\kappa_{1k}(x_k)\right) \right\}\dxj\dxk.
\end{align*}
Note that $U_{n,jk}^{(2)} = \sumiiprime W_{n,jk}((X_{j}^i, X_{k}^i), (X_{j}^{i'}, X_{k}^{i'}))$ is a degenerate $U$-statistic of order 2. Since the result of Lemma~\ref{lem:start-u-stat} holds without requiring structural assumptions on $\Wb$, we may apply it to obtain
\begin{align} \label{first-eq-biv}
\norm{U_{n,jk}^{(2)}}_\ell \le 48\normbig{\sumiiprime w_i W_{n,jk}((X_{j}^i, X_{k}^i), (X_{j}^{i'}, X_{k}^{i'}))w_{i'}}_\ell, \quad \ell\ge 2.
\end{align}
Here, $\{w_i\}_{i=1}^n$ is a Rademacher sequence independent of $\{(X_{j}^i, X_{k}^i)\}_{i=1}^n$, and $\{({X'}_{j}^i, {X'}_{k}^i)\}_{i=1}^\infty$ and $\{w_i'\}_{i=1}^n$ are decoupled random sequences corresponding to $\{(X_{j}^i, X_{k}^i)\}_{i=1}^n$ and $\{w_i\}_{i=1}^n$, respectively. For each $i \in [n]$, define $V_i := (X_{j}^i, X_{k}^i, w_i)$ and $V_i' := ({X'}_{j}^i, {X'}_{k}^i, w_i')$. Also define a function $h_{n,jk}$ by
\begin{align*}
h_{n,jk}(V_i, V_{i'}') := w_i W_{n,jk}((X_{j}^i, X_{k}^i), (X_{j}^{i'}, X_{k}^{i'}))w_{i'}.
\end{align*}
Then $\sumiiprime h_{n,jk}(V_i, V_{i'}')$ forms a decoupled and degenerate $U$-statistic of order 2. Let
\begin{align*}
\Uc_{n,jk}^{(2,1)} &:= \left(\sumiiprime \Eb(h_{jk,i,i'}^2)\right)^{\frac{1}{2}},\\
\Uc_{n,jk}^{(2,2)} &:= \Eb\left(\max_{i\in[n]}\Eb\left(\sum_{i'=1,\ne i}^n h_{jk,i,i'}^2 \Biggr| V_i\right)^{\frac{1}{2}}\right), \\
\Uc_{n,jk}^{(2,3)} &:= \norm{(h_{jk,i,i'})}_{L^2\to L^2}, \\
\Uc_{n,jk}^{(2,4)} &:= \Eb\left(\max_{i,i'}|h_{jk,i,i'}|^\ell\right)^{\frac{1}{\ell}},
\end{align*}
where, as in the statement of Lemma~\ref{lem:thm3.2gine}, we denote $h_{n,jk}(V_i, V_{i'}')$ simply by $h_{jk,i,i'}$. Then, applying Lemma~\ref{lem:thm3.2gine}, we obtain
\begin{align*}
\normbig{\sumiiprime h_{jk,i,i'}}_\ell \le C_2\left(\ell^{\frac{1}{2}}\Uc_{n,jk}^{(2,1)} + \ell^{\frac{3}{2}}\Uc_{n,jk}^{(2,2)} + \ell\Uc_{n,jk}^{(2,3)} + \ell^2\Uc_{n,jk}^{(2,4)}\right),
\end{align*}
for some absolute constant $0 < C_2 < \infty$. Notably, $C_2$ is independent of the choice of $(j,k)\in[d]^2$.

To bound the terms $\Uc_{n,jk}^{(2,1)}$--$\Uc_{n,jk}^{(2,4)}$, we proceed by analyzing the structural properties of $W_{n,jk}$, in the same spirit as our treatment of $W_{n,j}$ in the proof of Lemma~\ref{lem:u-stat1} (see also Figure~\ref{fig:Wn-prop}). Observe that
\begin{align*}
&W_{n,jk}((u_j,u_k),(u_j',u_k')) \\
&=\frac{1}{n^2} \int_0^1 \int_0^1 \left(K_{h_j}(x_j,u_j)K_{h_k}(x_k,u_k) - \Eb(K_{h_j}(x_j,X_j)K_{h_k}(x_k, X_k))\right) \\
&\hspace{5cm} \times \left(K_{h_j}(x_j,u_j')K_{h_k}(x_k,u_k') - \Eb(K_{h_j}(x_j,X_j)K_{h_k}(x_k, X_k))\right) \dxj\dxk \\
&= \frac{1}{n^2} \int_0^1 \int_0^1K_{h_j}(x_j,u_j)K_{h_k}(x_k,u_k)K_{h_j}(x_j,u_j')K_{h_k}(x_k,u_k') \dxj\dxk \\
&\quad + R_{n,jk}((u_j,u_k), (u_j',u_k')),
\end{align*}
where $\norm{\cdot}_{L^2\to L^2}$ is defined as in Lemma~\ref{lem:thm3.2gine}, and $R_{n,jk}$ denotes the remainder terms. A standard argument yields
\begin{align*}
\max_{(j,k)\in[d]^2}\sup_{(u_j,u_k), (u_j', u_k')\in[0,1]^2} |R_{n,jk}((u_j,u_k), (u_j',u_k'))| \lesssim \frac{1}{n^2}.
\end{align*}
Therefore, we obtain
\begin{align} \label{prop-wnjk}
\begin{aligned}
|W_{n,jk}((u_j,u_k),(u_j',u_k'))| \le \begin{cases}
\frac{C_3}{n^2h^2} &\quad \text{if } |u_j-u_j'|\le 2h_j \text{ and } |u_k-u_k'|\le 2h_k, \\
\frac{C_3}{n^2} &\quad \text{otherwise,}
\end{cases}
\end{aligned}
\end{align}
for some absolute constant $0 < C_3 < \infty$. Using this property along with the uniform boundedness of the bivariate density function $p_{jk}$, it follows directly that
\begin{align} \label{other-terms-biv}
\Uc_{n,jk}^{(2,1)} \le \frac{C_3}{nh}, \quad \Uc_{n,jk}^{(2,2)} \le \frac{C_3}{n^{3/2}h}, \quad \Uc_{n,jk}^{(2,4)} \le \frac{C_3}{n^2h^2}.
\end{align}
It remains to bound $\Uc_{n,jk}^{(2,3)}$. To this end, note that $\norm{(h_{jk,i,i'})}_{L^2\to L^2} = \norm{(|h_{jk,i,i'}|)}_{L^2\to L^2}$. Also, using \eqref{prop-wnjk}, we have
\begin{align*}
\max_{i}\Eb(|h_{jk,i,i'}||V_i) = \max_{i'}\Eb(|h_{jk,i,i'}||V_{i'}') \le \frac{C_3}{n^2}.
\end{align*}
Hence, we derive
\begin{align*}
\sumiiprime \Eb(\eta_i(V_i) |h_{jk,i,i'}|\zeta_{i'}(V_{i'}')) &\le \frac{1}{2}\sumiiprime \left\{\Eb(\eta_i(V_i)^2 |h_{jk,i,i'}|) + \Eb(\zeta_{i'}(V_{i'}')^2 |h_{jk,i,i'}|)\right\} \\
&\le \frac{C_3}{2n^2} \sumiiprime \left\{\Eb(\eta_i(V_i)^2) + \Eb(\zeta_{i'}(V_{i'}')^2)\right\} \\
&\le \frac{C_3}{n}.
\end{align*}
This gives
\begin{align} \label{third-term-biv}
\Uc_{n,jk}^{(2,3)} \le \frac{C_3}{n}.
\end{align}
Combining \eqref{other-terms-biv} and \eqref{third-term-biv}, we obtain
\begin{align} \label{ell-norm-biv}
\normbig{\sumiiprime h_{jk,i,i'}}_\ell \le C_4 \left(\ell^{\frac{1}{2}} \frac{1}{nh} + \ell^{3/2}\frac{1}{n^{3/2}h} + \ell\frac{1}{n} + \ell^2 \frac{1}{n^2h^2}\right),
\end{align}
for some absolute constant $0 < C_4 < \infty$.

Combining the result in \eqref{first-eq-biv} with \eqref{ell-norm-biv} and applying Markov's inequality, we may conclude that
\begin{align*}
\Pb\left(|U_{n,jk}^{(2)}| \ge C_5\left(t^{\frac{1}{2}} \frac{1}{nh} + t^{3/2}\frac{1}{n^{3/2}h} + t\frac{1}{n} + t^2 \frac{1}{n^2h^2}\right)\right) \le 2\exp(-t),
\end{align*}
for some absolute constant $0 < C_5 < \infty$. Since $C_5$ is independent of the choice of $(j,k) \in [d]^2$ and $\log d=o(nh)$, setting $t = C_6\log d$ for some absolute constant $0<C_6<\infty$ yields
\begin{align*}
\Pb\left(\max_{(j,k)\in[d]^2}|U_{n,jk}^{(2)}| \gtrsim B(n,h^2,d)\right) \lesssim d^{-1},
\end{align*}
which, together with \eqref{bound-sjk1}, completes the proof of the lemma.
\newpage

\bibliographystyle{apalike}
\bibliography{biblio}

\end{document}